\documentclass{article} 
\usepackage{iclr2027_conference,times}

\usepackage{amsmath,amsfonts,bm}

\def\eqref#1{equation~\ref{#1}}

\def\1{\bm{1}}

\DeclareMathAlphabet{\mathsfit}{\encodingdefault}{\sfdefault}{m}{sl}
\SetMathAlphabet{\mathsfit}{bold}{\encodingdefault}{\sfdefault}{bx}{n}

\usepackage{hyperref}
\hypersetup{hypertexnames=false}
\usepackage{url}
\usepackage{graphicx}
\usepackage[export]{adjustbox}
\usepackage{amsmath}
\usepackage{amssymb}
\usepackage{amsfonts}
\usepackage{booktabs}
\usepackage{multirow}
\usepackage{makecell}
\usepackage{enumitem}
\usepackage{array}
\usepackage{longtable}
\usepackage{wrapfig}
\usepackage{nicefrac}
\usepackage{siunitx}
\usepackage{xspace}
\usepackage{xcolor}
\usepackage{inconsolata}
\usepackage{fancyvrb}
\usepackage{fvextra}
\usepackage{CJKutf8}
\usepackage{placeins}
\usepackage{float}

\title{
    \centering
    VisCAD: A Foundation Model Suite with Multimodal Industrial CAD Intelligence
}

\newcommand{\viscad}{\texttt{VisCAD}\xspace}
\newcommand{\viscadm}{\texttt{VisCAD-M1}\xspace}
\newcommand{\viscadh}{\texttt{VisCAD-H1}\xspace}
\newcommand{\pcb}{\texttt{PubCADBench}\xspace}
\newcommand{\rcb}{\texttt{RealCADBench}\xspace}
\newcommand{\codex}{\texttt{Codex}\xspace}
\newcommand{\cc}{\texttt{Claude Code}\xspace}

\author{\parbox[t]{\dimexpr\textwidth-2\tabcolsep\relax}{%
\centering\normalfont\small
\textbf{Guanlin Li, Zhichao Huang, Huimu Yu, Yichen Long, Hongsen Liu\thanks{Corresponding to Hongsen Liu (\href{mailto:liuhongsen3@jd.com}{\texttt{liuhongsen3@jd.com}}), and Yuchen Wang (\href{mailto:wangyuchen.101@jd.com}{\texttt{wangyuchen.101@jd.com}}).}, Linxin Cai, Qiuhe Hong}\\
\textbf{Ziqi Liu, Luya Wang, Wenxiang Wu, Ning Zhang, Yuchen Wang\textsuperscript{*}}\\[0.5em]
JD Industrial
}}

\iclrfinalcopy 

\begin{document}

\maketitle
\lhead{Technical report of \texttt{VisCAD}}
\setlength{\parskip}{0.28pc plus 0.05pc minus 0.08pc}
\setlength{\parsep}{0.12em}

\begin{abstract}
AI-assisted computer-aided design (CAD) for industrial products involves two challenging phases.
Part-level generation maps diverse forms of user intent, including \textit{renders}, \textit{text descriptions}, \textit{2D drawings}, and \textit{real photographs}, to executable programs in a CAD domain-specific language.
Assembly-level generation must additionally handle interacting parts, plan mating relations, estimate poses, and place all parts correctly.
Existing specialized CAD models are commonly trained on narrow input domains (renders or texts) and often generalize poorly, while general-purpose frontier models cover broader inputs but perform inconsistently across CAD domains.
We present \viscad, a foundation model suite designed to provide both broad generalization and strong CAD capability for realistic industrial products.
In its core is \viscadm, a 27B model trained through mid-training and post-training for part-level design generation.
On \pcb and \rcb, \viscadm achieves the highest average part-level score among the evaluated models, reaching 0.5540 compared with 0.5496 for the strongest frontier model.
Reusing \viscadm as a test-time verifier can further raise the score to 0.5797, an approximately 5 percent relative improvement over the previous state of the art.
\viscad also includes a domain-specific harness that leverages frontier models for complex assembly generation and demonstrates advantages over general-purpose harnesses in both quantitative and qualitative evaluations.
\end{abstract}

\section{Introduction}
\label{sec:introduction}

Frontier and open-source multimodal foundation models have made rapid progress in code understanding, generation, and tool use, and are increasingly deployed as coding assistants for real-world software development~\citep{gpt5systemcard,qwen3vl,codexproduct,claudecode}.
This amazing progress suggests a natural route to CAD intelligence.
At its core, Computer-Aided Design (CAD) maps a user's design intent to a \emph{generalized parametric CAD program}, with two typical \textit{dialects}.
Such a program could be represented as a sequence of commands that mirrors human operations in a CAD software-specific GUI, or as geometric operations exposed by a CAD system or geometry kernel.
CAD generation can therefore be formulated as multimodal program generation: given intent expressed through language or visual evidence, a model produces an executable program under the API of a target Domain Specific Language (DSL) of a CAD environment.

The recent two years have consequently witnessed a growing body of large-model-based AI CAD research~\citep{cadsurvey2026,llm4cad,cadllama,openecad,text2cad,cadrecode,cadvlm,picasso}.
However, existing specialized models remain limited in model capacity, supervision scale, and application scope.
Most focus on single-part generation and accept only one intent modality, such as text or sketches or rendered views.
2D engineering drawings and in-the-wild product photographs---two common forms of industrial design intents---remain substantially underrepresented, and no existing specialized model that we study supports the full spectrum from text and renders to engineering drawings and real product images through a unified interface.
Moreover, many methods are trained and evaluated primarily on ABC, the DeepCAD subset and its derivatives, Fusion 360 Gallery~\citep{abc,deepcad,fusion360,sketchgraphs}.
Although these datasets have enabled important progress, their CAD-native and largely synthetic distributions provide limited coverage of heterogeneous industrial inputs.
Consequently, specialized AI CAD models often generalize poorly beyond their training domains and can trail frontier general-purpose models on diverse CAD tasks.

To address these limitations, we introduce \viscad, a foundation model suite for industrial multimodal CAD modeling (Figure~\ref{fig:viscad_overview}).
At the part level, \viscadm is a 27B multimodal generator that maps abundant design intents, ranging from text descriptions, engineering drawings, real product photographs, to rendered views, to executable FreeCAD Python programs.
Its data-curation and multi-stage training pipelines are designed to combine large-scale CAD programs with heterogeneous industrial visual evidence, improving both modality coverage and broader domain generalization.
At the assembly level, \viscadh is a CAD-native harness that decomposes complex design intent into a bill of materials, per-part representations, mating relations, and global placements.
Its CAD-DSL-agnostic intermediate representation (IR) separates assembly reasoning from backend syntax, allowing the same workflow to be realized in various CAD environments.

On the part-level leaderboards of \pcb and \rcb, \viscadm achieves an average profile score of $0.5540$, exceeding the best compared frontier model score of $0.5496$; parallel test-time scaling further raises the score to $0.5797$.
On a 50-instance assembly-generation study drawn from the same benchmarks, \viscadh obtains an assembly judge score of approximately $85.0$, compared with $68.0$ for \codex and $52.0$ for \cc.
Together, these results show that domain-specific training can provide broad, high-quality part generation, while a CAD-native harness can extend frontier CAD intelligence to complex assemblies and multiple CAD backends.


Our main contributions are:

\begin{figure}[!t]
    \centering
    \includegraphics[width=\linewidth]{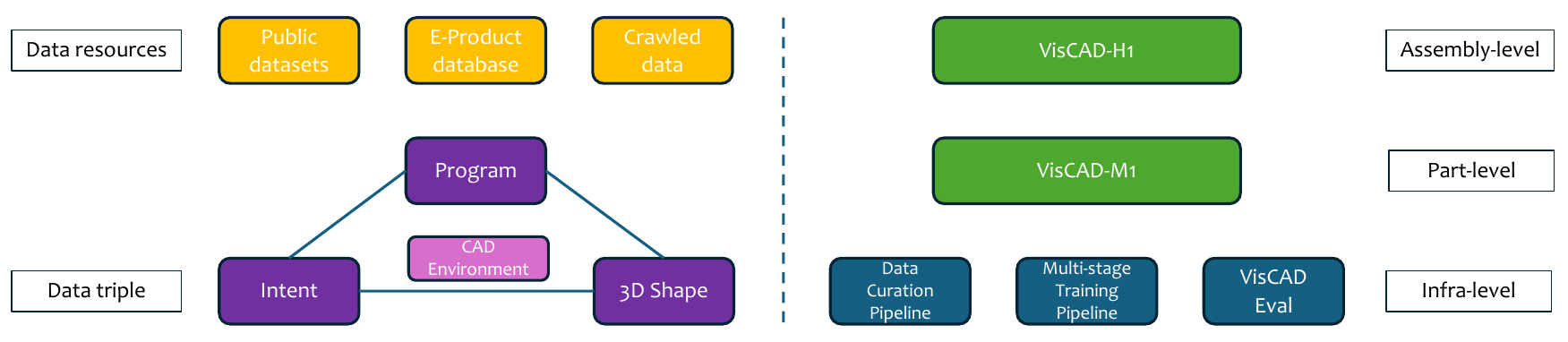}
    \caption{
        Overview of \viscad.
        Heterogeneous data resources are organized around a shared intent--program--shape abstraction and support data curation, multi-stage training, and evaluation.
        \viscadm generates individual parts, while \viscadh extends CAD intelligence to assembly-level design through a structured harness.
    }
    \label{fig:viscad_overview}
\end{figure}

\begin{itemize}
    \item
        We present \viscad, a foundation model suite with a 27B foundation model
        \viscadm that excels at \textit{part-level} design generation
        with a average score that is better than SOTA frontier models
        like \texttt{Gemini-3.1-Pro}
        (Ours $0.5540$ vs SOTA $0.5496$).
        This suite also \textit{previews} our assembly-level design
        generation harness which exhibits more delicate design
        intelligence for complex assemblies than general harnesses
        (\codex, \cc).

    \item
        We develop a test-time scaling method (\texttt{parallel-tts}). With this technique, the 27B model can itself be
        used as a reranker to improve its final prediction given several
        rollouts of the same input design intent.
        The \textit{self-}reranker can further achieve $2.7$ points
        improvements over itself and obtain more than $5\%$ relative
        improvements w.r.t. SOTA frontier model.

    \item
        To attack challenges of the scarcity of diverse general-domain data and the golden programs, we develop a comprehensive data curation pipeline that leverages off-the-shelf image editing and image-to-3D models to support raw data processing, data augmentation and recursive self-improvement based program search for obtaining more accurate CAD program supervision.
        Besides, we also propose a multi-stage training pipeline that maximizes the utilities of CAD data that varies in amount and quality scattered in different stages.

    \item
        We design a DSL-agnostic IR for assembly-level generation in \viscadh that explicitly represents Bills of Materials (BoMs), part-level geometries, mating relations, and global placements etc..
        This IR enables transferring of the same assembly representation across various CAD environments, namely \texttt{FreeCAD}, \texttt{SolidWorks}, \texttt{Autodesk Fusion}.
\end{itemize}

\section{Related Work}
\label{sec:related_work}

Learning-based CAD has moved from generating meshes toward executable programs.
DeepCAD, SketchGraphs, Text2CAD, and CAD-Recode recover parametric command sequences or code from textual inputs, sketches, or point clouds~\citep{deepcad,sketchgraphs,text2cad,cadrecode}.
BRepNet, BrepGen, CAD-Llama, and ParaCAD-RL treat B-Reps and large-model or RL training as direct targets~\citep{brepnet,brepgen,cadllama,paracadrl}.
These methods establish program-native CAD.
They remain part-centric: one solid, usually from CAD-native or synthetic supervision.

A second line of works widens the input modalities and domains.
OpenECAD, CadVLM, ChatCAD, PICASSO, and orthographic-reconstruction work condition on drawings, sketches, or mixed visual-language evidence~\citep{openecad,cadvlm,chatcad,picasso,orthodraw2023,orthodrawrl2025,bitscad}.
Engineering drawings and product photographs remain underrepresented relative to CAD-native renders, and the output is still typically one part rather than a verified assembly.
\viscadm trains for that coverage: factory photographs and drawings sit in the same intent--program--shape map as public code.

\begin{figure}[t]
    \centering
    \includegraphics[width=0.90\linewidth]{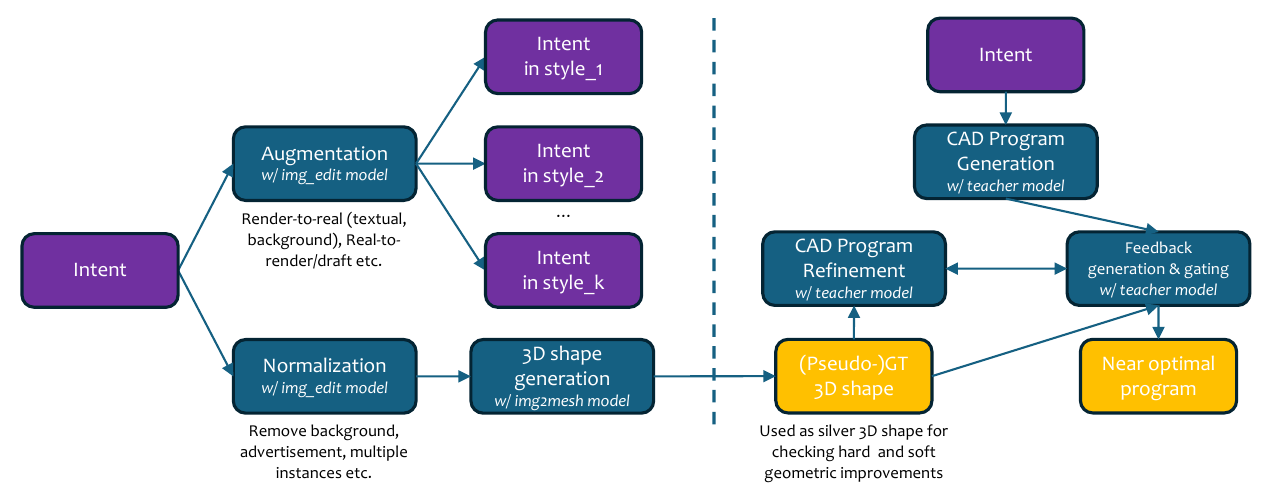}
    \caption{
        Our proposed comprehensive data curation pipeline.
    }
    \label{fig:data_curation_pipeline}
\end{figure}

Benchmarks follow similar split.
ABC and Fusion 360 Gallery supply CAD-native geometry and feature construction histories~\citep{abc,fusion360}.
BenchCAD, CADBench, and P3D-Bench score public program-generation slices~\citep{benchcad,cadbench,p3dbench}; we aggregate those benchmark datasets as \pcb to align with the current state of the research community.
\rcb~\citep{realcadbench} scores real world design intents related to factory automation products under any CAD environment of interest, with executability, Solid IoU, Surface IoU, and visual-semantic judge scores.
We adopt that metric vector (Appendix~\ref{sec:appendix_metrics}) and the same frozen prompts (Appendix~\ref{sec:appendix_judge_prompts}).
\rcb is the measurement contract; \viscad is a model-and-harness suite scored under it.

Harness work shows that execution-time revision changes the delivered system~\citep{gencad3d,gencadselfrepairing,multiagentcad,codexproduct,claudecode}.
A general coding harness can execute, observe errors, and rewrite.
It is not built around structured BoMs, per-part IR gating, or mating review, and it is usually tied to one tool stack.
Scoring a sampled part program and a harnessed loop in one table would mix objects of study.
\viscadh treats the outer loop as the assembly-level object and compares against \codex and \cc as delivered systems.

\section{Overview of \viscad and basics}
\label{sec:overview}

As illustrated in Figure~\ref{fig:viscad_overview}, \viscad is a foundation model suite for industrial CAD modeling or generation, organized into \textit{infrastructure-level}, \textit{part-level}, and \textit{assembly-level} components.
At the infra level, it provides a data curation pipeline that transforms heterogeneous resources into aligned triples of design intent, executable CAD programs, and corresponding 3D shapes.
A dedicated multi-stage training pipeline takes in these curated data to train \viscadm for part-level design modeling or generation.
At the assembly level, \viscadh complements the part generator with a domain-specific harness for planning, part positioning, and verifying multi-part assemblies.
\texttt{VisCAD\_Eval}, at the infra-level, provides a unified framework for evaluating generated CAD programs or designs and catching improvement signals at both part and assembly-level.

\paragraph{Shared abstraction and infrastructure.}
We treat the collection of all available data resources as a collection of triples $t = (i, p, s)$, which may have empty entries.
$i \in \mathcal{I}$ is the \textit{intent space}, which is multimodal in nature.
$ p \in \mathcal{P}$ is the \textit{program space} which can be further indexed by $l$ indicating certain CAD DSL.
$ s \in \mathcal{S}$ is the \textit{3D model space} or \textit{shape space}.
Thus, parametric CAD modeling by VLMs, as our central focus, can be defined as a mapping from intent space to program space where code verification happens, with shape space where visual verification happens.
This abstraction guides the design of our comprehensive data curation pipeline.
Besides, the varied data quality of $t$s also motivates the role it plays for training \viscadm in our multi-stage training pipeline.
\viscad suite also contains a systematic evaluation module that not only evaluates generated $\hat{p}$ with geometry similarity between its exported 3D shape $\hat{s}$ and ground-truth, but also with visual-based judge score that complements \textit{easy-to-hack} weaknesses of geometry metrics.

\paragraph{Part design generation.}
\viscadm, a multimodal foundation model, takes texts, 2D drawings, renders, or product photographs and generates python programs that reconstruct user design intents, obeying the \textit{spec} of \texttt{FreeCAD} API, that can run and export a 3D model.~\footnote{
    \textit{Note that our suite is not restricted to one CAD DSL, our data curation pipeline and harness is essentially DSL-agnostic.
    Our openness to any DSL is an ambitious move towards unifying the CAD software ecosystem.}
}

\paragraph{Assembly design generation.}
\viscadh takes multi-part intent and delivers an assembly through agentic workflow, with our built-in DSL-agnostic intermediate representation (IR) to fecilitate stable modeling process.
General harnesses are more flexible given open-ended tasks, however they may not follow the best domain know-hows to decompose complex intents into structured BoMs (Bill of Materials), reason about their part-level IRs, propose mating relationships and estimate global placements, which are essential atomic abilities required for assembly generation~\citep{multiagentcad}.


\begin{figure}[t]
    \centering
    \includegraphics[width=0.90\linewidth]{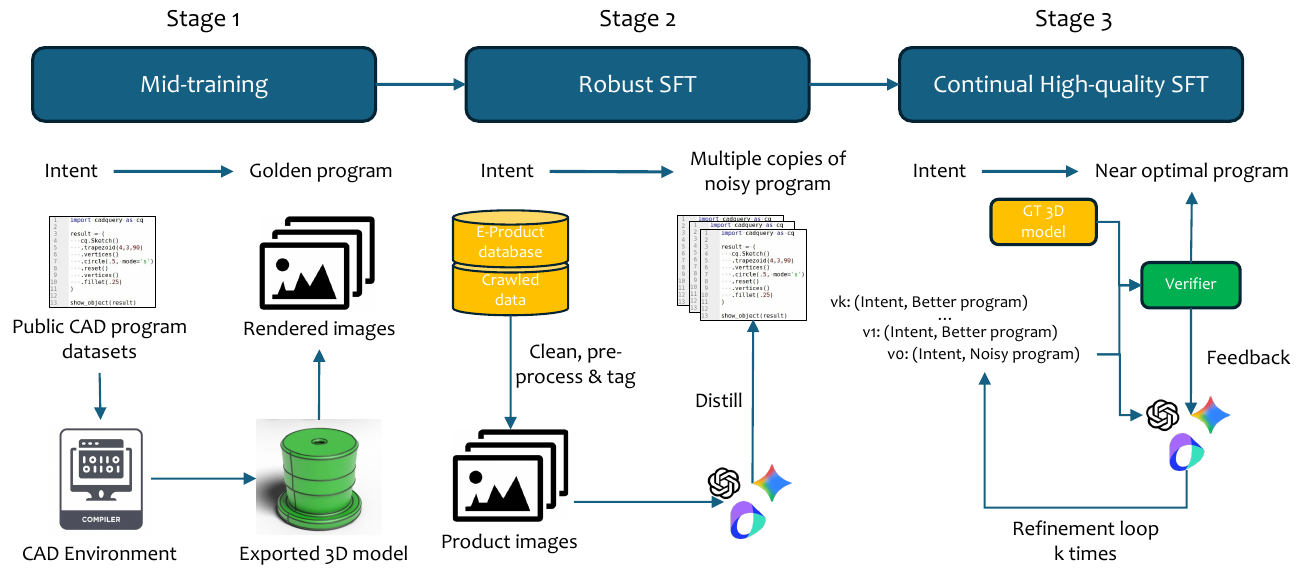}
    \caption{
        Our proposed multistage training pipeline of \viscadm.
    }
    \label{fig:m1_architecture}
\end{figure}

\section{\viscad: Data curation, model training and harness design}
\label{sec:method}

In this section, we introduce in details \textit{three} core components of our \viscad suite: 1) the data curation pipeline, 2) the model training pipeline, and 3) the domain-specific harness design respectively.

\subsection{Data curation pipeline}
\label{sec:data_curation_pipeline}

As mentioned in the overview, our data curation pipeline regards and transforms heterogeneous CAD relevant data items into a unified $(i, p, s)$ representation (Figure~\ref{fig:data_curation_pipeline}).
The raw sources span three families (Appendix~\ref{sec:appendix_data}): public corpora provide executable CAD programs and as well as render images; factory-automation catalogs provide realistic product photographs at SKU scale; purchased and collected industrial assets provide STEP/STL models and engineering drawings, together covering a wide spectrum of design intents.
To select data instances that are highly relevant for mechanical CAD design, a four-step category filter retains rigid mechanical products with clear modeling, mating, or drafting value and removes software, electronics, consumables, and weak-CAD categories, yielding 21 major categories and 128 subcategories (Figure~\ref{fig:appendix_category_flow}; Table~\ref{tab:appendix_category_sample}).

The pipeline over $t$ has two pathways, the forward one $i \rightarrow p \rightarrow s$ and the backward one $s \rightarrow p \rightarrow i$.
The motivation is to completes missing elements conditioned on other entries, and the goal is to obtain near optimal $p$ and diverse $i$.
If we start from $i$, which might be close to realistic scenarios (diverse), we should firstly normalize $i$ to reduce its complexity via off-the-shelf image edit models, and then reconstruct its shape $s'$, which might not be perfect, via off-the-shelf image-to-3D models.
Then, we can take RSI-based processes for searching the near optimal program $p'$ given $i, s'$, where we have a fast one and a slow one that handles trade-off between quantity and quality.
If we start from $s$ or $p$, it is important to get diverse $i$ through viewpoint-projection of $s$ and image editing.
And then we can run the forward process to get near optimal $p$ for supervised training.


\subsection{Multi-stage training of \viscadm}
\label{sec:viscad_m1}

\viscadm is a 27B multimodal model that maps natural-language descriptions, engineering drawings, rendered views, or product photographs to executable \texttt{FreeCAD} Python for single-part generation.
The program must compile, reconstruct the intended solid, and preserve identity-bearing features such as holes, bosses, and mounting faces etc..
Because data instance triples might be intrinsically incomplete, training cannot assume that every input intent is paired with a native program.
We therefore organize learning into three stages that progressively move from syntactic grounding and massive atomic geometric coverage to domain coverage and geometric quality (Figure~\ref{fig:m1_architecture}).

In mid-training, nearly 1M public available CAD programs in \texttt{CadQuery} are executed in its environment to export solids and projected from different viewpoints as multi-view images as input intents.
This reversible construction provides golden intent--program pairs and teaches CAD syntax, API usage, and the correspondence between visual elements of rendered image and code snippets.
In robust SFT, roughly 150k product images from e-commerce and crawled sources are cleaned, preprocessed, and tagged.
The teacher models distill multiple candidate programs for the same intent, exposing the model to broad industrial coverage while reducing sensitivity to any single noisy trajectory.
Finally, continual high-quality SFT uses roughly 10k curated examples in a generate--verify--feedback loop.
The stages are cumulative: mid-training supplies executable priors, robust SFT bridges the visual domain gap, and continual SFT restores fine-grained geometric fidelity without discarding coverage.


\begin{figure}[t]
    \centering
    \includegraphics[width=\linewidth]{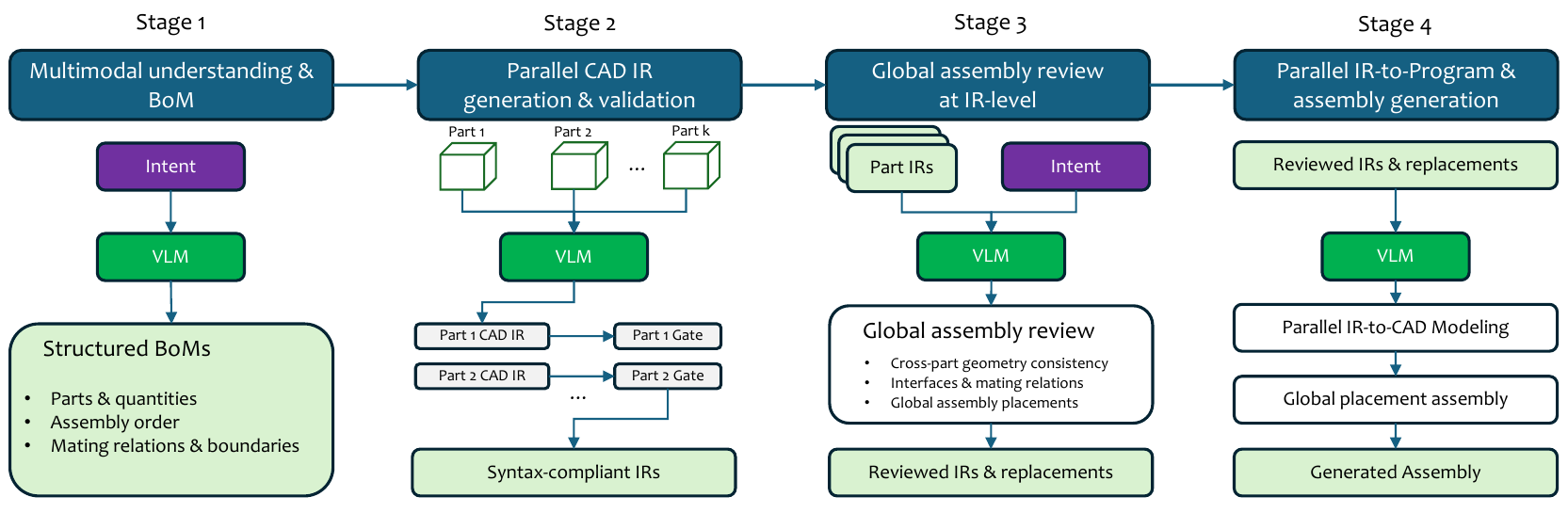}
    \caption{\viscadh. A CAD-native outer loop extracts structured BoMs, generates and gates per-part IRs, reviews global placement, and realizes an editable assembly on a CAD backend.}
    \label{fig:h1_architecture}
\end{figure}

\subsection{The design of \viscadh}
\label{sec:viscad_h1}

Assembly CAD is not a larger part-generation problem. API errors cascade across parts; a missing mating part is not a missing fillet.
Ordinary code-assistant loops execute and rewrite, but they do not expose BoMs, a per-part plan, or a global placement review~\citep{multiagentcad}.
\viscadh is a domain-specific harness around a frontier vision-language model.

The workflow has four gated stages (Figure~\ref{fig:h1_architecture}).
The \emph{Multimodal understanding \& BoM} stage turns product images, multi-view drawings, and text into parts and counts, a suggested assembly order, and mating or boundary hints; later stages consume this structure rather than re-parsing the raw evidence.
The \emph{Parallel part IR} stage assigns each part an intermediate representation---local geometry, interfaces, and constraints independent of FreeCAD, Fusion, or SolidWorks syntax---and gates syntax and local validity before anything is treated as a solid.
The \emph{Global IR review} stage reads all part IRs with the reference views, checks cross-part geometry, interfaces, and placements, and emits a reviewed IR with poses.
Local programs can all execute while the assembly is wrong---a crank without a pin, five pistons reduced to a hub.
The \emph{CAD realization} translates the reviewed IR into backend modeling calls, builds parts in parallel, and assembles the solids.
The VLM fills an IR-to-API slot rather than inventing the plan in one shot.

The IR is the portability layer (Figure~\ref{fig:ir_cases_main}).
Intent understanding and assembly reasoning stay in the IR; \texttt{FreeCAD}, \texttt{Fusion}, and \texttt{SolidWorks} consume translations of the same plan.
\viscadh is scored as an agent using assembly-level judge score instead of the part-level profile average of Eq.~\ref{eq:profile_average_main}.

\section{Experiments and main performances}
\label{sec:experiments}

\viscadm is evaluated as a standalone part-level design generator, with dedicated experiments on using itself as a reranker for test-time scaling, while \viscadh is evaluated as an assembly-level design generator (qualitative evaluations are in Sec.~\ref{sec:analy_h1})

\begin{table}[!t]
\centering
\footnotesize
\caption{Part-level profile average on \pcb and \rcb. Slice-level Success Rate, IoU, and Judge columns are in Tables~\ref{tab:appendix_pcb_main_results} and~\ref{tab:appendix_rcb_main_results}.}
\label{tab:rankings_main}
\setlength{\tabcolsep}{8pt}
\begin{tabular}{@{}lccc@{}}
\toprule
Model & AVG & \pcb & \rcb \\
\midrule
\texttt{Claude-Opus-4.8}~\citep{claudeopus48} & 0.5430 & 0.5579 & 0.5280 \\
\texttt{Gemini-3.1-Pro}~\citep{gemini31pro} & 0.5496 & 0.5533 & 0.5459 \\
\texttt{GPT-5.4}~\citep{gpt5systemcard} & 0.4913 & 0.5044 & 0.4782 \\
\texttt{GPT-5.5}~\citep{gpt5systemcard} & 0.5450 & 0.5519 & 0.5380 \\
\texttt{Kimi-K3}~\citep{kimi3} & 0.4489 & 0.4833 & 0.4144 \\
\texttt{Doubao-Seed-2.0-pro}~\citep{seed20} & 0.4046 & 0.4287 & 0.3804 \\
\midrule
\texttt{Qwen3-VL-8B}~\citep{qwen3vl} & 0.2680 & 0.3060 & 0.2300 \\
\texttt{Qwen3-VL-32B}~\citep{qwen3vl} & 0.3199 & 0.3496 & 0.2902 \\
\texttt{Qwen3.8-27B}~\citep{qwen38} & 0.4759 & 0.4943 & 0.4575 \\
\midrule
\viscadm & \textbf{0.5540} & \textbf{0.5596} & \textbf{0.5485} \\
\viscadm + \texttt{parallel-tts} & \textbf{0.5797} & \textbf{0.5833} & \textbf{0.5761} \\
\bottomrule
\end{tabular}
\end{table}

\subsection{Setup}
\label{sec:exp_scope}

\paragraph{Part-level sets.}
\pcb aggregates representative subsets of five public benchmarks totaling 1,100 tasks: BenchCAD (200), CADBench (300), Orthographic Reconstruction (200), P3D-Text (200), and P3D-Image (200) (Table~\ref{tab:appendix_pcb_slices}; Figure~\ref{fig:appendix_pcb_gallery}).
\rcb has four industrial part-level slices totaling 1,745 tasks: Text (568), 2D drawing (236), Real Picture (568), and Rendered Image (373) (Table~\ref{tab:appendix_rcb_slices}; Figure~\ref{fig:appendix_rcb_gallery}).
Compared systems are
\texttt{Claude-Opus-4.8}~\citep{claudeopus48},
\texttt{Gemini-3.1-Pro}~\citep{gemini31pro},
\texttt{GPT-5.4} and
\texttt{GPT-5.5}~\citep{gpt5systemcard},
\texttt{Kimi-K3}~\citep{kimi3},
\texttt{Doubao-Seed-2.0-pro}~\citep{seed20},
\texttt{Qwen3-VL-8} and
\texttt{Qwen3-VL-32B}~\citep{qwen3vl}, and
\texttt{Qwen3.8-27B}~\citep{qwen38}.
\viscadm is reported as a single design generation model and with test-time scaling (TTS) that reuses itself as a verifier.

\paragraph{Assembly-level sets.}
The assembly comparison uses a 50-instance subset from assembly data in \pcb and \rcb, not the full assembly tracks of either family.
Compared systems are \viscadh, \codex~\citep{codexproduct}, and \cc~\citep{claudecode}.
Valid-sample counts sit next to the Judge because a harness can refuse or fail to export.
Figure~\ref{fig:h1_cases_main} shows six source-report cases; Table~\ref{tab:appendix_h1_grid} lists all fifty on the same four-column sheet.
Assembly scores include an outer loop and are not compute-matched to Table~\ref{tab:rankings_main} or to each other.

\paragraph{Design artifacts and metrics.}
The metrics follow \rcb~\citep{realcadbench}.
We use \texttt{FreeCAD} API as our target CAD DSL.
The shared runtime executes the program and exports \texttt{result.stl}.
We score the executed solid, not a particular code path.
The vector is executability (Success Rate), Solid IoU, Surface IoU, and Judge Score from a judge model (\texttt{Kimi K2.6}).
Executability uses every assigned task.
Each quality column uses artifacts available to its evaluator; missing quality values are not zero-filled.
After a deterministic PCA-signed alignment that removes pose and uniform scale, Solid IoU measures occupied-volume overlap and Surface IoU emphasizes boundaries.
The judge is GT-free, and it compares renders of the exported 3D artifact with the input intent.
The Part Judge scores identity-bearing features of the target part.
The Assembly Judge scores component geometry $Q$, assembly accuracy $F$, and system design $D$ on a $0$--$100$ scale.
The three scores are combined into a single score.
Part ranking uses the profile average
\begin{equation}
P_c=\tfrac{1}{4}\bigl(E_c+G_c^{\mathrm{solid}}+G_c^{\mathrm{surface}}+J_c/100\bigr),
\label{eq:profile_average_main}
\end{equation}
an equal-weight mean of delivery, two geometric overlaps, and a normalized Judge. Alignment and voxelization of the predicted 3D mesh, aggregation of metric values and the judge prompts are in Appendices~\ref{sec:appendix_metrics} and~\ref{sec:appendix_judge_prompts}.
Slice-level four-column vectors are in Appendix~\ref{sec:appendix_part_tables}.

\subsection{Part-level results}
\label{sec:exp_m1_main}

Table~\ref{tab:rankings_main} reports the profile average on \pcb and \rcb. \viscadm attains $0.5540$ overall, above \texttt{Gemini-3.1-Pro} at $0.5496$ and \texttt{GPT-5.5} at $0.5450$.
The lead holds on both benchmarks: $0.5596$ vs.\ $0.5533$ on \pcb and $0.5485$ vs.\ $0.5459$ on \rcb.
TTS raises the same model to $0.5797$ ($0.5833$ / $0.5761$).
Open-weight baselines remain below the frontier band: \texttt{Qwen3-VL-8B} at $0.2680$, \texttt{Qwen3-VL-32B} at $0.3199$, \texttt{Qwen3.8-27B} at $0.4759$.
The base margin is small and obtained by a domain 27B model on a mix that includes industrial photographs and drawings, not only CAD-native renders.
TTS is the larger move. Section~\ref{sec:analysis} reads the slice columns.

\subsubsection{Detailed performances on each slice}
\label{sec:analy_part}

To save space, we have kept the slice-wise scores of \pcb and \rcb in Table~\ref{tab:appendix_pcb_main_results} and Table~\ref{tab:appendix_rcb_main_results} respectively in appendix.
On \pcb, \viscadm is stronger at IoU-based scores more often than visual judge scores.
It records the best Solid IoU on BenchCAD ($0.4422$) and CADBench ($0.5484$), and the best Surface IoU on same slices ($0.1789$, $0.2765$).
Judge scores on those slices are lower than \texttt{GPT-5.5} and \texttt{Gemini} BenchCAD $73.65$ vs.\ $77.71$ / $76.69$; CADBench $81.77$ vs.\ $87.68$ / $87.00$).
Executability is already very high (almost all above $0.95$) for every frontier model on every slices as well as \viscadm.
Our model overlaps the reference solid more closely; judge scores do not always follow that geometry lead.
The same geometry lead appears on orthographic reconstruction (Solid IoU $0.4604$).
On P3D-Image, executability is higher than \texttt{Gemini} ($0.9250$ vs.\ $0.8600$) while judge score remains lower ($53.28$ vs.\ $64.58$).

On \rcb, coverage under realistic design intent is what separates the evaluated models.
On Text with much longer text descriptions than P3D-Text, executability is $0.9331$ for \viscadm, against $0.8451$ for \texttt{Gemini}, $0.9067$ for \texttt{GPT-5.5}, and $0.6356$ for \texttt{Claude-Opus-4.8}; \texttt{Kimi-K3} reaches the highest observed text Solid IoU ($0.4584$) while has the lowest executability $0.0581$, meaning that geometry on the artifacts that exist can look strong while coverage collapses.
On real pictures, \viscadm has the highest executability ($0.9877$) and Solid IoU ($0.4456$).
On rendered views it has the highest Solid and Surface IoU ($0.5858$, $0.2581$), while \texttt{GPT-5.5} keeps a higher judge score ($77.80$ vs.\ $75.56$).
2D drawings remain a place where \texttt{Gemini}'s judge score ($79.08$) exceeds \viscadm ($69.13$) by a large margin even though executability is comparable.
Industrial photographs and product text separate models by whether a solid is exported successfully; slices in \pcb compress that difference because of very high executability.

\subsection{Assembly-level results}
\label{sec:exp_h1_main}

Appendix Table~\ref{tab:harness_main} reports the results of a 50-instance study.
\viscadh scores $85.0$ via the assembly judge, against $68.0$ for \codex and $52.0$ for \cc.
Valid samples are $49$, $49$, and $50$: almost every configuration exports an assembly, and the judge still separates them by $17$ and $33$ points.

\section{Analyses}
\label{sec:analysis}

\subsection{The effectiveness of mid-training and robust SFT}
\label{sec:analy_mid_training}

\begin{table}[H]
\centering
\scriptsize
\caption{The effectiveness of mid-training across five eval slices of \pcb. Each dataset slice reports Success Rate (Succ.), Solid IoU (Sol.), Surface IoU (Sur.), and Judge Score (Judge).}
\label{tab:mid_training_effectiveness}
\setlength{\tabcolsep}{2.8pt}
\resizebox{\textwidth}{!}{%
\begin{tabular}{l*{20}{c}}
\toprule
& \multicolumn{4}{c}{BenchCAD (200)}
& \multicolumn{4}{c}{CADBench (300)}
& \multicolumn{4}{c}{Orthographic Reconstruction (200)}
& \multicolumn{4}{c}{P3D-Text (200)}
& \multicolumn{4}{c}{P3D-Image (200)} \\
\cmidrule(lr){2-5}\cmidrule(lr){6-9}\cmidrule(lr){10-13}\cmidrule(lr){14-17}\cmidrule(lr){18-21}
Model
& Succ. & Sol. & Sur. & Judge
& Succ. & Sol. & Sur. & Judge
& Succ. & Sol. & Sur. & Judge
& Succ. & Sol. & Sur. & Judge
& Succ. & Sol. & Sur. & Judge \\
\midrule
\texttt{8B init. qw3vl}
& 0.9650 & 0.3421 & 0.1310 & 57.6067
& 0.9200 & 0.4507 & 0.1981 & 74.5601
& 0.9550 & 0.2810 & 0.1024 & 57.6431
& 0.9200 & 0.1886 & 0.0623 & 48.1332
& 0.8600 & 0.2110 & 0.0896 & 35.7332 \\
\texttt{8B init. mid\_trn\_ckpt}
& 0.9550 & 0.3675 & 0.1467 & 62.7553
& 0.9367 & 0.4842 & 0.2113 & 80.9614
& 0.9500 & 0.2862 & 0.0898 & 61.5796
& 0.9700 & 0.1996 & 0.0602 & 48.5658
& 0.8150 & 0.2179 & 0.1013 & 42.4418 \\
\midrule
\texttt{32B init. qw3vl}
& 0.9650 & 0.3481 & 0.1624 & 58.9577
& 0.9467 & 0.4841 & 0.2326 & 80.5915
& 0.9750 & 0.3295 & 0.1385 & 64.3183
& 0.9750 & 0.1968 & 0.0636 & 52.6929
& 0.8400 & 0.2278 & 0.1034 & 44.2780 \\
\texttt{32B init. mid\_trn\_ckpt}
& 0.9600 & 0.3574 & 0.1532 & 63.4973
& 0.9433 & 0.4888 & 0.2206 & 82.3735
& 0.9700 & 0.3502 & 0.1413 & 67.9330
& 0.9650 & 0.2170 & 0.0648 & 53.6341
& 0.8950 & 0.2311 & 0.1132 & 48.2573 \\
\bottomrule
\end{tabular}%
}
\end{table}

\begin{figure*}[!t]
\centering
\includegraphics[width=\textwidth]{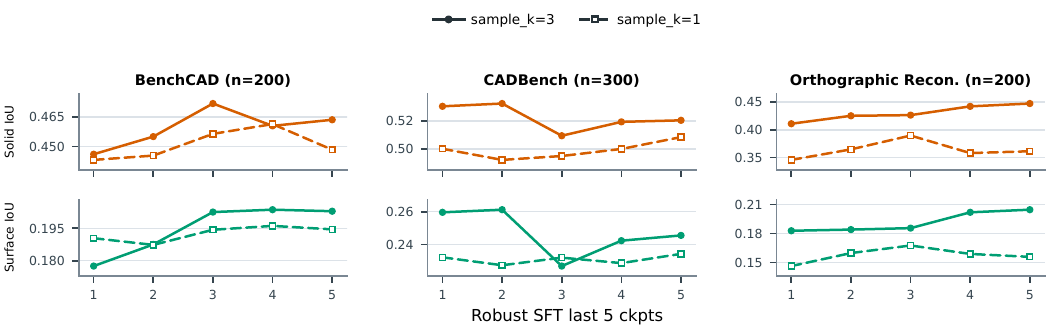}
\caption{Solid and surface IoU learning curves across the last five robust SFT ckpts for \texttt{sample\_k=3} and \texttt{sample\_k=1}. Each subplot uses an independent y-axis range to highlight ckpt-level differences.}
\label{fig:robust_sft_ckpt_comparison}
\end{figure*}

This section demonstrates the \textit{effectiveness} of mid-training and robust SFT in the proposed multi-stage training pipeline.
For mid-training, recall that it uses millions of programs to reversely curate multi-view images as input intent from the multiple viewpoints of the exported 3D meshes, so the corresponding mapping from input to program is golden.
The advantage is to increase the \textit{upper limit} of the model's final capability after the overall training pipeline.
Table~\ref{tab:mid_training_effectiveness} proves that with mid-training, many IoU-related entries increase about 1 point, while judge score entries increase about 5 points.
For robust SFT, recall that it leverages a fast distillation pipeline to get multiple target CAD programs w.r.t. the same input for learning.
This simple data augmentation strategy might regularize supervised training to be less prone to \textit{noises} in the teacher's trajectories.
Figure~\ref{fig:robust_sft_ckpt_comparison} demonstrates the advantage of robust SFT:
\textit{three} copies per input is better than \textit{one} towards the end of training.

\subsection{The effectiveness of test-time scaling}
\label{sec:analy_tts}

Test-time scaling (TTS) is a natural extension of the trained model under the vibe of recursive self-improvements (RSI).
We have tried \textit{two} direct and simple ideas under TTS and RSI.
The first is to use \viscadm as a sequential refiner that iteratively refine the initial program after seeing the multi-view images from the exported mesh as well as the input intent.
The second is to use \viscadm as a parallel reranker that chooses among $k$ rollouts from \viscadm for the same input.
We find that \texttt{sequential-tts} does not lead to improvements, however, \texttt{parallel-tts} works very well, which can improve further when scaling the rollout parameter $k$ from $8$ to $16$.
This indicates that the model itself cannot generatively leads to better program, but can discriminatively select what it thinks are the best prediction among several candidates.
Table~\ref{tab:test_time_scaling_effectiveness} shows the improvement from \viscadm through \texttt{parallel-tts} on \pcb.
The reranker simply takes multiple rollouts of \viscadm (program and an image of the exported 3D model of the program with 6-view), then uses a listwise reranking strategy to score each rollouts and pick the highest scored one.
With $k=8$, the listwise reranker w/o thinking raises 1.3 points, while raises 1.7 points w/ thinking; with $k=16$, the score can further reach $0.5833$, more than 2.4 points above the naive \viscadm.

\begin{table*}[!t]
\centering
\scriptsize
\caption{The effectiveness of test-time scaling across five evaluation slices of \pcb. AVG denotes the profile average; each dataset slice reports Success Rate (Succ.), Solid IoU (Sol.), Surface IoU (Sur.), and Judge Score (Judge).}
\label{tab:test_time_scaling_effectiveness}
\setlength{\tabcolsep}{2.8pt}
\resizebox{\textwidth}{!}{%
\begin{tabular}{l*{21}{c}}
\toprule
&
& \multicolumn{4}{c}{BenchCAD (200)}
& \multicolumn{4}{c}{CADBench (300)}
& \multicolumn{4}{c}{Orthographic Reconstruction (200)}
& \multicolumn{4}{c}{P3D-Text (200)}
& \multicolumn{4}{c}{P3D-Image (200)} \\
\cmidrule(lr){3-6}\cmidrule(lr){7-10}\cmidrule(lr){11-14}\cmidrule(lr){15-18}\cmidrule(lr){19-22}
Model & AVG
& Succ. & Sol. & Sur. & Judge
& Succ. & Sol. & Sur. & Judge
& Succ. & Sol. & Sur. & Judge
& Succ. & Sol. & Sur. & Judge
& Succ. & Sol. & Sur. & Judge \\
\midrule
\viscadm
& 0.5596
& 0.9900 & 0.4422 & 0.1789 & 73.6503
& 0.9667 & 0.5484 & 0.2765 & 81.7749
& 0.9700 & 0.4604 & 0.2100 & 75.2626
& 0.9800 & 0.2330 & 0.0779 & 55.6098
& 0.9250 & 0.2678 & 0.1364 & 53.2831 \\
\texttt{Listwise no-think @ 8}
& 0.5725
& 1.0000 & 0.4464 & 0.1823 & 76.3918
& 0.9967 & 0.5304 & 0.2625 & 86.8655
& 1.0000 & 0.4628 & 0.2130 & 81.6922
& 1.0000 & 0.2229 & 0.0729 & 57.5865
& 1.0000 & 0.2992 & 0.1486 & 58.8119 \\
\texttt{Listwise low-think @ 8}
& 0.5764
& 1.0000 & 0.4571 & 0.1987 & 76.1894
& 1.0000 & 0.5501 & 0.2674 & 86.7932
& 1.0000 & 0.4766 & 0.2323 & 81.6218
& 1.0000 & 0.2281 & 0.0752 & 58.5800
& 0.9950 & 0.2843 & 0.1446 & 58.8449 \\
\texttt{Listwise low-think @ 16}
& 0.5833
& 1.0000 & 0.4458 & 0.1966 & 79.0896
& 1.0000 & 0.5484 & 0.2803 & 87.8387
& 0.9950 & 0.4920 & 0.2420 & 81.1256
& 1.0000 & 0.2419 & 0.0798 & 61.0735
& 1.0000 & 0.3004 & 0.1529 & 60.0035 \\
\bottomrule
\end{tabular}%
}
\end{table*}

\begin{figure}[!t]
\centering
\small
\setlength{\tabcolsep}{2.2pt}
\begin{tabular}{@{}cccc@{}}
\toprule
Input & \viscadh & \codex & \cc \\
\midrule

\includegraphics[width=0.16\linewidth,height=0.8in,keepaspectratio,valign=c]{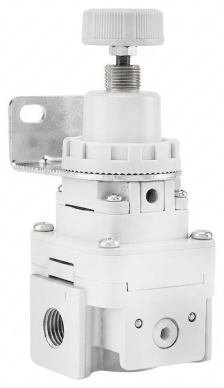} &
\includegraphics[width=0.16\linewidth,height=0.8in,keepaspectratio,valign=c]{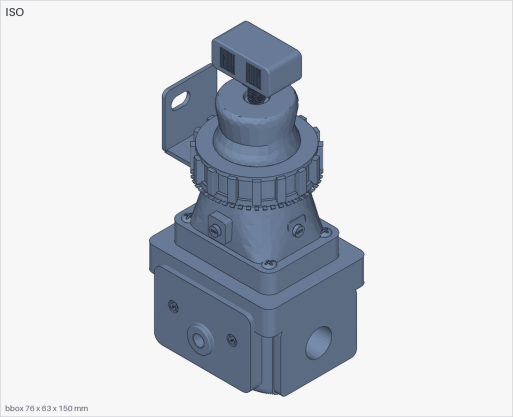} &
\includegraphics[width=0.16\linewidth,height=0.8in,keepaspectratio,valign=c]{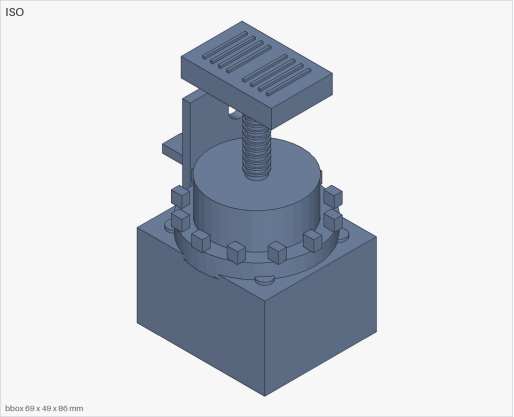} &
\includegraphics[width=0.16\linewidth,height=0.8in,keepaspectratio,valign=c]{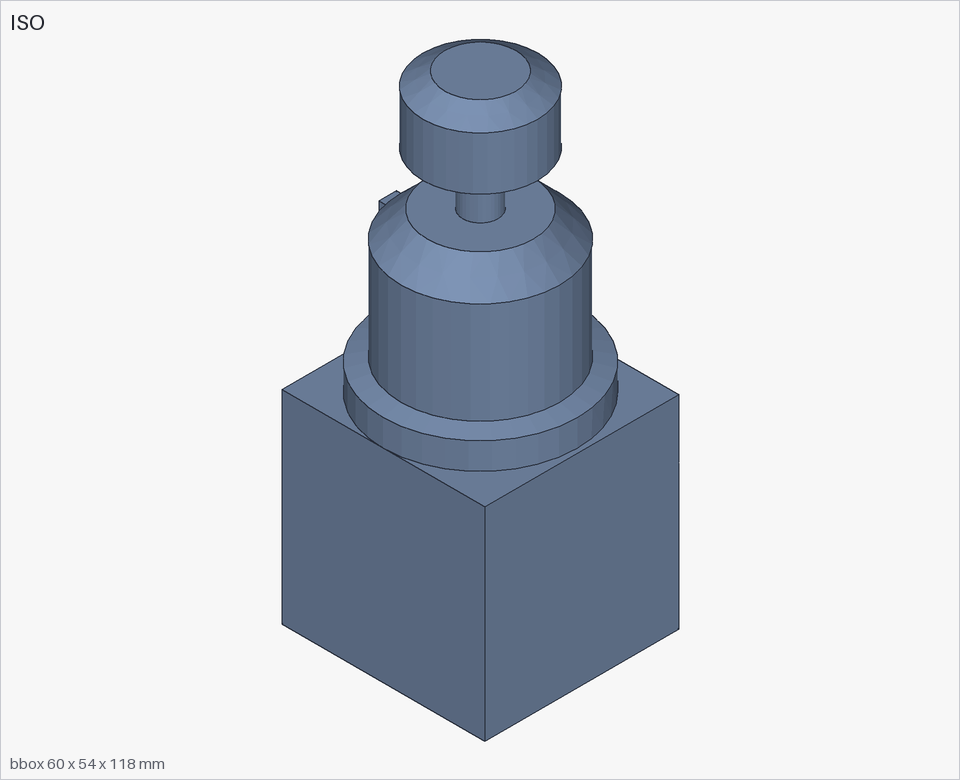} \\

\includegraphics[width=0.16\linewidth,height=0.8in,keepaspectratio,valign=c]{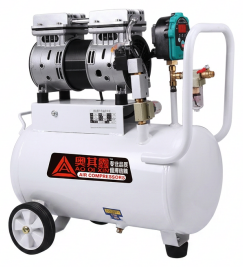} &
\includegraphics[width=0.16\linewidth,height=0.8in,keepaspectratio,valign=c]{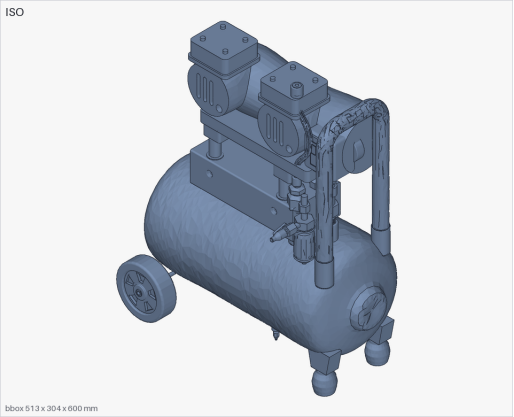} &
\includegraphics[width=0.16\linewidth,height=0.8in,keepaspectratio,valign=c]{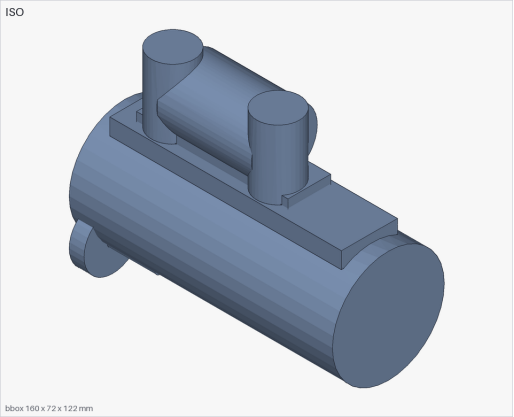} &
\includegraphics[width=0.16\linewidth,height=0.8in,keepaspectratio,valign=c]{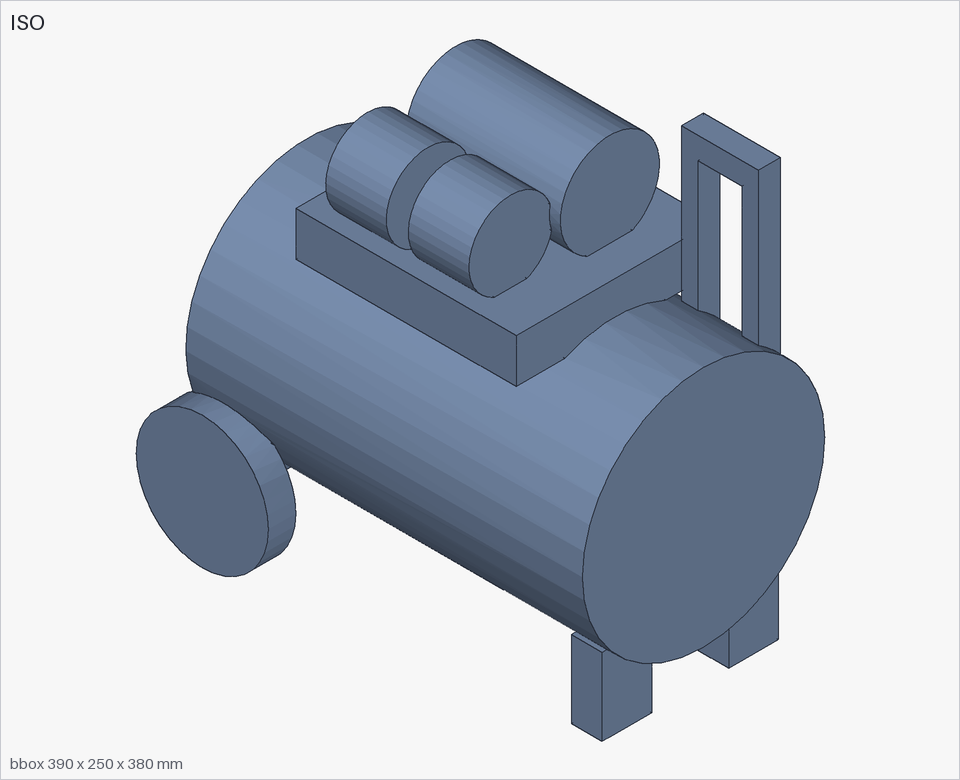} \\

\bottomrule
\end{tabular}
\caption{
    Two representative assembly Test cases for a qualitative comparison among \viscadh, \codex and \cc.
    The results of all 50 instances are listed in Table~\ref{tab:appendix_h1_grid} of appendix.
}
\label{fig:h1_cases_main}
\vspace{0.35em}
\end{figure}

\subsection{Assembly design case study and IR stability}
\label{sec:analy_h1}

\begin{figure}[!t]
\centering
\small
\setlength{\tabcolsep}{2.4pt}
\begin{tabular}{@{}cccc@{}}
\toprule
Input & \viscadh FreeCAD & \viscadh Fusion & \viscadh SolidWorks \\
\midrule
\includegraphics[width=0.15\linewidth,height=1.30in,keepaspectratio,valign=c]{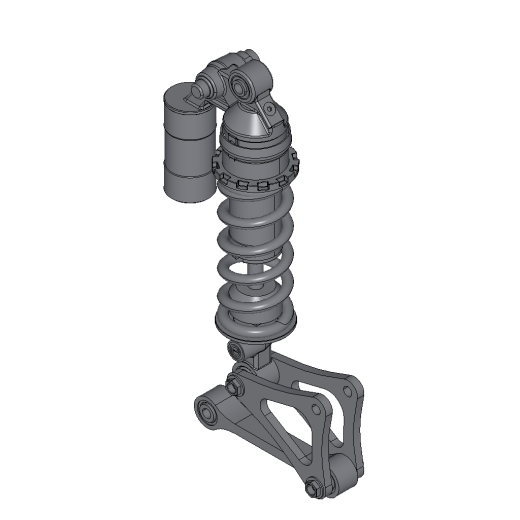} &
\includegraphics[width=0.15\linewidth,height=1.46in,keepaspectratio,valign=c]{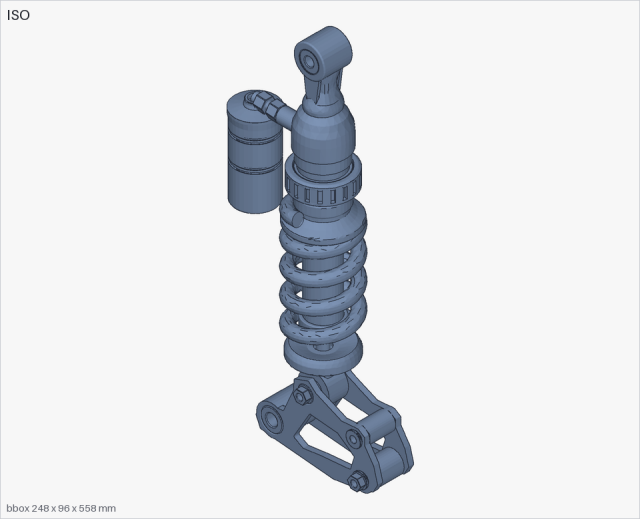} &
\includegraphics[width=0.15\linewidth,height=0.80in,keepaspectratio,valign=c]{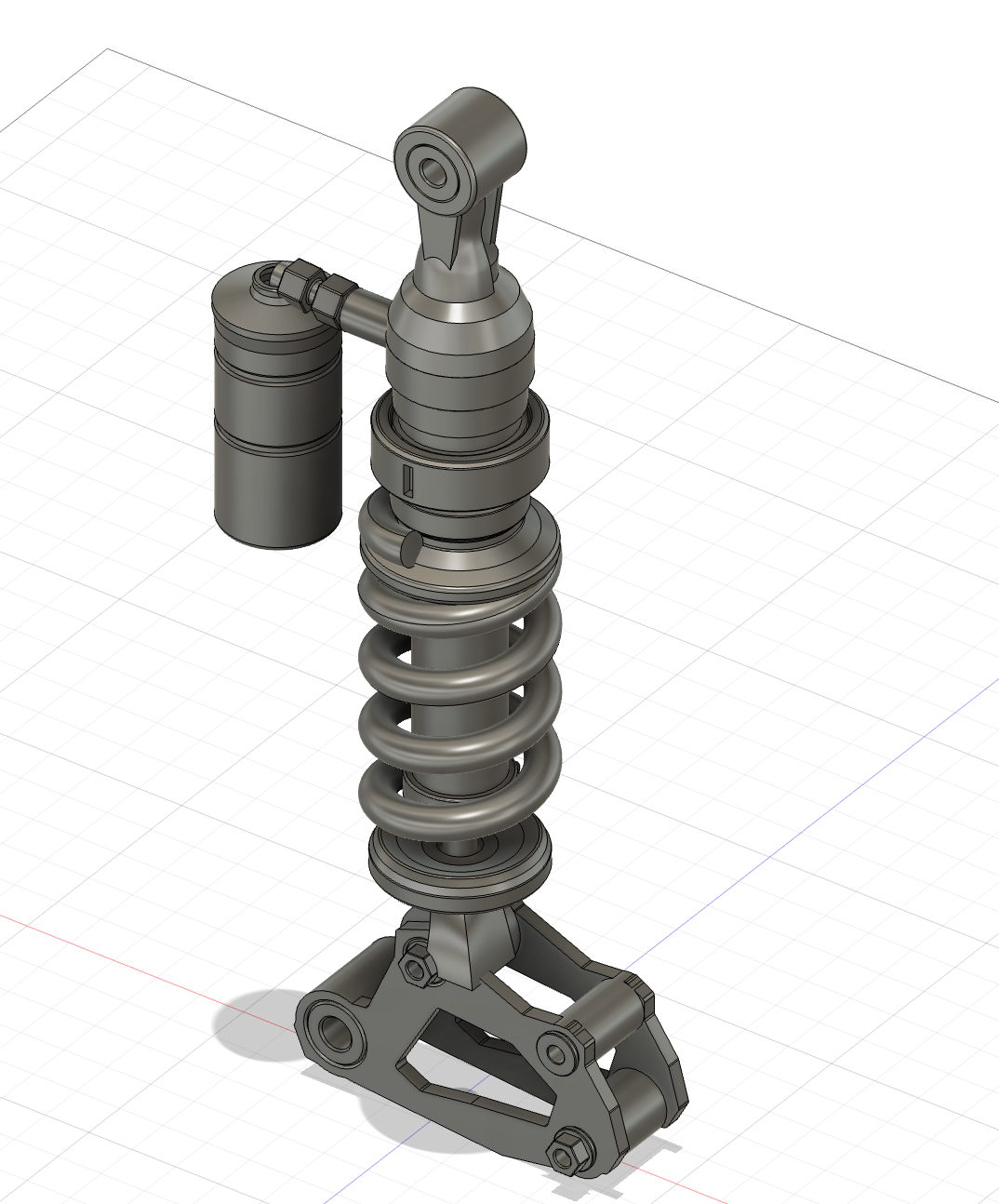} &
\includegraphics[width=0.15\linewidth,height=1.46in,keepaspectratio,valign=c]{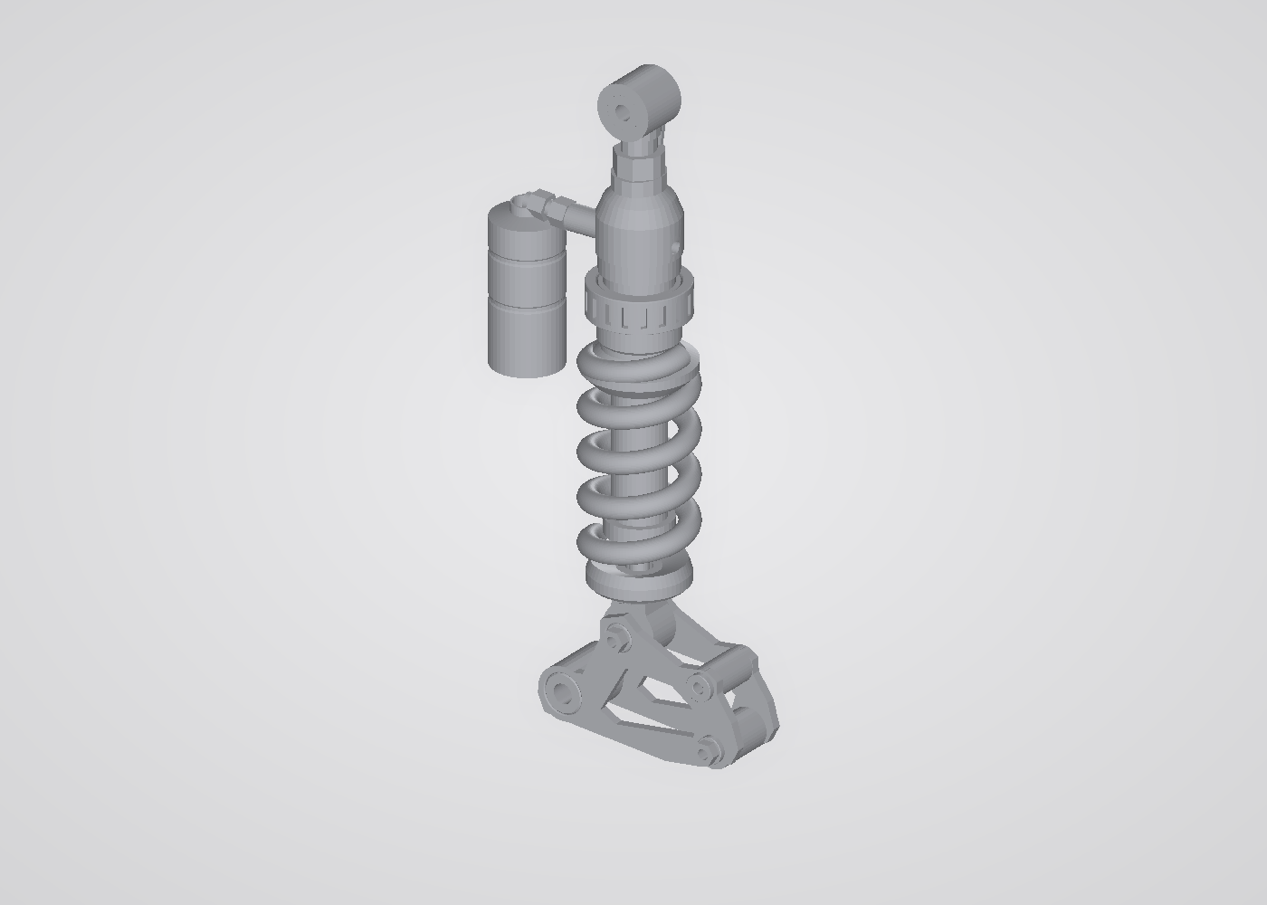} \\
\midrule
& \codex FreeCAD & \codex Fusion & \codex SolidWorks \\
\midrule
&
\includegraphics[width=0.15\linewidth,height=1.46in,keepaspectratio,valign=c]{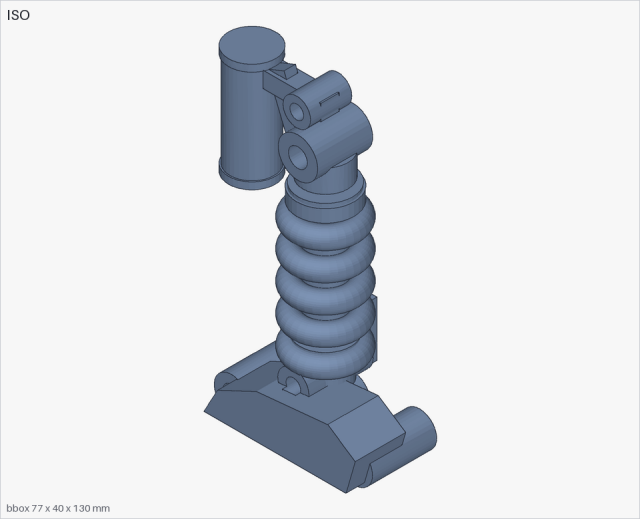} &
\includegraphics[width=0.15\linewidth,height=0.80in,keepaspectratio,valign=c]{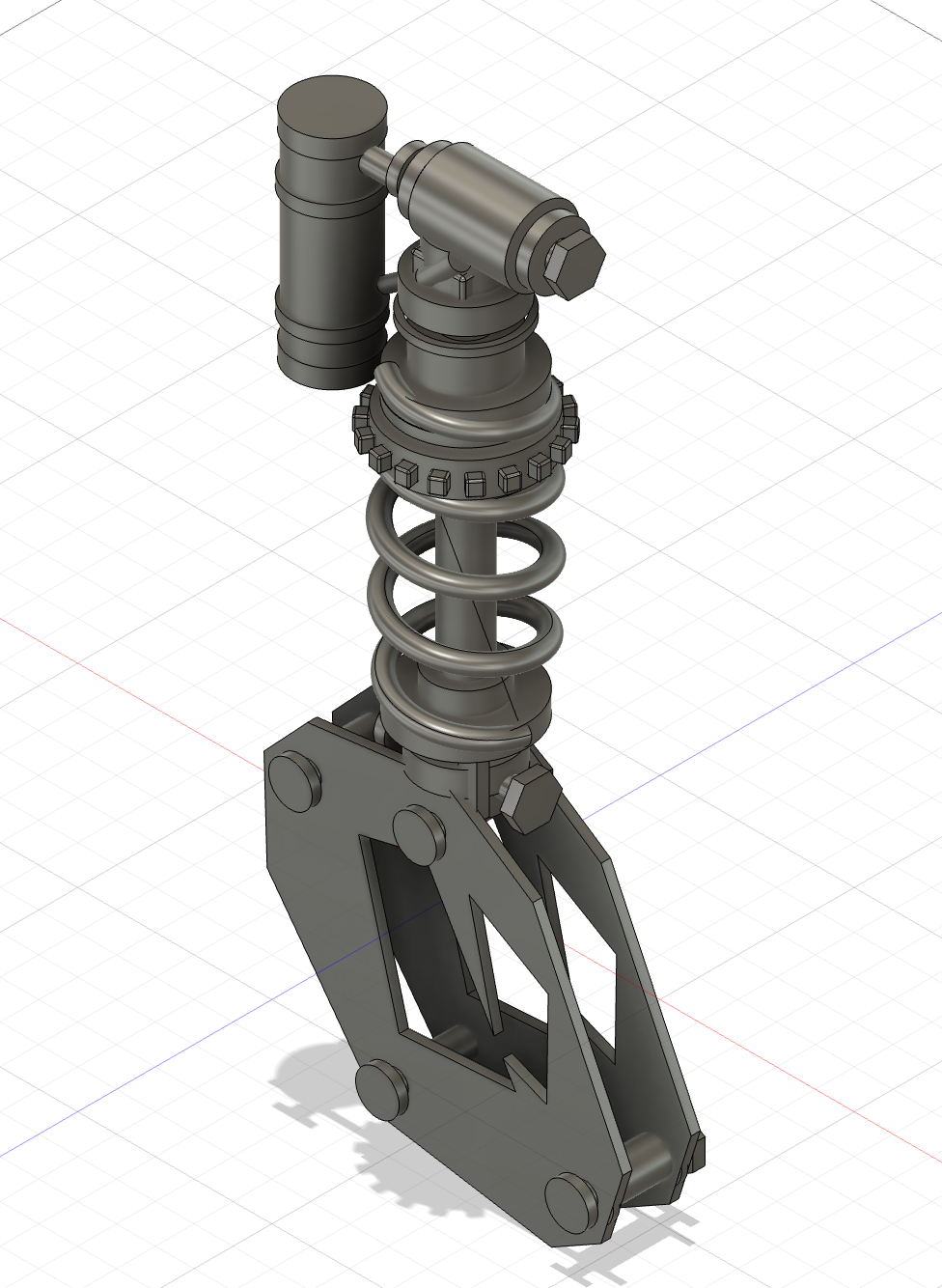} &
\includegraphics[width=0.20\linewidth,height=1.80in,keepaspectratio,valign=c]{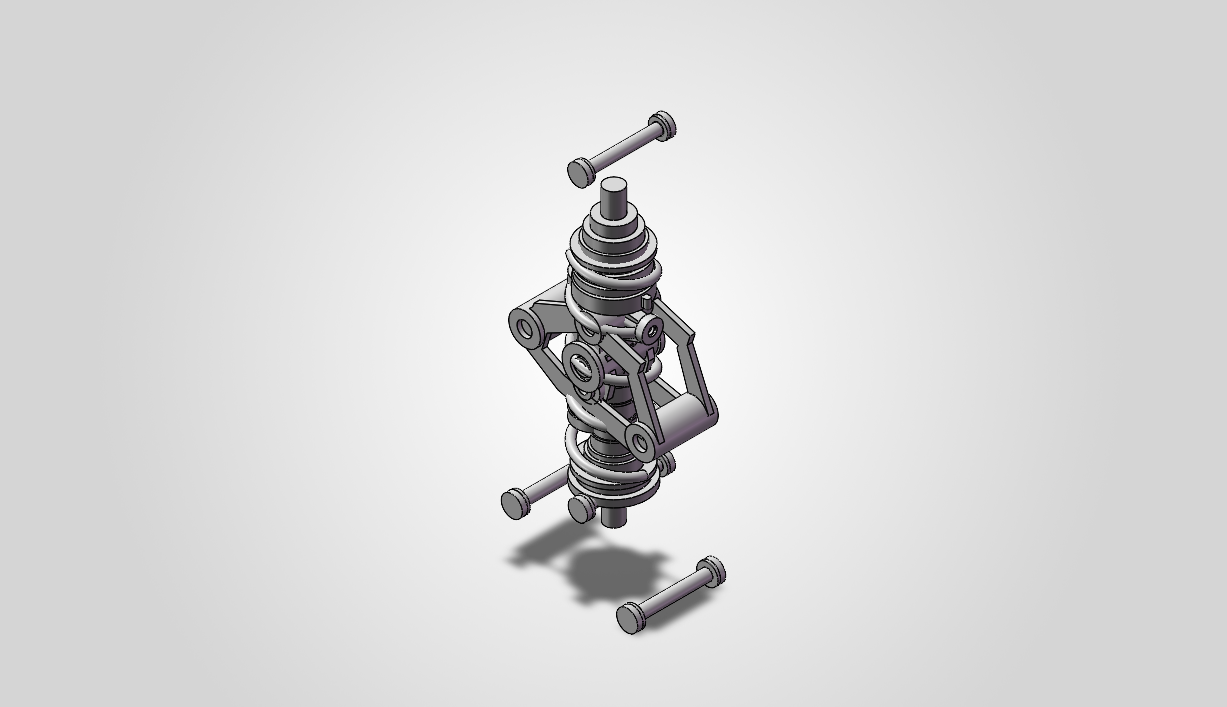} \\


\bottomrule
\end{tabular}
\caption{
    Design stability based on our proposed IR across platforms compared to \codex.
    The same assembly is realized in 3 CAD environments, namely \texttt{FreeCAD}, \texttt{Autodesk Fusion}, and \texttt{SolidWorks}.
}
\label{fig:ir_cases_main}
\end{figure}

In Figure~\ref{fig:h1_cases_main}, we can induce that \codex often recovers a coarse envelope---a stadium base without the linkage, a radial hub without pistons---and then stops.
\cc more often distorts topology or fails to export.
\viscadh more often keeps local parts after the envelope is in place: crank pins and guides, piston count, cooling fins, a CMM gantry, regulator ports, a compressor tank and pumps.
The BoM and IR stages force the system to name parts and gate them before the backend is asked for solids; a generic coding loop can emit one plausible program and stop.
Figure~\ref{fig:ir_cases_main} demonstrates the stability of the final generated assembly design across CAD platforms.
It shows that \viscadh, which produces the platform-agnostic IR first and then translate the IR to program for specific DSL, can produce more stable final design compared to \codex.

\section{Conclusion}
\label{sec:conclusion}

We present \viscad, an industrial CAD foundation model suite spanning general part generation with \viscadm and structured assembly modeling with \viscadh.
\viscadm can map a broad range of \textit{texts}, \textit{engineering drawings}, \textit{product photographs}, and \textit{renders} through one visual foundation model interface to executable CAD programs.
On \pcb and \rcb, it achieves the best part-level profile average among the evaluated systems ($0.5540$ versus $0.5496$ for the strongest state-of-the-art frontier models), with further gains from parallel test-time scaling.

This suite also contributes \textit{general} and \textit{reusable} data and training infrastructure.
Its data curation pipeline converts heterogeneous, incomplete evidence into \textit{intent--program--shape} triples and synthetically fills missing intents, programs or geometry.
Its multi-stage training recipe maximizes the utility of different data sources with varied quantity and quality through large-scale narrow-domain mid-training, general-domain robust supervised fine-tuning, and continual high-quality fine-tuning.
These pipelines are portable across data sources, intent modalities, model backbones, and CAD environments rather than tied to one benchmark or DSL.

At the assembly level, \viscadh demonstrates that to what extent a domain-specific agent can elevate frontier-model CAD capability, scoring $85.0$ versus $68.0$ and $52.0$ for general-purpose harnesses (\codex, \cc), under our assembly judge.
We formulate assembly generation as \emph{intent--IR--program}: an intent is grounded in a \textit{backend-agnostic} IR of parts, geometry, mates, and placements before translation into executable code.
This IR can act as a \emph{world language} that connects otherwise isolated CAD dialects, as demonstrated across \texttt{FreeCAD}, \texttt{Fusion}, and \texttt{SolidWorks}.

We are on the road to fully integrate \viscadm and \viscadh into one holistic agent-native CAD intelligence, unifying part design generation, assembly reasoning, execution, verification, and feedback in one model--agent loop to bring the best user experience across abundant design intent modalities, user interactions and CAD softwares.

\clearpage

\bibliographystyle{iclr2027_conference}
\bibliography{iclr2027_conference}

\clearpage
\appendix



\section{Data curation and training details}
\label{sec:appendix_data}

This section records material used to build \viscadm that is not required to follow Sections~\ref{sec:method}--\ref{sec:analysis}.

\paragraph{Sources.}
Three families supply raw data.
Catalog imagery (\texttt{src\_1}) comes from factory-automation e-commerce SKUs, mainly product photographs.
Purchased industrial CAD (\texttt{src\_2}) supplies STEP/STL assets and drawings.
Public corpora (\texttt{src\_3}) supply CAD code and drawings from open datasets.
Catalog construction expands each product term with an LLM, retrieves related SKUs through a search API, and keeps the main product image.
Purchased CAD is rendered to single- and multi-view images.
Public code is kept when it compiles to a non-empty solid.

\begin{figure}[htbp]
\centering
\includegraphics[width=0.98\textwidth]{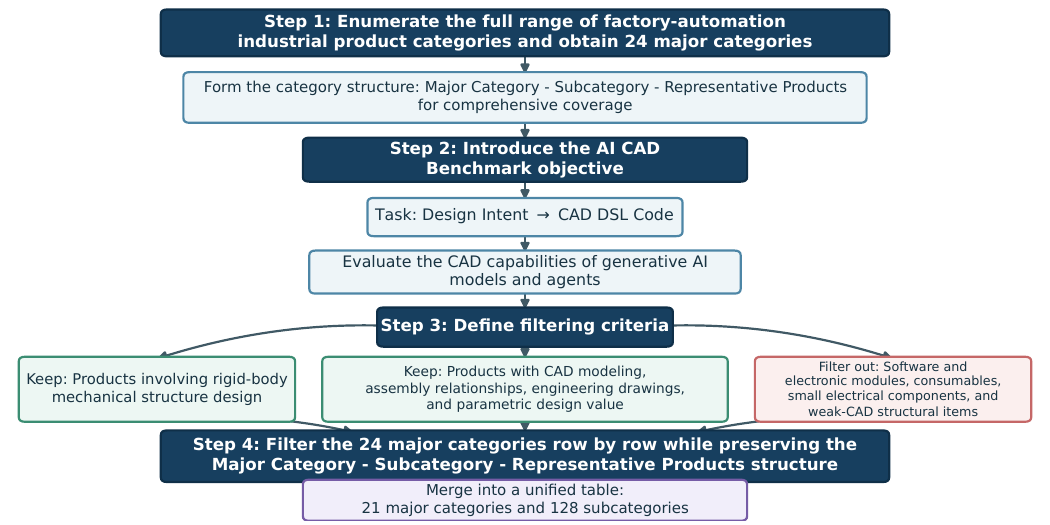}
\caption{Category-construction flow from the source report (21 major categories after filtering).}
\label{fig:appendix_category_flow}
\end{figure}

\paragraph{Category filter.}
A four-step filter enumerates factory-automation categories, introduces the design-intent$\to$CAD-code objective, and drops software, electronics, consumables, and weak-CAD items.
The retained taxonomy has 21 major categories and 128 subcategories.
Table~\ref{tab:appendix_category_sample} shows a fragment; Figure~\ref{fig:appendix_category_flow} shows the construction flow from the source report.

\begin{table}[htbp]
\centering
\small
\caption{Sample of the factory-automation category filter used to collect \viscadm data.}
\label{tab:appendix_category_sample}
\setlength{\tabcolsep}{5pt}
\begin{tabular}{p{3.2cm}p{3.4cm}p{6.2cm}}
\toprule
Major category & Subcategory & Typical products \\
\midrule
Electric Actuators & Electric Cylinders & Rod-type electric cylinders, slide-table cylinders \\
Robots and End Effectors & Industrial Robots & Six-axis robots, SCARA robots, Delta robots \\
Linear Motion & Linear Modules & Ball-screw modules, belt-driven modules \\
\bottomrule
\end{tabular}
\end{table}

\section{Evaluation protocol}
\label{sec:appendix_metrics}

\viscad uses the same metric vector and judge prompts as \rcb~\citep{realcadbench}.
This section restates the definitions needed to read Tables~\ref{tab:rankings_main}--\ref{tab:appendix_rcb_main_results}; the complete prompts are reproduced in Appendix~\ref{sec:appendix_judge_prompts}.

\paragraph{Artifact contract.}
A system emits \texttt{FreeCAD} Python. The shared runtime executes the program and exports \texttt{result.stl}.
We score the executed solid, not a particular code path.
Syntax errors, runtime failures, missing files, and empty geometry fail executability.

\paragraph{Pairwise geometry.}
Let \(\tilde C_i\) be the aligned prediction for task \(i\) and \(C_i^\star\) its reference STL.
After a deterministic PCA-signed alignment that removes pose and uniform scale, Solid IoU and Surface IoU share one transform and a voxel grid of resolution \(R=96\) with \(2\%\) padding:
\begin{equation}
\begin{aligned}
g_i^{\mathrm{solid}} &=
\frac{|V_{\mathrm{filled}}(\tilde C_i)\cap V_{\mathrm{filled}}(C_i^\star)|}
{|V_{\mathrm{filled}}(\tilde C_i)\cup V_{\mathrm{filled}}(C_i^\star)|},\\
g_i^{\mathrm{surface}} &=
\frac{|V_{\mathrm{boundary}}(\tilde C_i)\cap V_{\mathrm{boundary}}(C_i^\star)|}
{|V_{\mathrm{boundary}}(\tilde C_i)\cup V_{\mathrm{boundary}}(C_i^\star)|}.
\end{aligned}
\label{eq:geometry_metrics}
\end{equation}
Filled occupancy measures volume; boundary occupancy is more sensitive to thin structures.
Because alignment includes uniform scaling, drawing results do not measure millimetre-accurate dimension recovery.

\paragraph{Aggregates.}
Let \(c\) be a reported condition, \(\mathcal{A}_c\) its assigned tasks, \(\mathcal{G}_c^q\) the tasks with a geometry value from evaluator \(q\in\{\mathrm{solid},\mathrm{surface}\}\), and \(\mathcal{J}_c\) the tasks with a Judge result. Executability is \(e_i=\mathbb{I}[p_i\text{ executes and yields a valid }\hat C_i]\). The reported aggregates are
\begin{equation}
\begin{aligned}
E_c &= \frac{1}{|\mathcal{A}_c|}\sum_{i\in\mathcal{A}_c} e_i,\\
G_c^q &= \frac{1}{|\mathcal{G}_c^q|}\sum_{i\in\mathcal{G}_c^q} g_i^q,\\
J_c &= \frac{1}{|\mathcal{J}_c|}\sum_{i\in\mathcal{J}_c} j_i,\qquad j_i\in[0,100].
\end{aligned}
\label{eq:aggregate_metrics}
\end{equation}
Executability uses every assigned task; each quality column uses evaluator-available artifacts. Missing quality values are not zero-filled. The part-level AVG in Table~\ref{tab:rankings_main} is the profile average
\begin{equation}
P_c=\tfrac{1}{4}\bigl(E_c+G_c^{\mathrm{solid}}+G_c^{\mathrm{surface}}+J_c/100\bigr).
\label{eq:profile_average}
\end{equation}
Judge columns stay on the \(0\)--\(100\) scale and are divided by 100 only in Equation~\ref{eq:profile_average}. The assembly table reports \(J_c\) from the Assembly Judge directly.

\paragraph{Judge.}
The judge model is \texttt{Kimi-K2.6}, with prompts ratified by regime-owning CAD practitioners in \rcb before scoring.
The part judge scores only the target part (identity and salient features).
The assembly judge scores component geometry \(Q\), assembly accuracy \(F\), and system design \(D\).
The judge compares renders of the delivered artifact with the original input and has no access to the reference STL. Geometry and Judge can therefore disagree; both are reported.
A batch visibility pass labels assembly components as \texttt{full}, \texttt{partial}, or \texttt{invisible}; those labels are metadata and do not change scoring weights.

\section{Part-level detailed results}
\label{sec:appendix_part_tables}

\begin{figure*}[htbp]
\centering
\includegraphics[width=0.98\textwidth]{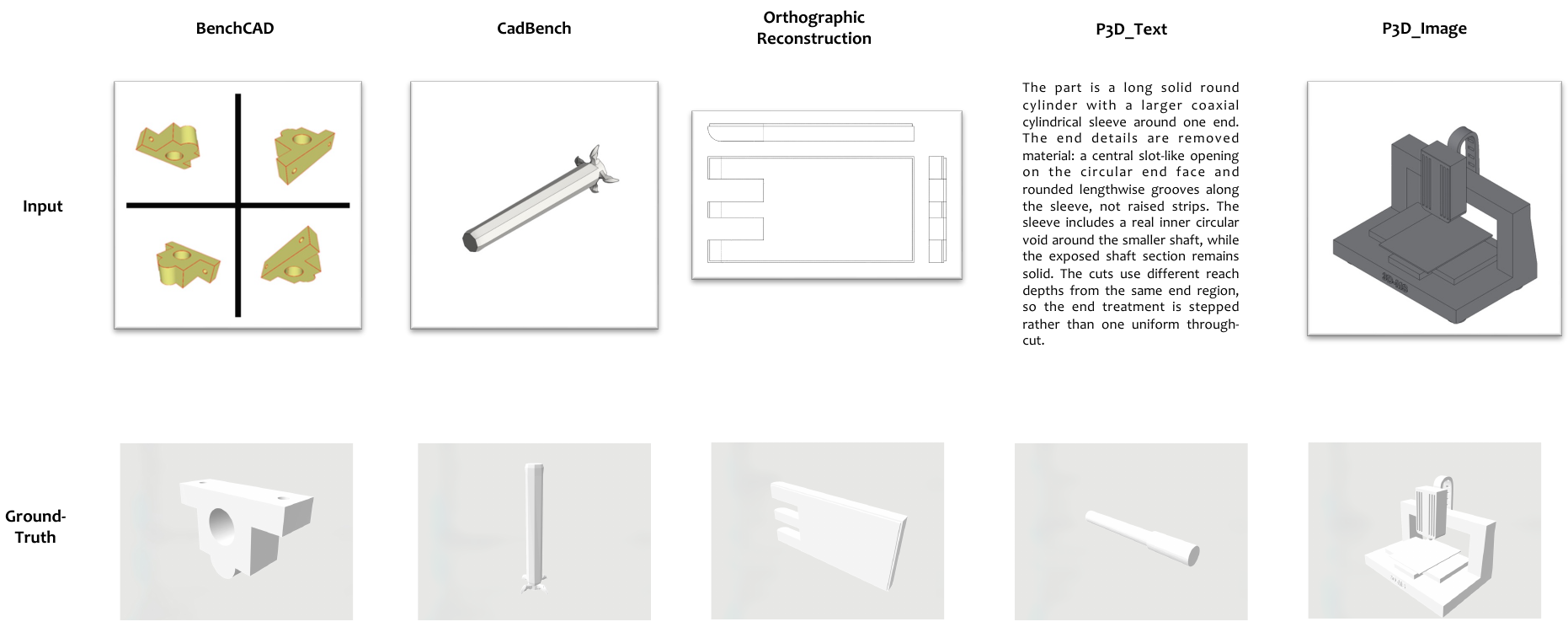}
\caption{\pcb gallery from the source report: public slices covering text, renders, textured views, multi-view, and orthographic settings.}
\label{fig:appendix_pcb_gallery}
\end{figure*}

\begin{figure*}[htbp]
\centering
\includegraphics[width=0.98\textwidth]{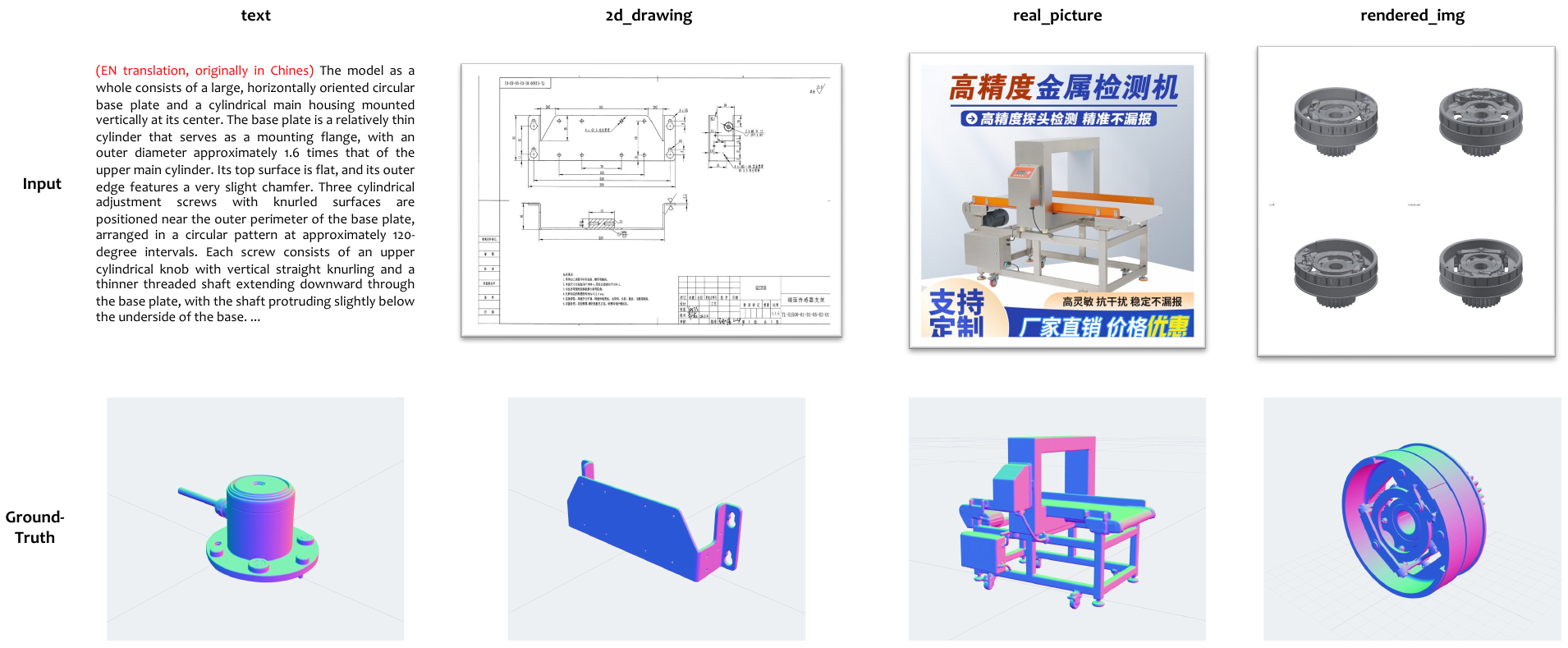}
\caption{\rcb gallery from the source report: text, drawings, real photographs, and rendered CAD views.}
\label{fig:appendix_rcb_gallery}
\end{figure*}

Table~\ref{tab:appendix_pcb_slices} and Table~\ref{tab:appendix_rcb_slices} define the slices behind Table~\ref{tab:rankings_main}. Figure~\ref{fig:appendix_pcb_gallery} and Figure~\ref{fig:appendix_rcb_gallery} reproduce the source-report galleries for those families. Table~\ref{tab:appendix_pcb_main_results} and Table~\ref{tab:appendix_rcb_main_results} report the full four-column vectors.

\begin{table}[htbp]
\centering
\scriptsize
\caption{\pcb slices (1,100 tasks).}
\label{tab:appendix_pcb_slices}
\setlength{\tabcolsep}{6pt}
\begin{tabular}{lcc}
\toprule
Slice & Emphasis & Size \\
\midrule
BenchCAD~\citep{benchcad} & public part-level CAD & 200 \\
CADBench~\citep{cadbench} & rendered-image program generation & 300 \\
Orthographic Reconstruction~\citep{orthodraw2023,orthodrawrl2025} & three-view / drawing reconstruction & 200 \\
P3D-Text~\citep{p3dbench} & text-conditioned parametric generation & 200 \\
P3D-Image~\citep{p3dbench} & image-conditioned parametric generation & 200 \\
\bottomrule
\end{tabular}
\end{table}

\begin{table}[htbp]
\centering
\scriptsize
\caption{\rcb part-level slices (1,745 tasks)~\citep{realcadbench}.}
\label{tab:appendix_rcb_slices}
\setlength{\tabcolsep}{6pt}
\begin{tabular}{lcc}
\toprule
Slice & Emphasis & Size \\
\midrule
Text & industrial text-to-CAD & 568 \\
2D Drawing & engineering-drawing-to-CAD & 236 \\
Real Picture & product-photo-to-CAD & 568 \\
Rendered Image & CAD-native rendered-view reconstruction & 373 \\
\bottomrule
\end{tabular}
\end{table}

\begin{table*}[htbp]
\centering
\scriptsize
\caption{Detailed per-slice performance on \pcb. AVG is the profile average in Equation~\ref{eq:profile_average}.}
\label{tab:appendix_pcb_main_results}
\setlength{\tabcolsep}{2.8pt}
\resizebox{\textwidth}{!}{%
\begin{tabular}{lccccccccccccccccccccc}
\toprule
& & \multicolumn{4}{c}{BenchCAD (200)} & \multicolumn{4}{c}{CADBench (300)} & \multicolumn{4}{c}{Orthographic Reconstruction (200)} & \multicolumn{4}{c}{P3D-Text (200)} & \multicolumn{4}{c}{P3D-Image (200)} \\
\cmidrule(lr){3-6}\cmidrule(lr){7-10}\cmidrule(lr){11-14}\cmidrule(lr){15-18}\cmidrule(lr){19-22}
Model & AVG & Succ. & Sol. & Sur. & Judge & Succ. & Sol. & Sur. & Judge & Succ. & Sol. & Sur. & Judge & Succ. & Sol. & Sur. & Judge & Succ. & Sol. & Sur. & Judge \\
\midrule
\texttt{Claude-Opus-4.8} & 0.5579 & 0.9800 & 0.3963 & 0.1422 & 74.6260 & 0.9733 & 0.5110 & 0.2465 & 85.5550 & 0.9850 & 0.4613 & 0.2283 & 80.3076 & 0.9950 & 0.2298 & 0.0732 & 60.4361 & 0.9500 & 0.2773 & 0.1369 & 56.2036 \\
\texttt{Gemini-3.1-Pro} & 0.5533 & 0.9700 & 0.4404 & 0.1735 & 76.6926 & 0.9833 & 0.5265 & 0.2424 & 87.0011 & 0.9700 & 0.3805 & 0.1474 & 75.9129 & 0.9850 & 0.2206 & 0.0735 & 60.9738 & 0.8600 & 0.2861 & 0.1548 & 64.5813 \\
\texttt{GPT-5.4} & 0.5044 & 0.9400 & 0.3583 & 0.1168 & 69.8584 & 0.9300 & 0.5148 & 0.2476 & 83.9089 & 0.9600 & 0.2585 & 0.0875 & 69.7562 & 0.9250 & 0.2359 & 0.0816 & 55.8663 & 0.7100 & 0.2525 & 0.1414 & 53.3569 \\
\texttt{GPT-5.5} & 0.5519 & 0.9700 & 0.3929 & 0.1600 & 77.7077 & 0.9767 & 0.4631 & 0.1737 & 87.6771 & 0.9900 & 0.4583 & 0.2126 & 78.9228 & 0.9750 & 0.2220 & 0.0815 & 57.9285 & 0.8700 & 0.2694 & 0.1403 & 66.0496 \\
\texttt{Kimi-K3} & 0.4833 & 0.8650 & 0.4051 & 0.1417 & 76.5975 & 0.8600 & 0.4688 & 0.1896 & 87.2364 & 0.8300 & 0.4164 & 0.2088 & 72.7213 & 0.5800 & 0.2440 & 0.0797 & 64.4241 & 0.3300 & 0.2790 & 0.1318 & 62.5581 \\
\texttt{Doubao-Seed-2.0-pro} & 0.4287 & 0.6900 & 0.3933 & 0.1579 & 69.2558 & 0.7633 & 0.5052 & 0.2403 & 80.9379 & 0.7200 & 0.2489 & 0.1029 & 66.3439 & 0.6000 & 0.2167 & 0.0716 & 53.3404 & 0.3600 & 0.2228 & 0.1322 & 44.9432 \\
\midrule
\texttt{Qwen3-VL-8B} & 0.3060 & 0.4950 & 0.2945 & 0.0792 & 30.1294 & 0.7033 & 0.3695 & 0.1149 & 61.4916 & 0.7900 & 0.2506 & 0.0643 & 46.6559 & 0.4350 & 0.2096 & 0.0499 & 27.2735 & 0.2250 & 0.1576 & 0.0601 & 16.5109 \\
\texttt{Qwen3-VL-32B} & 0.3496 & 0.5150 & 0.3376 & 0.1168 & 50.0717 & 0.6367 & 0.4246 & 0.1638 & 72.9576 & 0.7050 & 0.2499 & 0.0912 & 55.8546 & 0.5750 & 0.1823 & 0.0634 & 36.2067 & 0.3250 & 0.1401 & 0.0730 & 24.2472 \\
\texttt{Qwen3.8-27B} & 0.4943 & 0.8950 & 0.3453 & 0.1427 & 66.2510 & 0.9367 & 0.4527 & 0.1815 & 83.0687 & 0.9550 & 0.3413 & 0.1404 & 73.2573 & 0.8700 & 0.2245 & 0.0692 & 51.6725 & 0.7450 & 0.2244 & 0.1097 & 49.0839 \\
\midrule
\texttt{VisCAD-M1} & \textbf{0.5596} & \textbf{0.9900} & \textbf{0.4422} & \textbf{0.1789} & 73.6503 & 0.9667 & \textbf{0.5484} & \textbf{0.2765} & 81.7749 & 0.9700 & \textbf{0.4604} & 0.2100 & 75.2626 & 0.9800 & 0.2330 & 0.0779 & 55.6098 & \textbf{0.9250} & 0.2678 & 0.1364 & 53.2831 \\
\texttt{VisCAD-M1} + \texttt{parallel-tts} & \textbf{0.5833} & \textbf{1.0000} & \textbf{0.4458} & \textbf{0.1966} & \textbf{79.0896} & \textbf{1.0000} & \textbf{0.5484} & 0.2803 & \textbf{87.8387} & \textbf{0.9950} & \textbf{0.4920} & \textbf{0.2420} & \textbf{81.1256} & \textbf{1.0000} & \textbf{0.2419} & \textbf{0.0798} & \textbf{61.0735} & \textbf{1.0000} & \textbf{0.3004} & \textbf{0.1529} & \textbf{60.0035} \\
\bottomrule
\end{tabular}%
}
\end{table*}

\begin{table*}[htbp]
\centering
\scriptsize
\caption{Detailed per-slice performance on \rcb. AVG is the profile average in Equation~\ref{eq:profile_average}.}
\label{tab:appendix_rcb_main_results}
\setlength{\tabcolsep}{2.8pt}
\resizebox{\textwidth}{!}{%
\begin{tabular}{lccccccccccccccccc}
\toprule
& & \multicolumn{4}{c}{Text (568)} & \multicolumn{4}{c}{2D Drawing (236)} & \multicolumn{4}{c}{Real Picture (568)} & \multicolumn{4}{c}{Rendered Image (373)} \\
\cmidrule(lr){3-6}\cmidrule(lr){7-10}\cmidrule(lr){11-14}\cmidrule(lr){15-18}
Model & AVG & Succ. & Sol. & Sur. & Judge & Succ. & Sol. & Sur. & Judge & Succ. & Sol. & Sur. & Judge & Succ. & Sol. & Sur. & Judge \\
\midrule
\texttt{Claude-Opus-4.8} & 0.5280 & 0.6356 & 0.4041 & 0.1066 & 67.2654 & 0.9661 & 0.2811 & 0.1390 & 70.9259 & 0.9806 & 0.3945 & 0.1026 & 56.7468 & 0.9864 & 0.5521 & 0.2250 & 72.5600 \\
\texttt{Gemini-3.1-Pro} & 0.5459 & 0.8451 & 0.3995 & 0.1106 & 67.7365 & 0.8729 & 0.3378 & 0.1729 & 79.0802 & 0.9085 & 0.4410 & 0.1238 & 62.3418 & 0.9035 & 0.5379 & 0.2240 & 76.4951 \\
\texttt{GPT-5.4} & 0.4782 & 0.6984 & 0.3643 & 0.0982 & 60.8809 & 0.7839 & 0.2044 & 0.0959 & 51.3244 & 0.8539 & 0.3988 & 0.1155 & 55.6164 & 0.8660 & 0.5506 & 0.2254 & 71.5476 \\
\texttt{GPT-5.5} & 0.5380 & 0.9067 & 0.3509 & 0.0943 & 60.8809 & 0.9237 & 0.2817 & 0.1400 & 51.6176 & 0.9208 & 0.3754 & 0.1089 & 64.4221 & 0.9732 & 0.5234 & 0.2141 & 77.8014 \\
\texttt{Kimi-K3} & 0.4144 & 0.0581 & 0.4584 & 0.1685 & 69.6191 & 0.1271 & 0.3325 & 0.1632 & 69.5584 & 0.5722 & 0.4501 & 0.1218 & 62.3541 & 0.6113 & 0.5515 & 0.2091 & 79.0637 \\
\texttt{Doubao-Seed-2.0-pro} & 0.3804 & 0.2447 & 0.3451 & 0.0922 & 60.2508 & 0.3915 & 0.2672 & 0.1274 & 55.5862 & 0.5563 & 0.3826 & 0.1101 & 50.5330 & 0.5335 & 0.5118 & 0.2014 & 65.9351 \\
\midrule
\texttt{Qwen3-VL-8B} & 0.2300 & 0.0616 & 0.2575 & 0.0682 & 36.2247 & 0.1314 & 0.1653 & 0.0739 & 23.5440 & 0.2394 & 0.3311 & 0.0906 & 22.3909 & 0.4075 & 0.3794 & 0.1212 & 37.1693 \\
\texttt{Qwen3-VL-32B} & 0.2902 & 0.1391 & 0.2507 & 0.0730 & 48.1867 & 0.3220 & 0.1824 & 0.0956 & 31.0987 & 0.4736 & 0.3120 & 0.0841 & 30.5300 & 0.3941 & 0.4090 & 0.1455 & 50.4072 \\
\texttt{Qwen3.8-27B} & 0.4575 & 0.7359 & 0.3600 & 0.0921 & 57.4800 & 0.7
627 & 0.1973 & 0.1052 & 56.2100 & 0.8468 & 0.3802 & 0.0889 & 45.7700 & 0.8391 & 0.4964 & 0.1643 & 65.7200 \\
\midrule
\texttt{VisCAD-M1} & \textbf{0.5485} & \textbf{0.9331} & \textbf{0.3886} & \textbf{0.1026} & 65.6302 & 0.9068 & 0.2579 & 0.1349 & 69.1291 & \textbf{0.9877} & \textbf{0.4456} & \textbf{0.1243} & 58.1434 & 0.9652 & \textbf{0.5858} & \textbf{0.2581} & 75.5621 \\
\texttt{VisCAD-M1} + \texttt{parallel-tts} & \textbf{0.5761} & \textbf{1.0000} & \textbf{0.3945} & \textbf{0.1021} & \textbf{68.3647} & \textbf{1.0000} & \textbf{0.2685} & \textbf{0.1358} & \textbf{78.4878} & \textbf{1.0000} & \textbf{0.4481} & \textbf{0.1275} & \textbf{61.7992} & \textbf{0.9946} & \textbf{0.5987} & \textbf{0.2662} & \textbf{79.5955} \\
\bottomrule
\end{tabular}%
}
\end{table*}

\section{Assembly-level additional results}
\label{sec:appendix_h1_cases}

\begin{table}[htbp]
\centering
\footnotesize
\caption{Harness comparison on a representative subset of the assmebly slices from \pcb and \rcb (25 instances are from \pcb, while 25 from \rcb).}
\label{tab:harness_main}
\setlength{\tabcolsep}{10pt}
\begin{tabular}{@{}lcc@{}}
\toprule
Harness & Succ. Rate & Judge Score (Assm.) \\
\midrule
\codex~\citep{codexproduct} & 0.98 & 68.0 \\
\cc~\citep{claudecode} & 1.00 & 52.0 \\
\midrule
\viscadh & 0.98 & \textbf{85.0} \\
\bottomrule
\end{tabular}
\end{table}

Figure~\ref{fig:h1_cases_main} shows six instances from the 50-instance study.
Table~\ref{tab:appendix_h1_grid} extends that comparison to every instance, with the same column order: input, \viscadh, \codex, \cc.
Some evaluation-sheet cells contain stacked screenshots; we always take the topmost drawing in the sheet.
Missing exports are marked as such rather than imputed.
Numbers under system views are Assembly Judge scores on the $0$--$100$ scale; they are case-level readouts from the sheet, not a second ranking table.

\begingroup
\small
\setlength{\tabcolsep}{2.2pt}
\setlength{\LTcapwidth}{\linewidth}
\begin{longtable}{@{}>{\centering\arraybackslash}p{0.72in}>{\centering\arraybackslash}p{1.10in}>{\centering\arraybackslash}p{1.10in}>{\centering\arraybackslash}p{1.10in}>{\centering\arraybackslash}p{1.10in}@{}}
\caption{Full 50-instance assembly comparison, extending Figure~\ref{fig:h1_cases_main}. Each cell uses the topmost screenshot from the evaluation sheet. Missing exports are marked as such. Numbers under system views are Assembly Judge scores on $0$--$100$.}
\label{tab:appendix_h1_grid}\\
\toprule
{\scriptsize Case} & Input & \viscadh & Codex & Claude Code \\
\midrule
\endfirsthead
\caption[]{Full 50-instance assembly comparison (continued).}\\
\toprule
{\scriptsize Case} & Input & \viscadh & Codex & Claude Code \\
\midrule
\endhead
\midrule
\multicolumn{5}{r}{\scriptsize Continued on next page.}\\
\endfoot
\bottomrule
\endlastfoot
\texttt{\scriptsize pcb\_102410} & \makecell{\includegraphics[width=\linewidth,height=0.58in,keepaspectratio,valign=c]{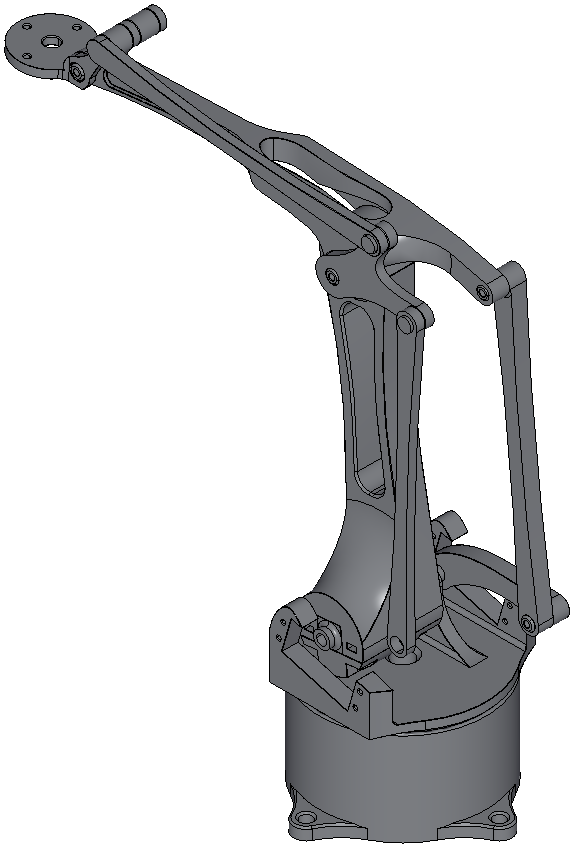}} & \makecell{\includegraphics[width=\linewidth,height=0.58in,keepaspectratio,valign=c]{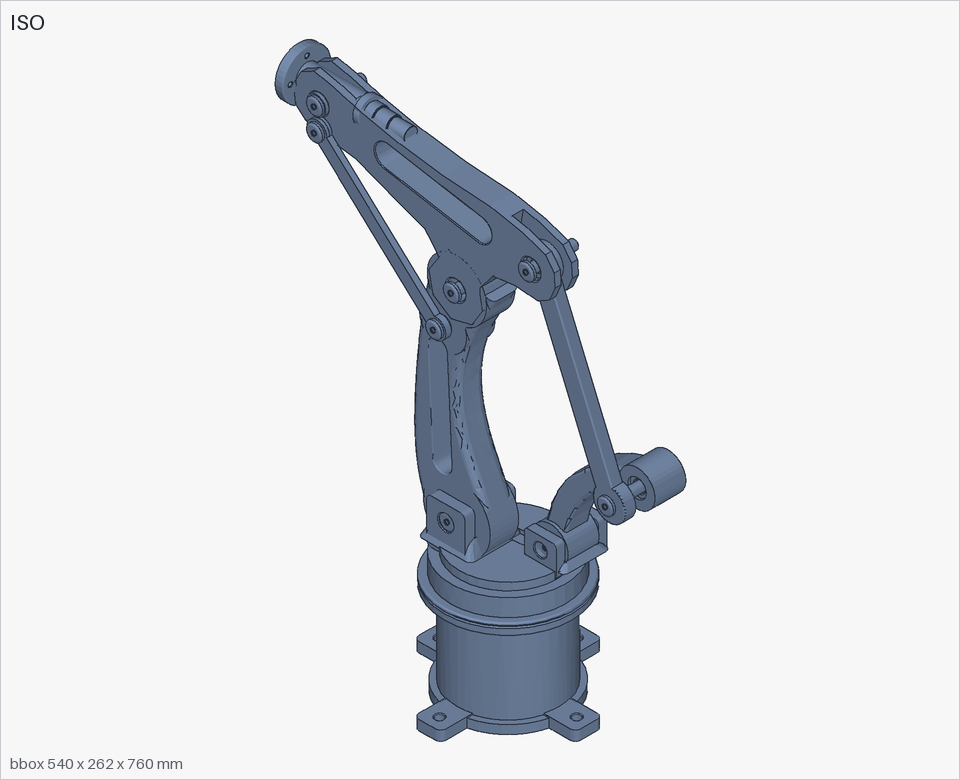}\\[-0.2em]{\scriptsize 88.9}} & \makecell{\includegraphics[width=\linewidth,height=0.58in,keepaspectratio,valign=c]{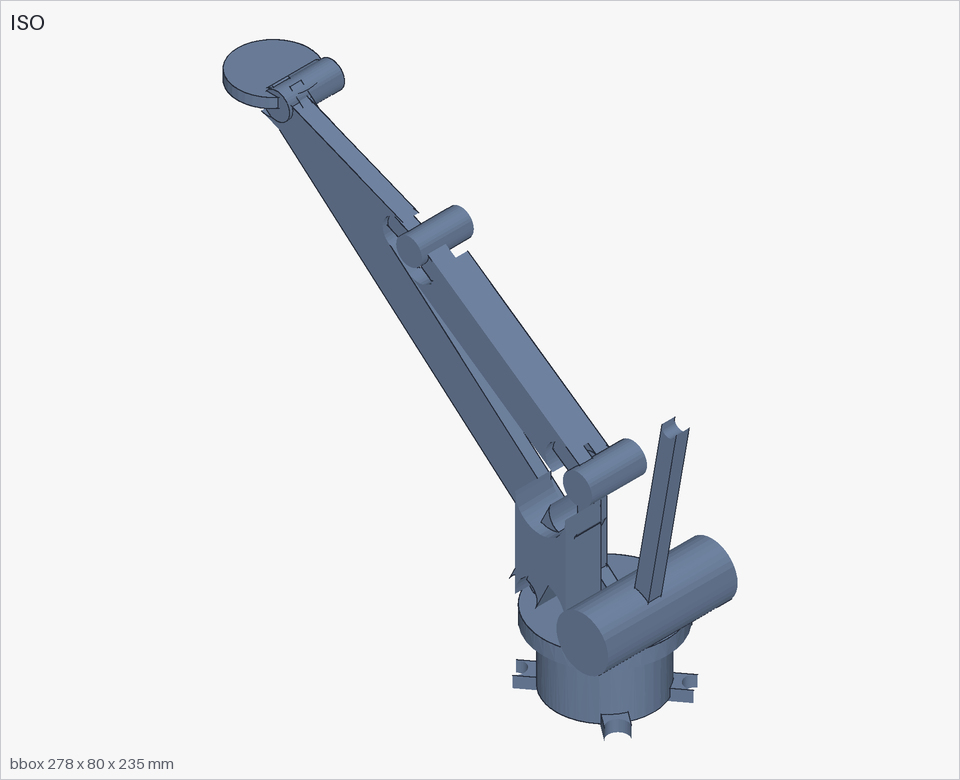}\\[-0.2em]{\scriptsize 56.4}} & \makecell{\includegraphics[width=\linewidth,height=0.58in,keepaspectratio,valign=c]{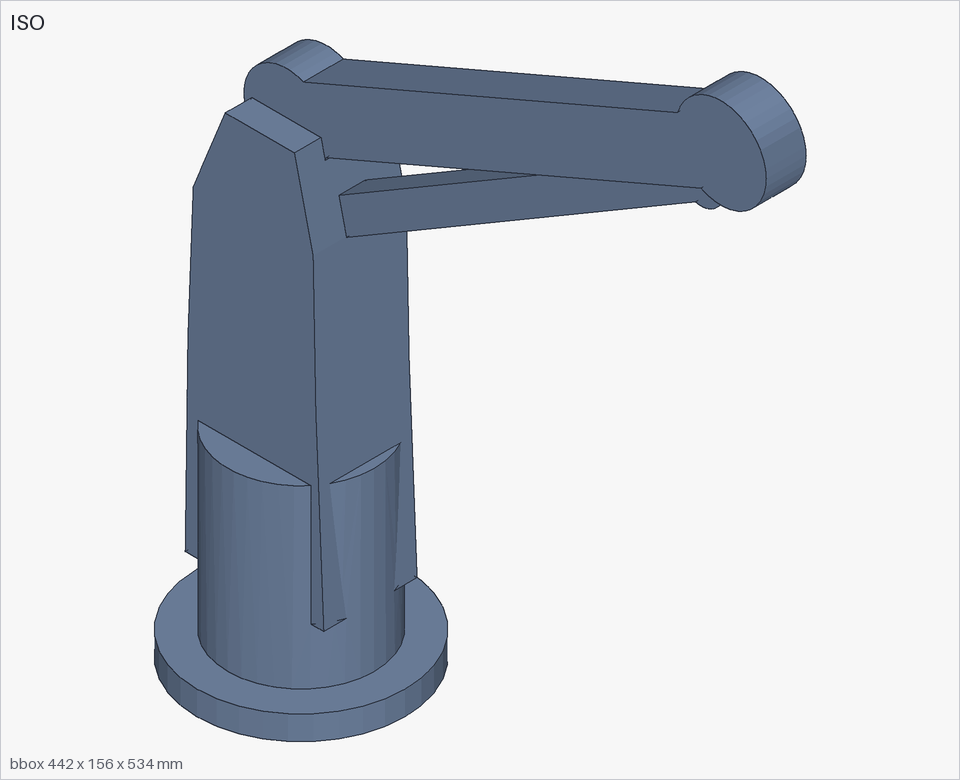}\\[-0.2em]{\scriptsize 47.9}} \\
\texttt{\scriptsize pcb\_110965} & \makecell{\includegraphics[width=\linewidth,height=0.58in,keepaspectratio,valign=c]{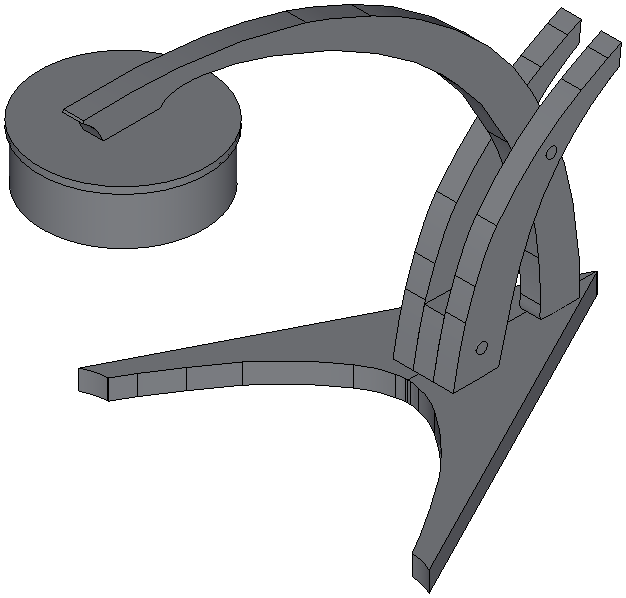}} & \makecell{\includegraphics[width=\linewidth,height=0.58in,keepaspectratio,valign=c]{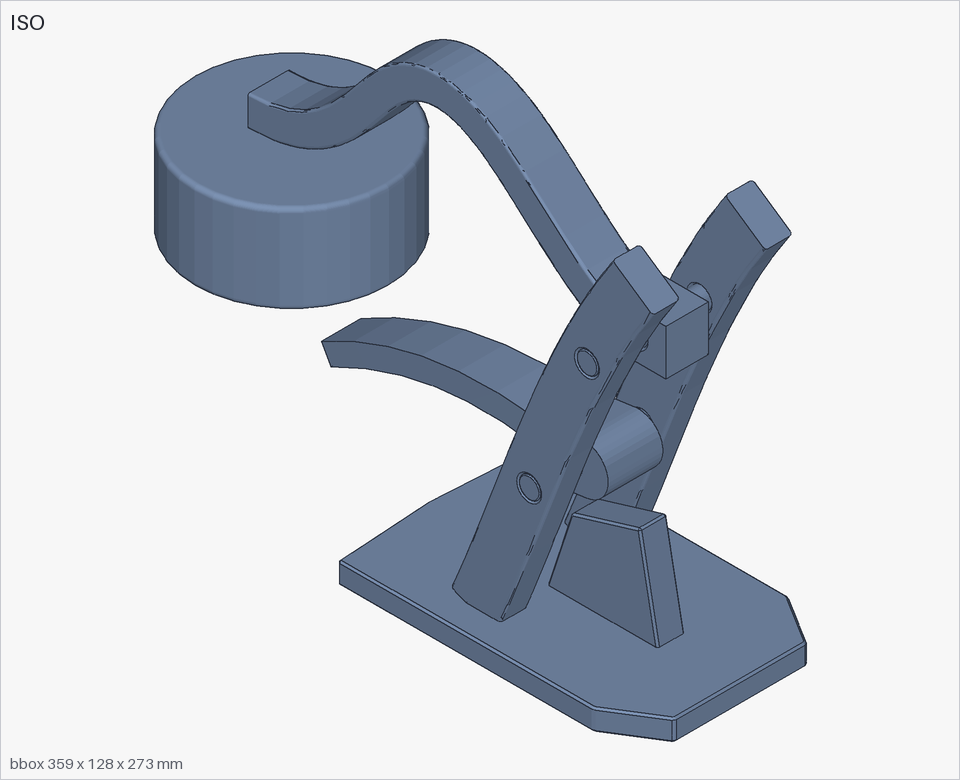}\\[-0.2em]{\scriptsize 79.0}} & \makecell{\includegraphics[width=\linewidth,height=0.58in,keepaspectratio,valign=c]{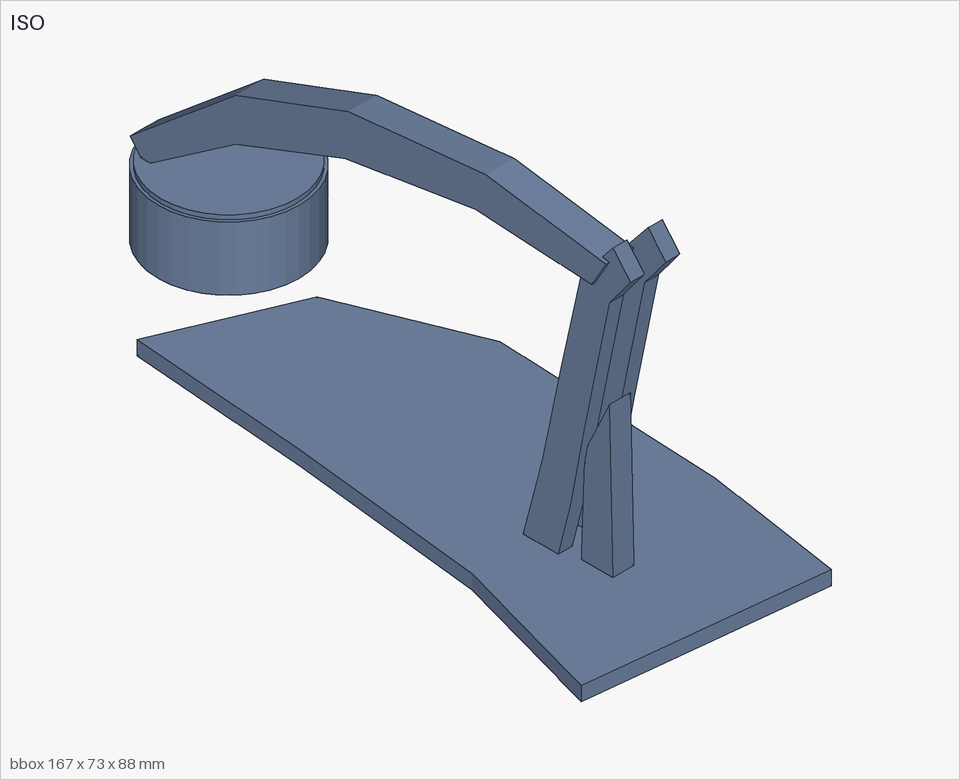}\\[-0.2em]{\scriptsize 71.7}} & \makecell{\includegraphics[width=\linewidth,height=0.58in,keepaspectratio,valign=c]{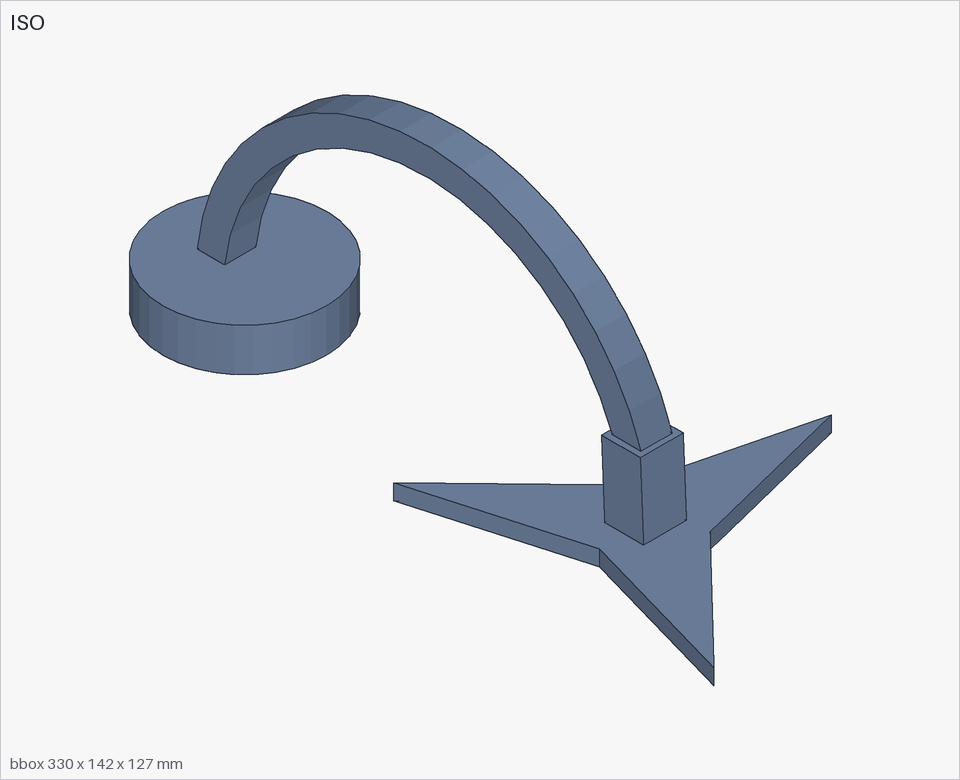}\\[-0.2em]{\scriptsize 60.9}} \\
\texttt{\scriptsize pcb\_111151} & \makecell{\includegraphics[width=\linewidth,height=0.58in,keepaspectratio,valign=c]{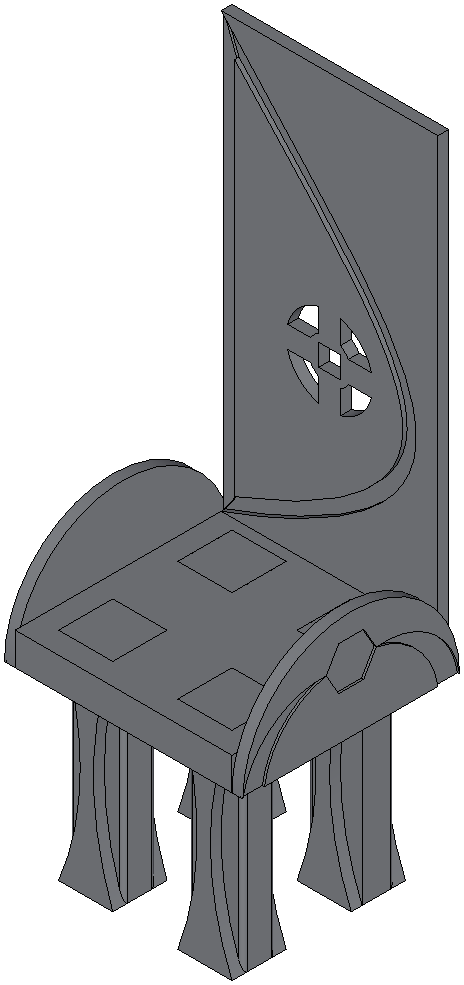}} & \makecell{\includegraphics[width=\linewidth,height=0.58in,keepaspectratio,valign=c]{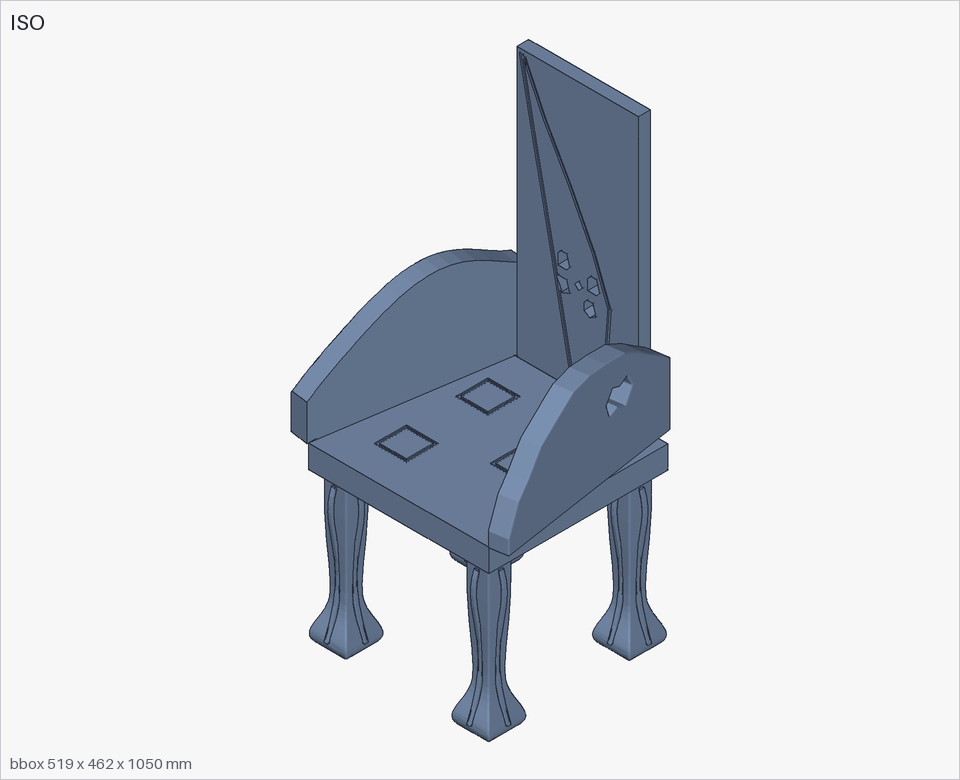}\\[-0.2em]{\scriptsize 92.5}} & \makecell{\includegraphics[width=\linewidth,height=0.58in,keepaspectratio,valign=c]{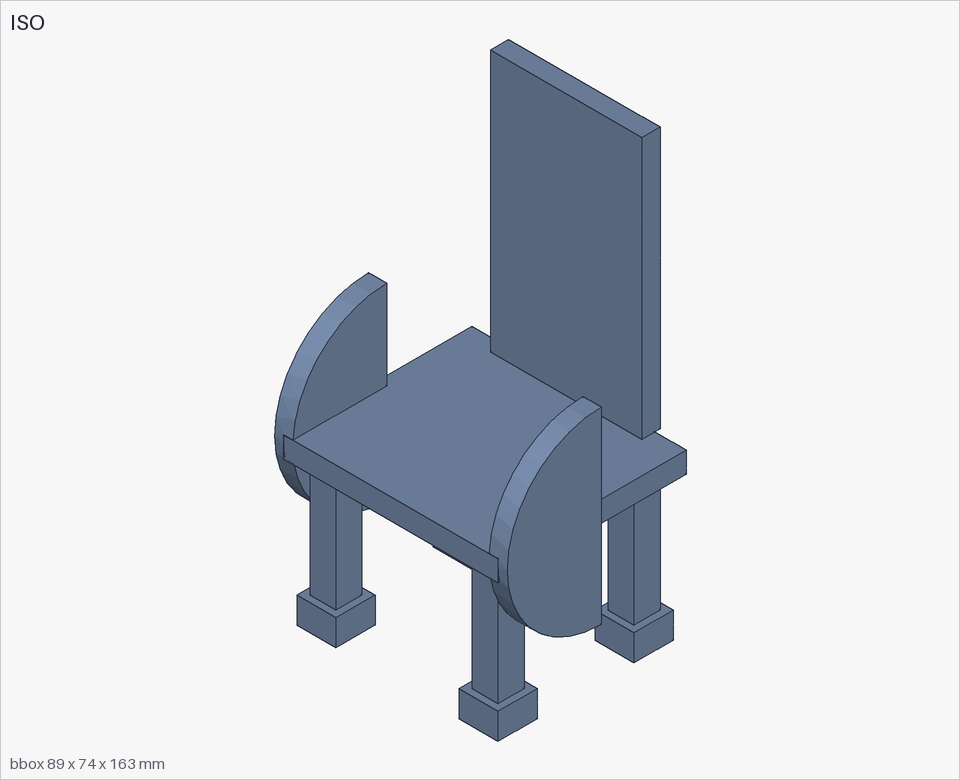}\\[-0.2em]{\scriptsize 69.8}} & \makecell{\includegraphics[width=\linewidth,height=0.58in,keepaspectratio,valign=c]{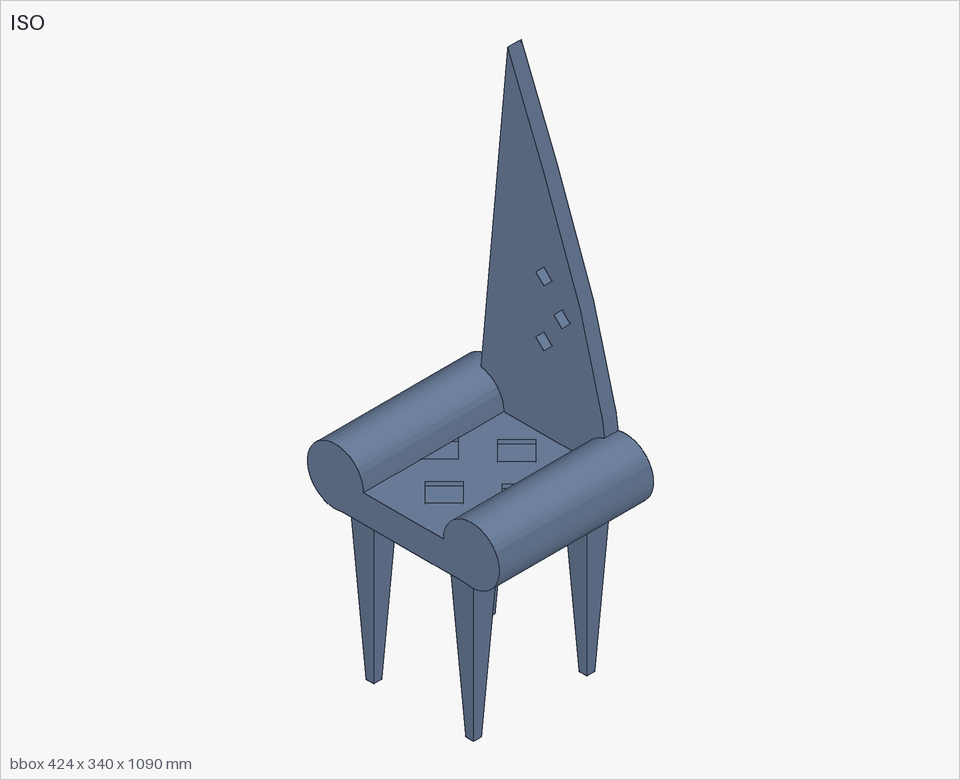}\\[-0.2em]{\scriptsize 64.9}} \\
\texttt{\scriptsize pcb\_116076} & \makecell{\includegraphics[width=\linewidth,height=0.58in,keepaspectratio,valign=c]{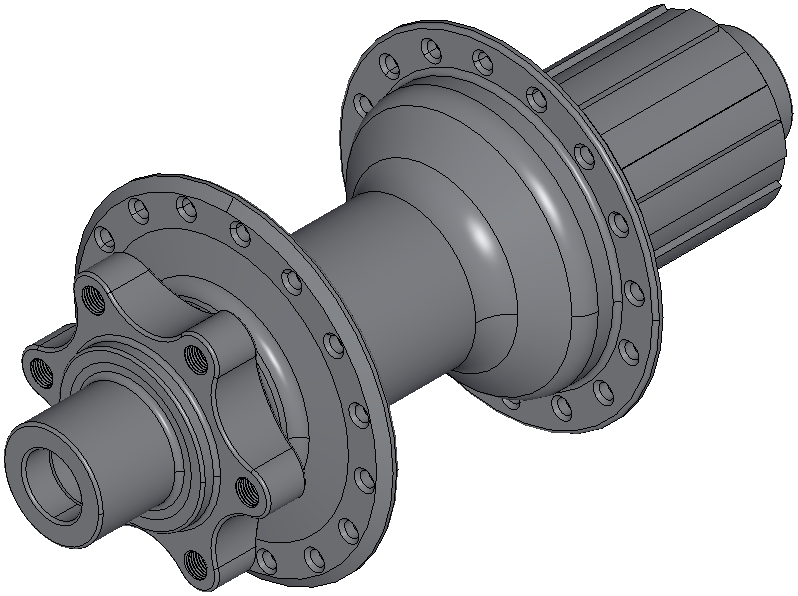}} & \makecell{\includegraphics[width=\linewidth,height=0.58in,keepaspectratio,valign=c]{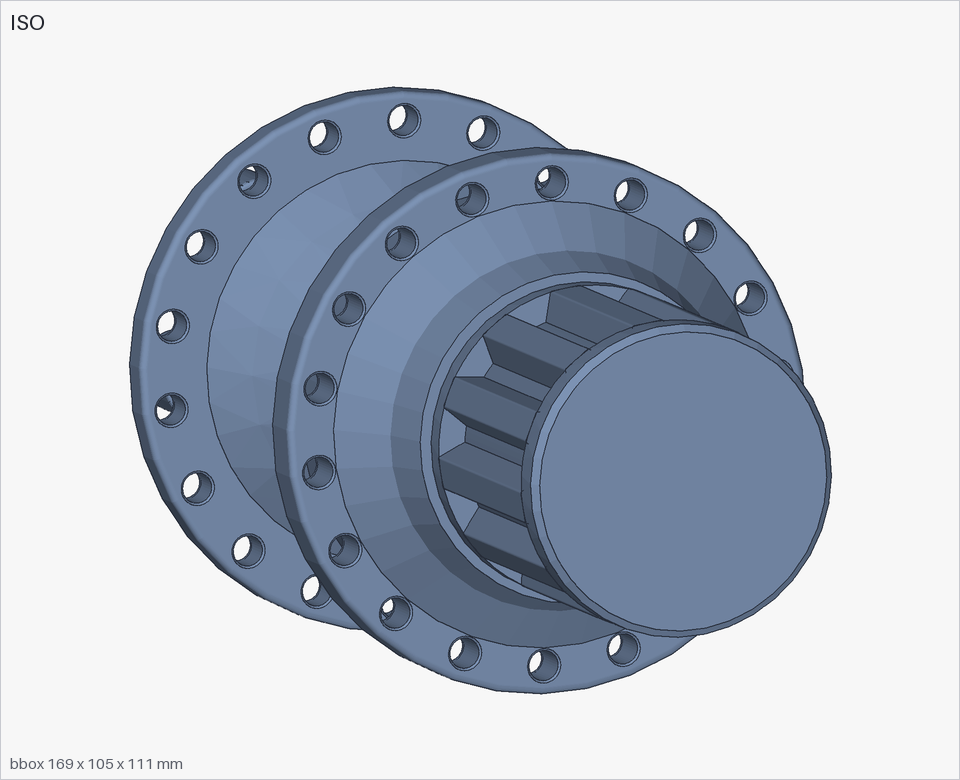}\\[-0.2em]{\scriptsize 87.0}} & \makecell{\includegraphics[width=\linewidth,height=0.58in,keepaspectratio,valign=c]{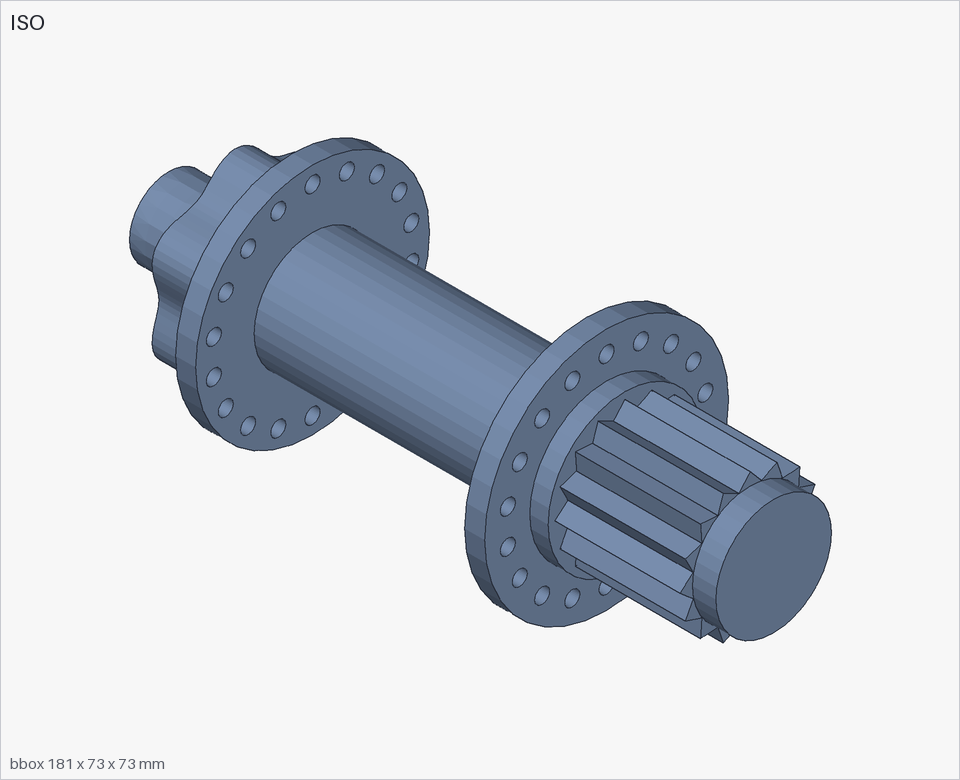}\\[-0.2em]{\scriptsize 75.8}} & \makecell{\includegraphics[width=\linewidth,height=0.58in,keepaspectratio,valign=c]{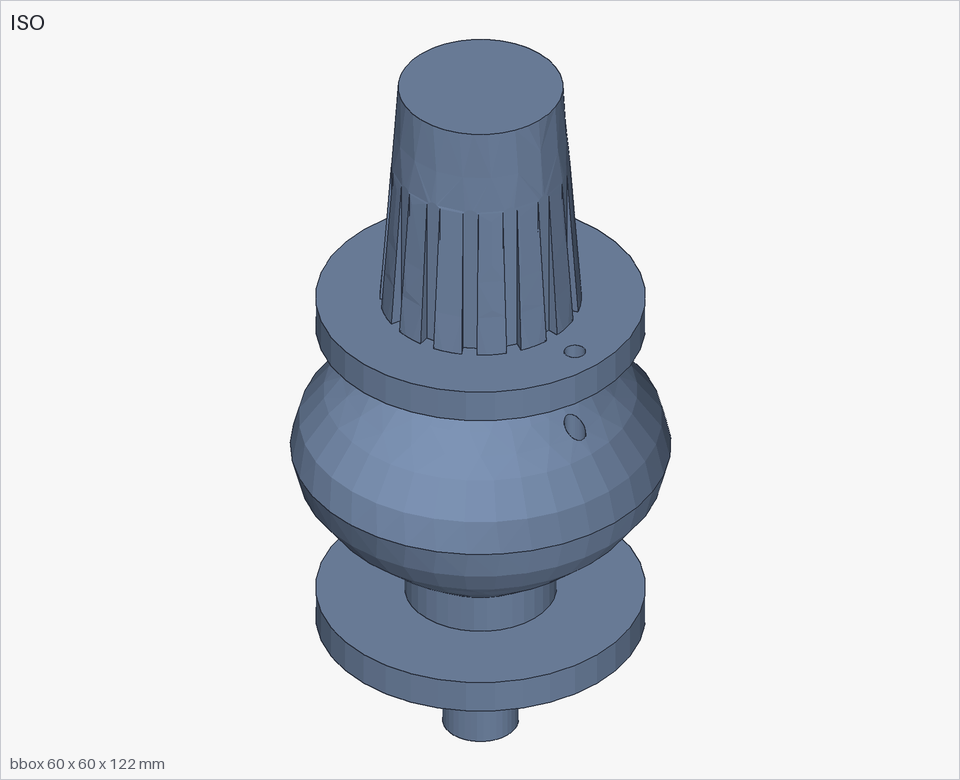}\\[-0.2em]{\scriptsize 45.5}} \\
\texttt{\scriptsize pcb\_126907} & \makecell{\includegraphics[width=\linewidth,height=0.58in,keepaspectratio,valign=c]{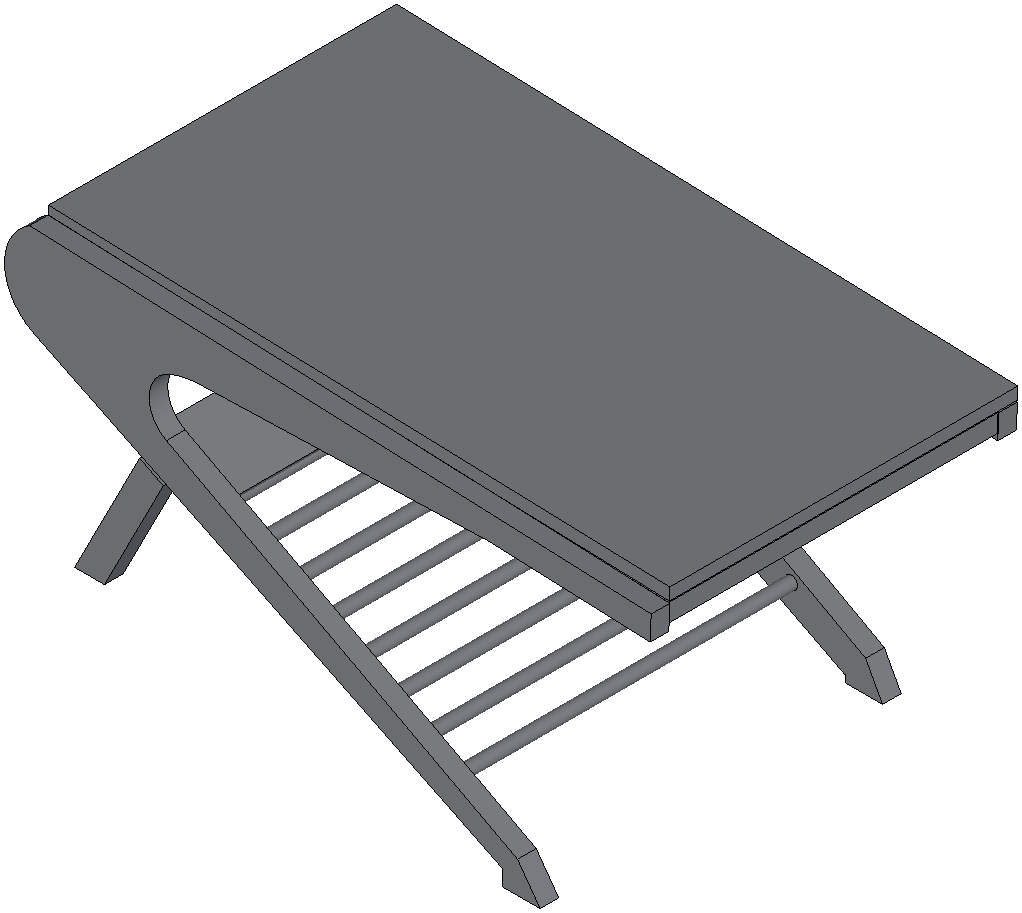}} & \makecell{\includegraphics[width=\linewidth,height=0.58in,keepaspectratio,valign=c]{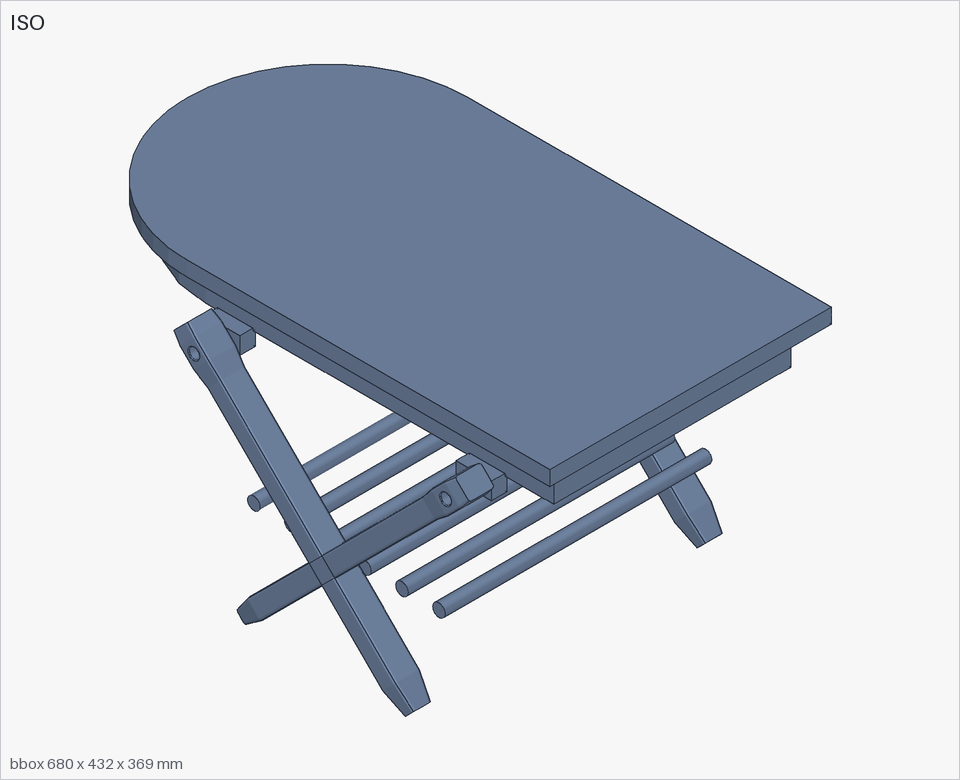}\\[-0.2em]{\scriptsize 87.8}} & \makecell{\includegraphics[width=\linewidth,height=0.58in,keepaspectratio,valign=c]{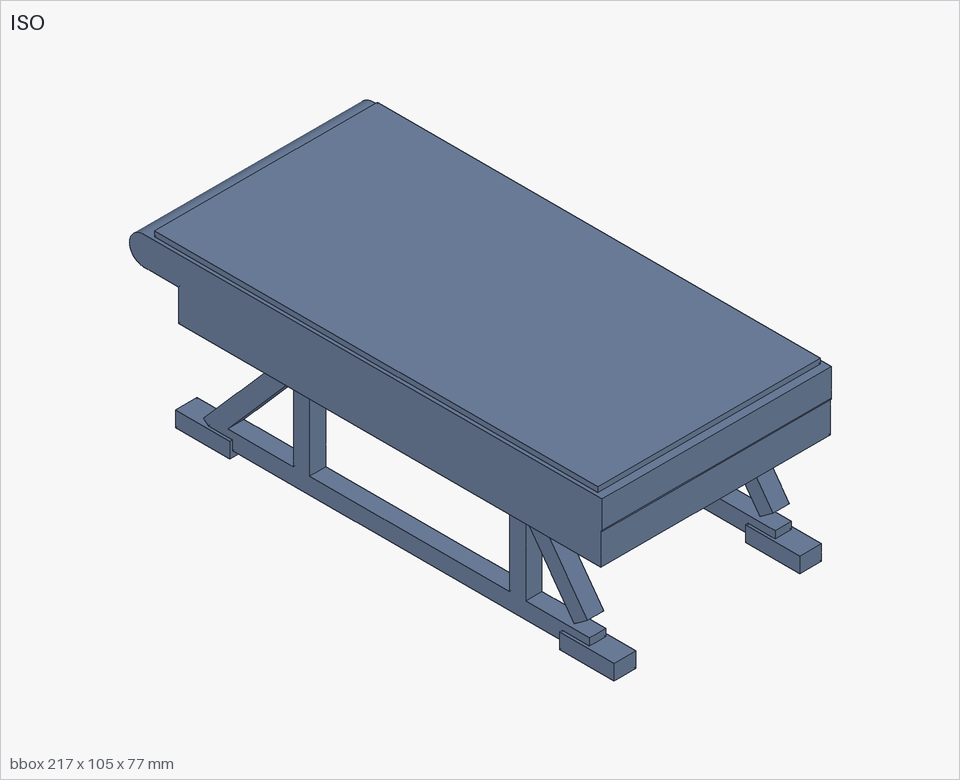}\\[-0.2em]{\scriptsize 65.1}} & \makecell{\includegraphics[width=\linewidth,height=0.58in,keepaspectratio,valign=c]{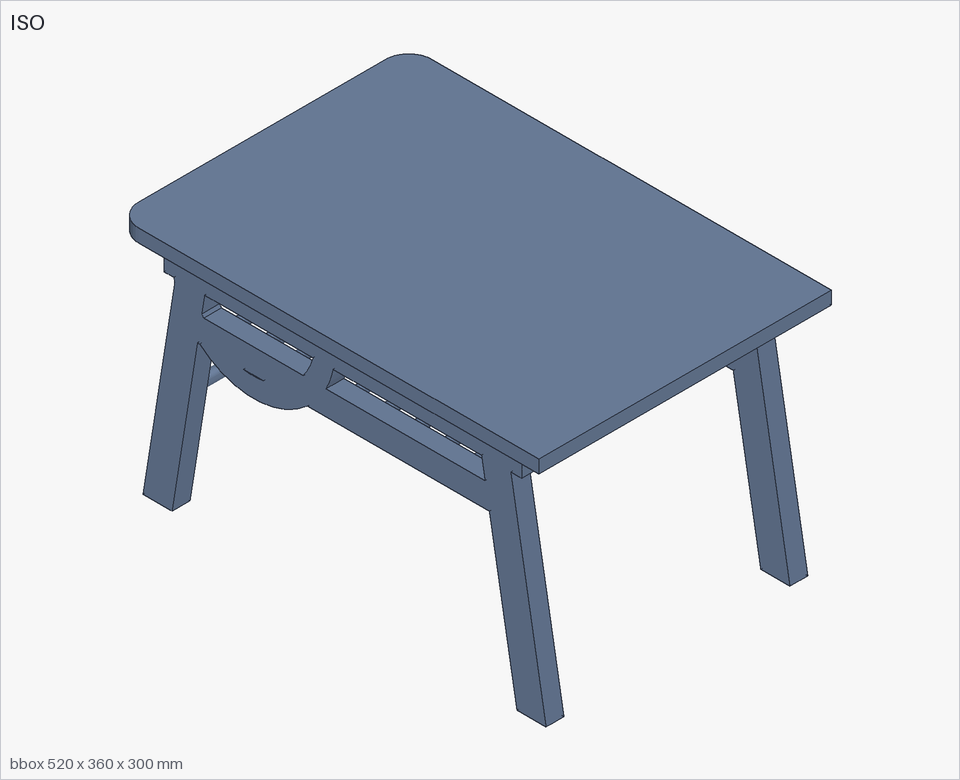}\\[-0.2em]{\scriptsize 69.5}} \\
\texttt{\scriptsize pcb\_137485} & \makecell{\includegraphics[width=\linewidth,height=0.58in,keepaspectratio,valign=c]{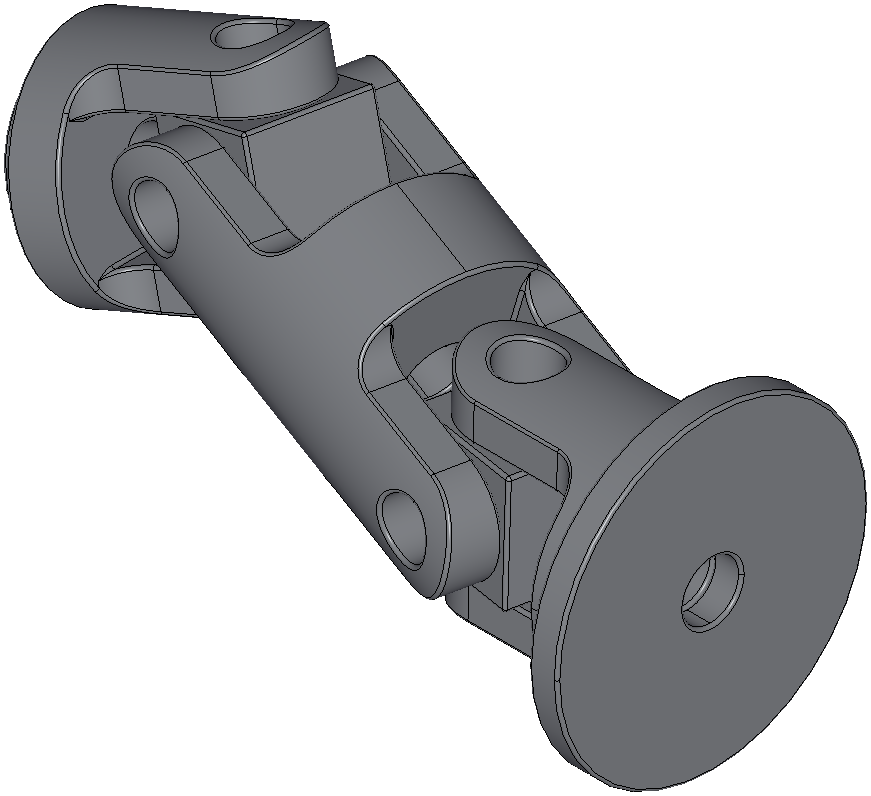}} & \makecell{\includegraphics[width=\linewidth,height=0.58in,keepaspectratio,valign=c]{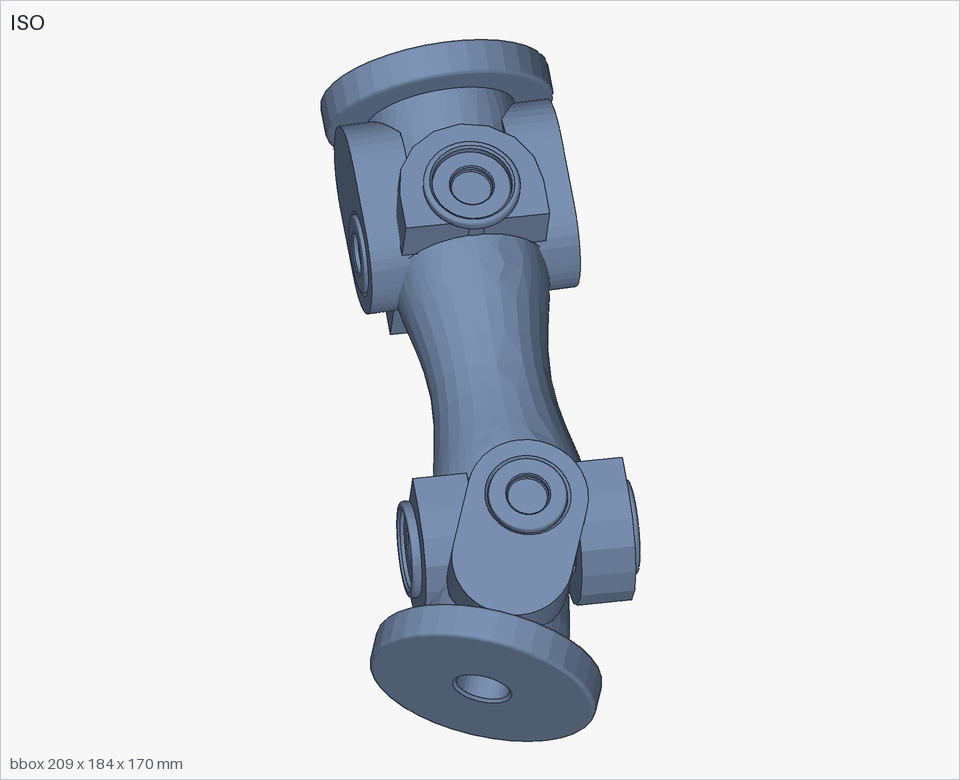}\\[-0.2em]{\scriptsize 87.9}} & \makecell{\includegraphics[width=\linewidth,height=0.58in,keepaspectratio,valign=c]{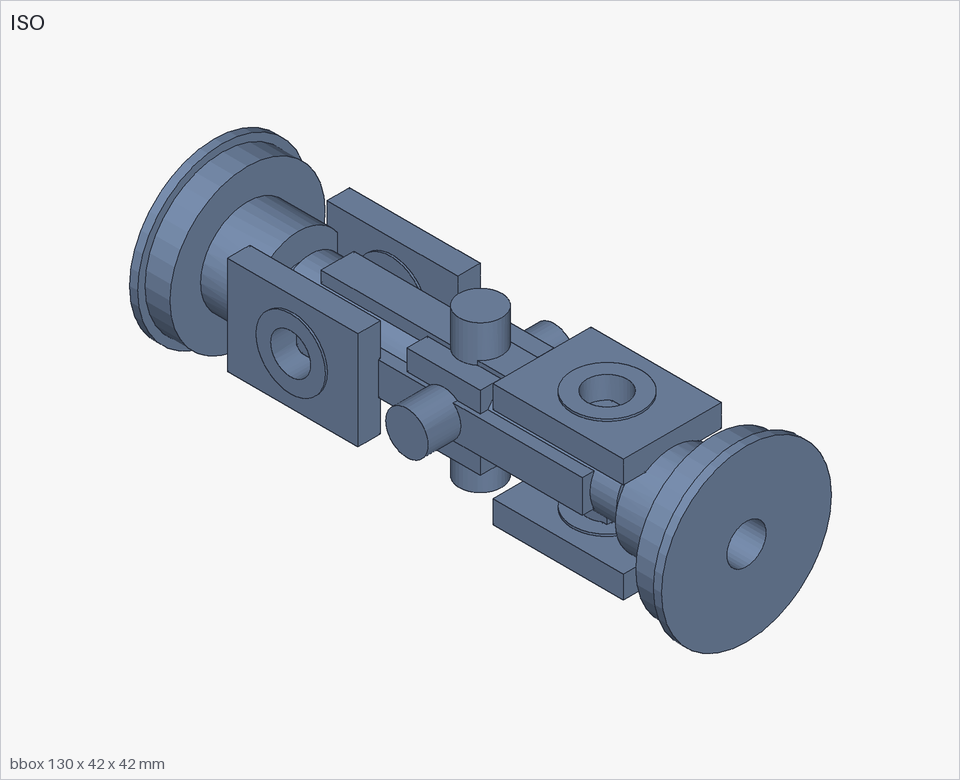}\\[-0.2em]{\scriptsize 73.3}} & \makecell{\includegraphics[width=\linewidth,height=0.58in,keepaspectratio,valign=c]{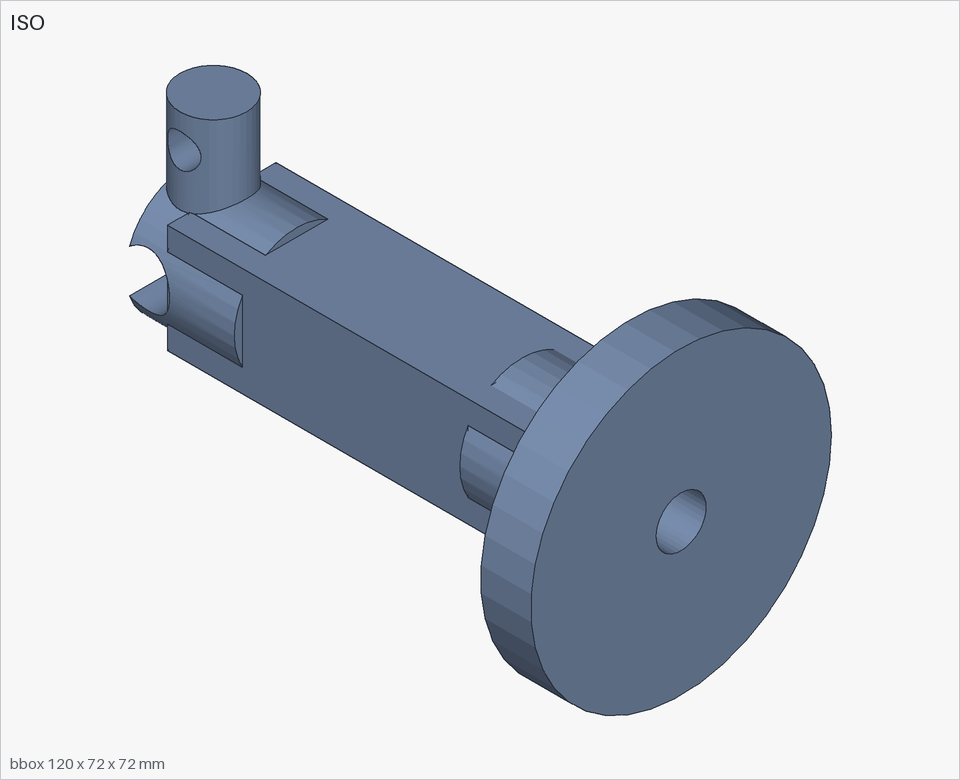}\\[-0.2em]{\scriptsize 59.7}} \\
\texttt{\scriptsize pcb\_139704} & \makecell{\includegraphics[width=\linewidth,height=0.58in,keepaspectratio,valign=c]{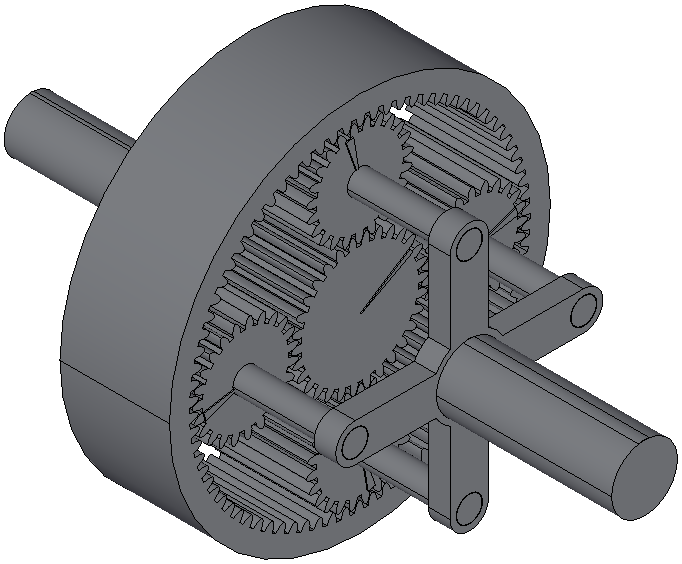}} & \makecell{\includegraphics[width=\linewidth,height=0.58in,keepaspectratio,valign=c]{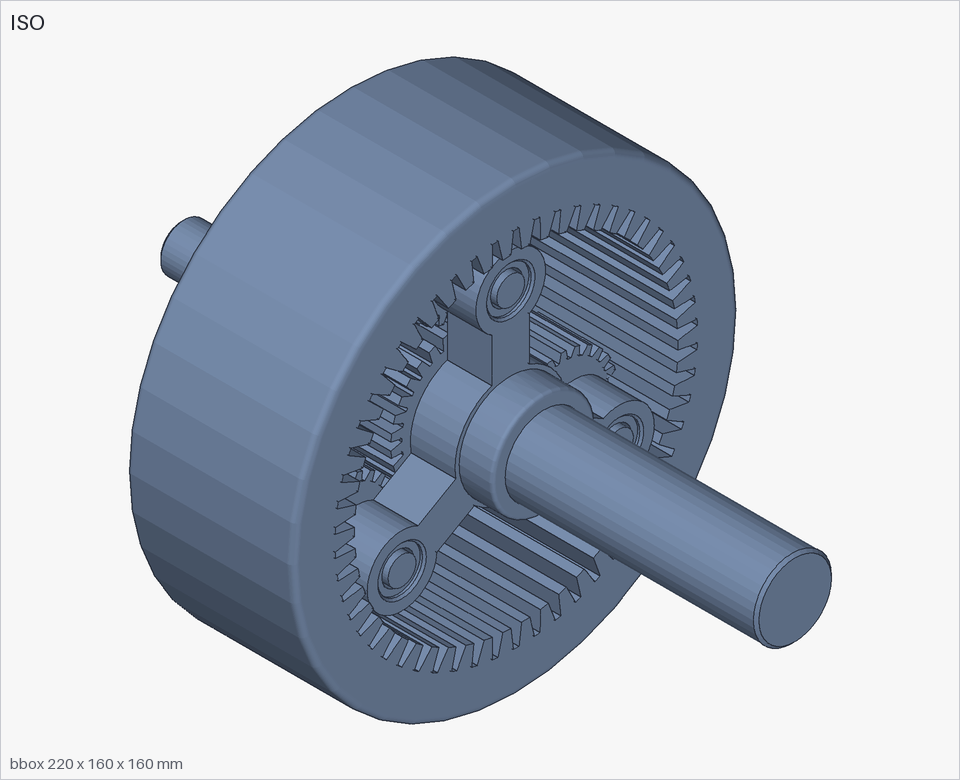}\\[-0.2em]{\scriptsize 84.7}} & \makecell{\includegraphics[width=\linewidth,height=0.58in,keepaspectratio,valign=c]{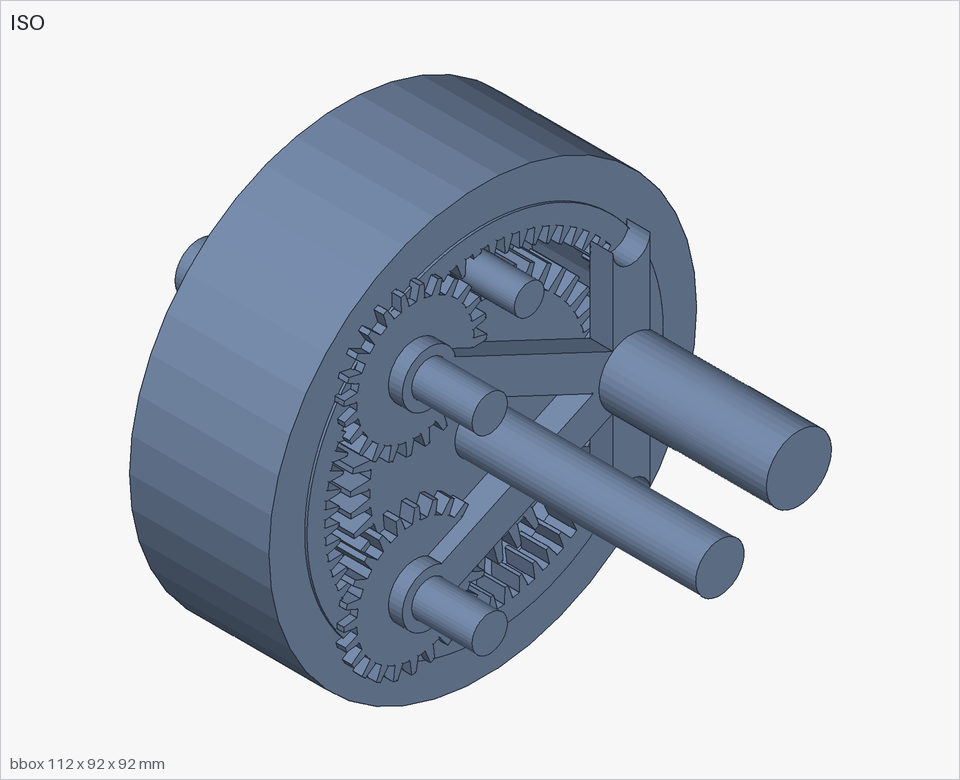}\\[-0.2em]{\scriptsize 80.5}} & \makecell{\raisebox{0.18in}{\color{gray}\scriptsize no export}} \\
\texttt{\scriptsize pcb\_142057} & \makecell{\includegraphics[width=\linewidth,height=0.58in,keepaspectratio,valign=c]{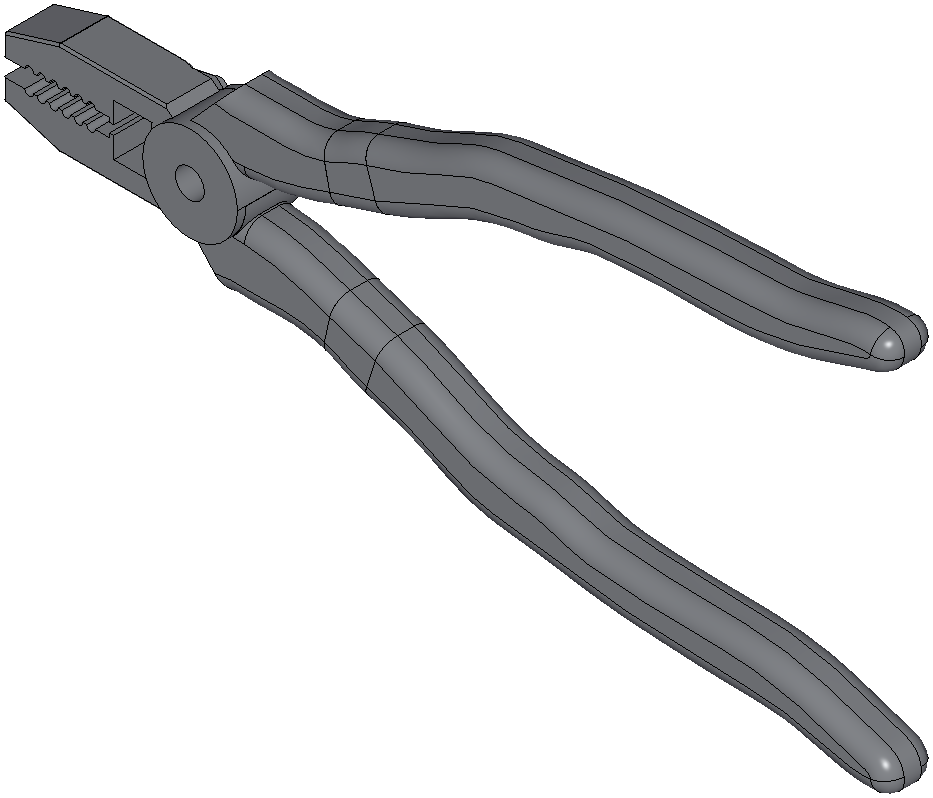}} & \makecell{\includegraphics[width=\linewidth,height=0.58in,keepaspectratio,valign=c]{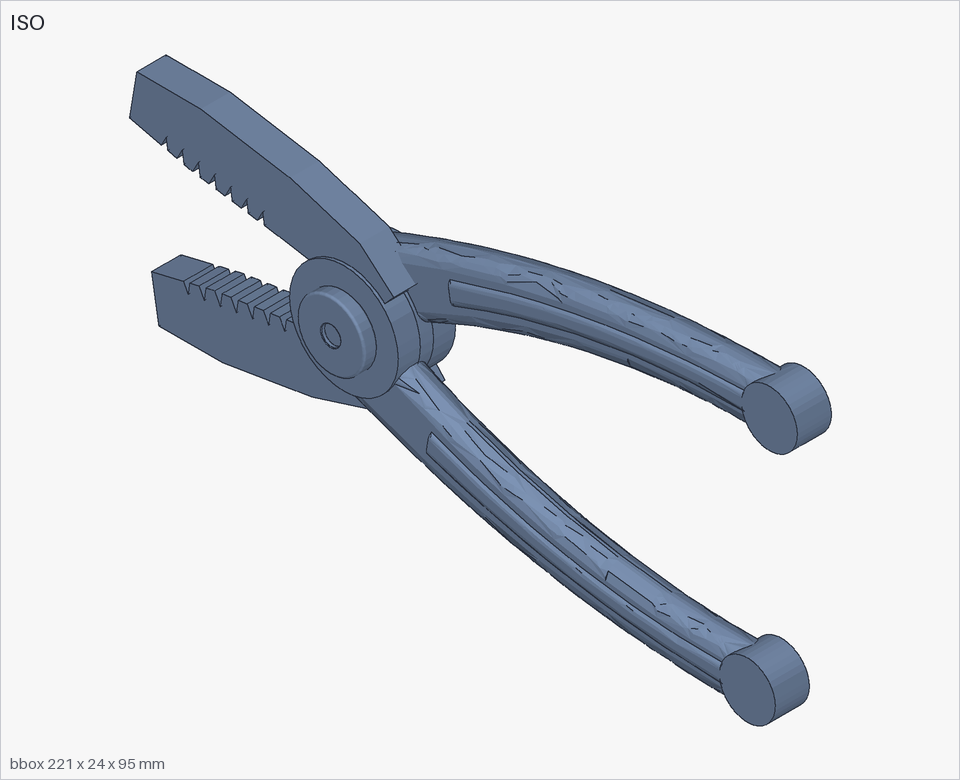}\\[-0.2em]{\scriptsize 85.7}} & \makecell{\includegraphics[width=\linewidth,height=0.58in,keepaspectratio,valign=c]{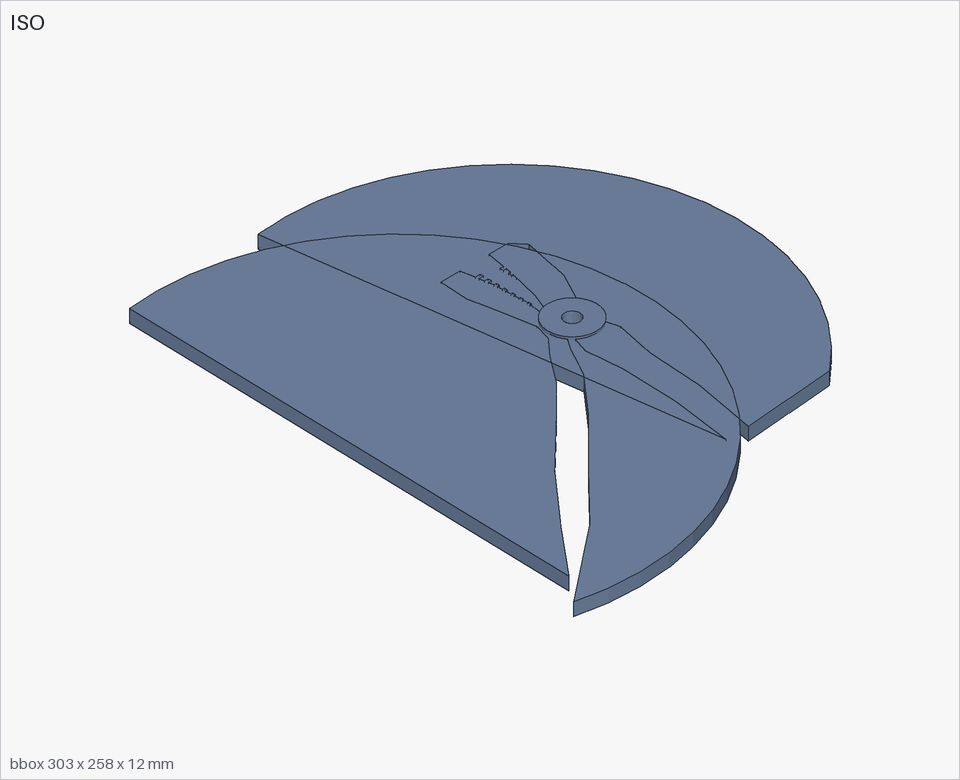}\\[-0.2em]{\scriptsize 49.2}} & \makecell{\includegraphics[width=\linewidth,height=0.58in,keepaspectratio,valign=c]{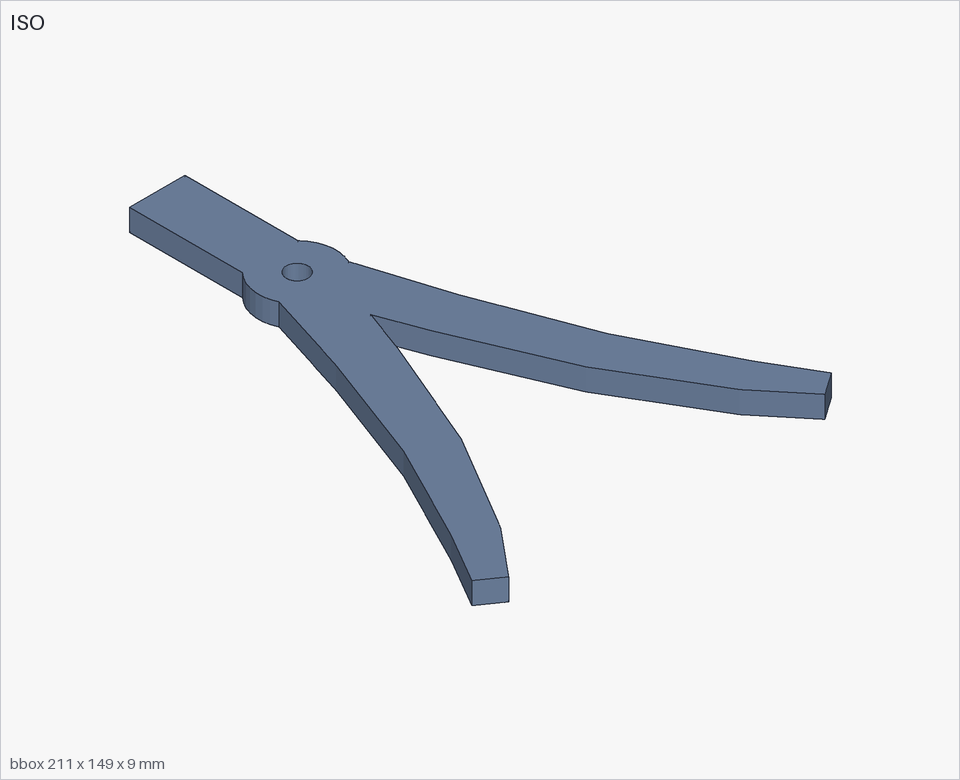}\\[-0.2em]{\scriptsize 50.0}} \\
\texttt{\scriptsize pcb\_143872} & \makecell{\includegraphics[width=\linewidth,height=0.58in,keepaspectratio,valign=c]{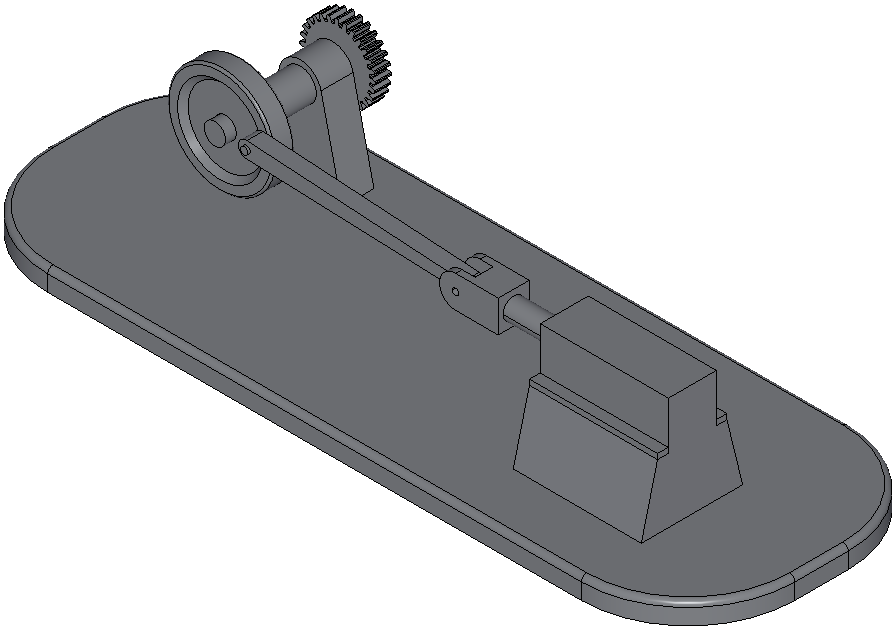}} & \makecell{\includegraphics[width=\linewidth,height=0.58in,keepaspectratio,valign=c]{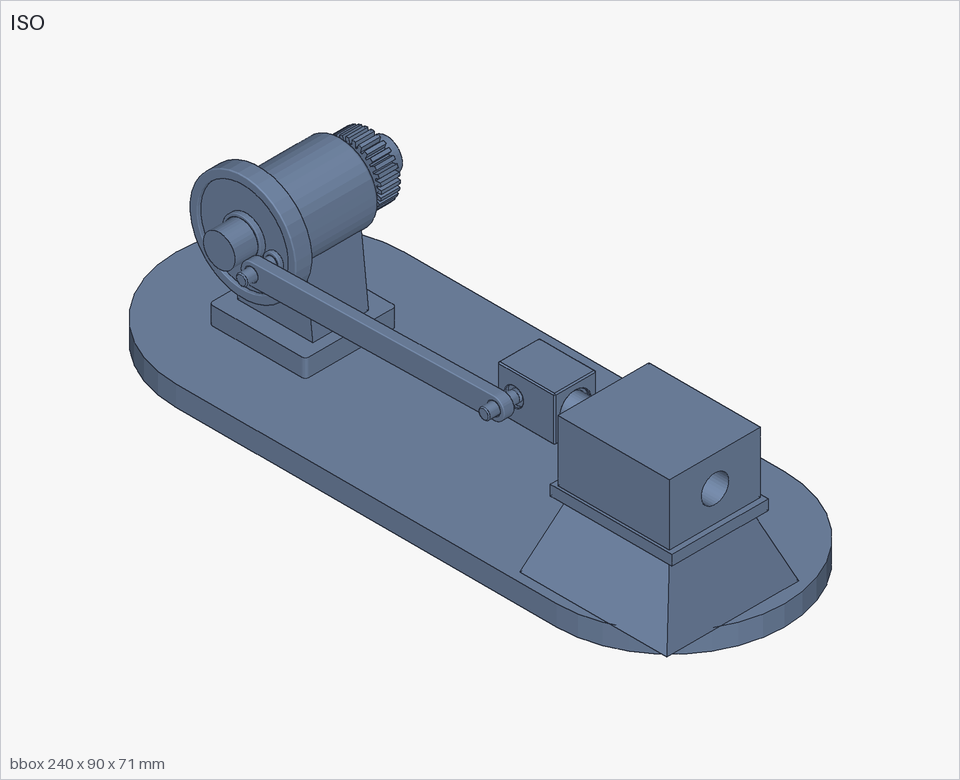}\\[-0.2em]{\scriptsize 89.0}} & \makecell{\includegraphics[width=\linewidth,height=0.58in,keepaspectratio,valign=c]{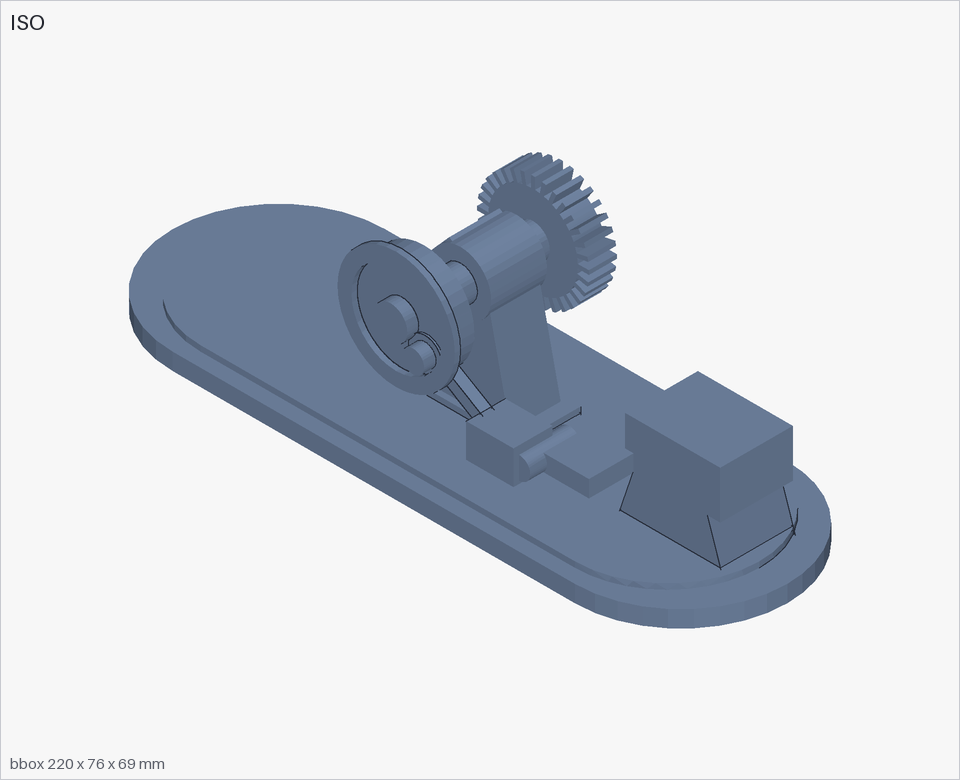}\\[-0.2em]{\scriptsize 85.9}} & \makecell{\raisebox{0.18in}{\color{gray}\scriptsize no export}} \\
\texttt{\scriptsize pcb\_144436} & \makecell{\includegraphics[width=\linewidth,height=0.58in,keepaspectratio,valign=c]{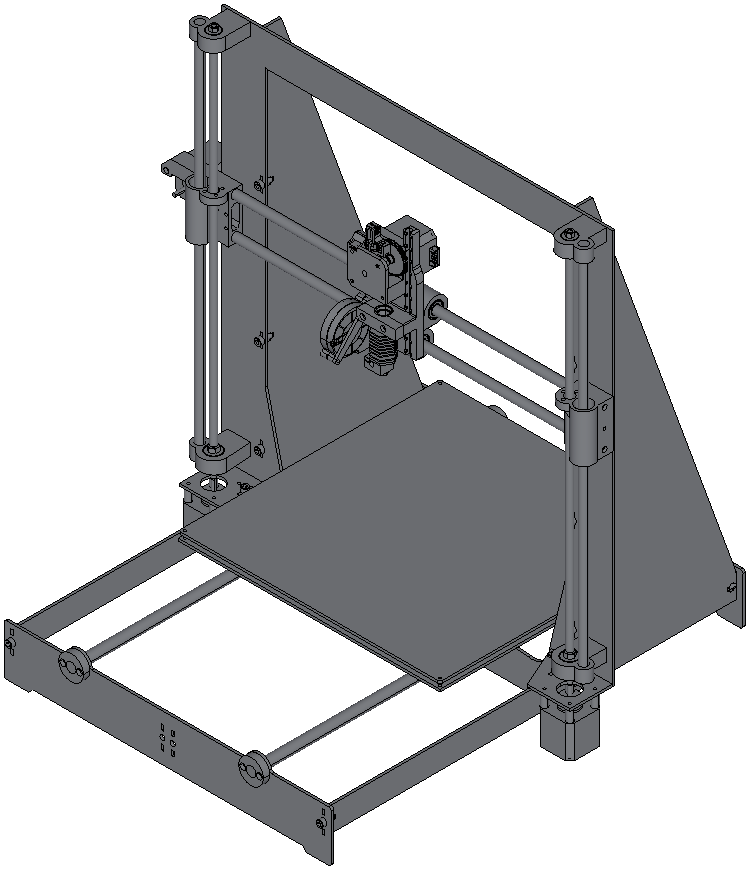}} & \makecell{\includegraphics[width=\linewidth,height=0.58in,keepaspectratio,valign=c]{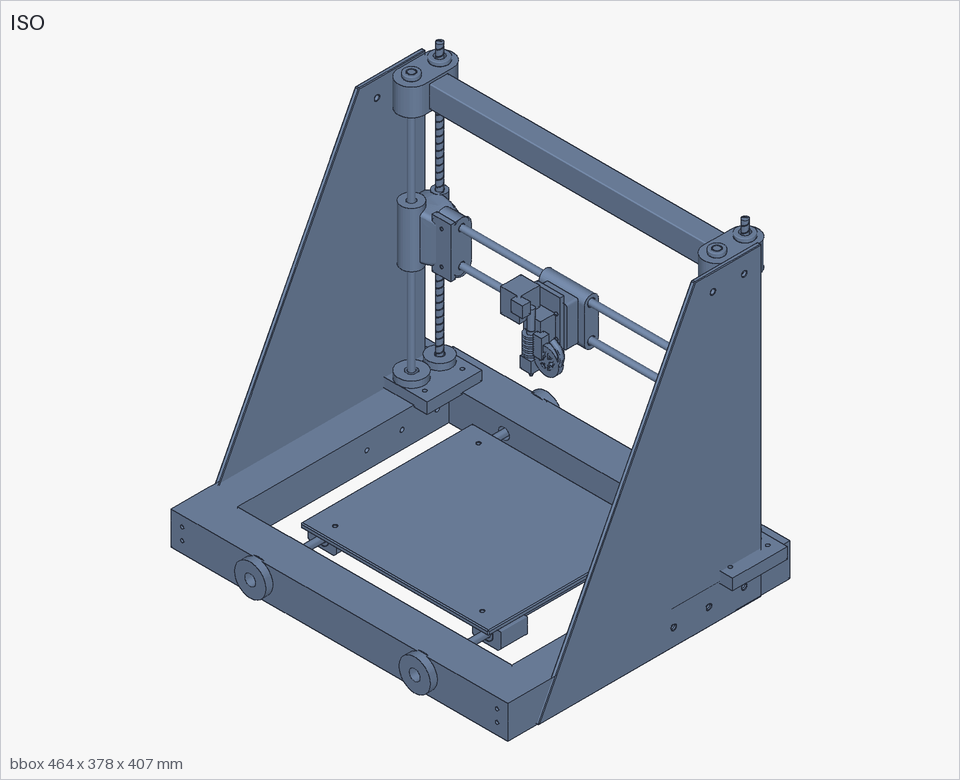}\\[-0.2em]{\scriptsize 89.2}} & \makecell{\raisebox{0.18in}{\color{gray}\scriptsize no export}} & \makecell{\raisebox{0.18in}{\color{gray}\scriptsize no export}} \\
\texttt{\scriptsize pcb\_145368} & \makecell{\includegraphics[width=\linewidth,height=0.58in,keepaspectratio,valign=c]{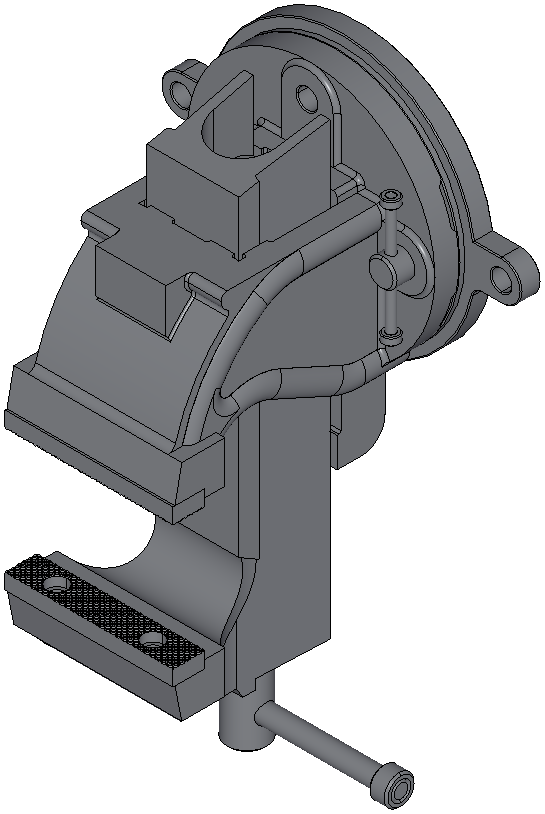}} & \makecell{\includegraphics[width=\linewidth,height=0.58in,keepaspectratio,valign=c]{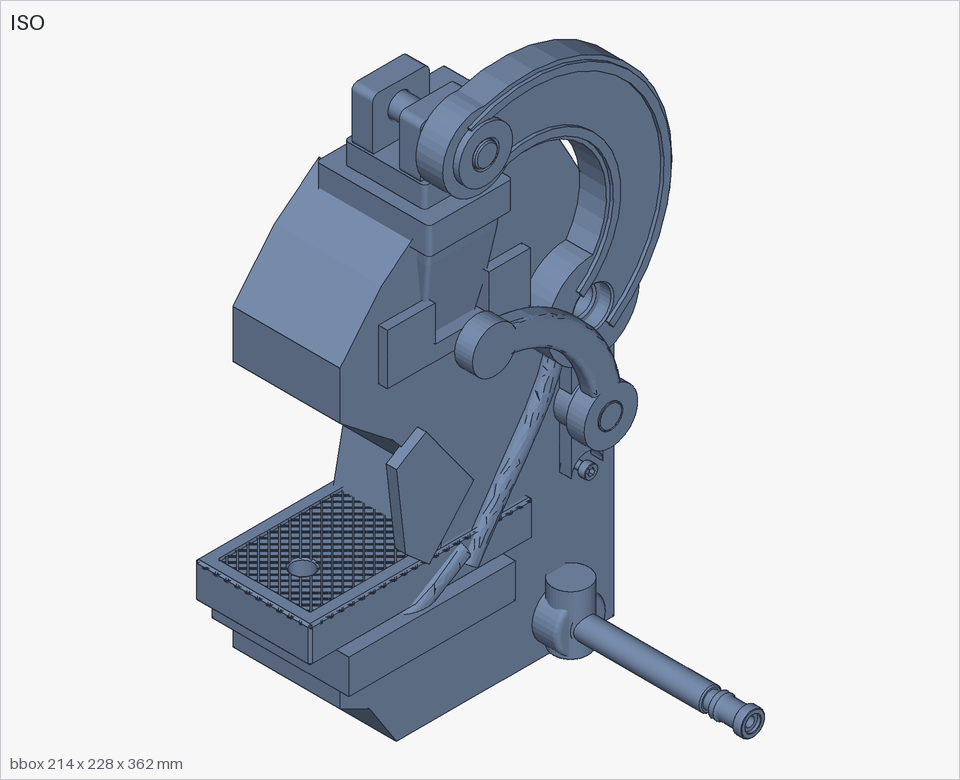}\\[-0.2em]{\scriptsize 84.5}} & \makecell{\includegraphics[width=\linewidth,height=0.58in,keepaspectratio,valign=c]{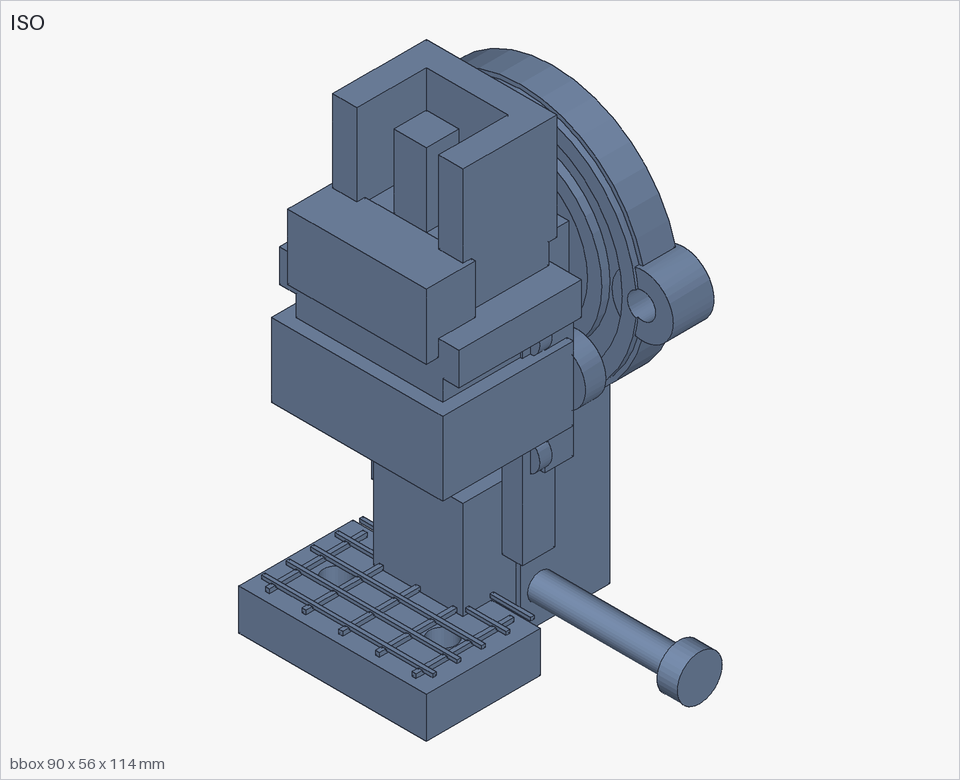}\\[-0.2em]{\scriptsize 78.0}} & \makecell{\raisebox{0.18in}{\color{gray}\scriptsize no export}} \\
\texttt{\scriptsize pcb\_21557} & \makecell{\includegraphics[width=\linewidth,height=0.58in,keepaspectratio,valign=c]{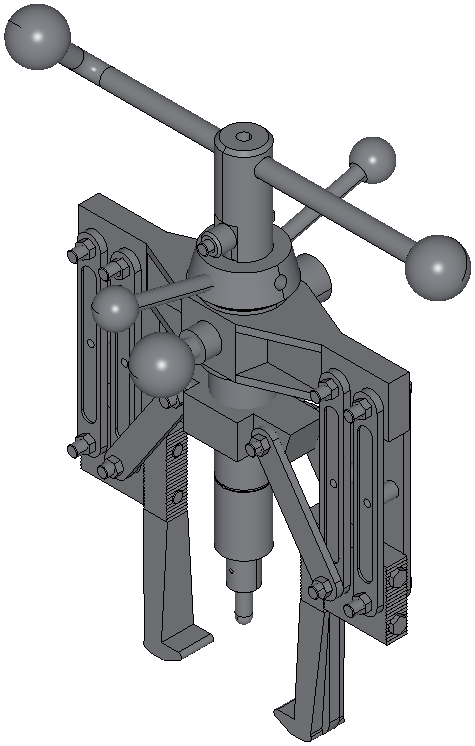}} & \makecell{\includegraphics[width=\linewidth,height=0.58in,keepaspectratio,valign=c]{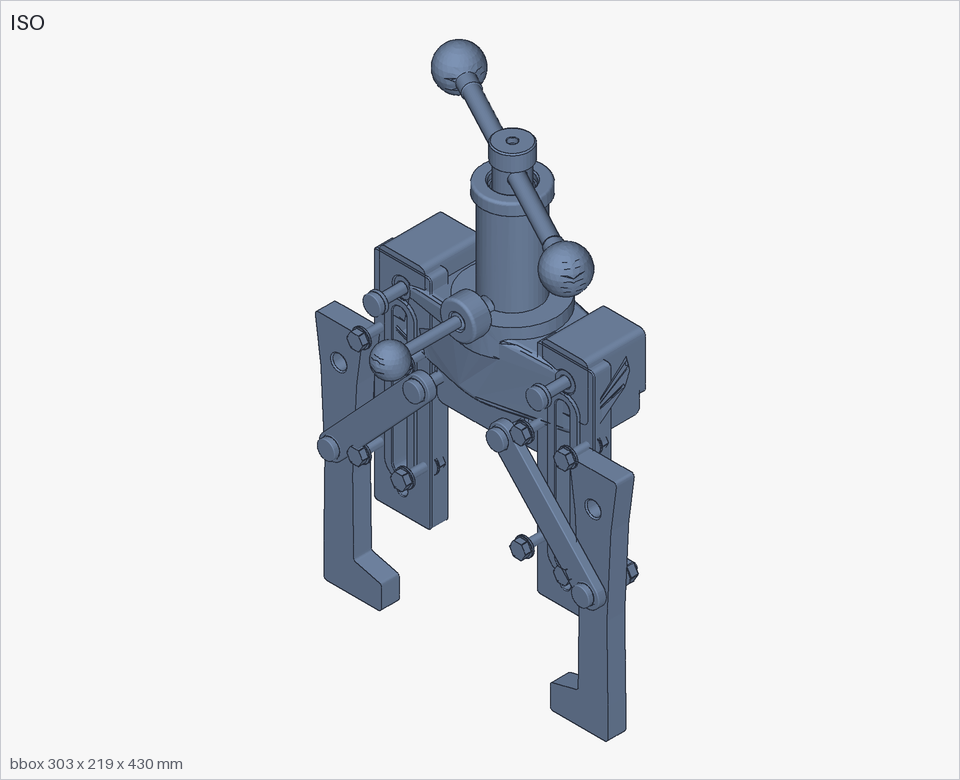}\\[-0.2em]{\scriptsize 84.9}} & \makecell{\includegraphics[width=\linewidth,height=0.58in,keepaspectratio,valign=c]{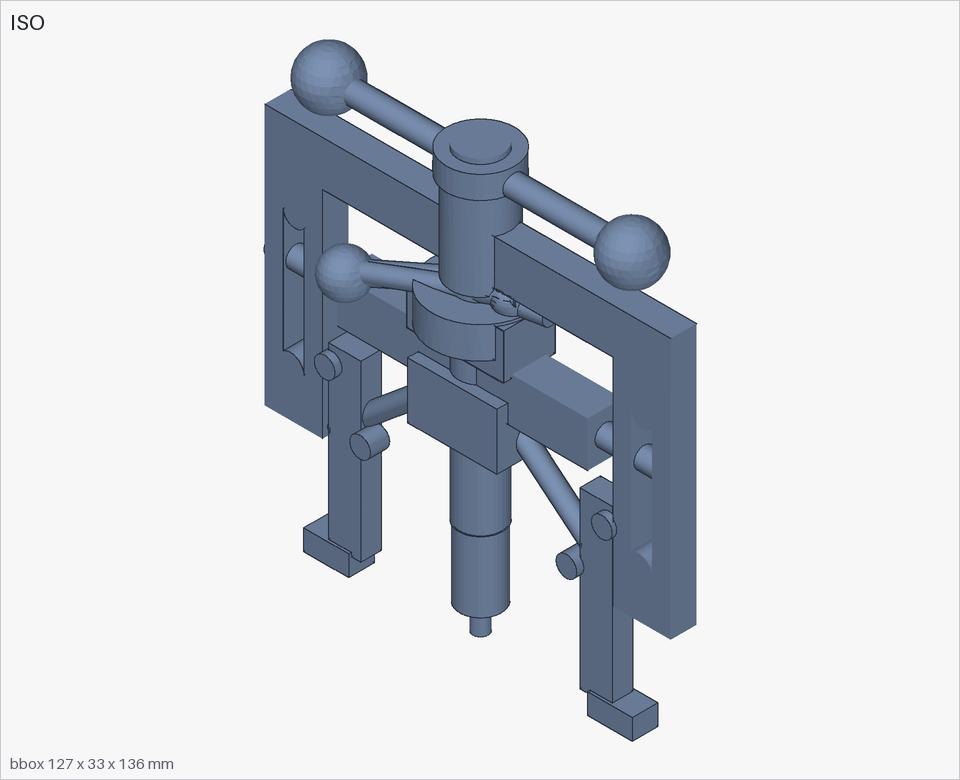}\\[-0.2em]{\scriptsize 71.8}} & \makecell{\includegraphics[width=\linewidth,height=0.58in,keepaspectratio,valign=c]{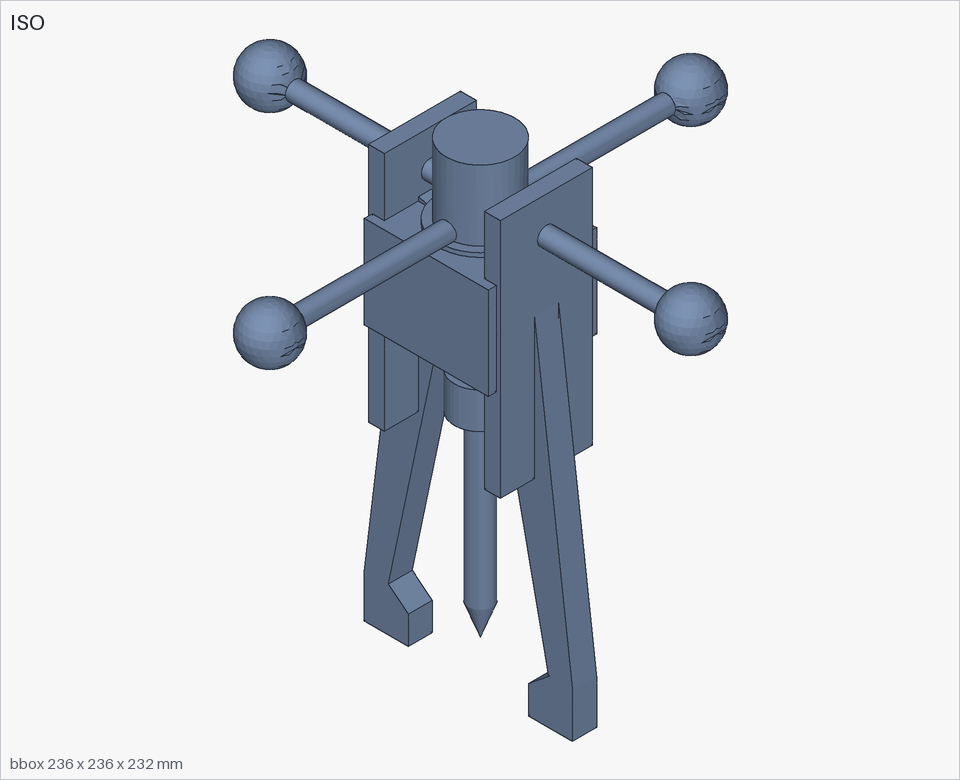}\\[-0.2em]{\scriptsize 56.1}} \\
\texttt{\scriptsize pcb\_22081} & \makecell{\includegraphics[width=\linewidth,height=0.58in,keepaspectratio,valign=c]{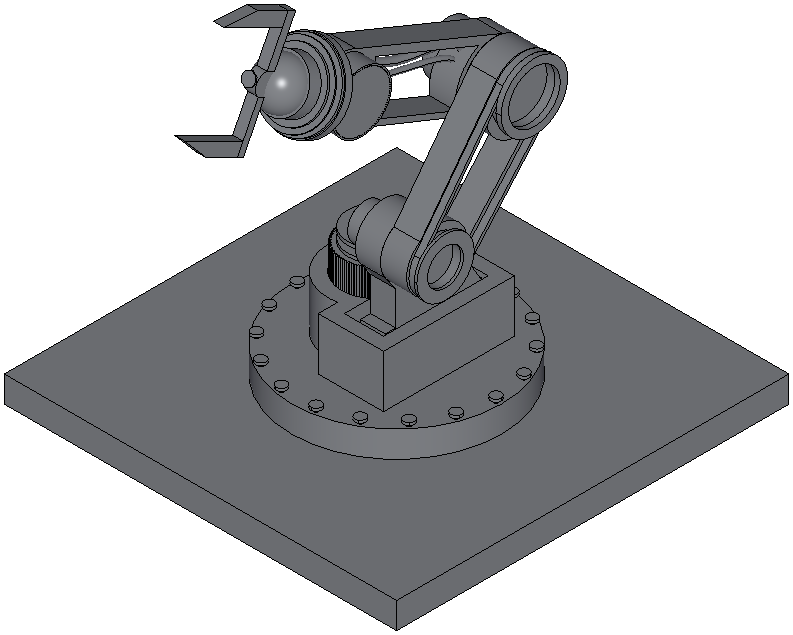}} & \makecell{\includegraphics[width=\linewidth,height=0.58in,keepaspectratio,valign=c]{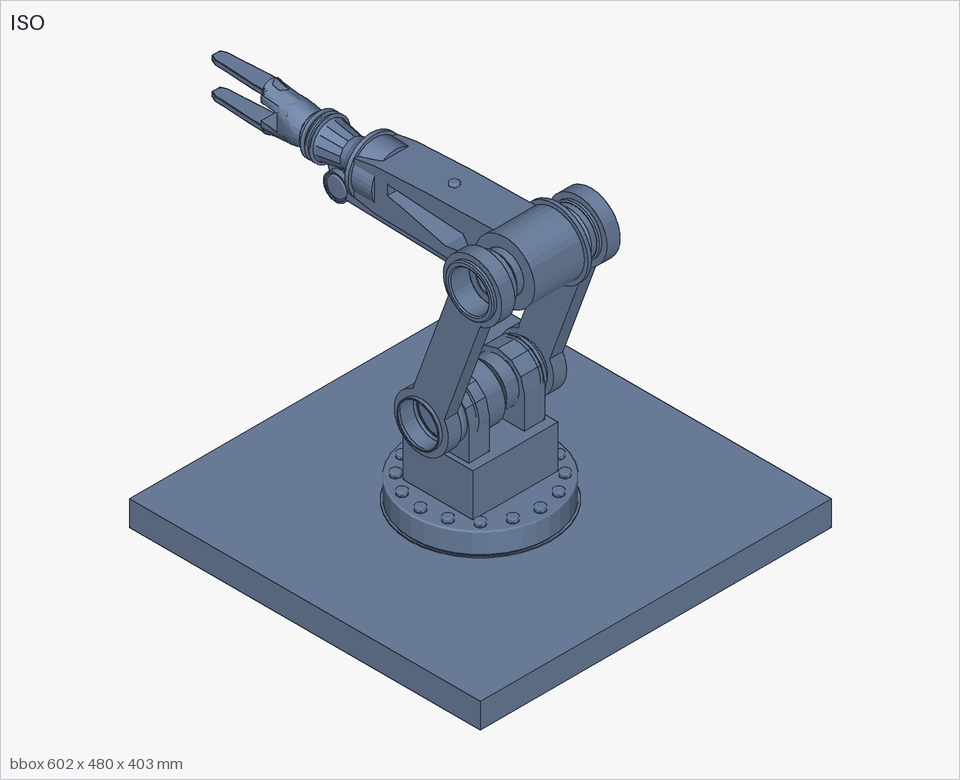}\\[-0.2em]{\scriptsize 83.2}} & \makecell{\includegraphics[width=\linewidth,height=0.58in,keepaspectratio,valign=c]{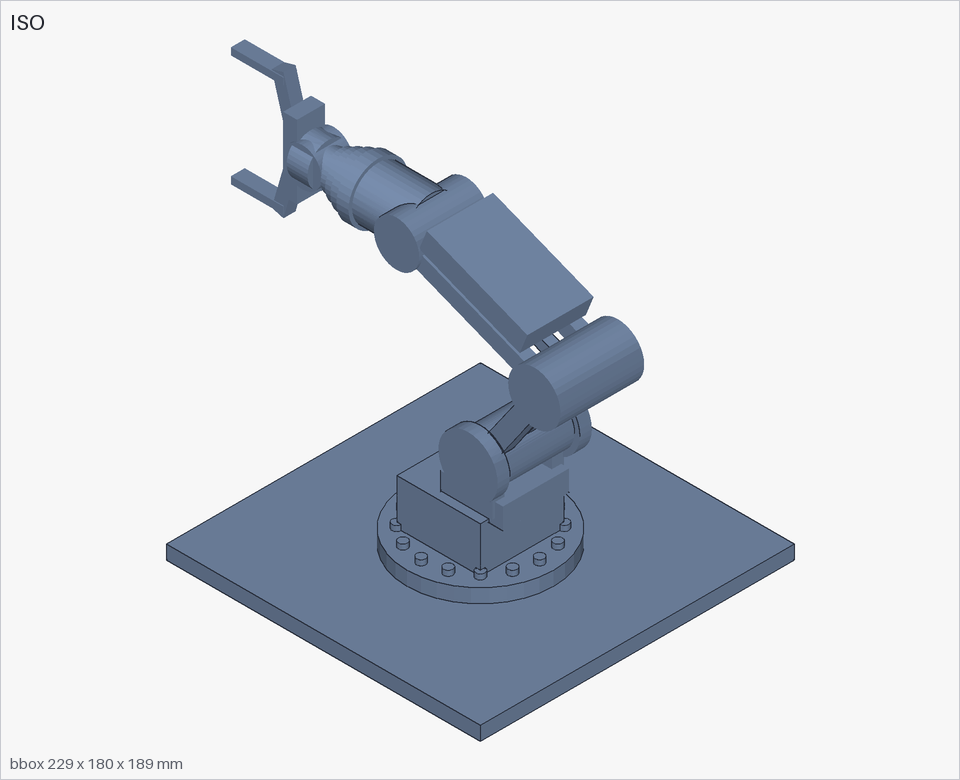}\\[-0.2em]{\scriptsize 80.8}} & \makecell{\includegraphics[width=\linewidth,height=0.58in,keepaspectratio,valign=c]{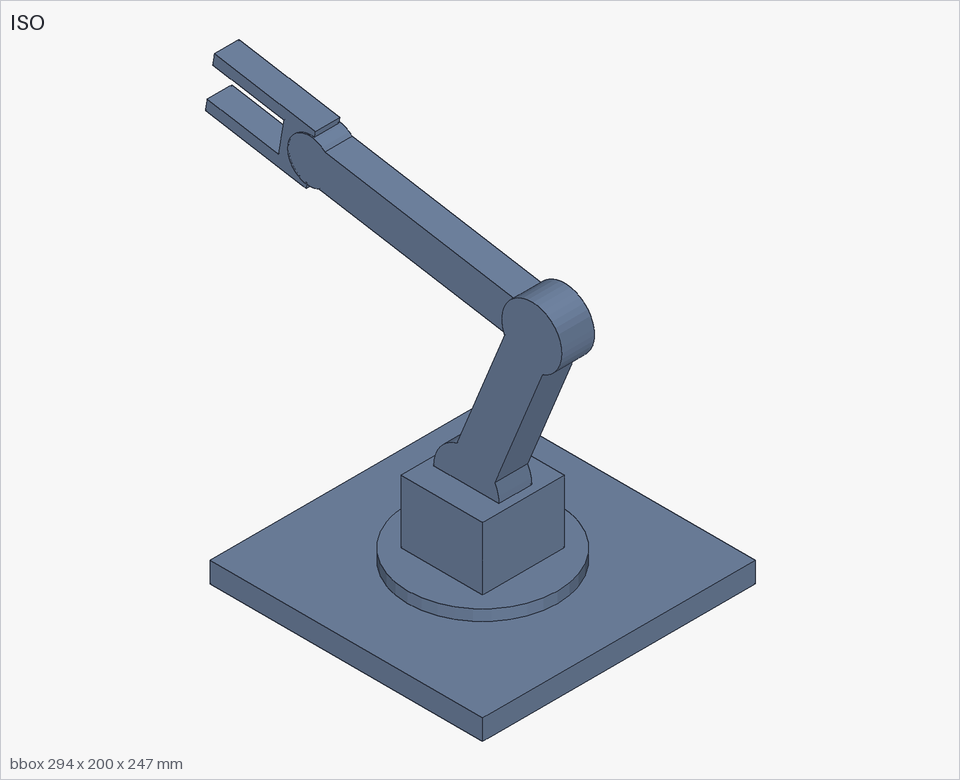}\\[-0.2em]{\scriptsize 62.2}} \\
\texttt{\scriptsize pcb\_22630} & \makecell{\includegraphics[width=\linewidth,height=0.58in,keepaspectratio,valign=c]{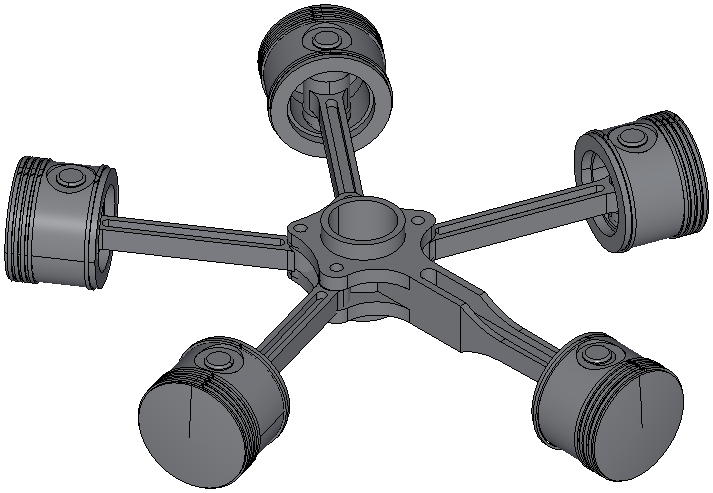}} & \makecell{\includegraphics[width=\linewidth,height=0.58in,keepaspectratio,valign=c]{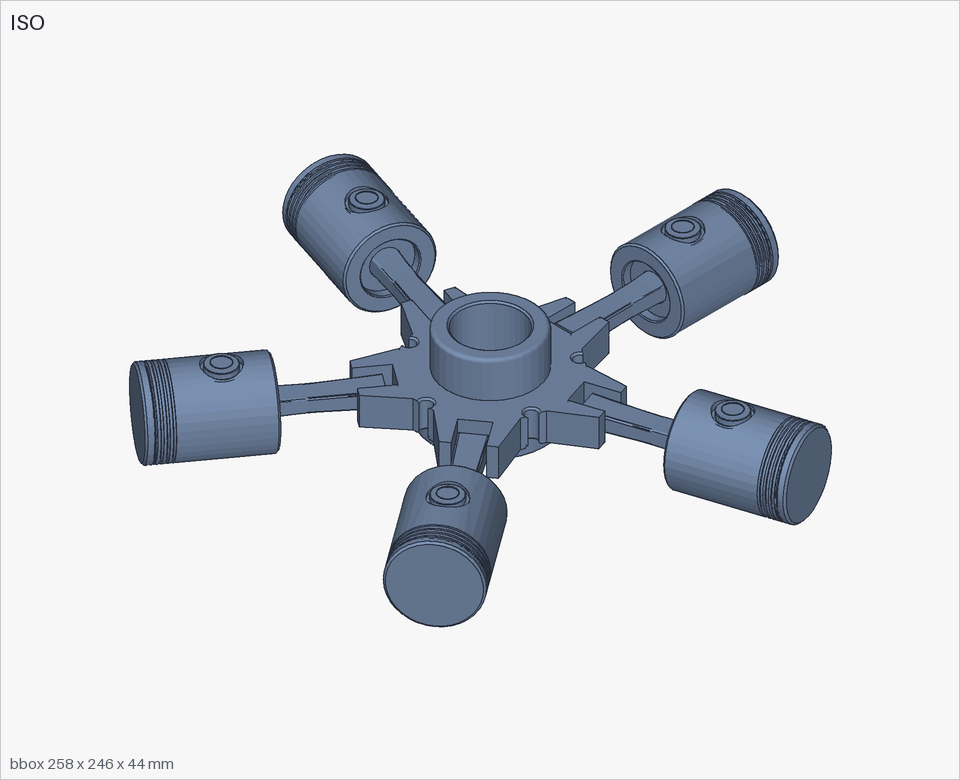}\\[-0.2em]{\scriptsize 90.1}} & \makecell{\includegraphics[width=\linewidth,height=0.58in,keepaspectratio,valign=c]{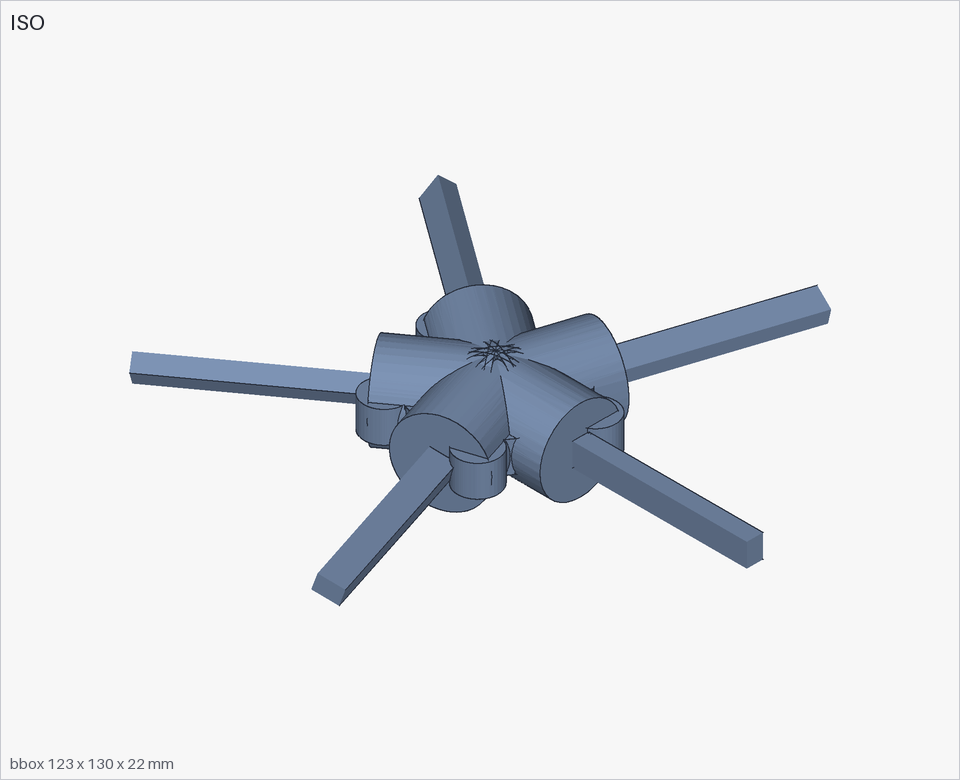}\\[-0.2em]{\scriptsize 45.6}} & \makecell{\includegraphics[width=\linewidth,height=0.58in,keepaspectratio,valign=c]{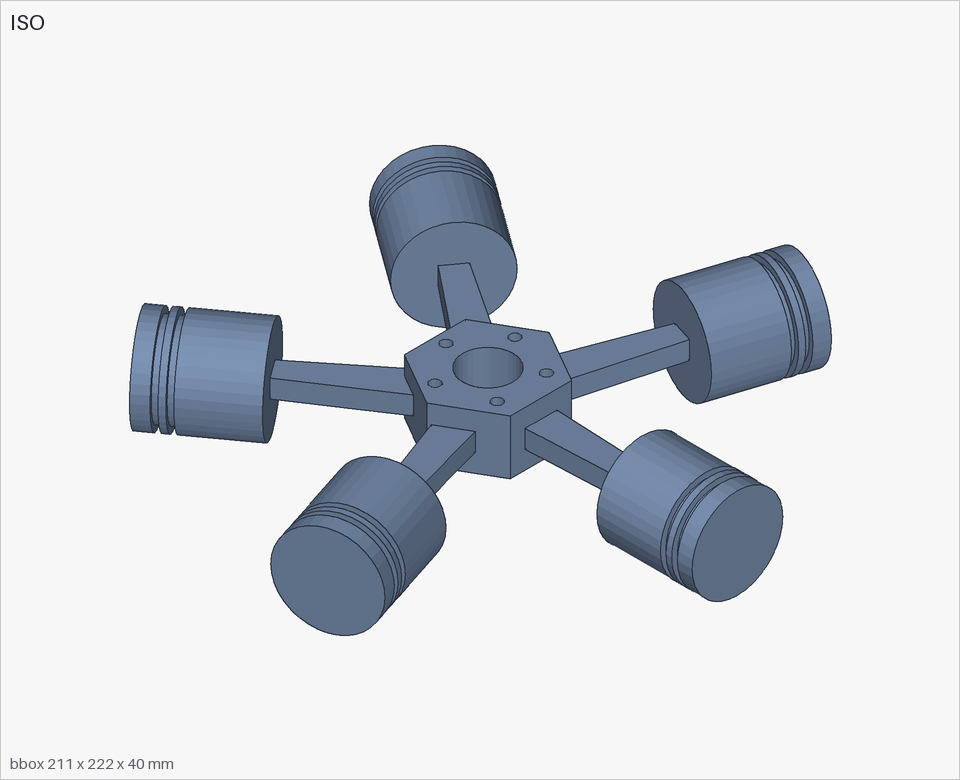}\\[-0.2em]{\scriptsize 80.7}} \\
\texttt{\scriptsize pcb\_23053} & \makecell{\includegraphics[width=\linewidth,height=0.58in,keepaspectratio,valign=c]{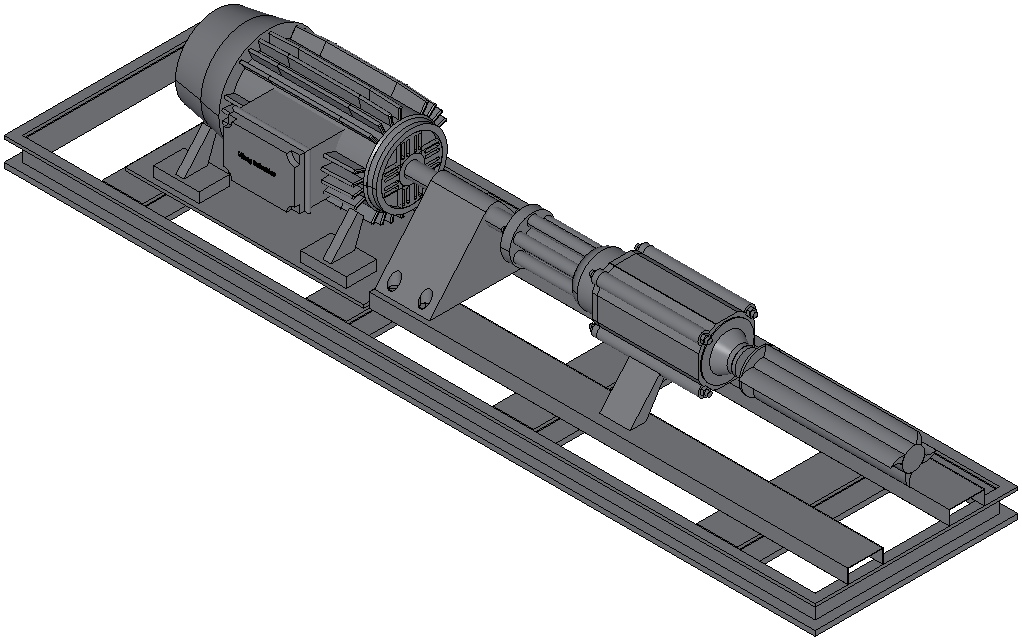}} & \makecell{\includegraphics[width=\linewidth,height=0.58in,keepaspectratio,valign=c]{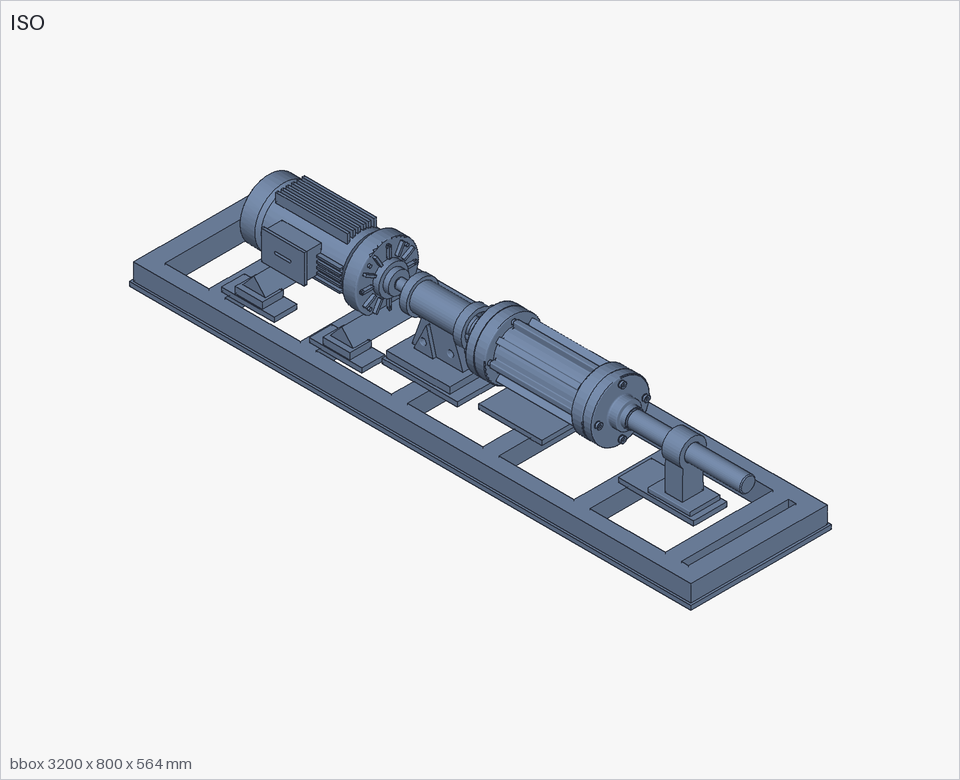}\\[-0.2em]{\scriptsize 85.5}} & \makecell{\includegraphics[width=\linewidth,height=0.58in,keepaspectratio,valign=c]{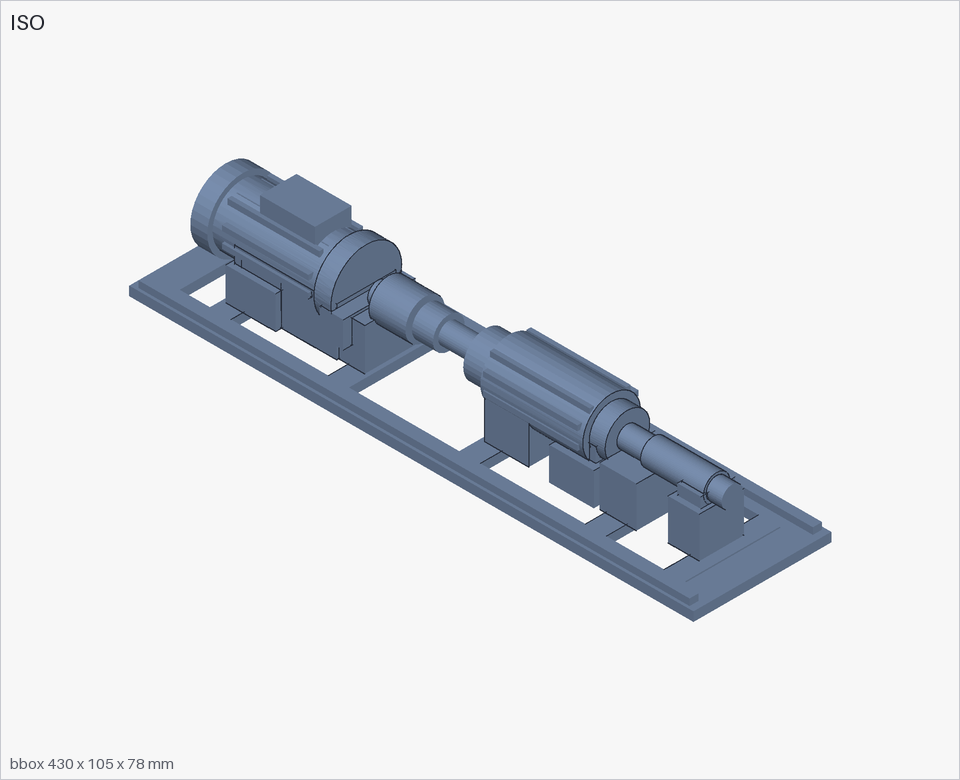}\\[-0.2em]{\scriptsize 81.1}} & \makecell{\includegraphics[width=\linewidth,height=0.58in,keepaspectratio,valign=c]{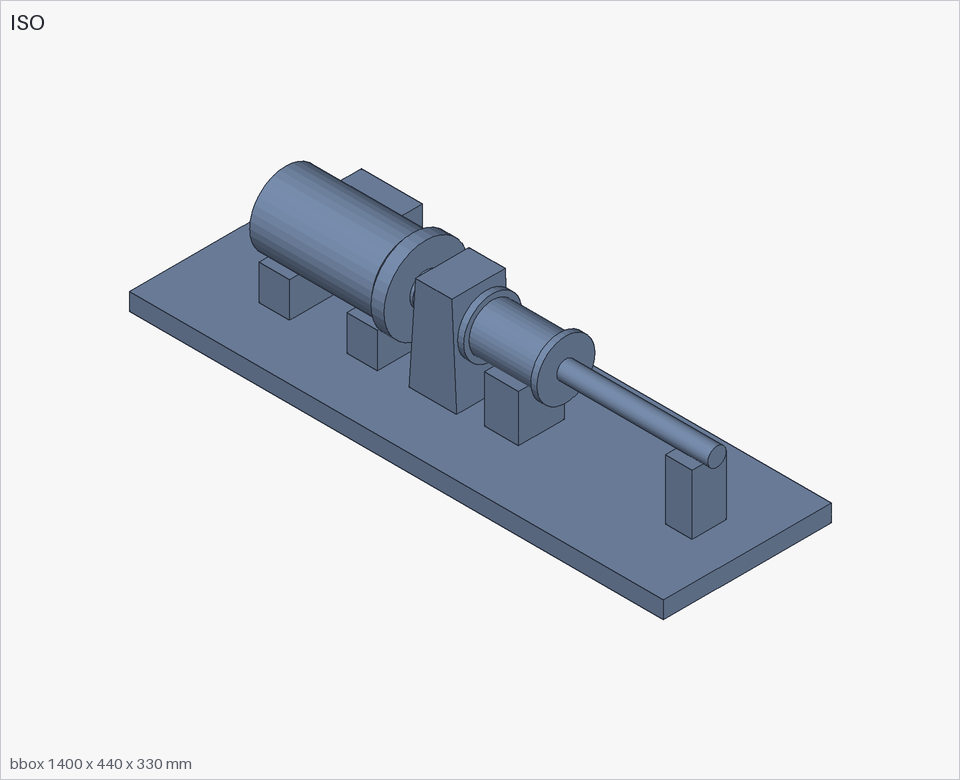}\\[-0.2em]{\scriptsize 68.9}} \\
\texttt{\scriptsize pcb\_23127} & \makecell{\includegraphics[width=\linewidth,height=0.58in,keepaspectratio,valign=c]{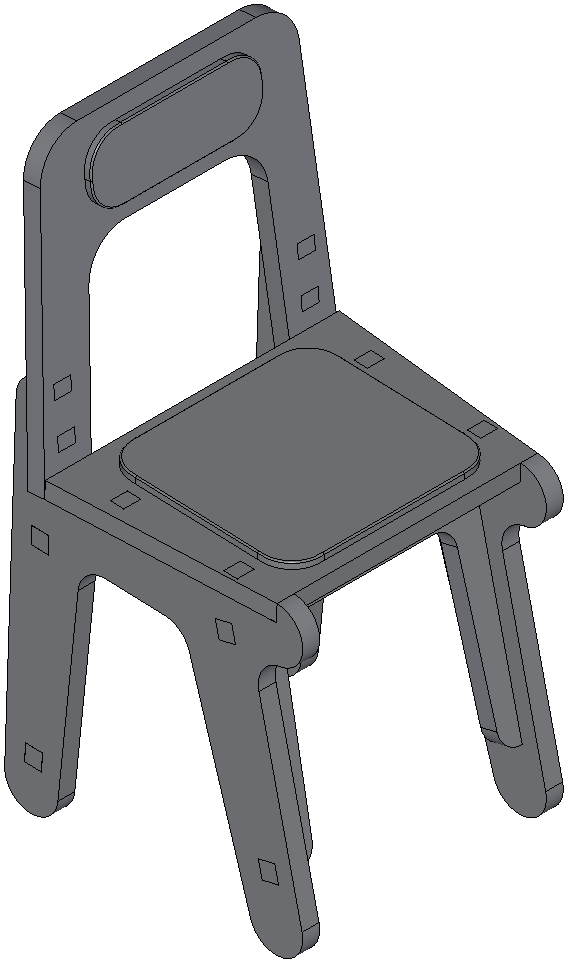}} & \makecell{\includegraphics[width=\linewidth,height=0.58in,keepaspectratio,valign=c]{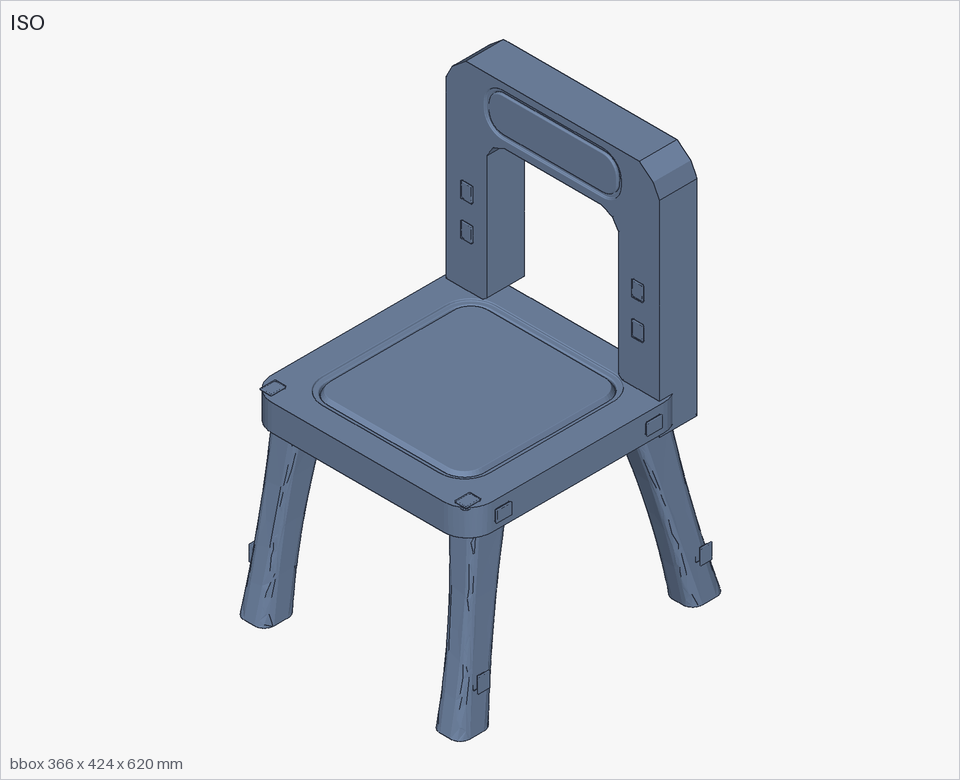}\\[-0.2em]{\scriptsize 88.4}} & \makecell{\includegraphics[width=\linewidth,height=0.58in,keepaspectratio,valign=c]{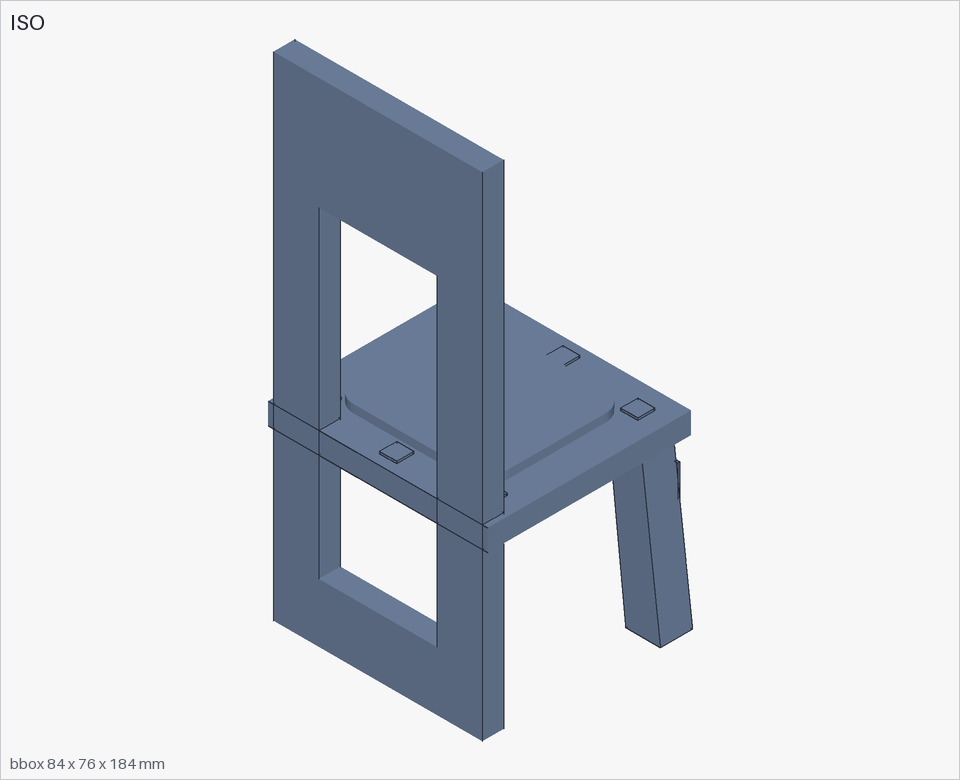}\\[-0.2em]{\scriptsize 76.8}} & \makecell{\includegraphics[width=\linewidth,height=0.58in,keepaspectratio,valign=c]{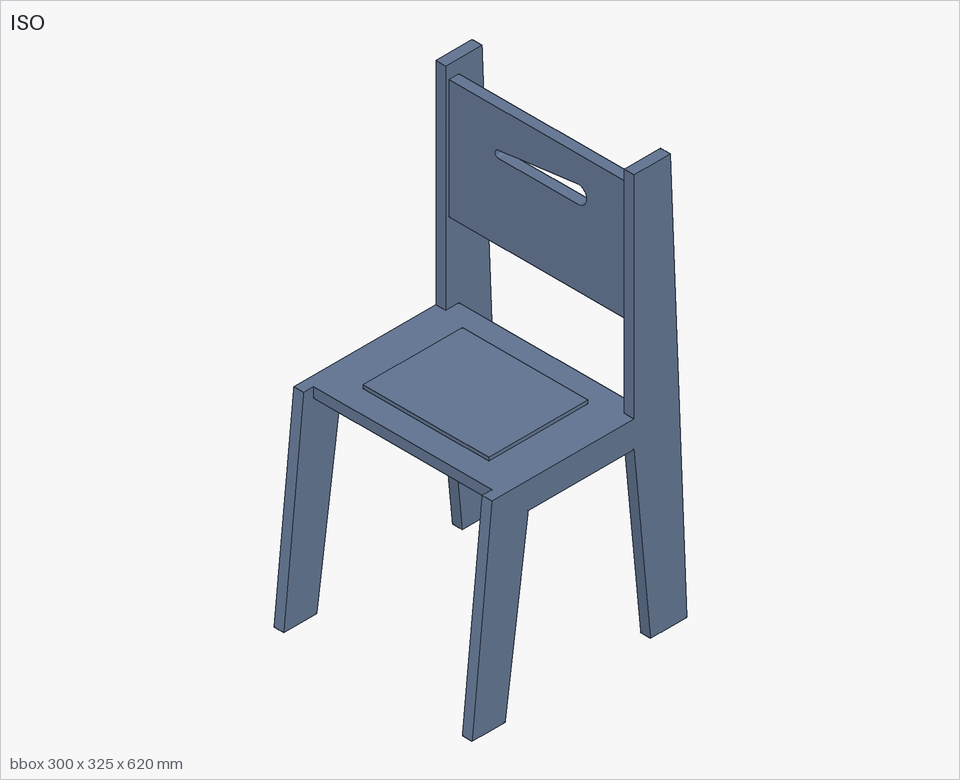}\\[-0.2em]{\scriptsize 67.6}} \\
\texttt{\scriptsize pcb\_23751} & \makecell{\includegraphics[width=\linewidth,height=0.58in,keepaspectratio,valign=c]{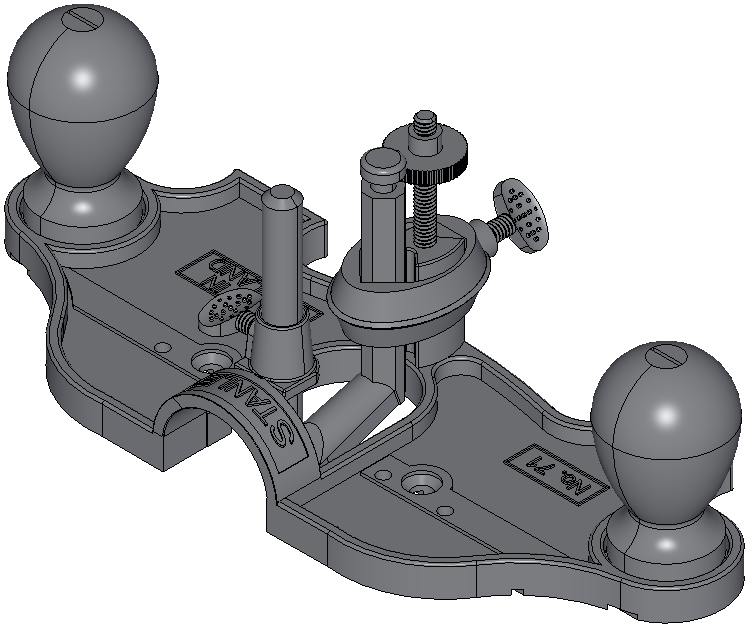}} & \makecell{\includegraphics[width=\linewidth,height=0.58in,keepaspectratio,valign=c]{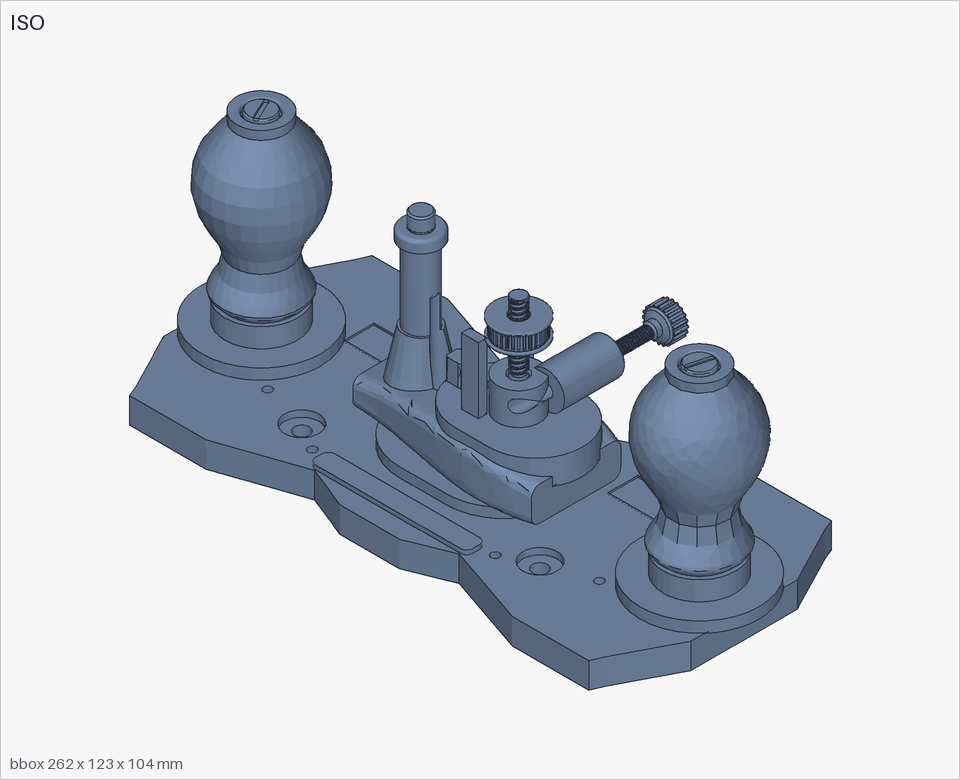}\\[-0.2em]{\scriptsize 74.9}} & \makecell{\includegraphics[width=\linewidth,height=0.58in,keepaspectratio,valign=c]{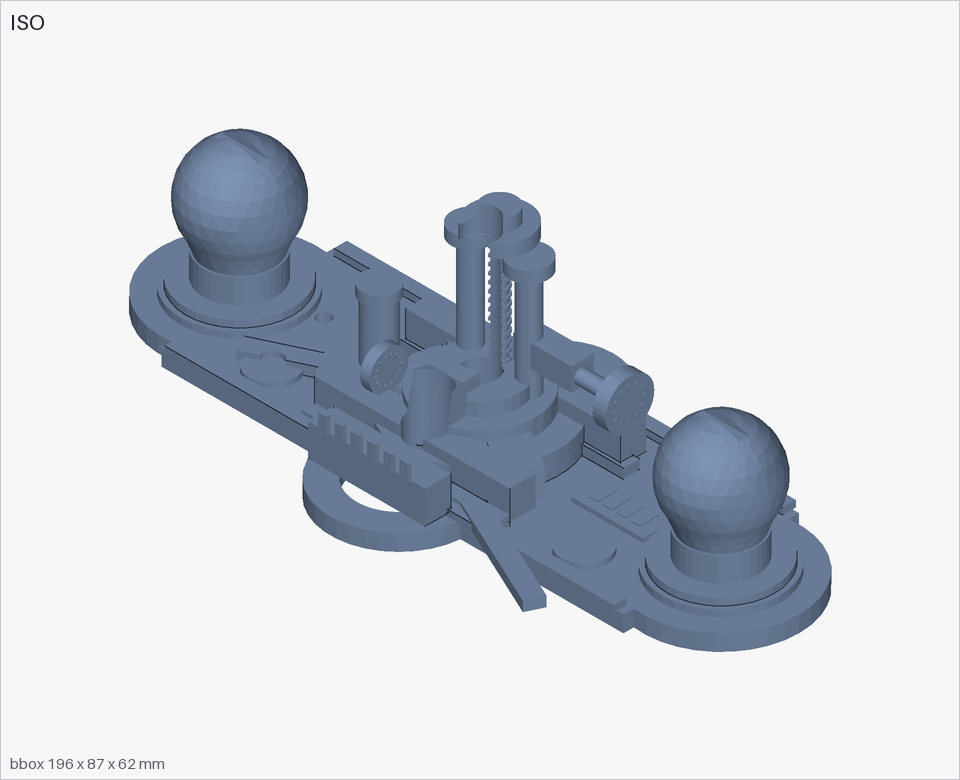}\\[-0.2em]{\scriptsize 74.4}} & \makecell{\includegraphics[width=\linewidth,height=0.58in,keepaspectratio,valign=c]{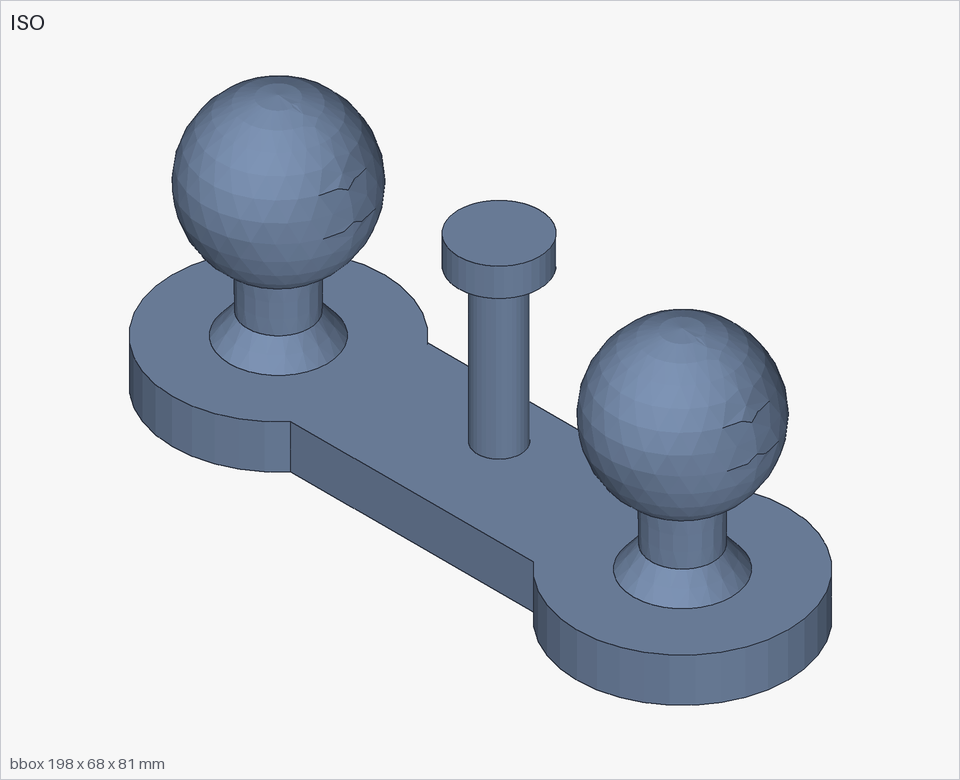}\\[-0.2em]{\scriptsize 40.1}} \\
\texttt{\scriptsize pcb\_24547} & \makecell{\includegraphics[width=\linewidth,height=0.58in,keepaspectratio,valign=c]{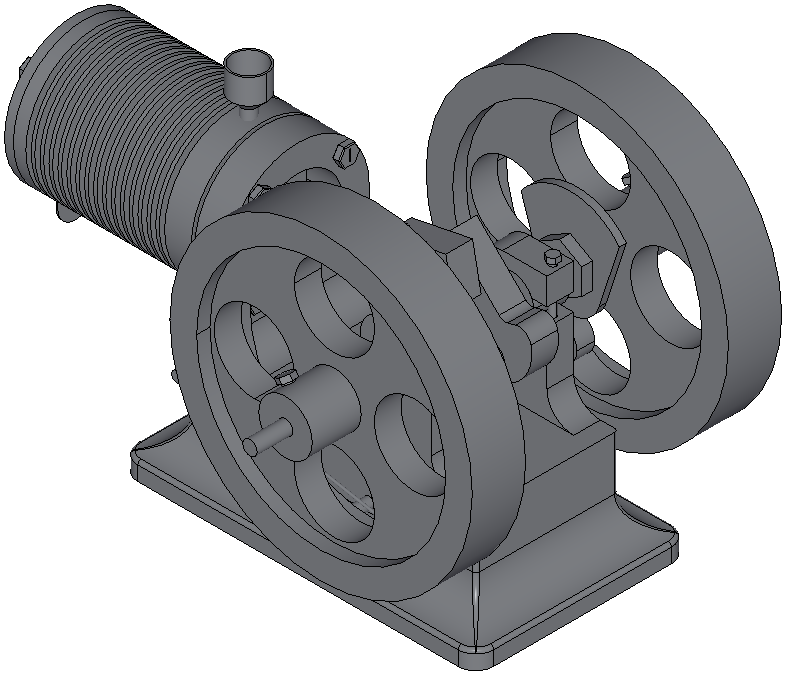}} & \makecell{\includegraphics[width=\linewidth,height=0.58in,keepaspectratio,valign=c]{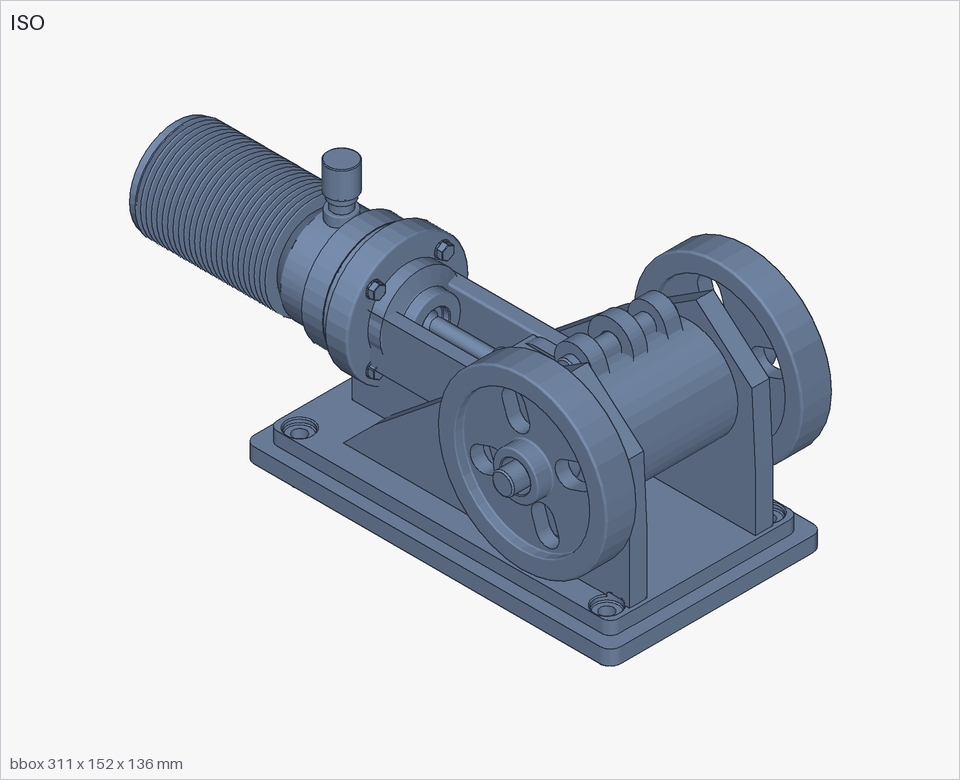}\\[-0.2em]{\scriptsize 83.3}} & \makecell{\includegraphics[width=\linewidth,height=0.58in,keepaspectratio,valign=c]{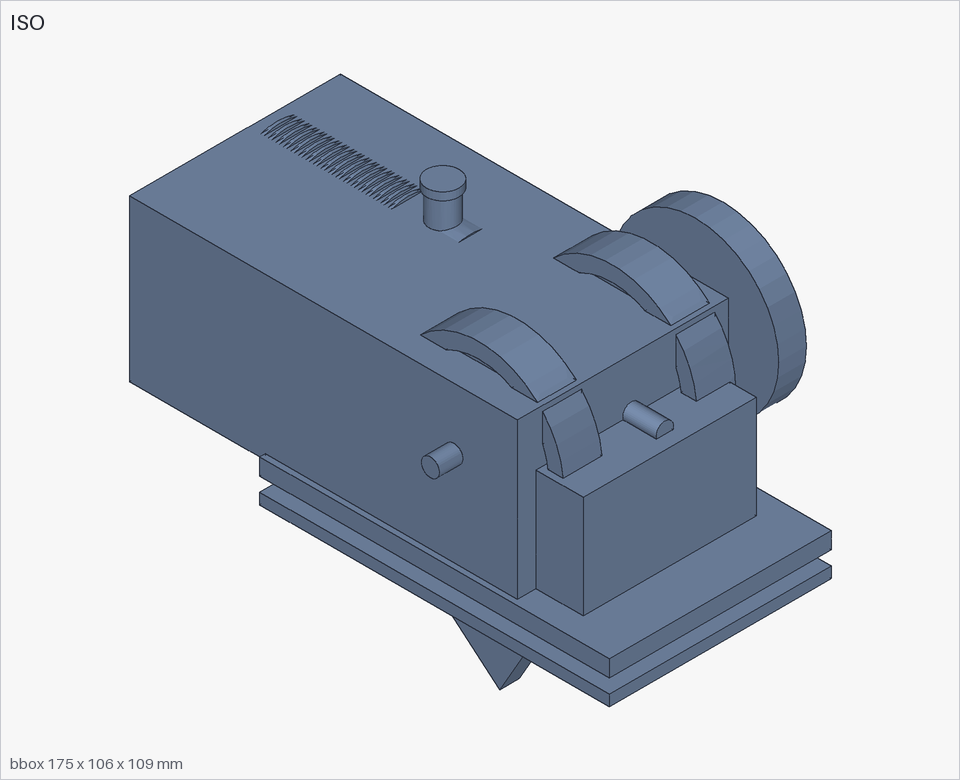}\\[-0.2em]{\scriptsize 46.0}} & \makecell{\includegraphics[width=\linewidth,height=0.58in,keepaspectratio,valign=c]{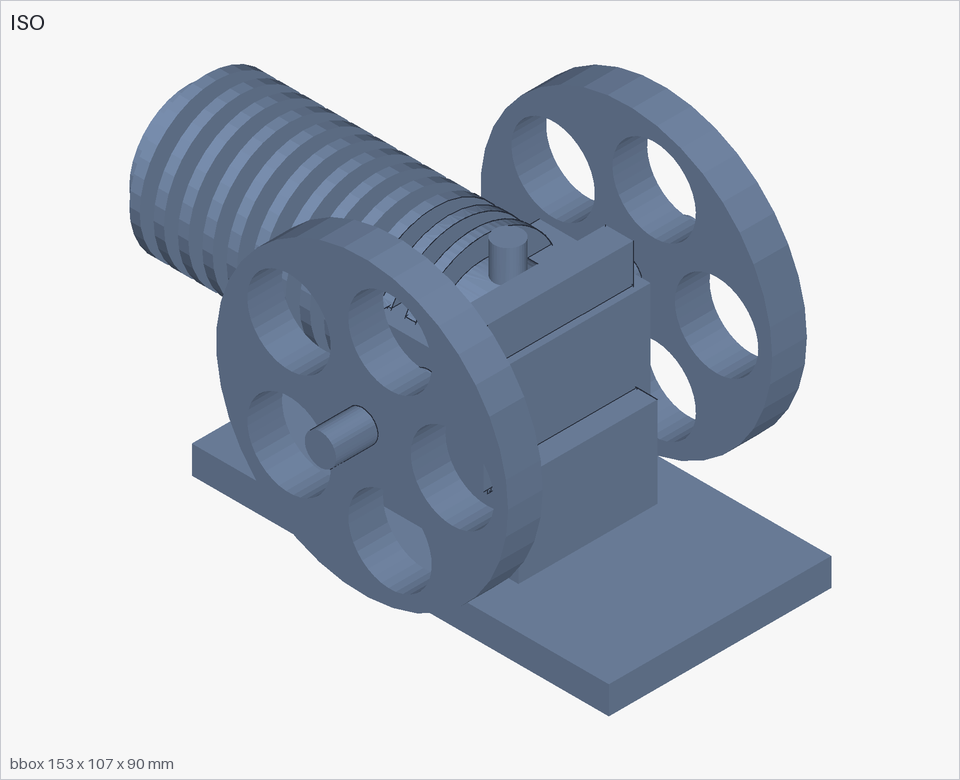}\\[-0.2em]{\scriptsize 73.1}} \\
\texttt{\scriptsize pcb\_25464} & \makecell{\includegraphics[width=\linewidth,height=0.58in,keepaspectratio,valign=c]{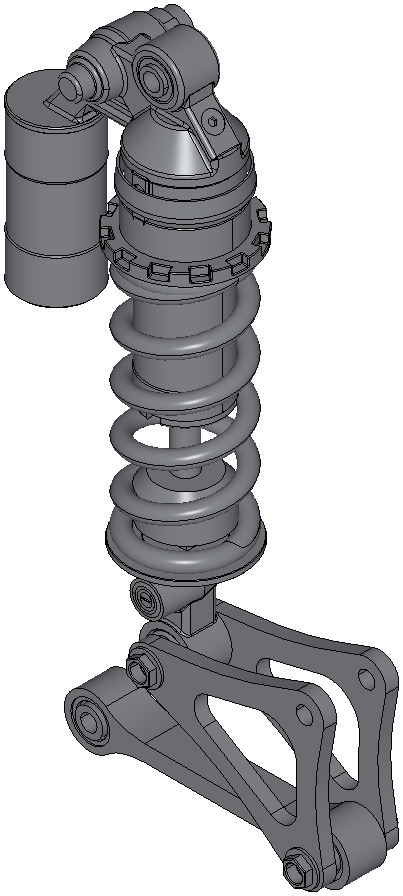}} & \makecell{\includegraphics[width=\linewidth,height=0.58in,keepaspectratio,valign=c]{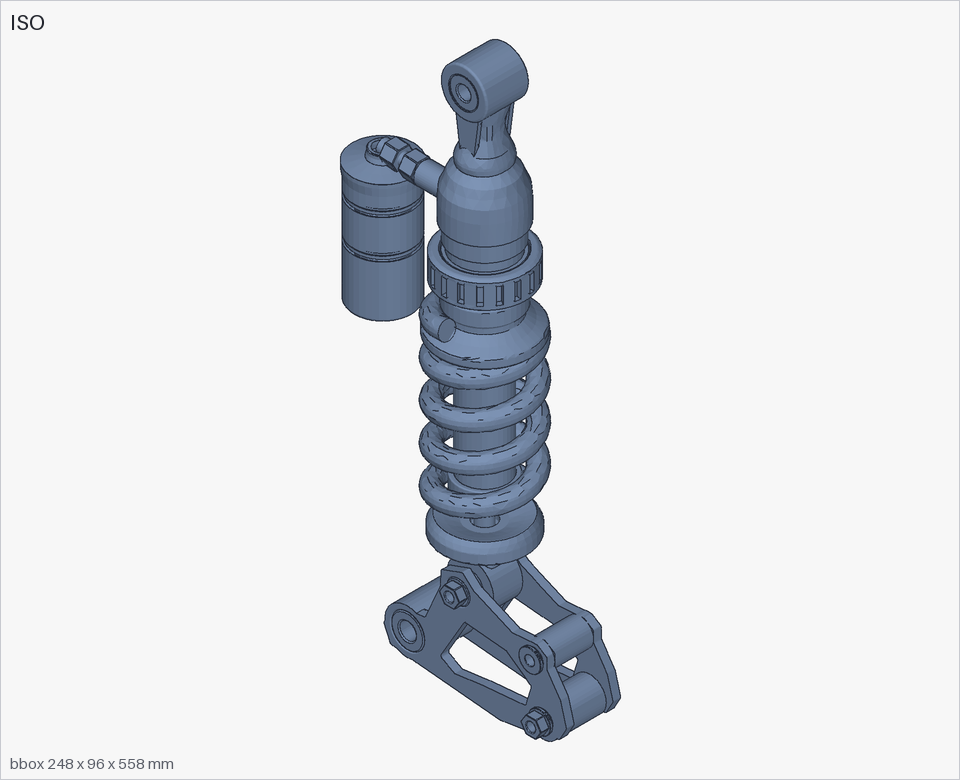}\\[-0.2em]{\scriptsize 83.6}} & \makecell{\includegraphics[width=\linewidth,height=0.58in,keepaspectratio,valign=c]{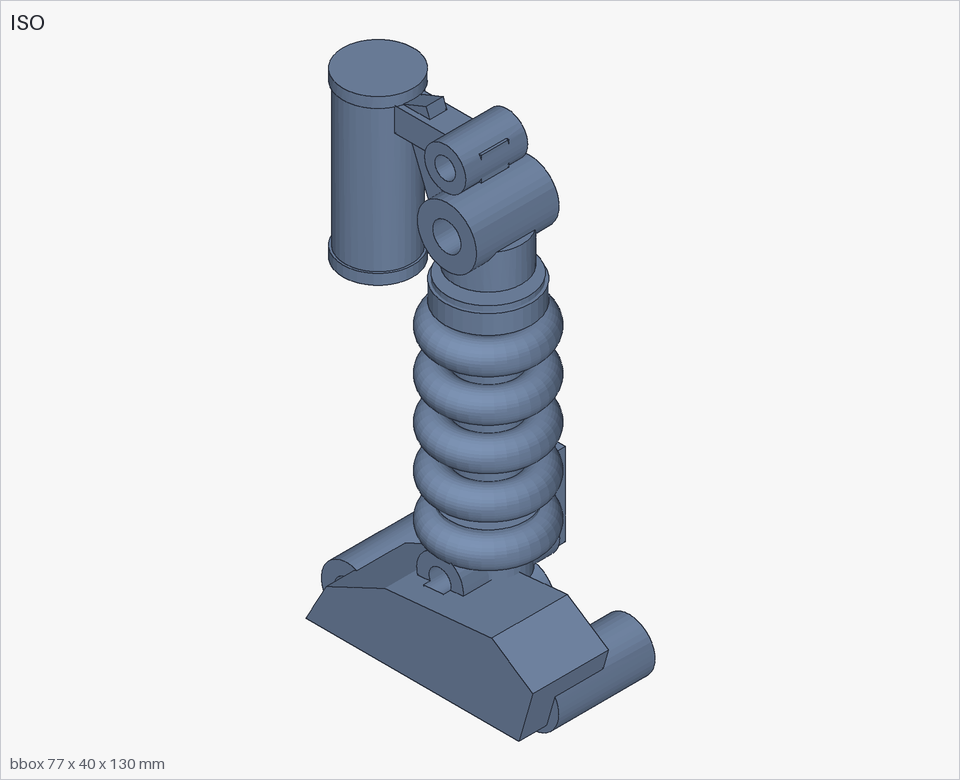}\\[-0.2em]{\scriptsize 65.7}} & \makecell{\includegraphics[width=\linewidth,height=0.58in,keepaspectratio,valign=c]{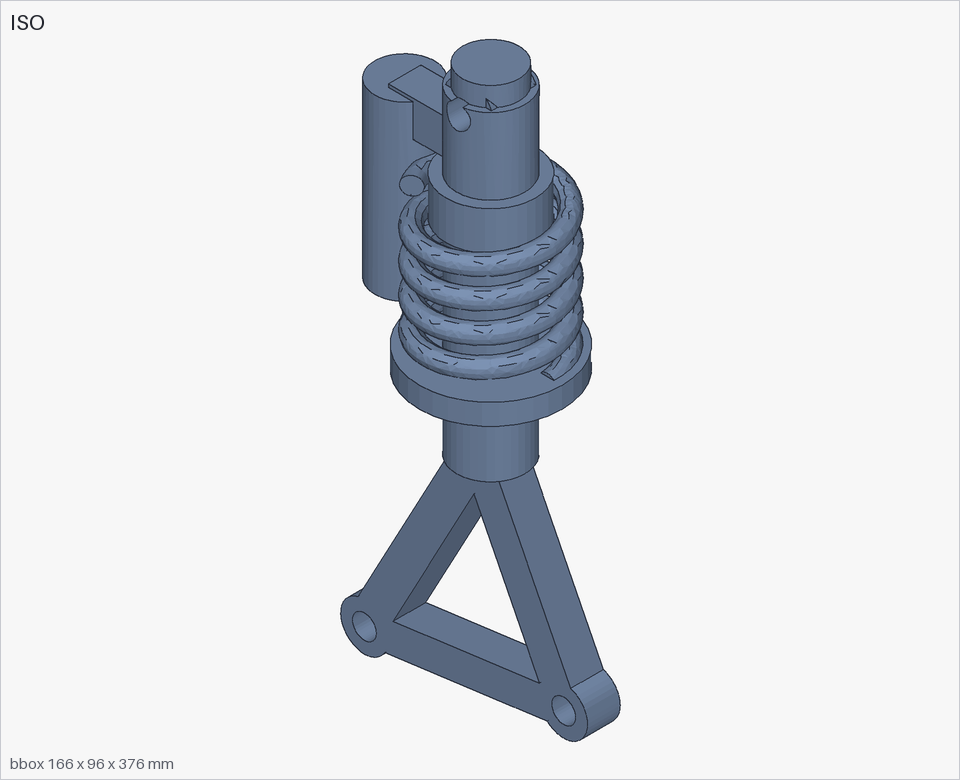}\\[-0.2em]{\scriptsize 63.8}} \\
\texttt{\scriptsize pcb\_31455} & \makecell{\includegraphics[width=\linewidth,height=0.58in,keepaspectratio,valign=c]{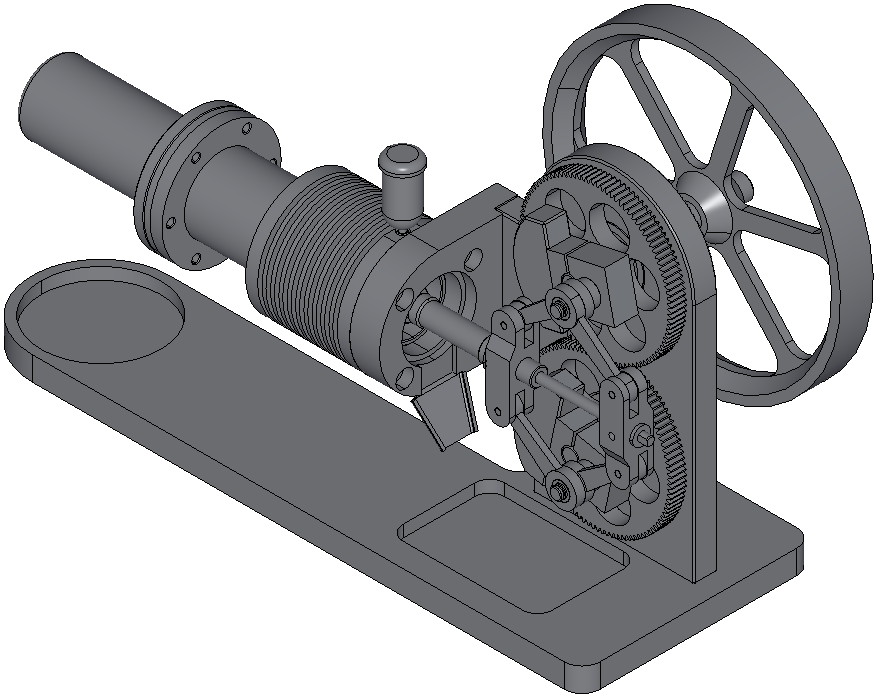}} & \makecell{\includegraphics[width=\linewidth,height=0.58in,keepaspectratio,valign=c]{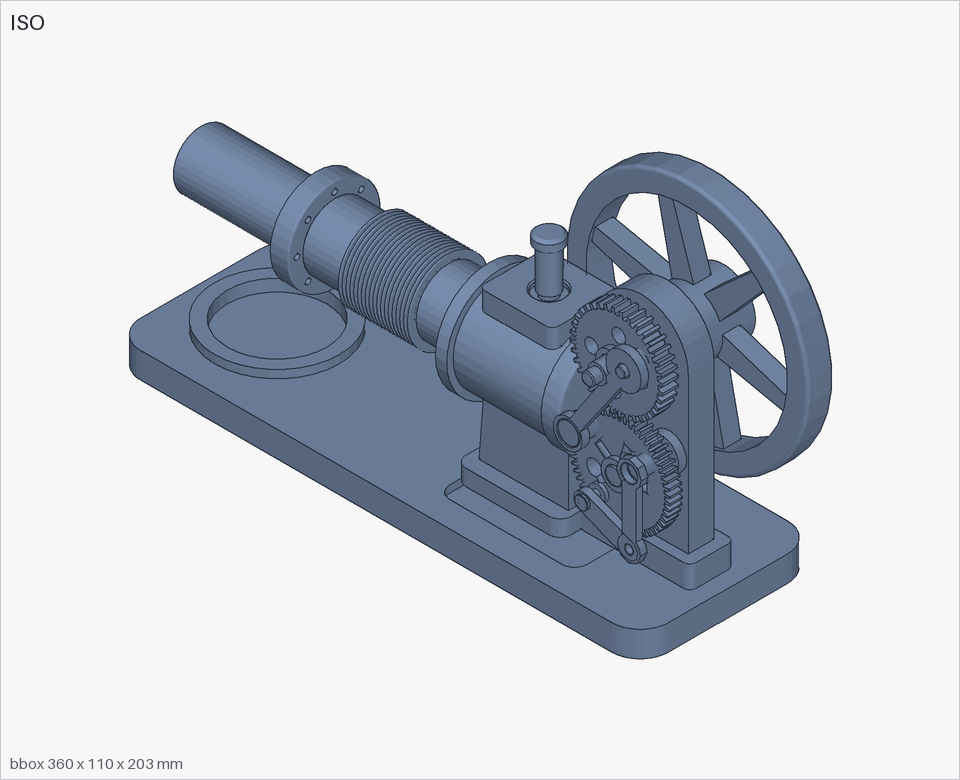}\\[-0.2em]{\scriptsize 87.1}} & \makecell{\includegraphics[width=\linewidth,height=0.58in,keepaspectratio,valign=c]{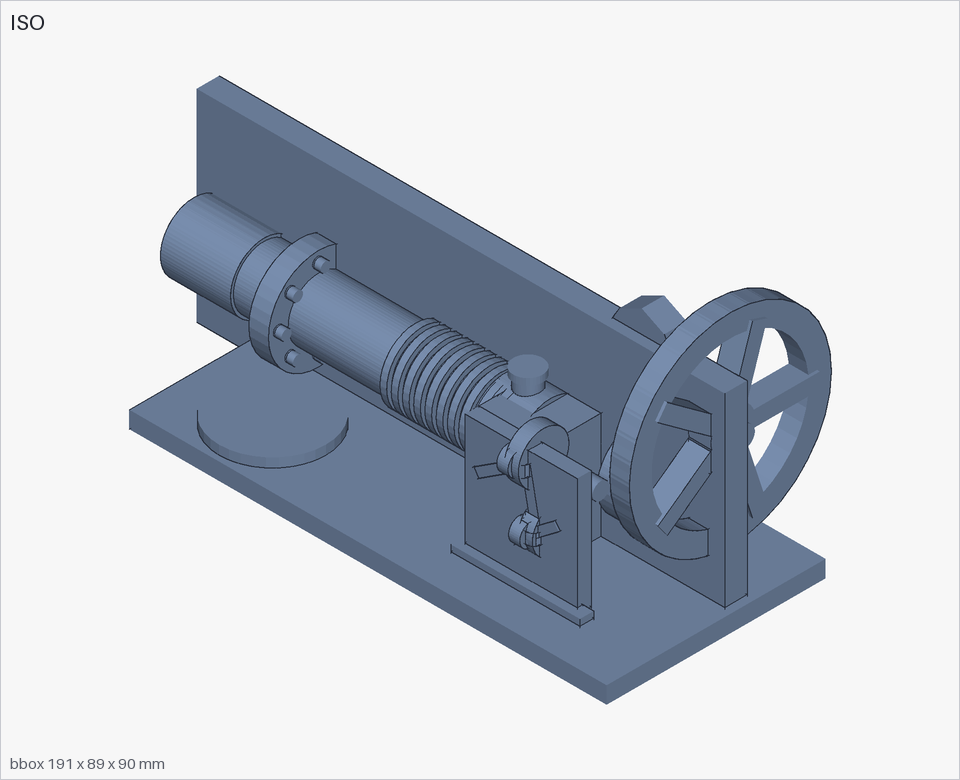}\\[-0.2em]{\scriptsize 75.3}} & \makecell{\includegraphics[width=\linewidth,height=0.58in,keepaspectratio,valign=c]{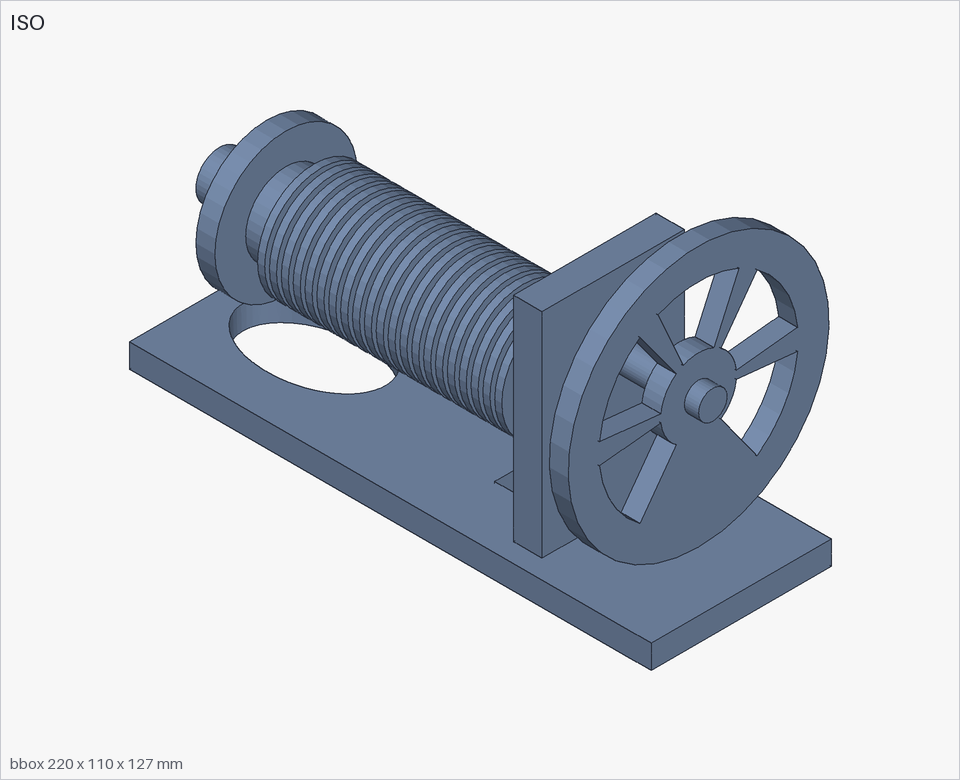}\\[-0.2em]{\scriptsize 52.4}} \\
\texttt{\scriptsize pcb\_31845} & \makecell{\includegraphics[width=\linewidth,height=0.58in,keepaspectratio,valign=c]{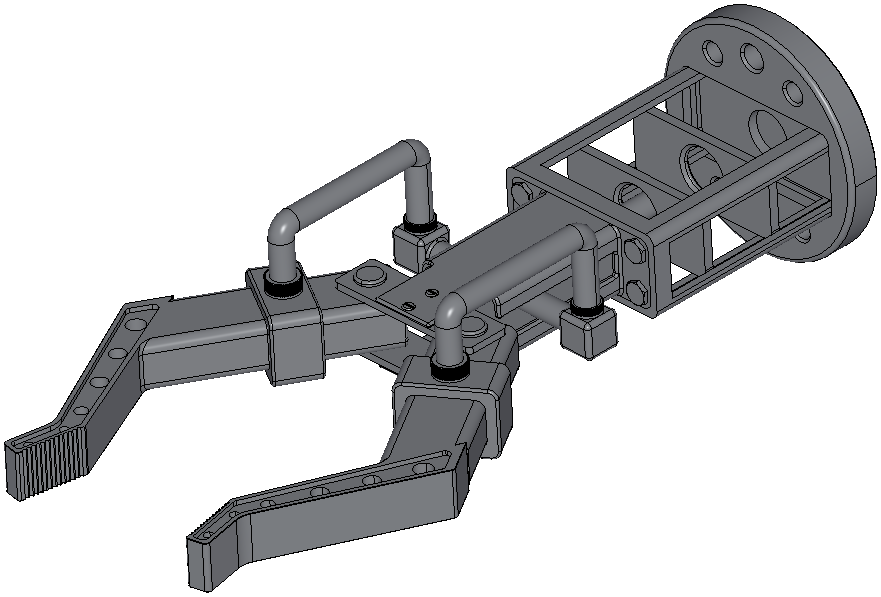}} & \makecell{\includegraphics[width=\linewidth,height=0.58in,keepaspectratio,valign=c]{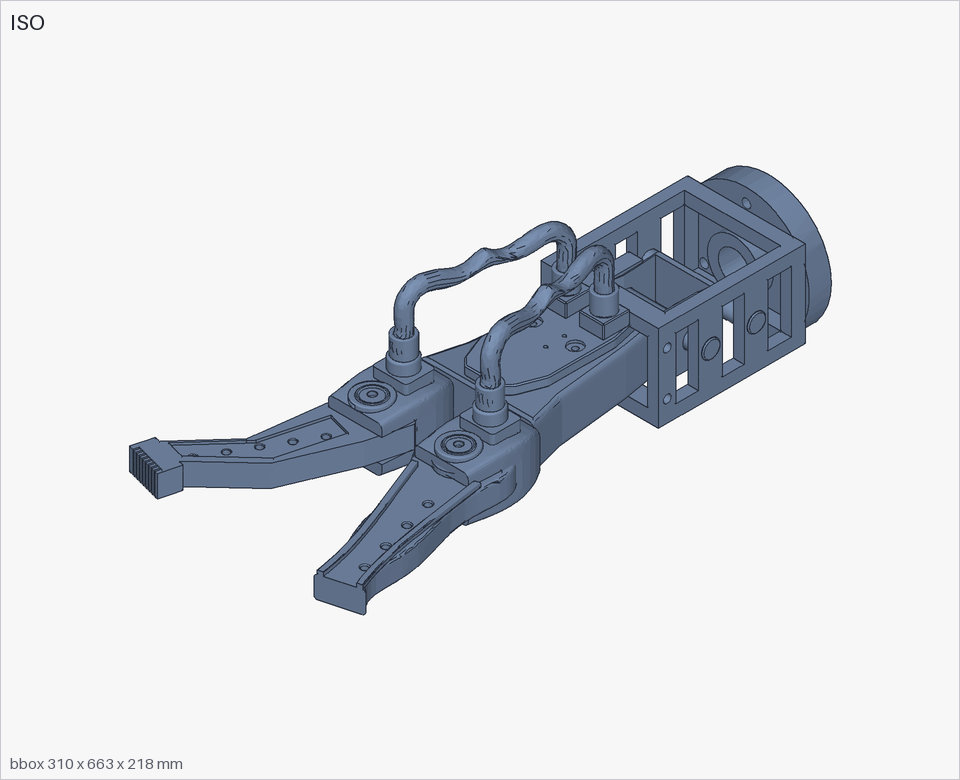}\\[-0.2em]{\scriptsize 85.4}} & \makecell{\includegraphics[width=\linewidth,height=0.58in,keepaspectratio,valign=c]{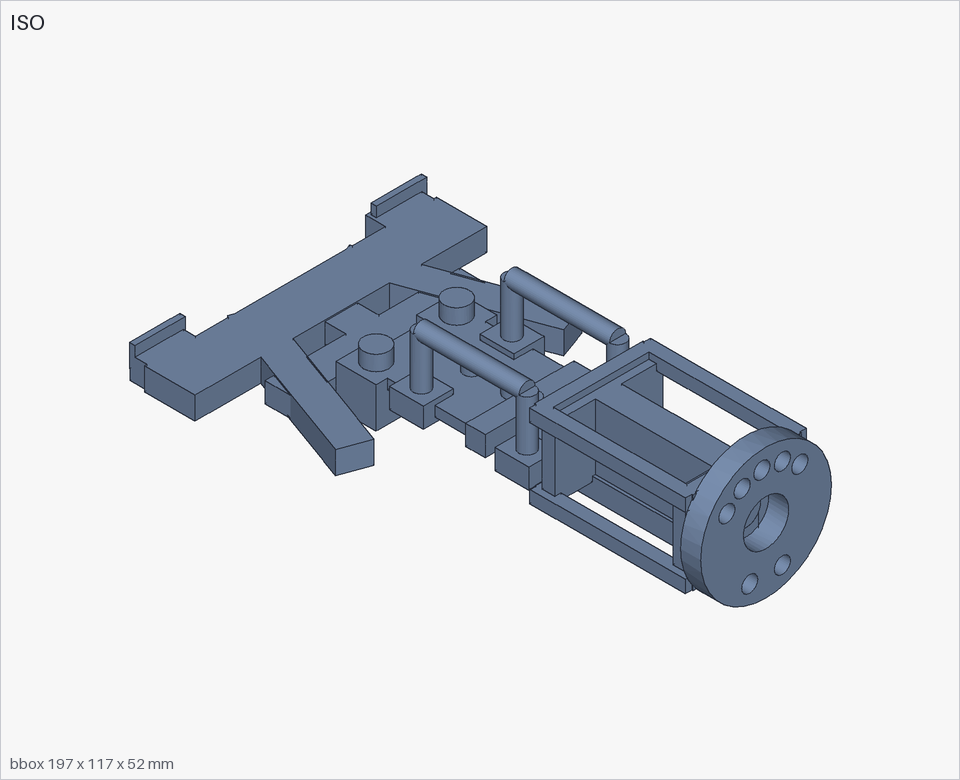}\\[-0.2em]{\scriptsize 75.5}} & \makecell{\includegraphics[width=\linewidth,height=0.58in,keepaspectratio,valign=c]{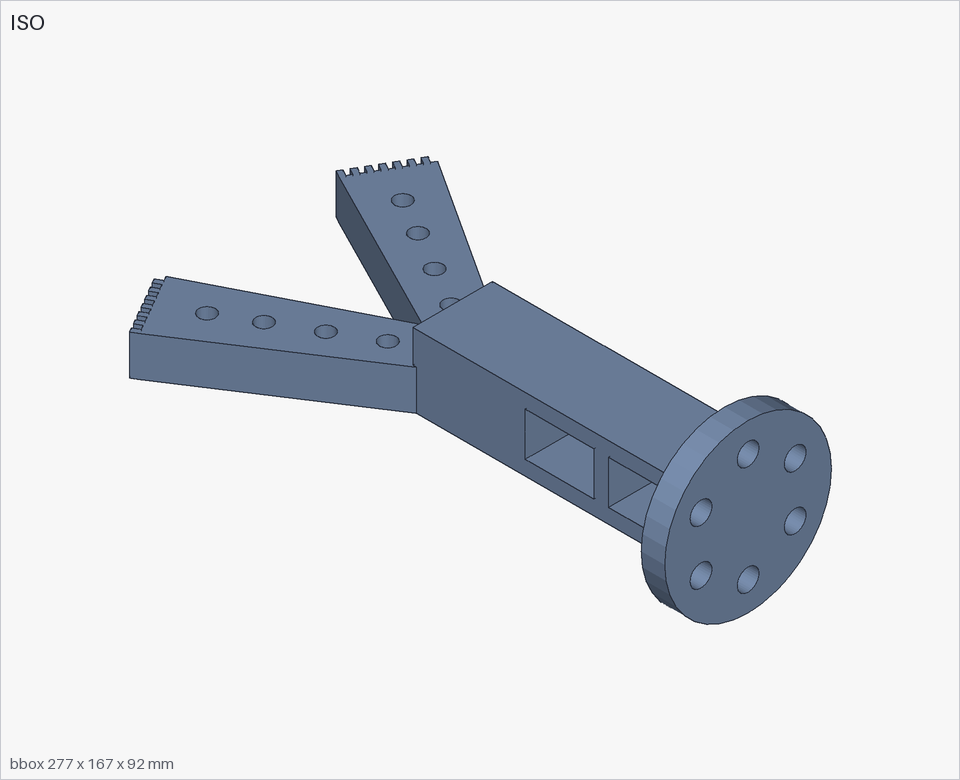}\\[-0.2em]{\scriptsize 50.7}} \\
\texttt{\scriptsize pcb\_34587} & \makecell{\includegraphics[width=\linewidth,height=0.58in,keepaspectratio,valign=c]{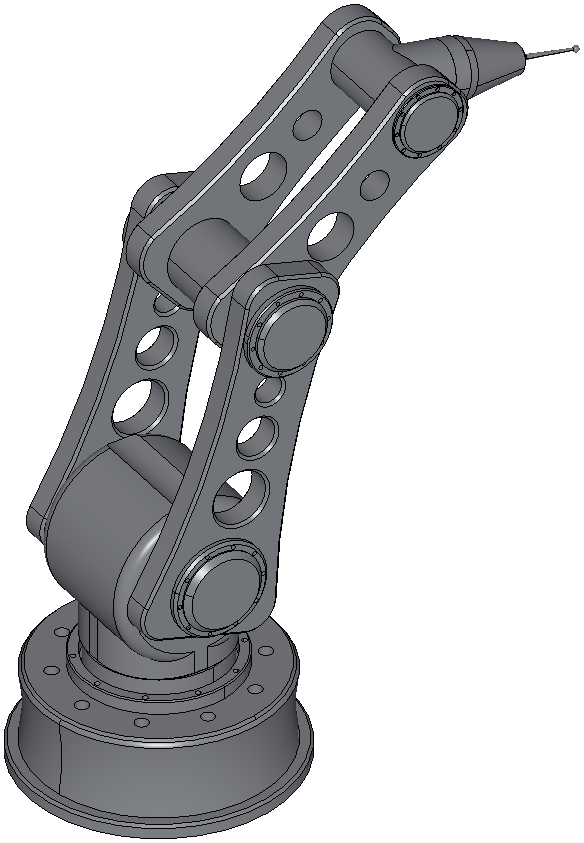}} & \makecell{\includegraphics[width=\linewidth,height=0.58in,keepaspectratio,valign=c]{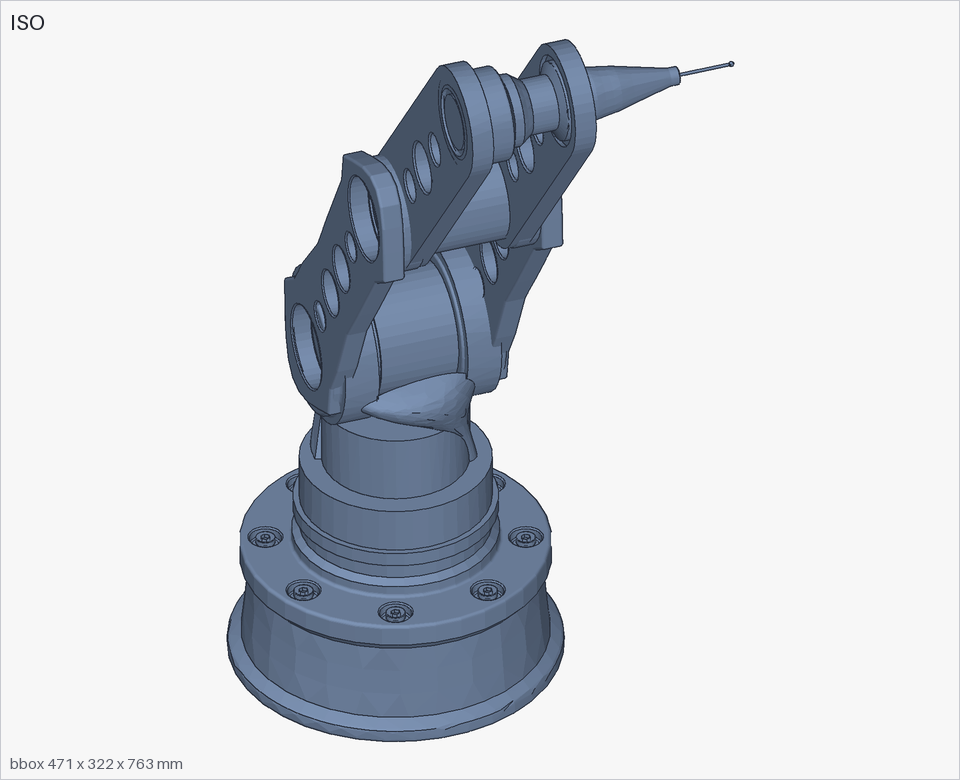}\\[-0.2em]{\scriptsize 87.0}} & \makecell{\includegraphics[width=\linewidth,height=0.58in,keepaspectratio,valign=c]{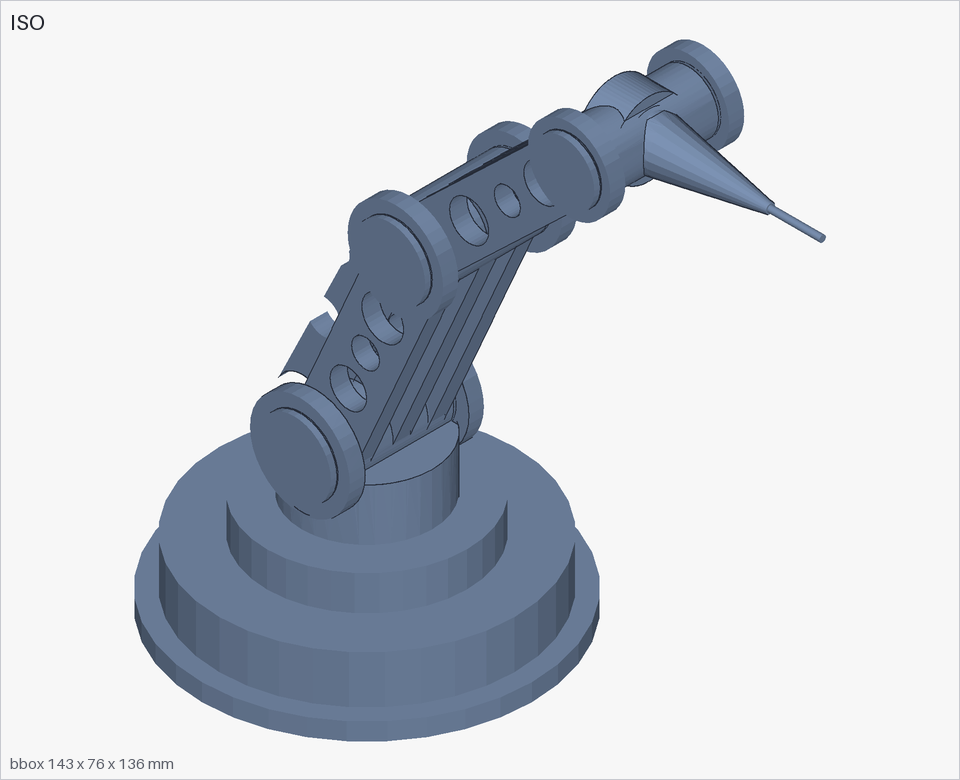}\\[-0.2em]{\scriptsize 77.3}} & \makecell{\includegraphics[width=\linewidth,height=0.58in,keepaspectratio,valign=c]{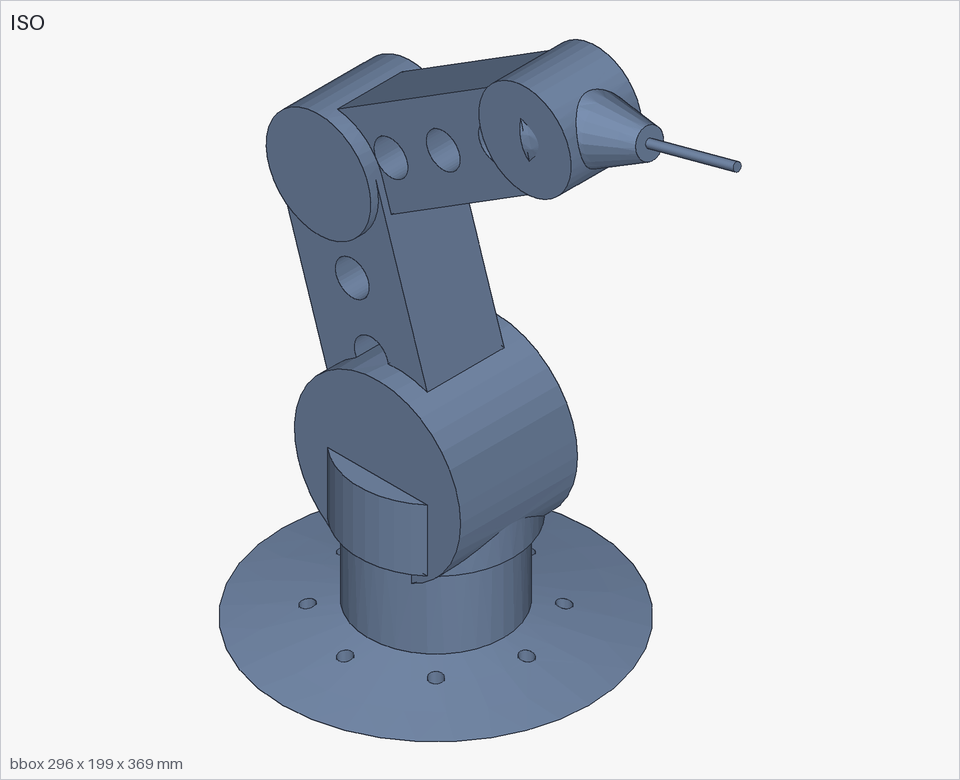}\\[-0.2em]{\scriptsize 70.3}} \\
\texttt{\scriptsize pcb\_52890} & \makecell{\includegraphics[width=\linewidth,height=0.58in,keepaspectratio,valign=c]{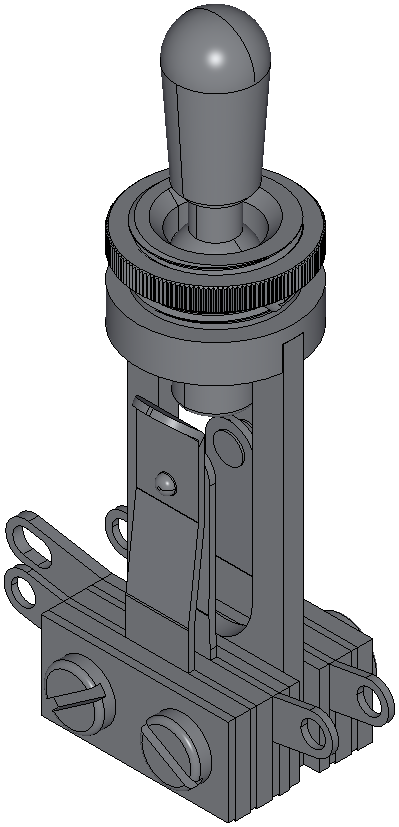}} & \makecell{\includegraphics[width=\linewidth,height=0.58in,keepaspectratio,valign=c]{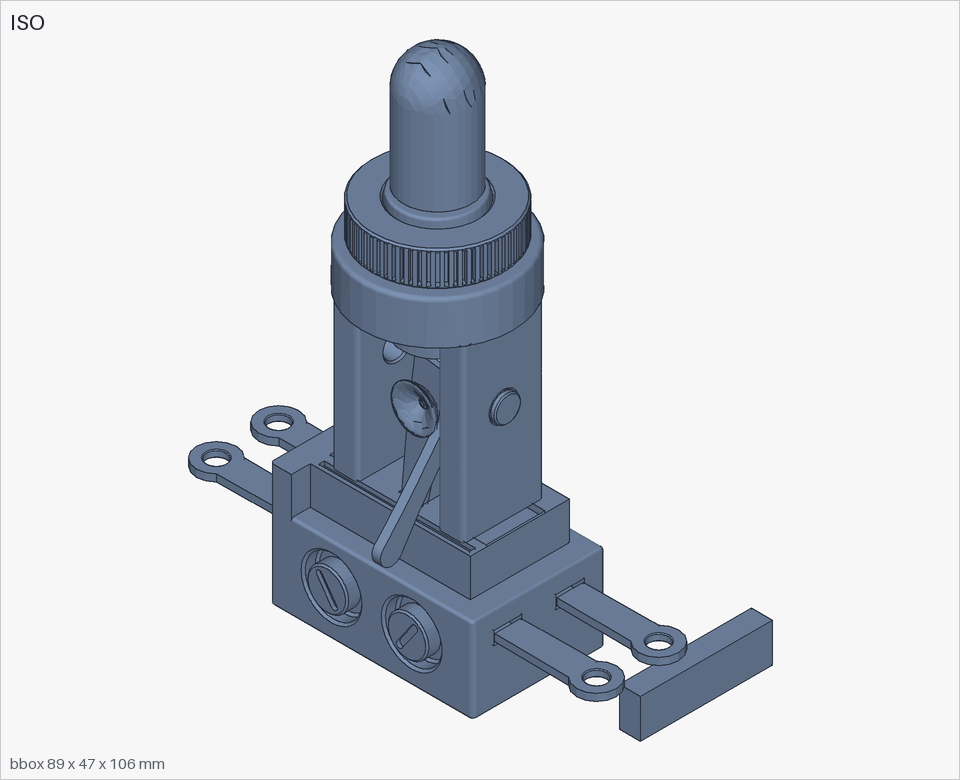}\\[-0.2em]{\scriptsize 84.9}} & \makecell{\includegraphics[width=\linewidth,height=0.58in,keepaspectratio,valign=c]{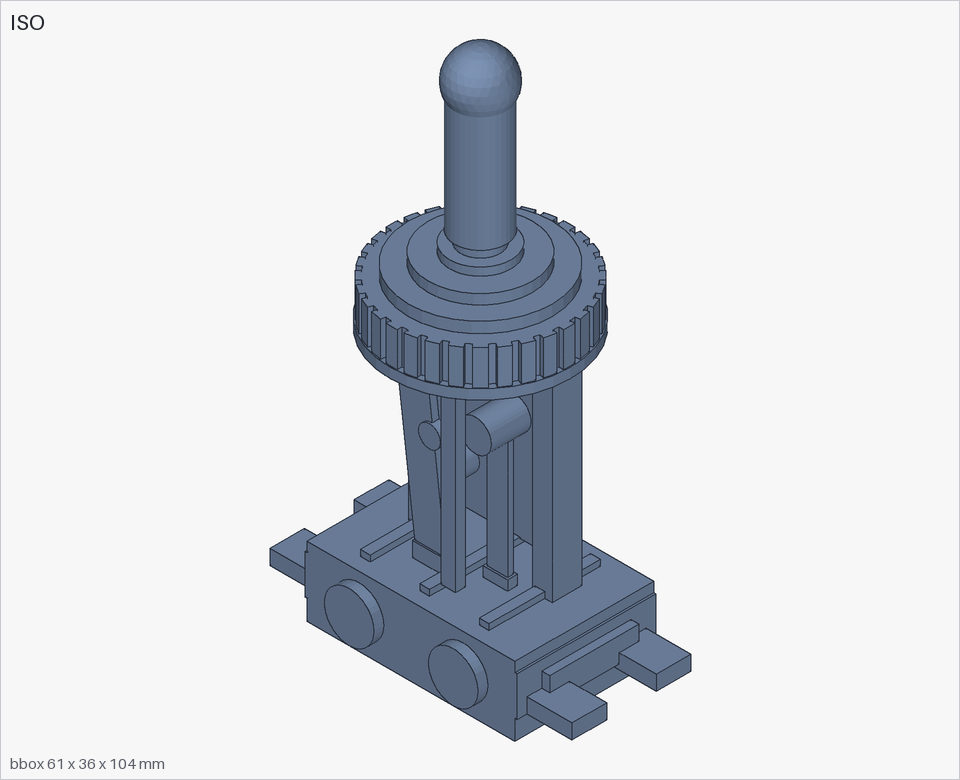}\\[-0.2em]{\scriptsize 81.9}} & \makecell{\includegraphics[width=\linewidth,height=0.58in,keepaspectratio,valign=c]{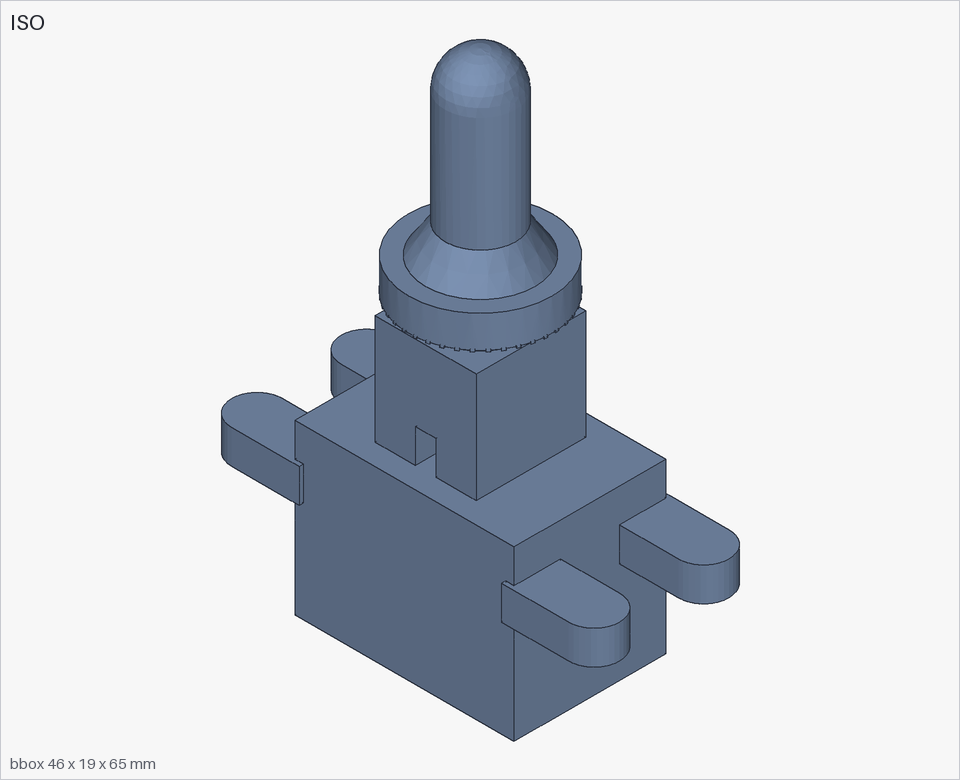}\\[-0.2em]{\scriptsize 55.2}} \\
\texttt{\scriptsize pcb\_69217} & \makecell{\includegraphics[width=\linewidth,height=0.58in,keepaspectratio,valign=c]{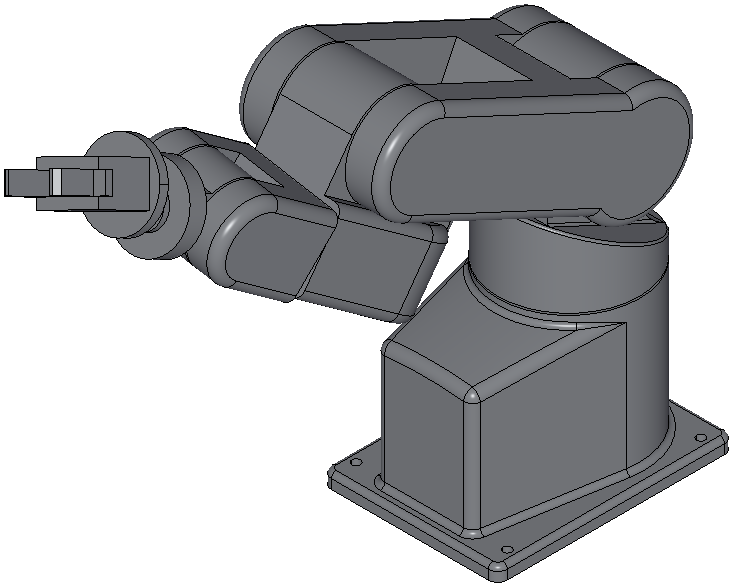}} & \makecell{\includegraphics[width=\linewidth,height=0.58in,keepaspectratio,valign=c]{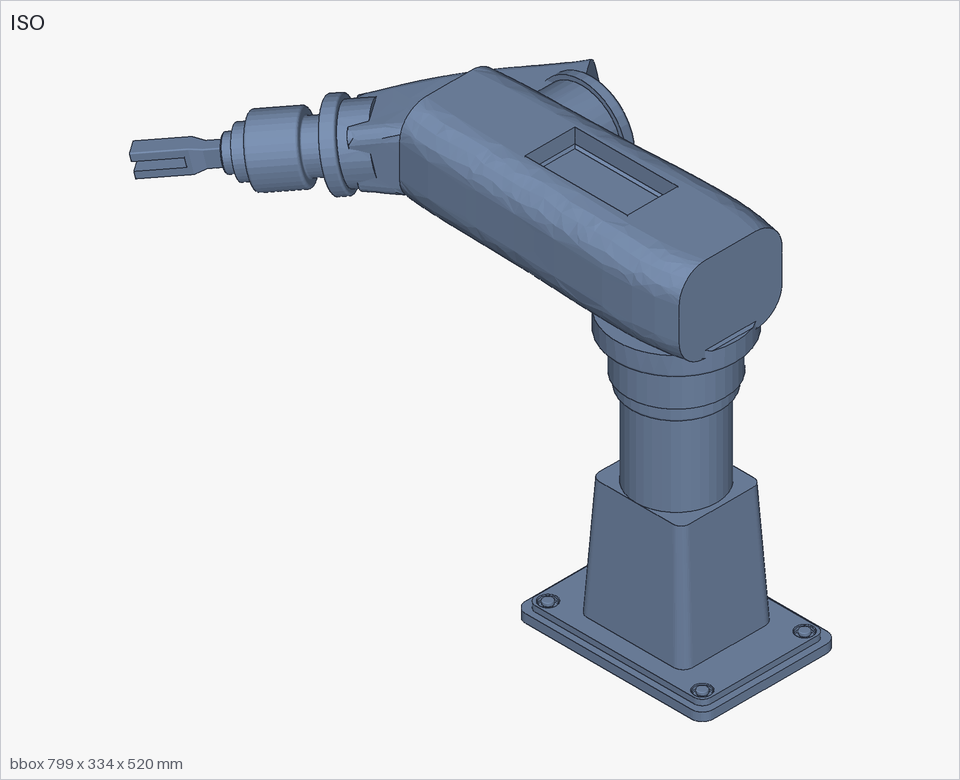}\\[-0.2em]{\scriptsize 81.5}} & \makecell{\includegraphics[width=\linewidth,height=0.58in,keepaspectratio,valign=c]{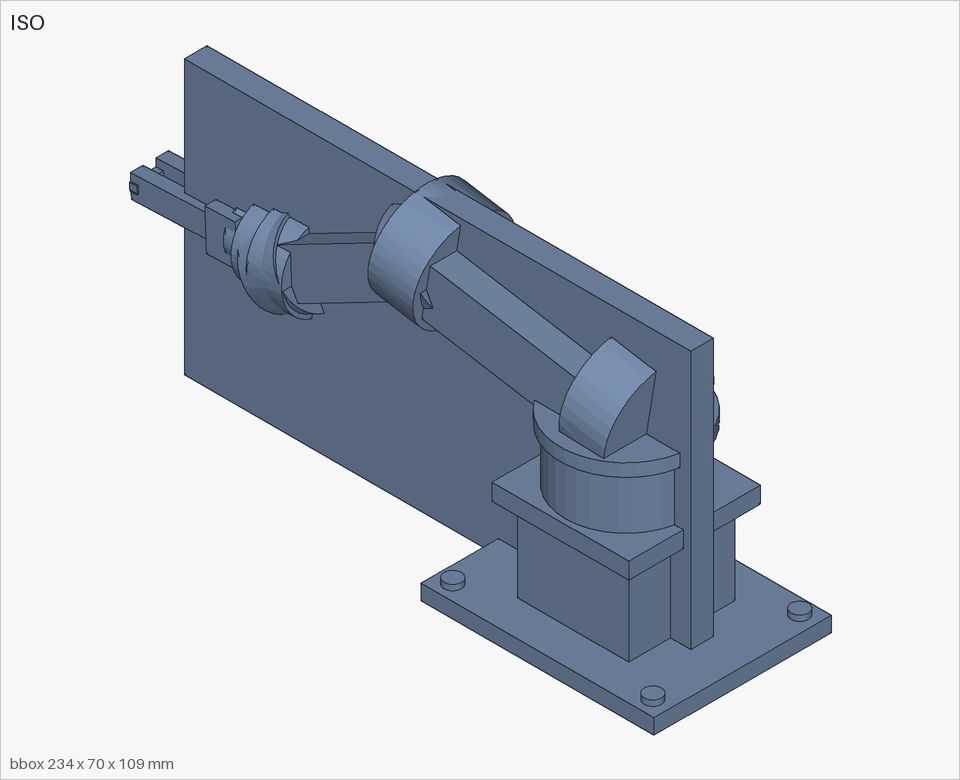}\\[-0.2em]{\scriptsize 74.0}} & \makecell{\includegraphics[width=\linewidth,height=0.58in,keepaspectratio,valign=c]{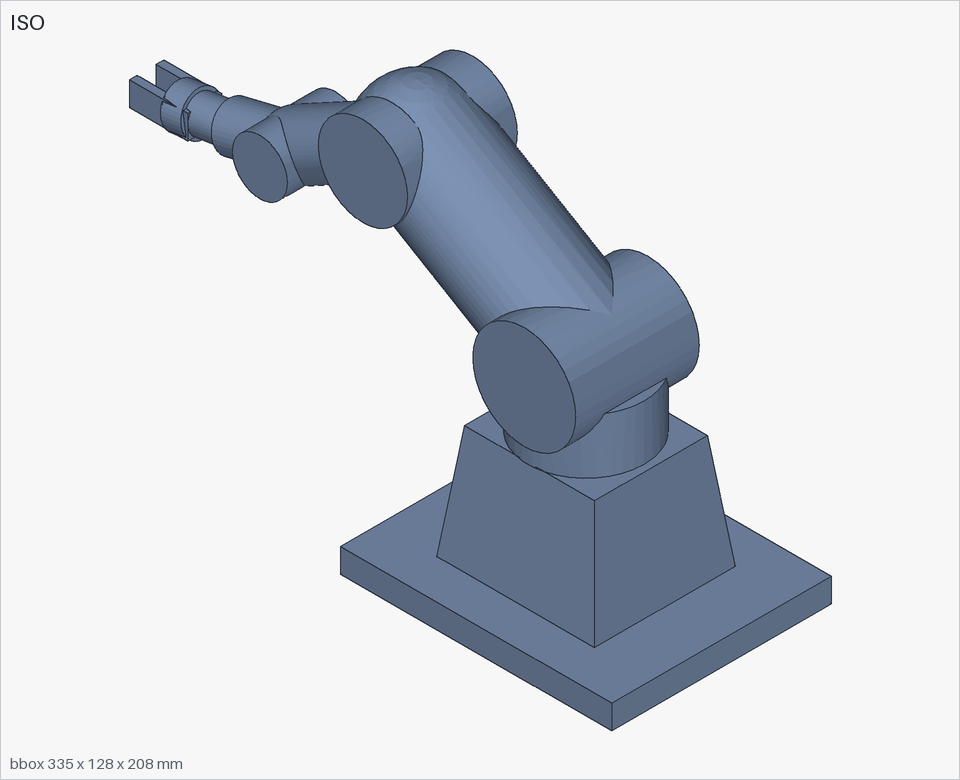}\\[-0.2em]{\scriptsize 70.2}} \\
\texttt{\scriptsize pcb\_85436} & \makecell{\includegraphics[width=\linewidth,height=0.58in,keepaspectratio,valign=c]{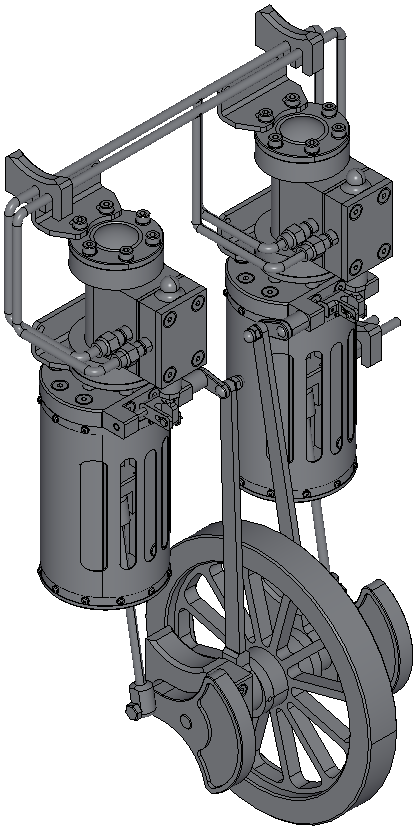}} & \makecell{\includegraphics[width=\linewidth,height=0.58in,keepaspectratio,valign=c]{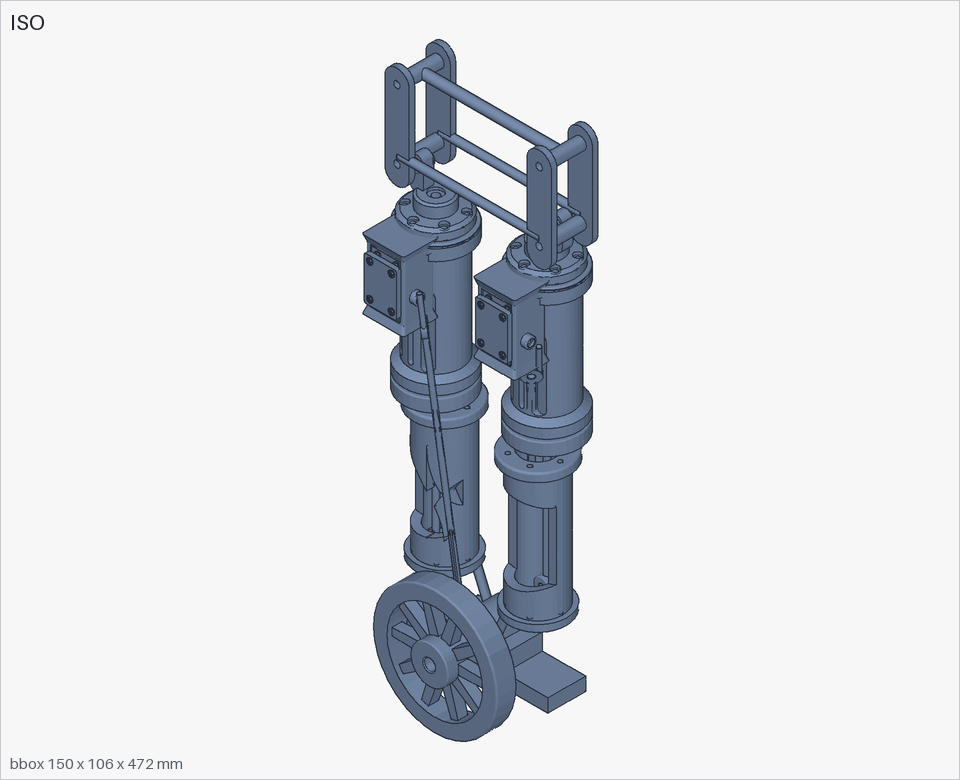}\\[-0.2em]{\scriptsize 78.4}} & \makecell{\includegraphics[width=\linewidth,height=0.58in,keepaspectratio,valign=c]{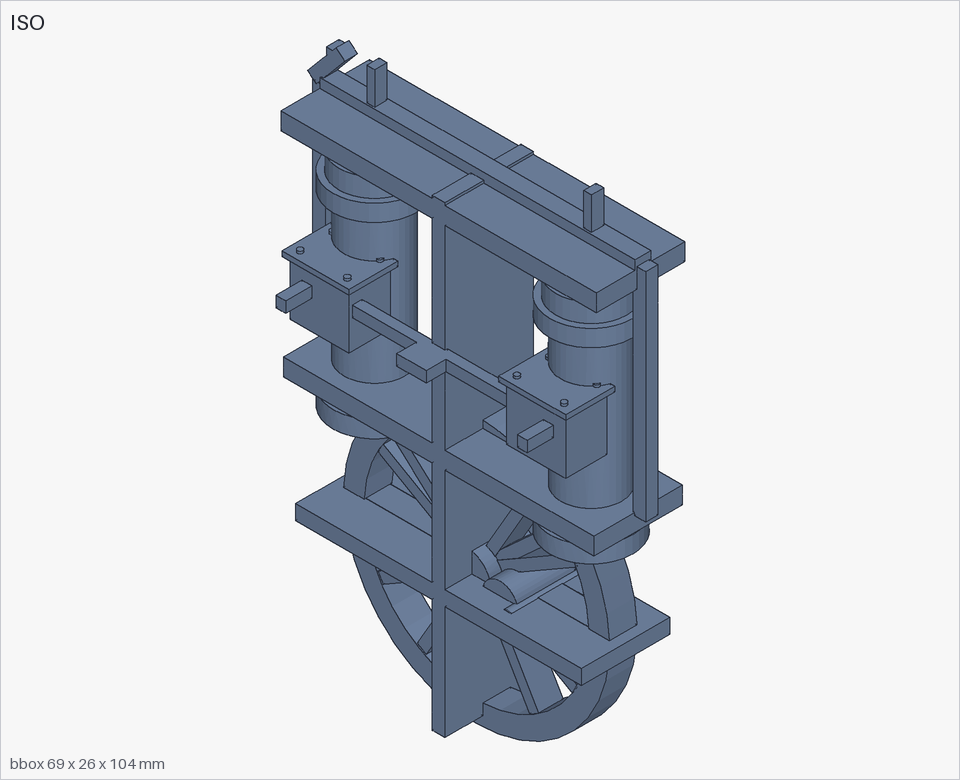}\\[-0.2em]{\scriptsize 67.0}} & \makecell{\includegraphics[width=\linewidth,height=0.58in,keepaspectratio,valign=c]{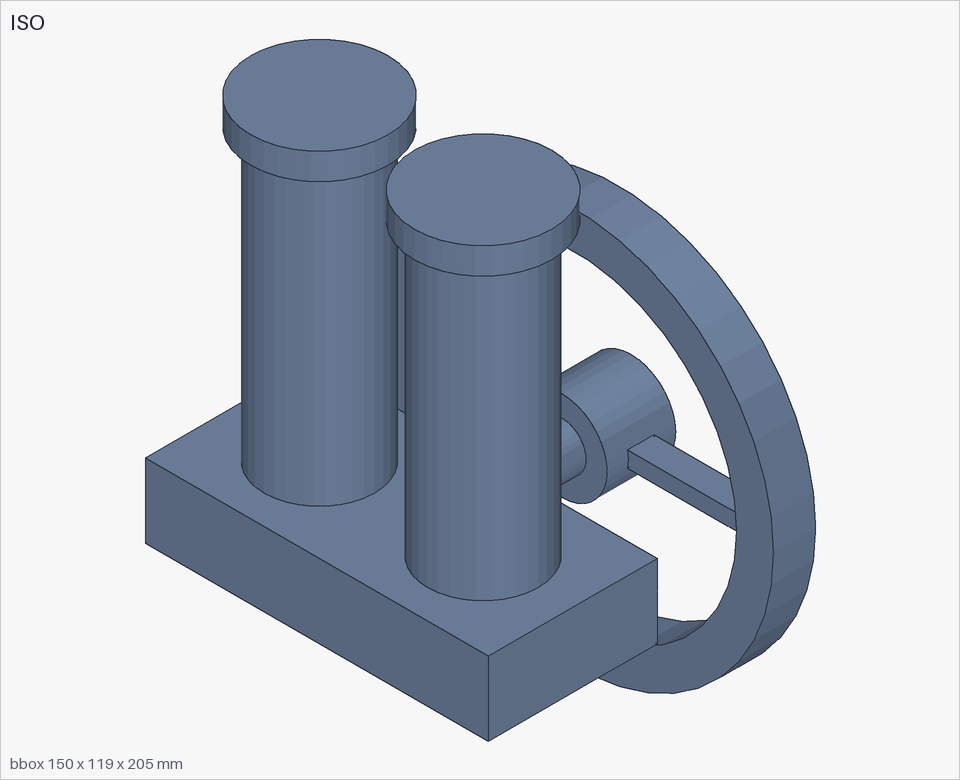}\\[-0.2em]{\scriptsize 41.4}} \\
\texttt{\scriptsize rcb\_000052132} & \makecell{\includegraphics[width=\linewidth,height=0.58in,keepaspectratio,valign=c]{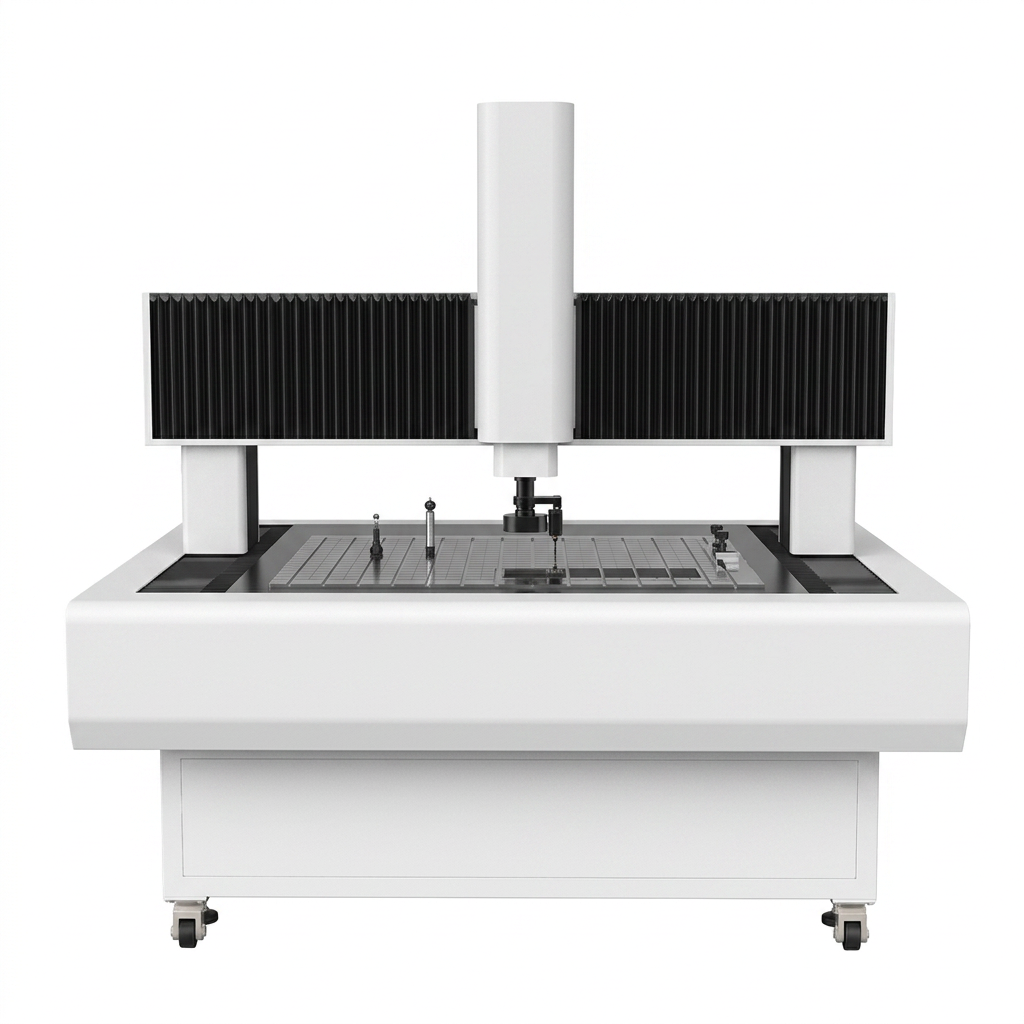}} & \makecell{\includegraphics[width=\linewidth,height=0.58in,keepaspectratio,valign=c]{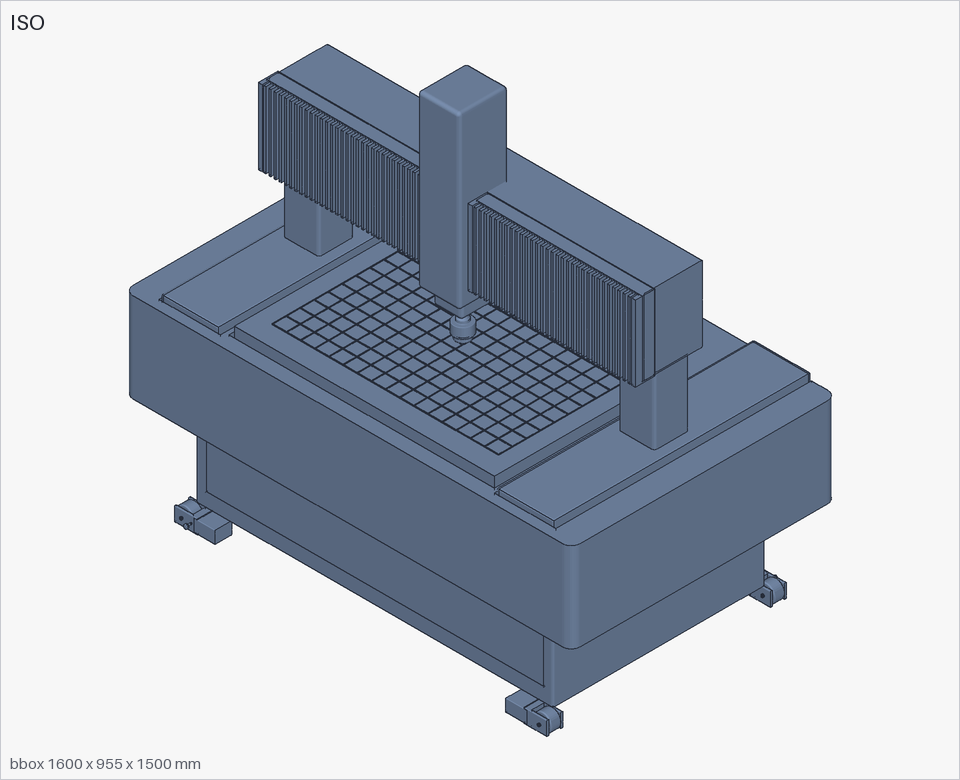}\\[-0.2em]{\scriptsize 86.1}} & \makecell{\includegraphics[width=\linewidth,height=0.58in,keepaspectratio,valign=c]{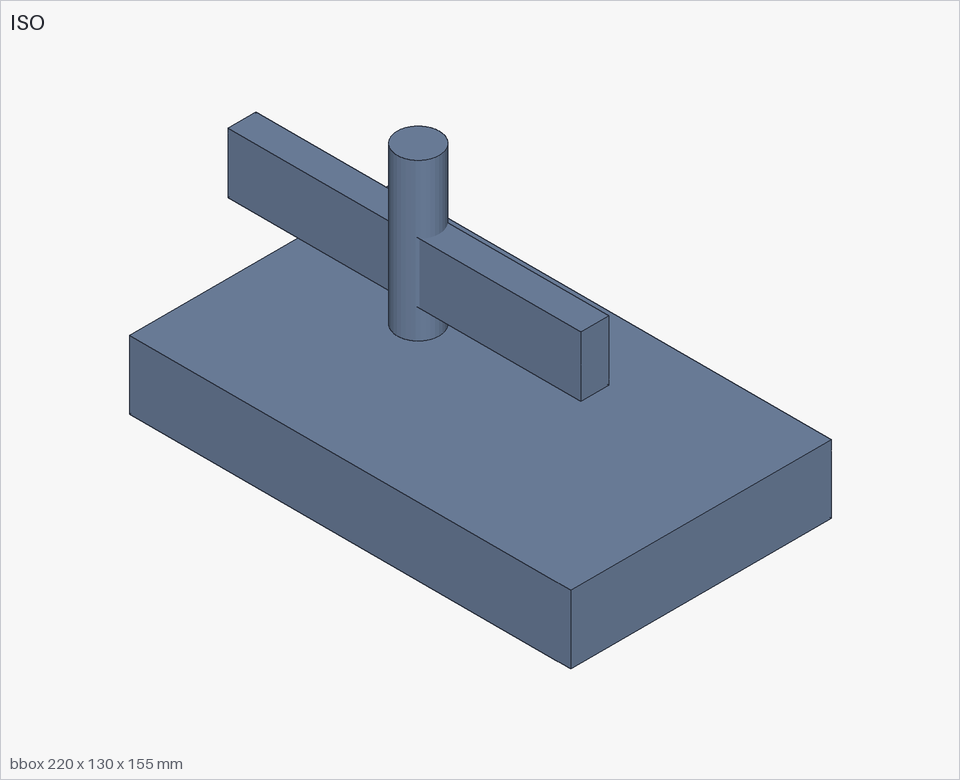}\\[-0.2em]{\scriptsize 30.6}} & \makecell{\includegraphics[width=\linewidth,height=0.58in,keepaspectratio,valign=c]{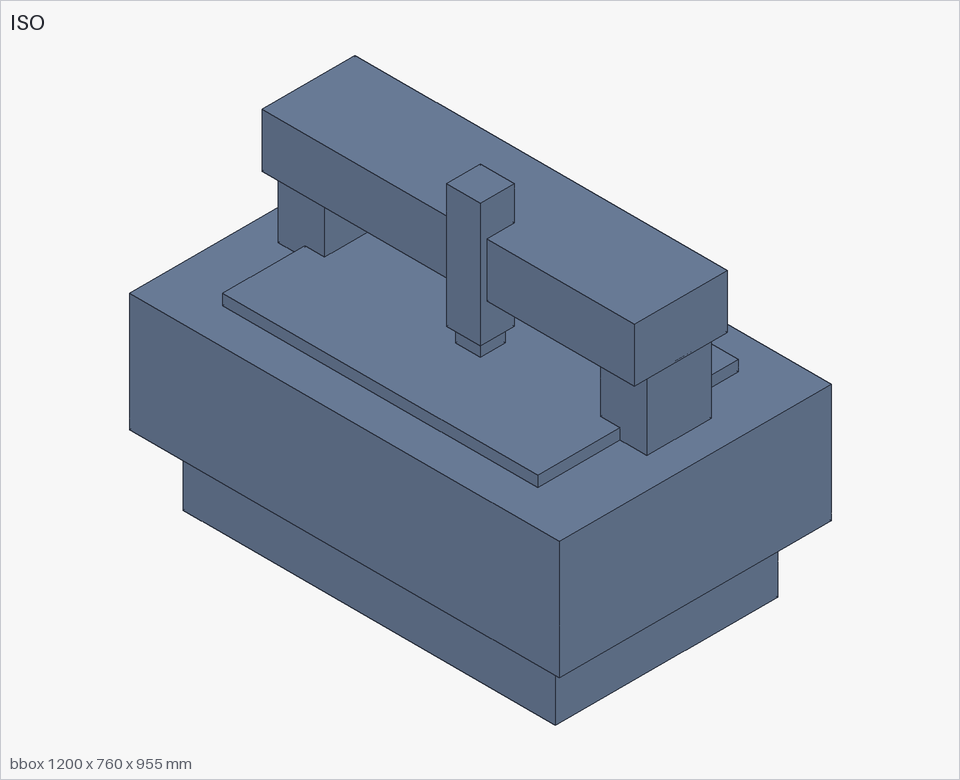}\\[-0.2em]{\scriptsize 65.0}} \\
\texttt{\scriptsize rcb\_000112974} & \makecell{\includegraphics[width=\linewidth,height=0.58in,keepaspectratio,valign=c]{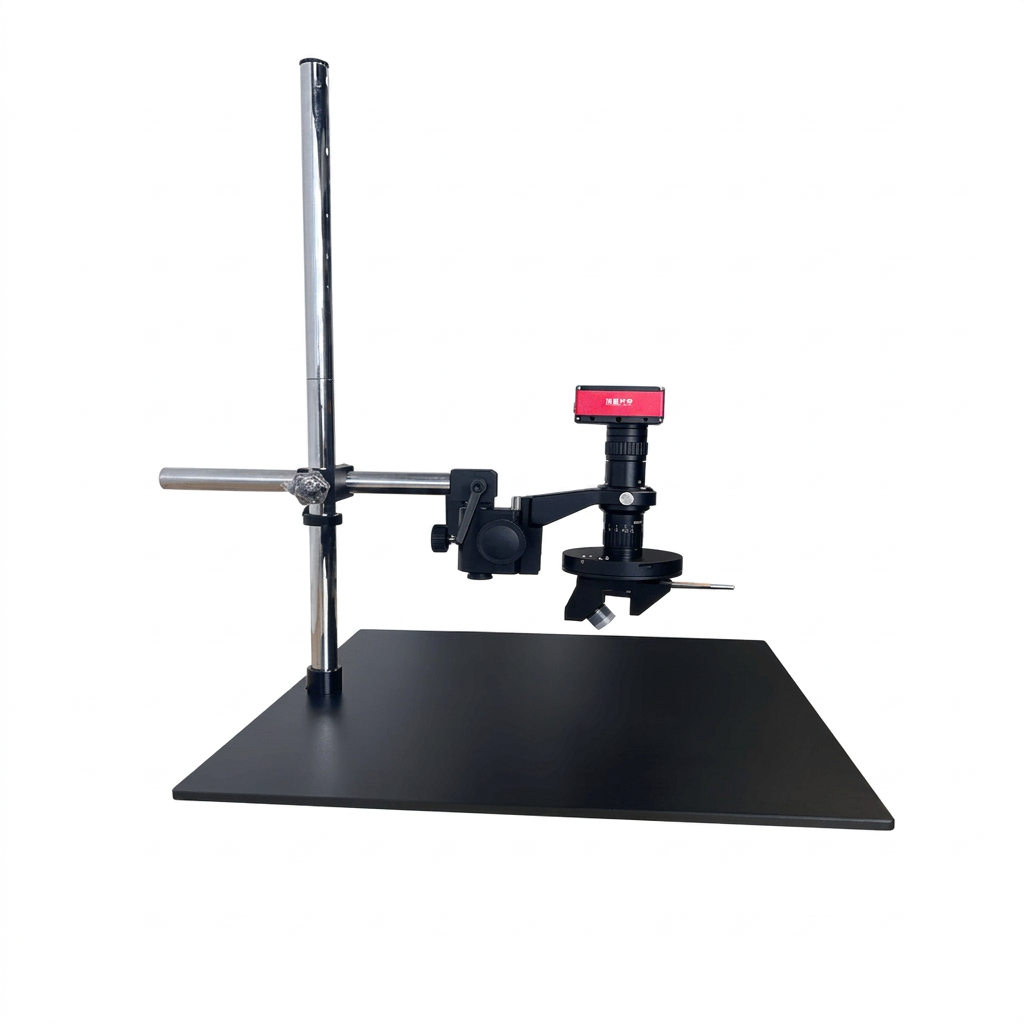}} & \makecell{\includegraphics[width=\linewidth,height=0.58in,keepaspectratio,valign=c]{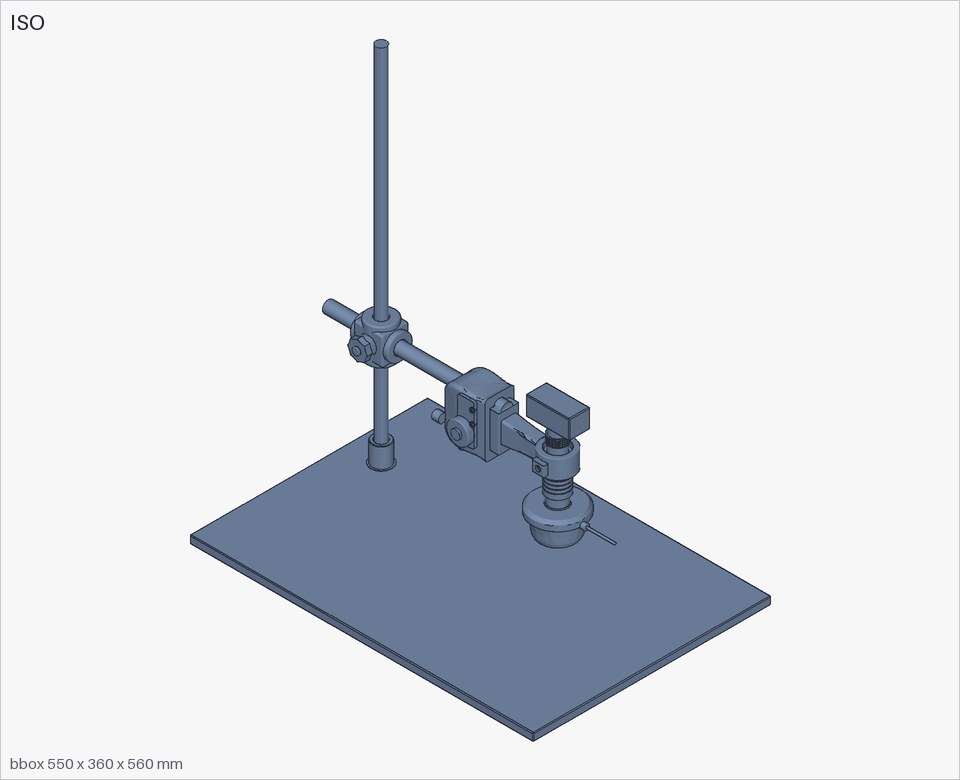}\\[-0.2em]{\scriptsize 86.1}} & \makecell{\includegraphics[width=\linewidth,height=0.58in,keepaspectratio,valign=c]{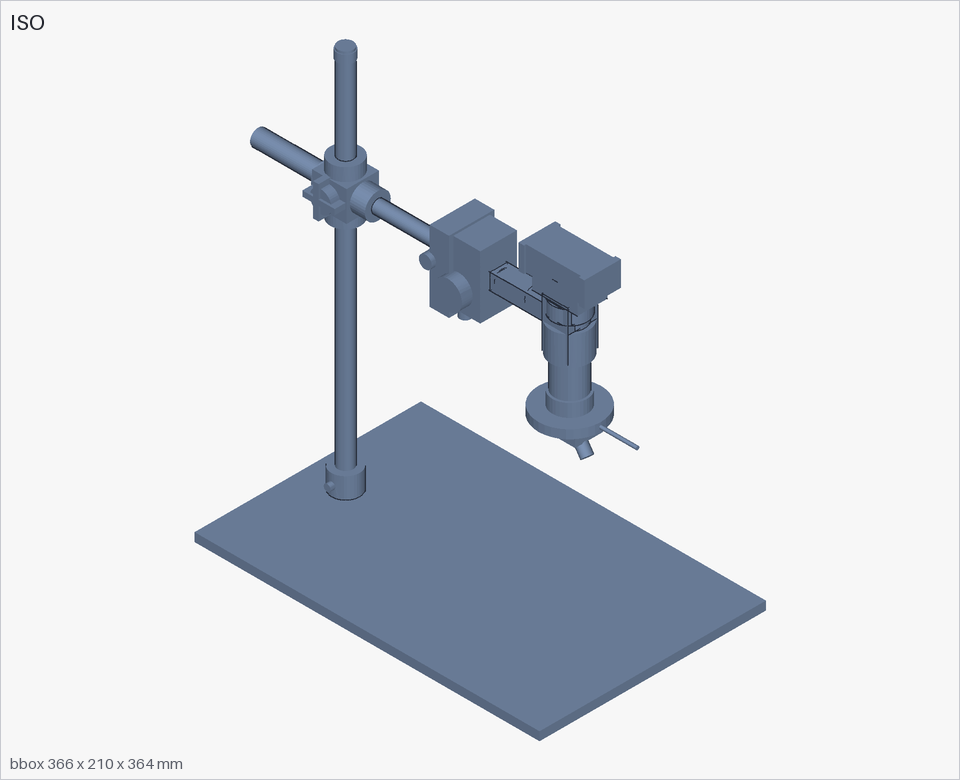}\\[-0.2em]{\scriptsize 83.4}} & \makecell{\includegraphics[width=\linewidth,height=0.58in,keepaspectratio,valign=c]{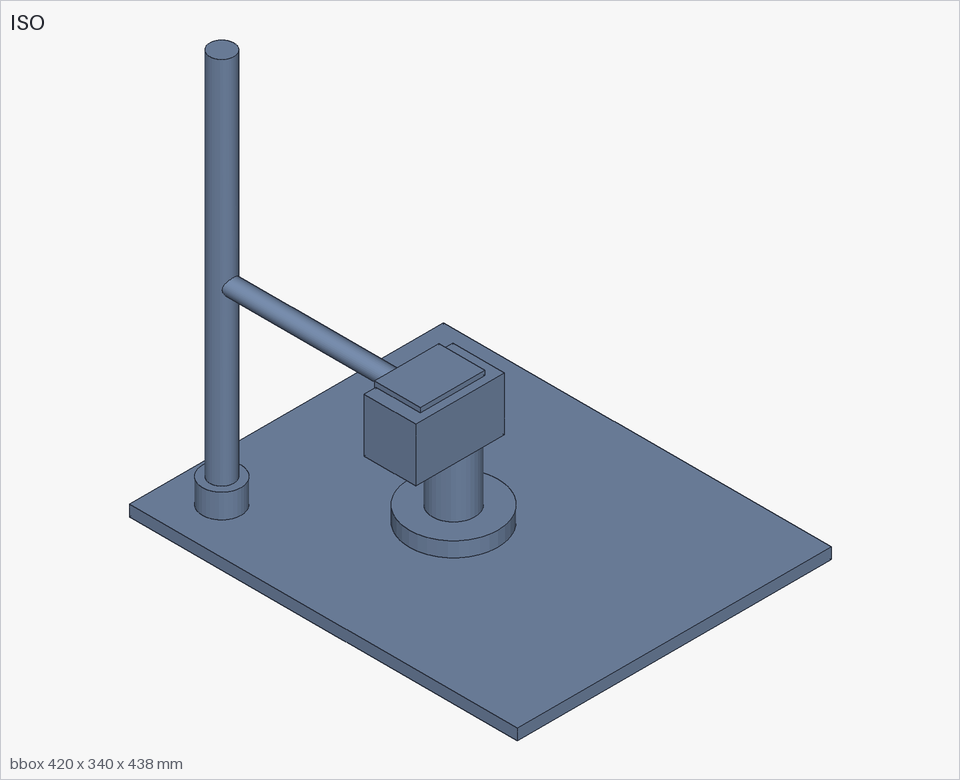}\\[-0.2em]{\scriptsize 57.9}} \\
\texttt{\scriptsize rcb\_000133697} & \makecell{\includegraphics[width=\linewidth,height=0.58in,keepaspectratio,valign=c]{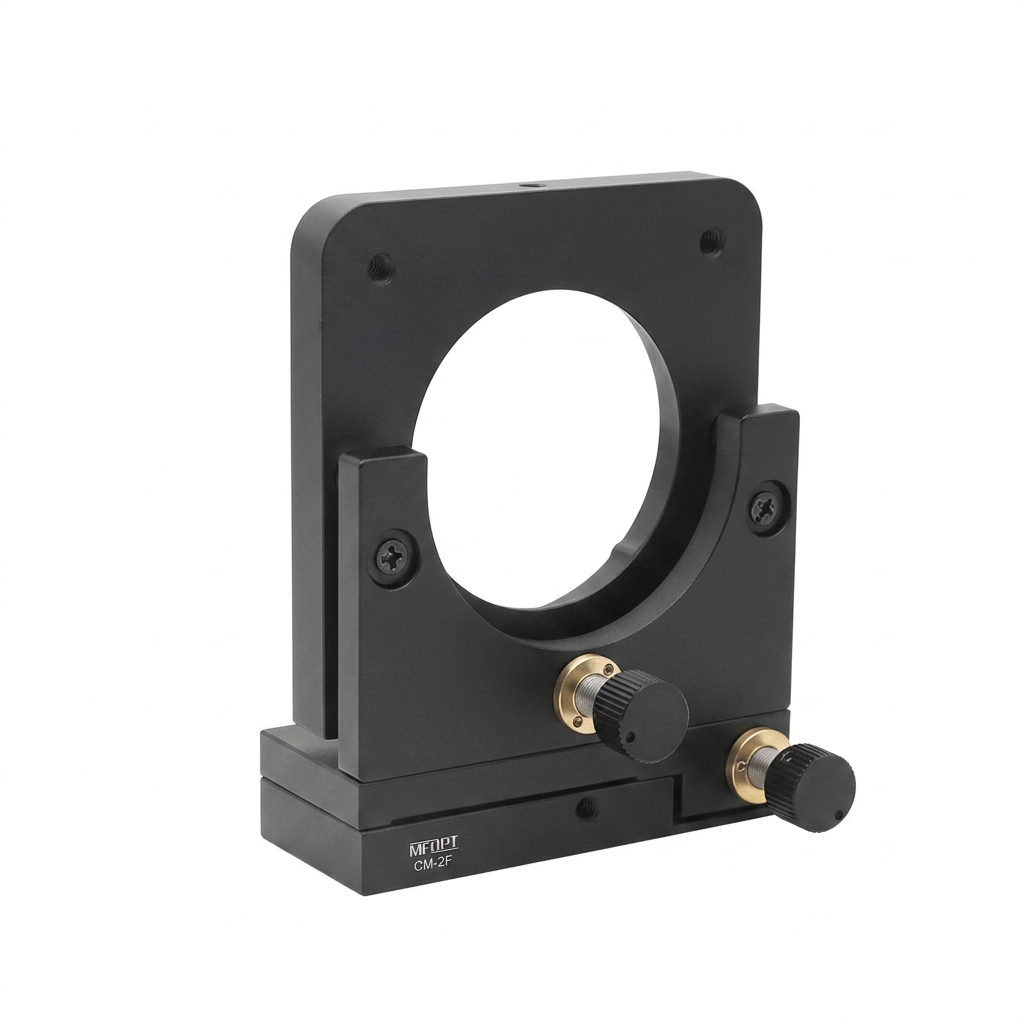}} & \makecell{\includegraphics[width=\linewidth,height=0.58in,keepaspectratio,valign=c]{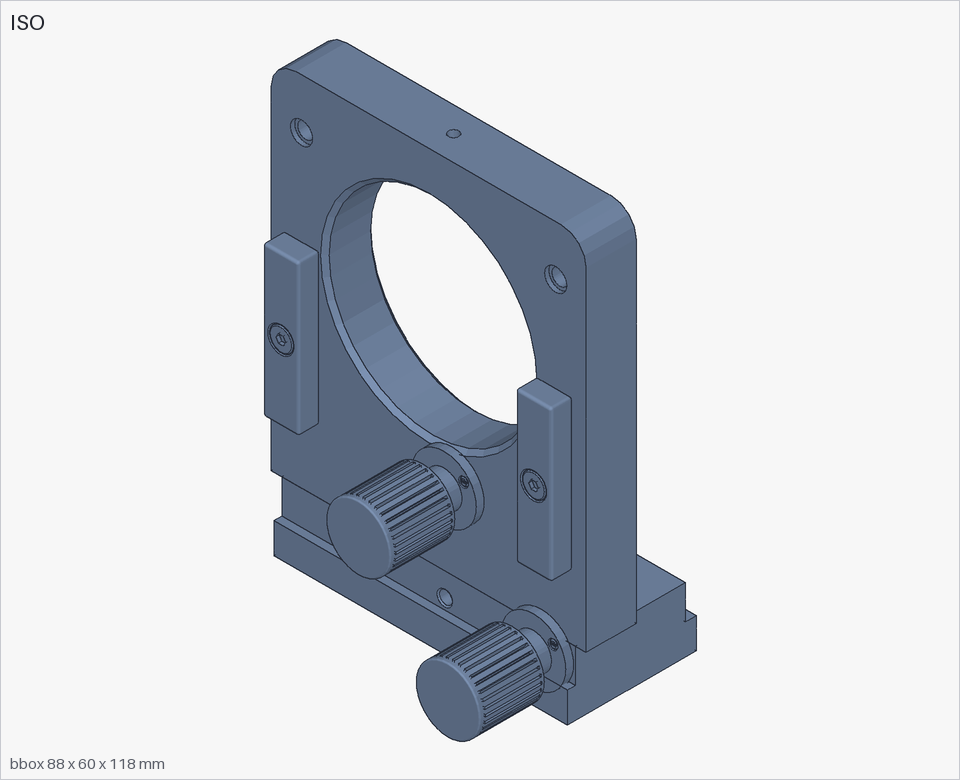}\\[-0.2em]{\scriptsize 80.3}} & \makecell{\includegraphics[width=\linewidth,height=0.58in,keepaspectratio,valign=c]{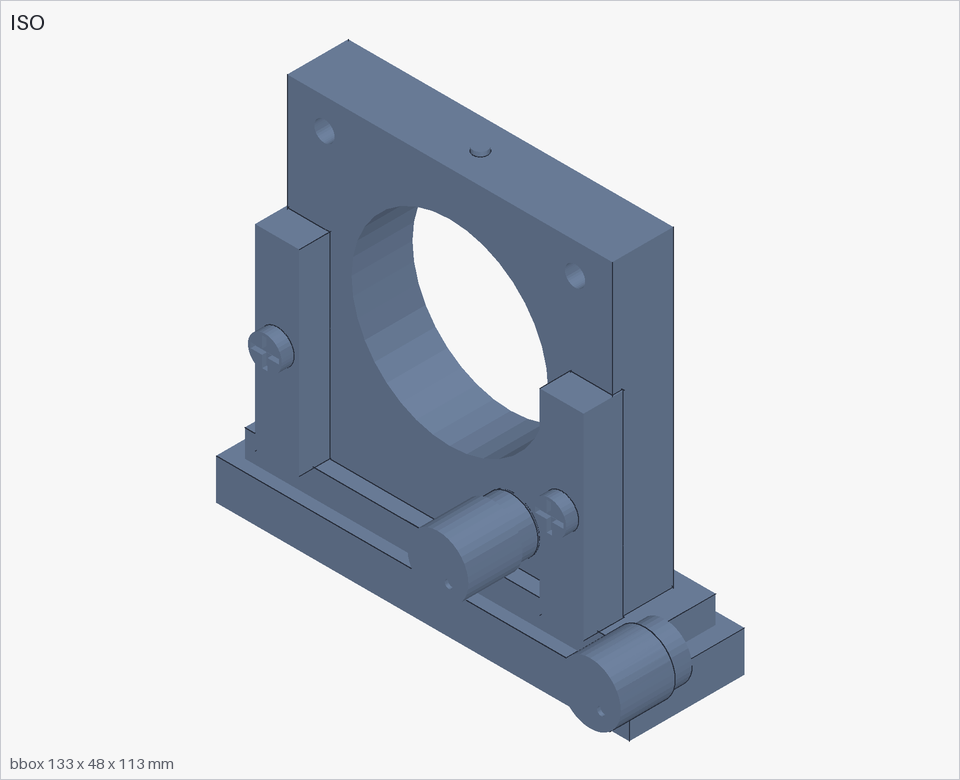}\\[-0.2em]{\scriptsize 77.2}} & \makecell{\includegraphics[width=\linewidth,height=0.58in,keepaspectratio,valign=c]{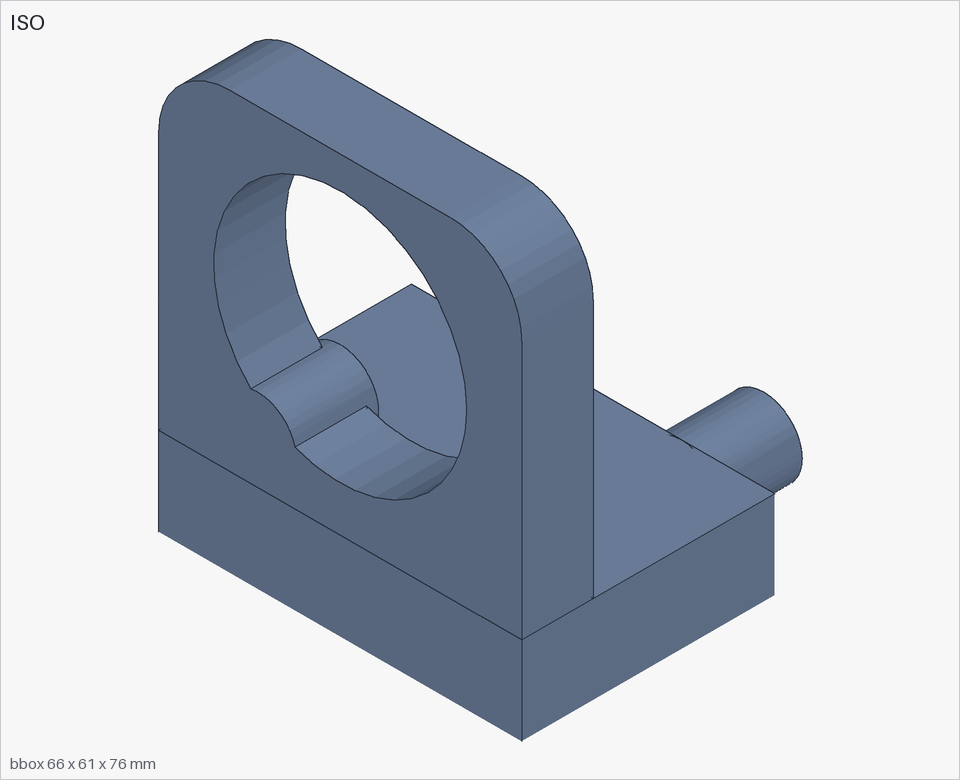}\\[-0.2em]{\scriptsize 54.5}} \\
\texttt{\scriptsize rcb\_000198365} & \makecell{\includegraphics[width=\linewidth,height=0.58in,keepaspectratio,valign=c]{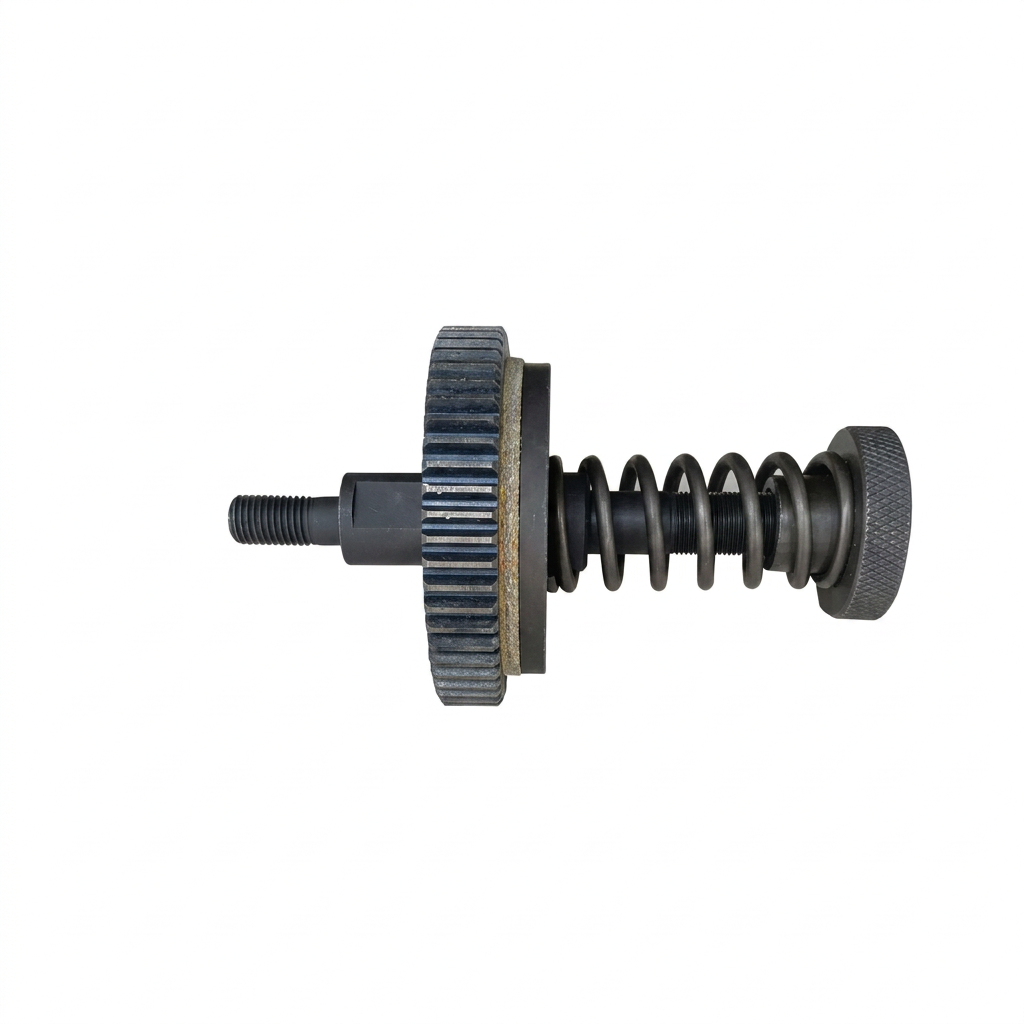}} & \makecell{\includegraphics[width=\linewidth,height=0.58in,keepaspectratio,valign=c]{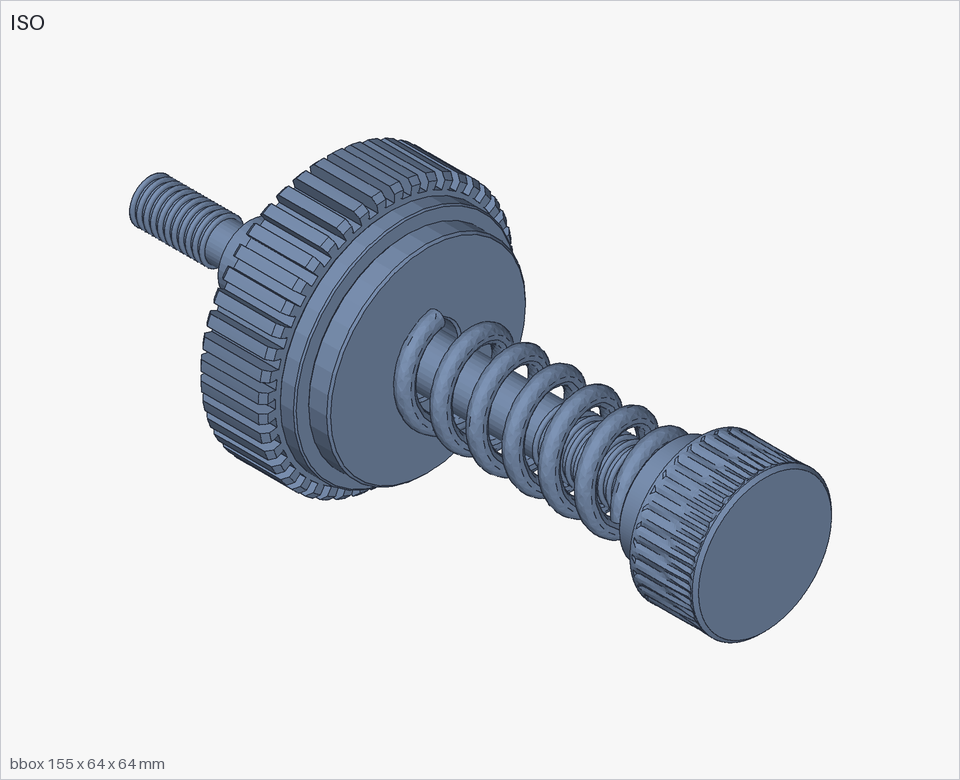}\\[-0.2em]{\scriptsize 88.2}} & \makecell{\includegraphics[width=\linewidth,height=0.58in,keepaspectratio,valign=c]{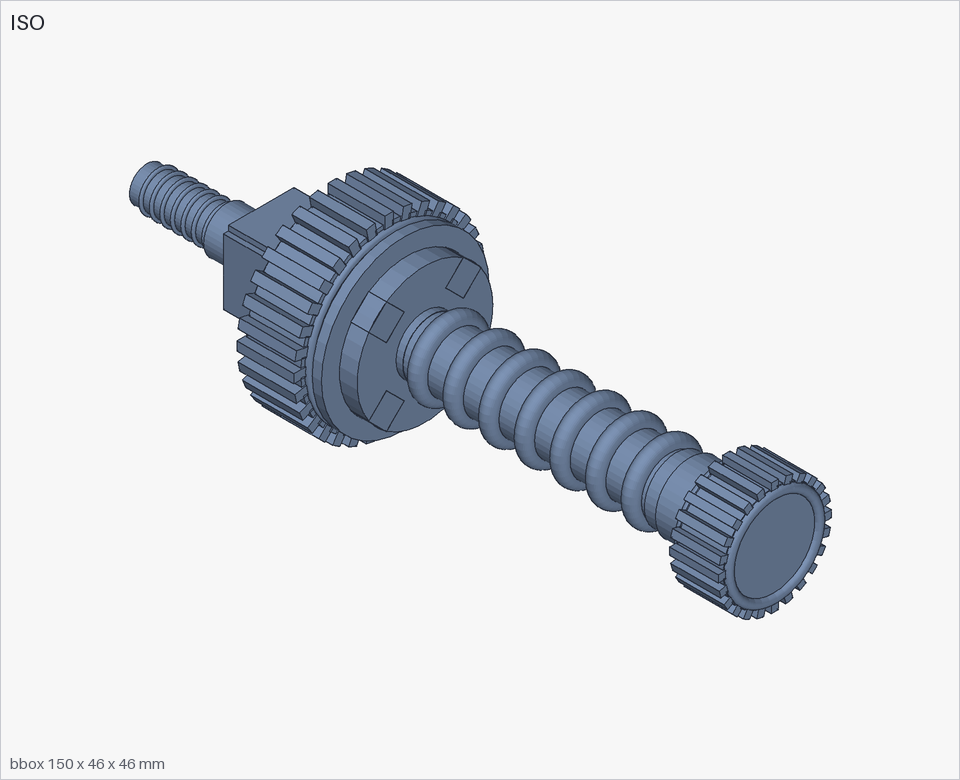}\\[-0.2em]{\scriptsize 88.8}} & \makecell{\includegraphics[width=\linewidth,height=0.58in,keepaspectratio,valign=c]{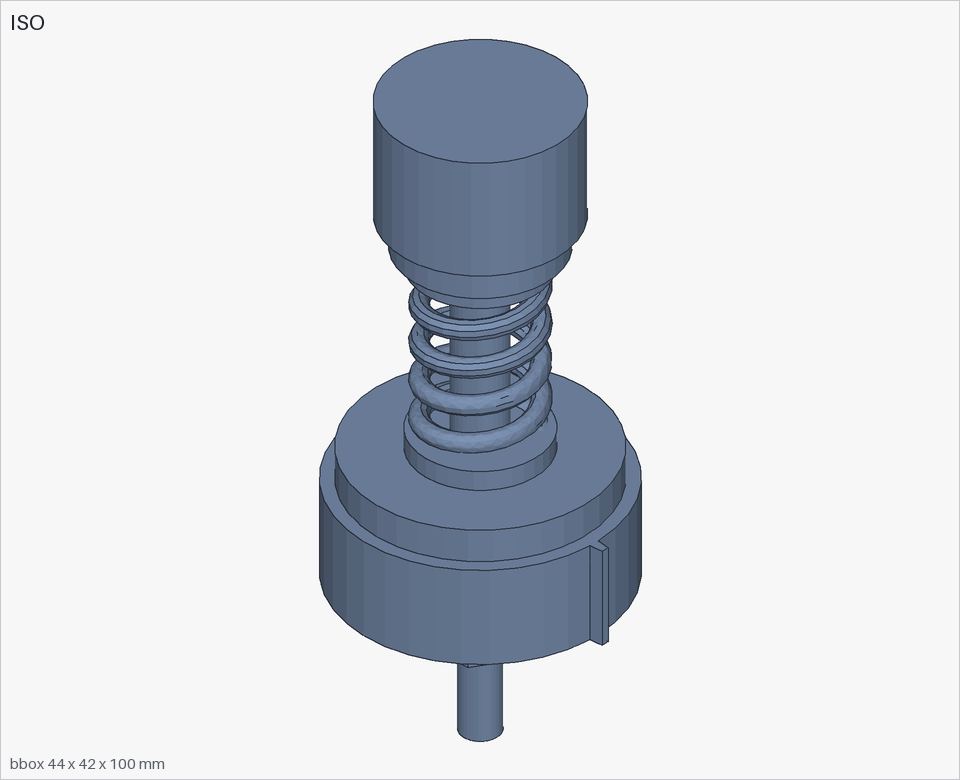}\\[-0.2em]{\scriptsize 67.9}} \\
\texttt{\scriptsize rcb\_000300010} & \makecell{\includegraphics[width=\linewidth,height=0.58in,keepaspectratio,valign=c]{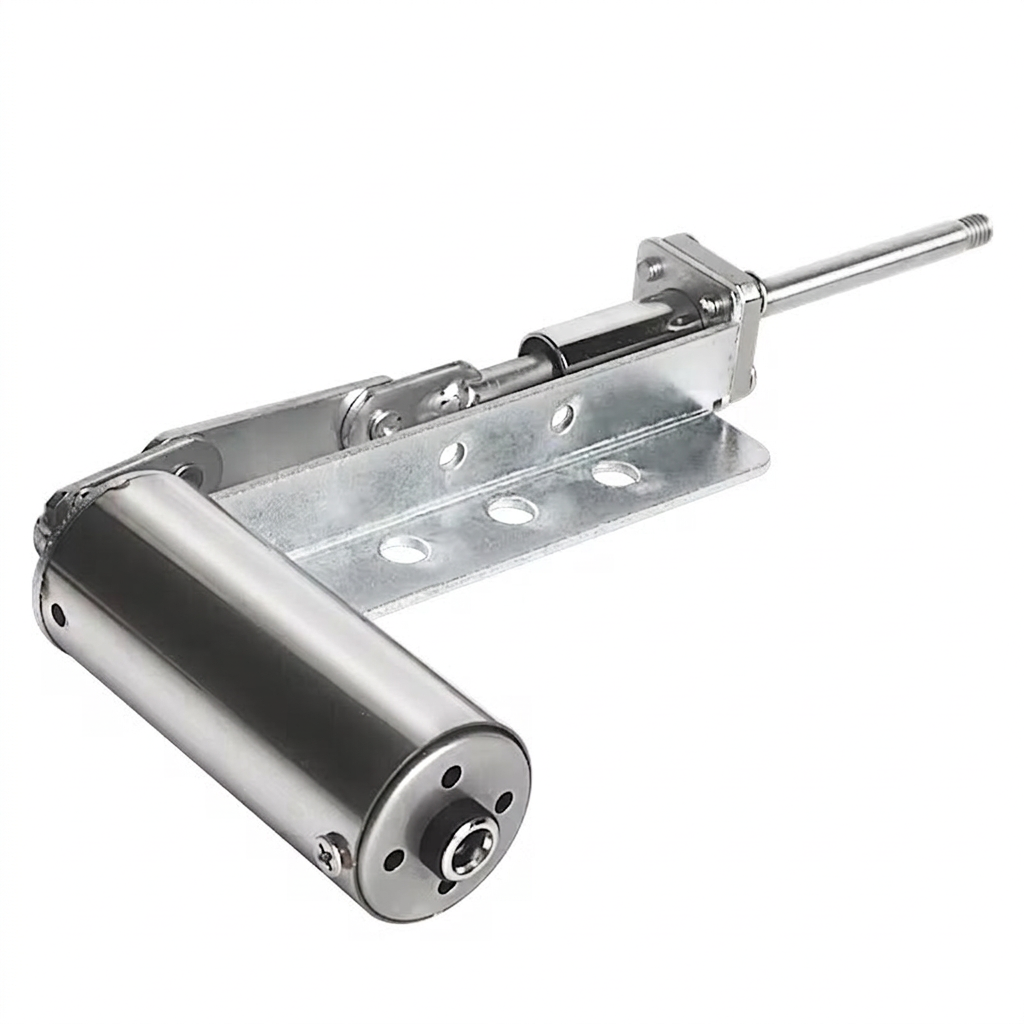}} & \makecell{\includegraphics[width=\linewidth,height=0.58in,keepaspectratio,valign=c]{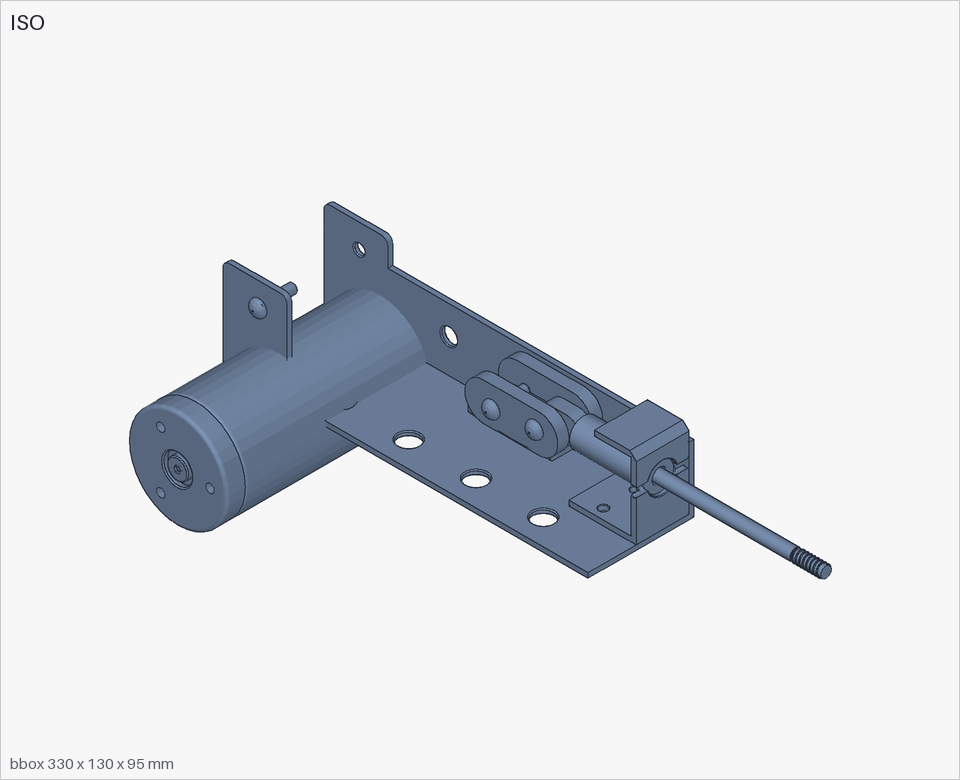}\\[-0.2em]{\scriptsize 80.4}} & \makecell{\raisebox{0.18in}{\color{gray}\scriptsize no export}} & \makecell{\includegraphics[width=\linewidth,height=0.58in,keepaspectratio,valign=c]{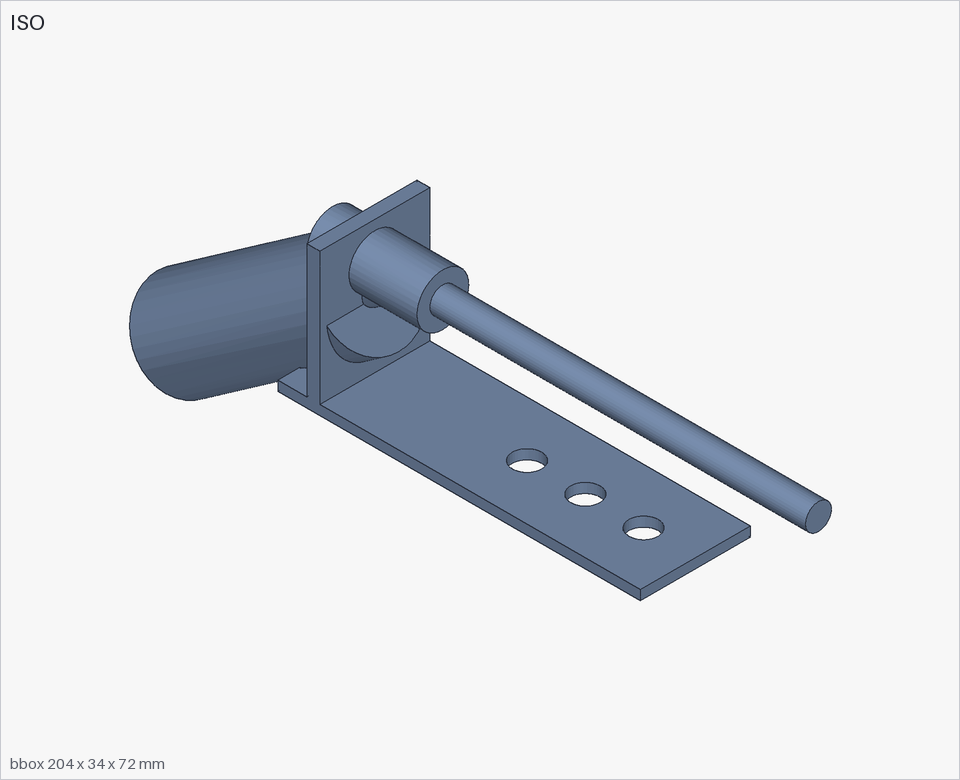}\\[-0.2em]{\scriptsize 54.0}} \\
\texttt{\scriptsize rcb\_000304481} & \makecell{\includegraphics[width=\linewidth,height=0.58in,keepaspectratio,valign=c]{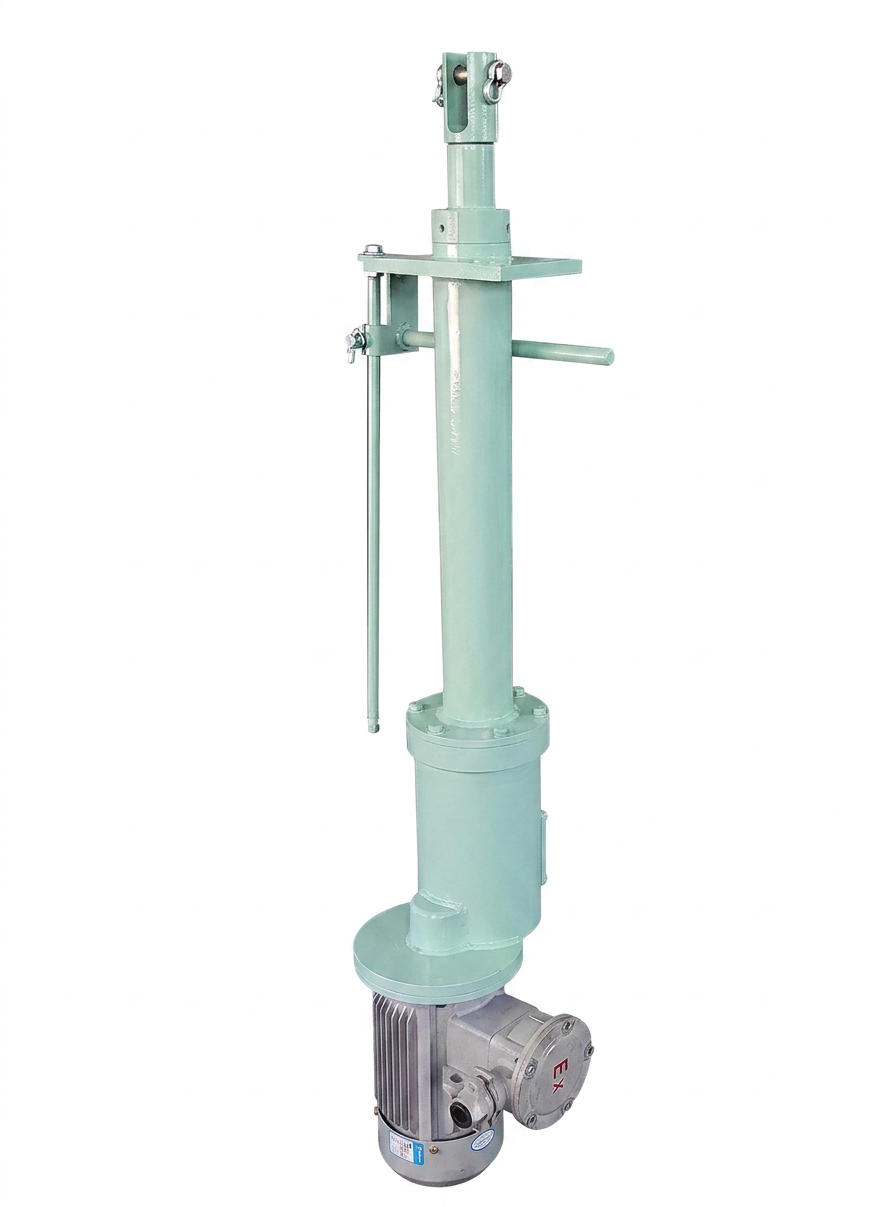}} & \makecell{\includegraphics[width=\linewidth,height=0.58in,keepaspectratio,valign=c]{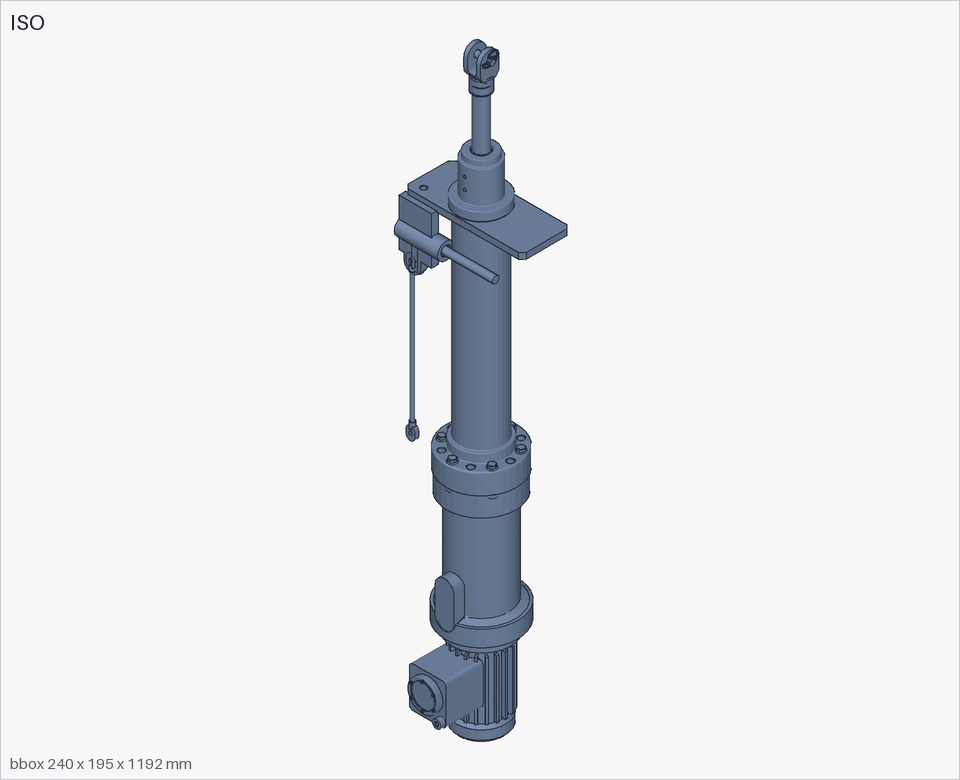}\\[-0.2em]{\scriptsize 83.0}} & \makecell{\includegraphics[width=\linewidth,height=0.58in,keepaspectratio,valign=c]{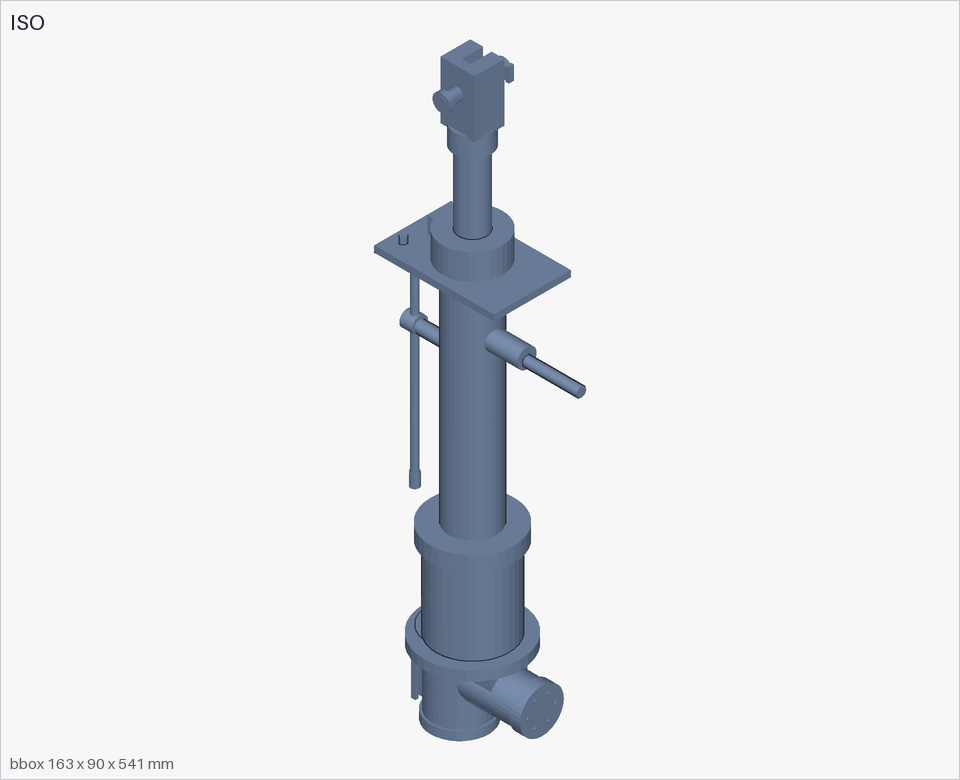}\\[-0.2em]{\scriptsize 73.5}} & \makecell{\includegraphics[width=\linewidth,height=0.58in,keepaspectratio,valign=c]{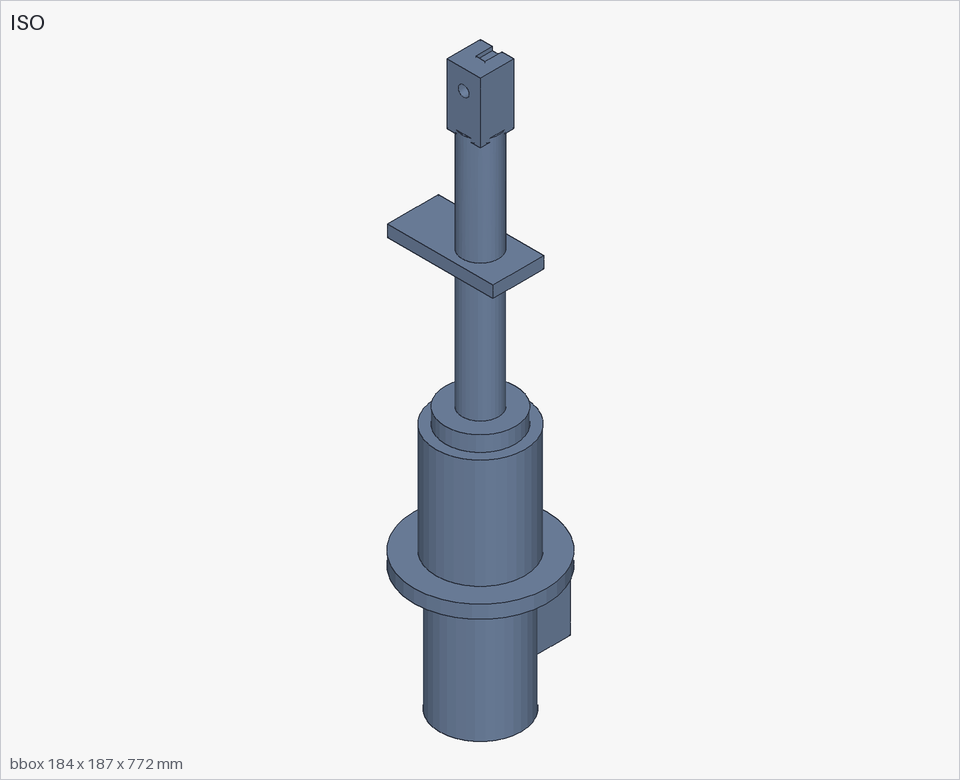}\\[-0.2em]{\scriptsize 47.0}} \\
\texttt{\scriptsize rcb\_000315240} & \makecell{\includegraphics[width=\linewidth,height=0.58in,keepaspectratio,valign=c]{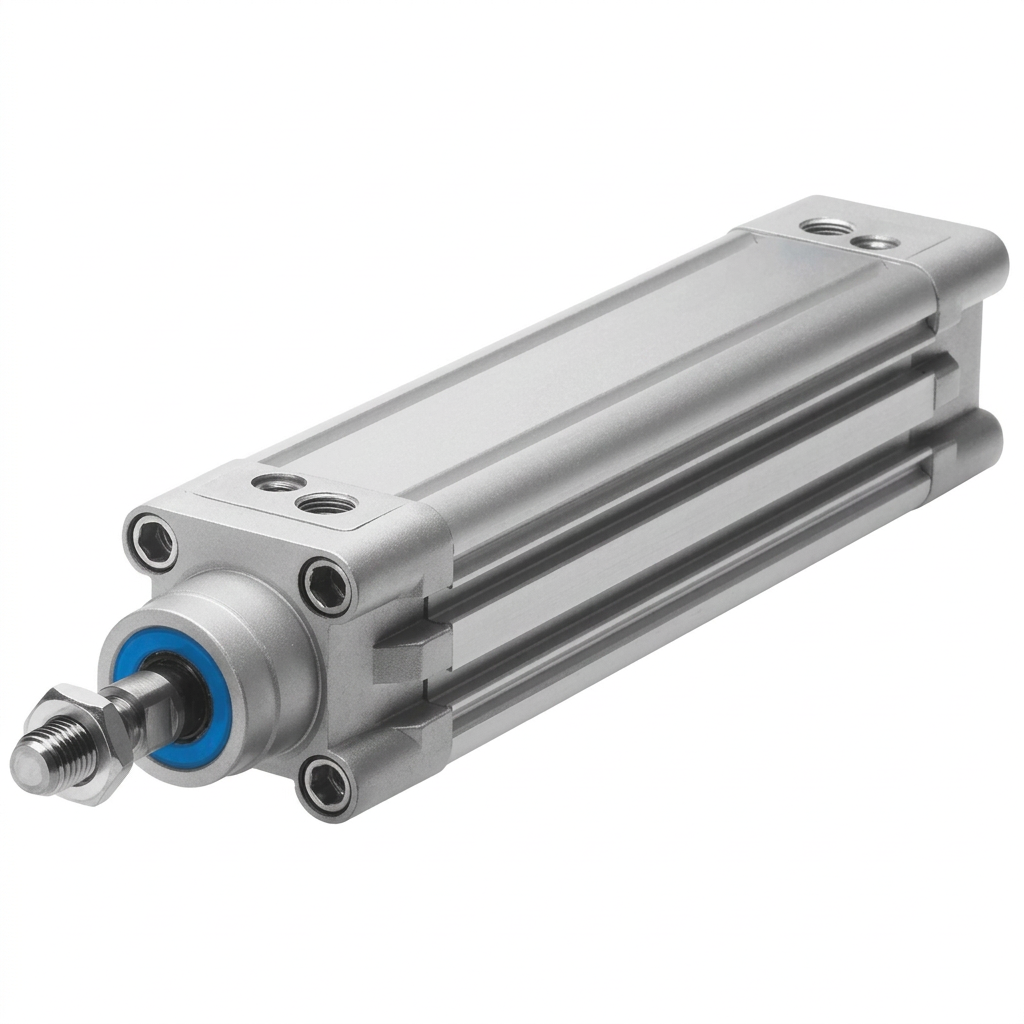}} & \makecell{\includegraphics[width=\linewidth,height=0.58in,keepaspectratio,valign=c]{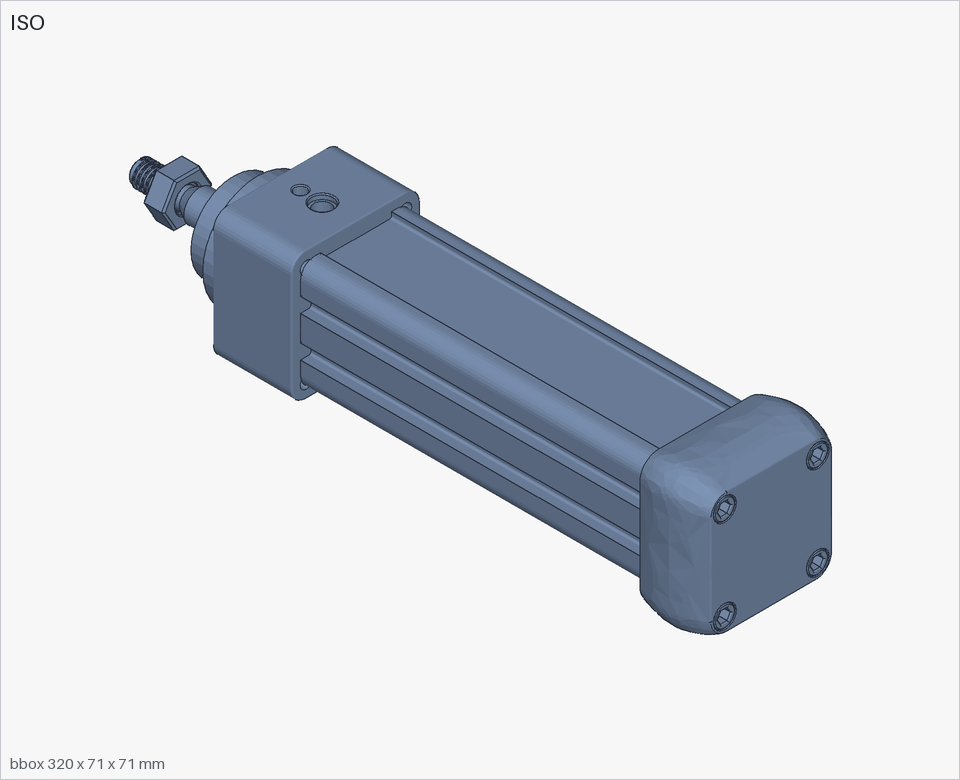}\\[-0.2em]{\scriptsize 85.3}} & \makecell{\includegraphics[width=\linewidth,height=0.58in,keepaspectratio,valign=c]{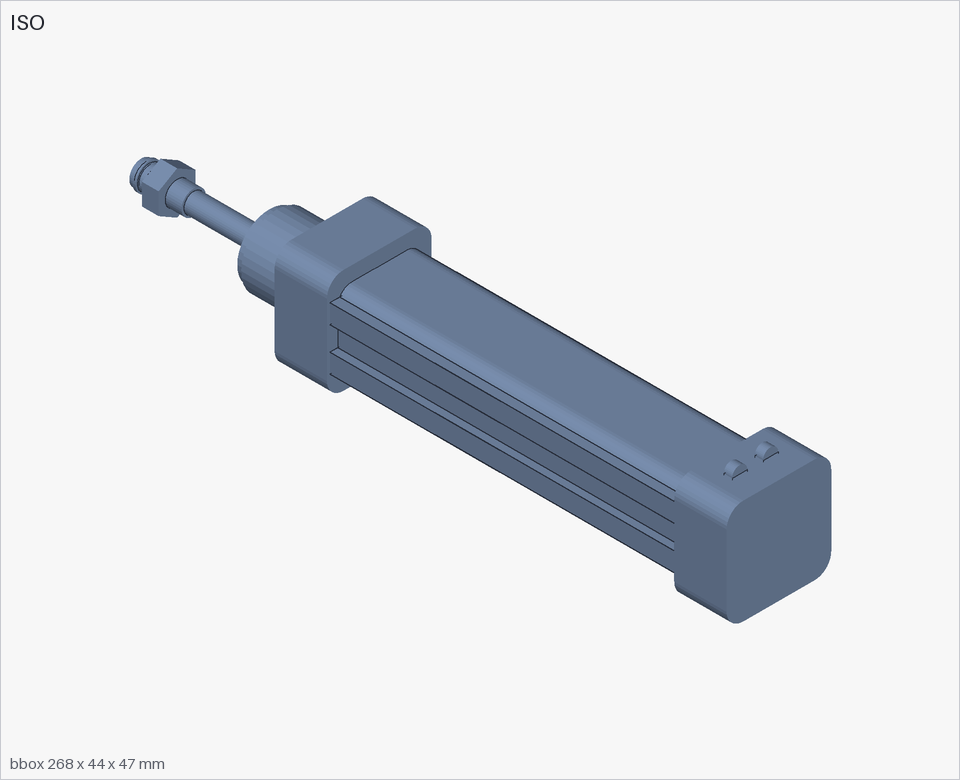}\\[-0.2em]{\scriptsize 78.6}} & \makecell{\includegraphics[width=\linewidth,height=0.58in,keepaspectratio,valign=c]{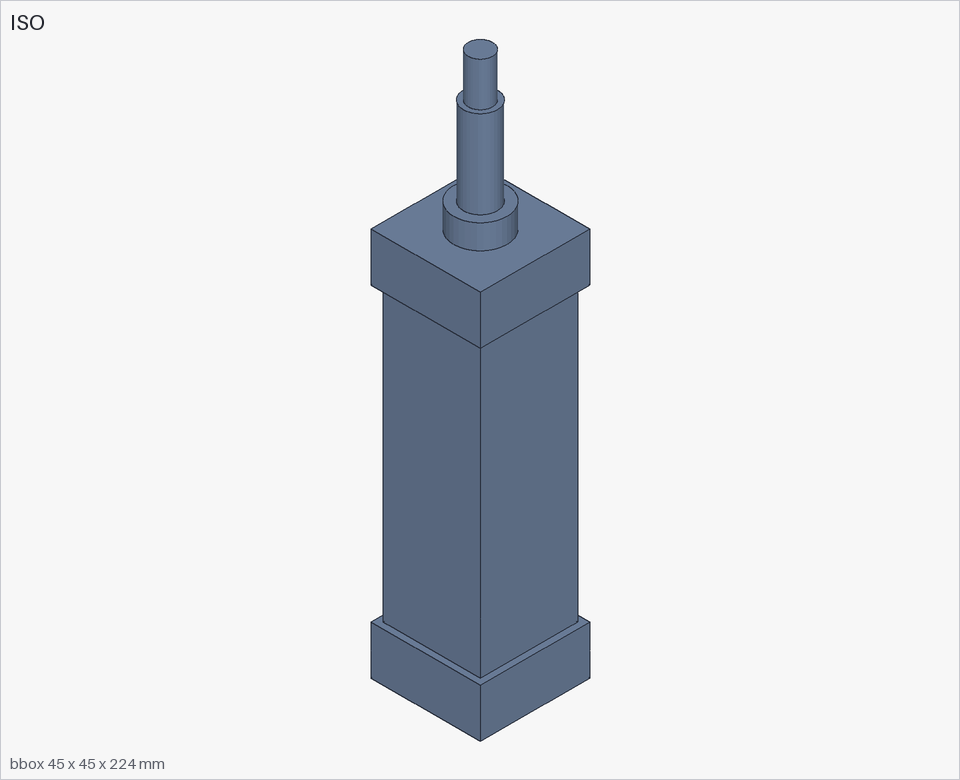}\\[-0.2em]{\scriptsize 54.0}} \\
\texttt{\scriptsize rcb\_000330472} & \makecell{\includegraphics[width=\linewidth,height=0.58in,keepaspectratio,valign=c]{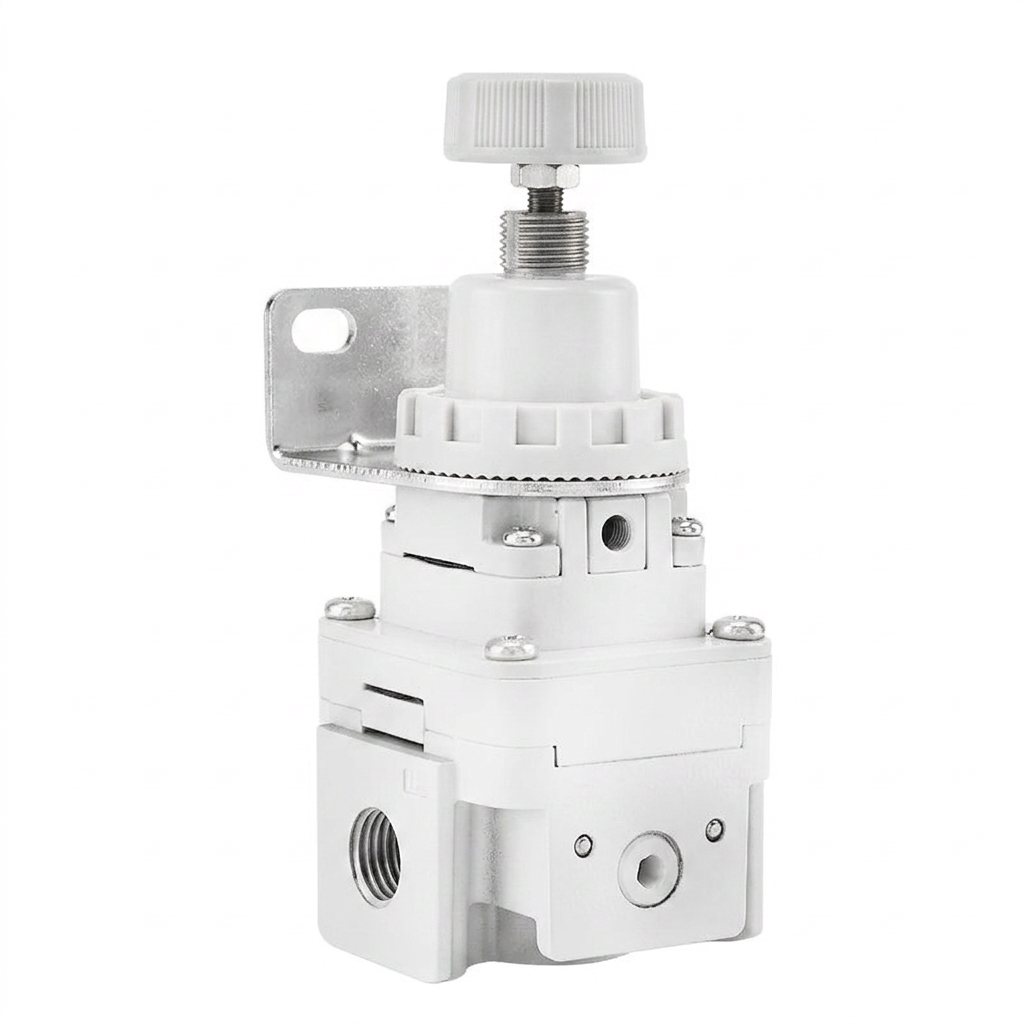}} & \makecell{\includegraphics[width=\linewidth,height=0.58in,keepaspectratio,valign=c]{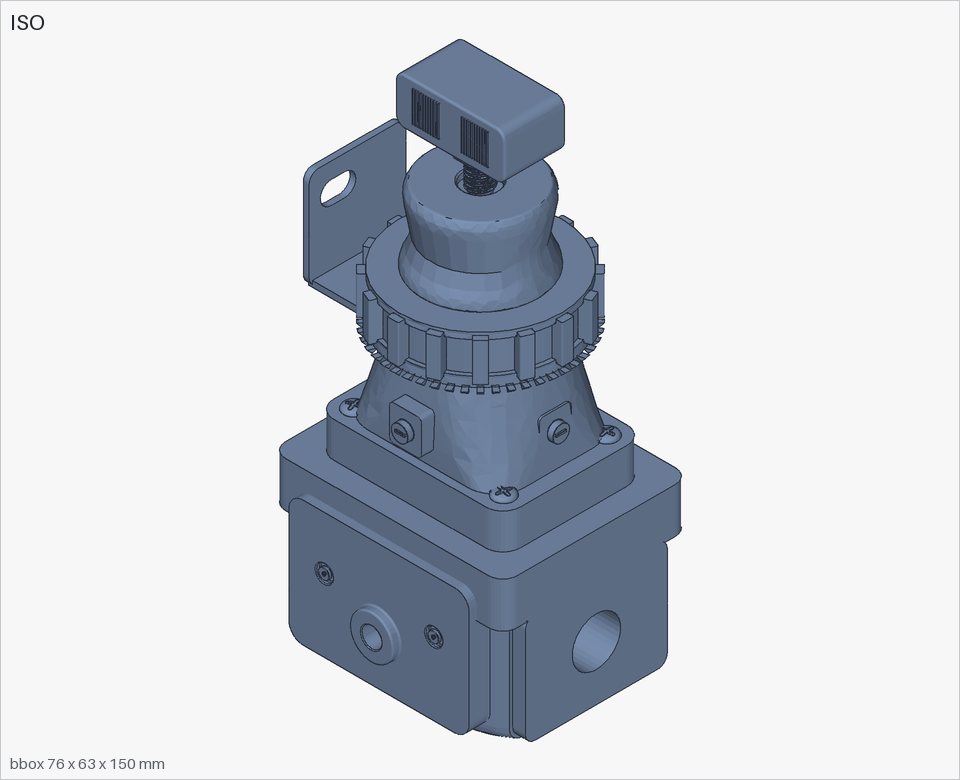}\\[-0.2em]{\scriptsize 83.7}} & \makecell{\includegraphics[width=\linewidth,height=0.58in,keepaspectratio,valign=c]{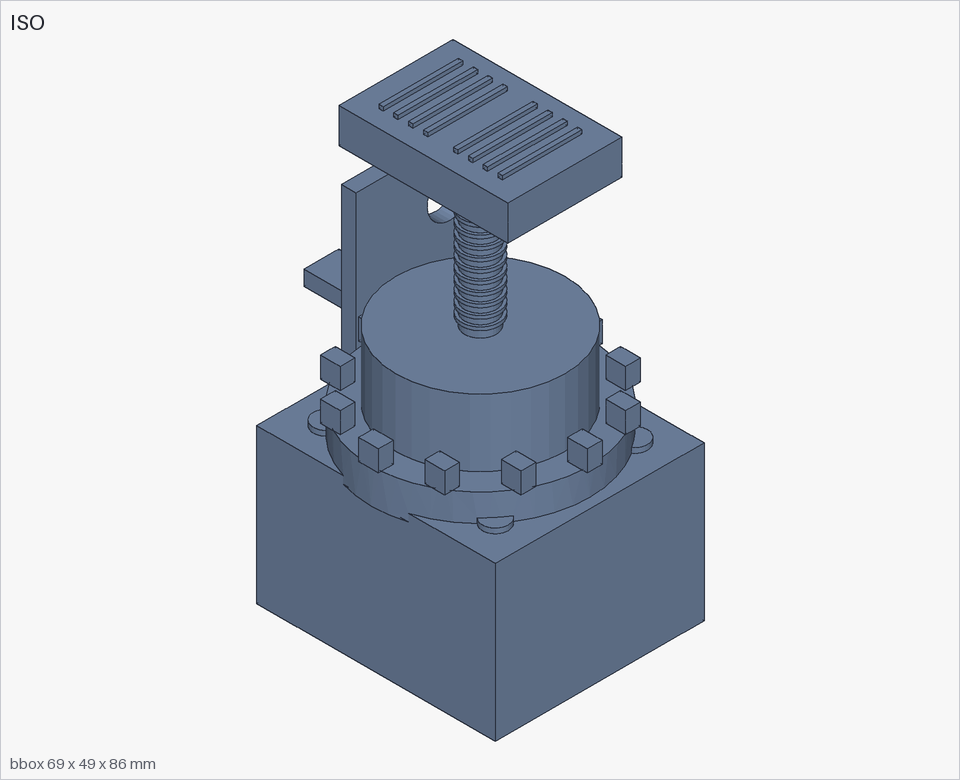}\\[-0.2em]{\scriptsize 64.0}} & \makecell{\includegraphics[width=\linewidth,height=0.58in,keepaspectratio,valign=c]{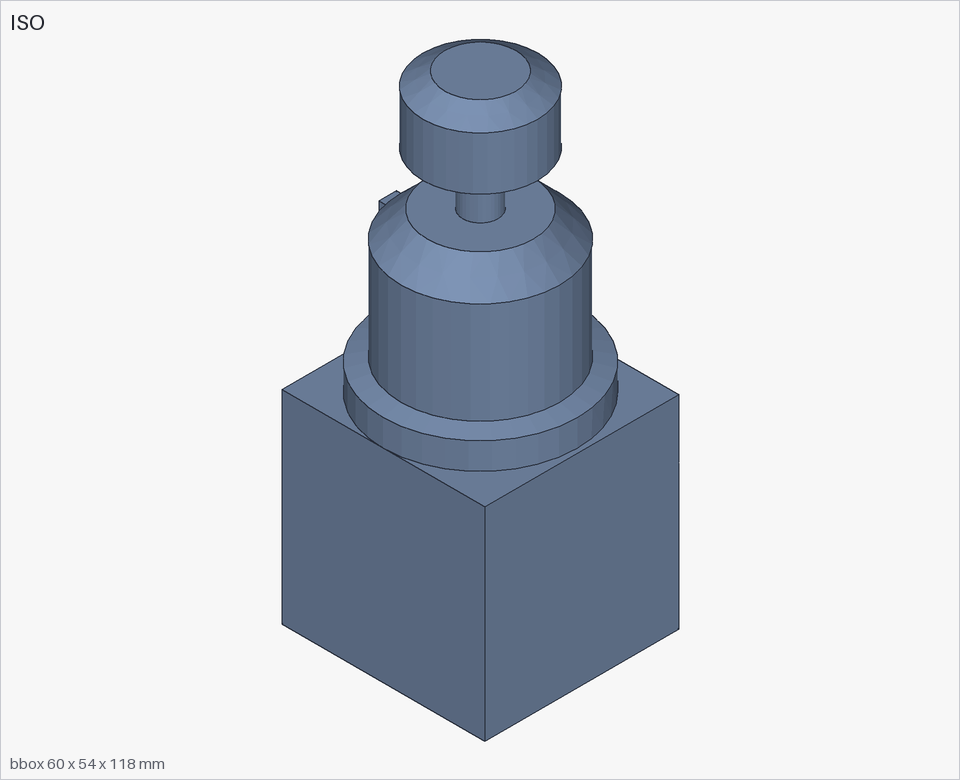}\\[-0.2em]{\scriptsize 50.5}} \\
\texttt{\scriptsize rcb\_000342241} & \makecell{\includegraphics[width=\linewidth,height=0.58in,keepaspectratio,valign=c]{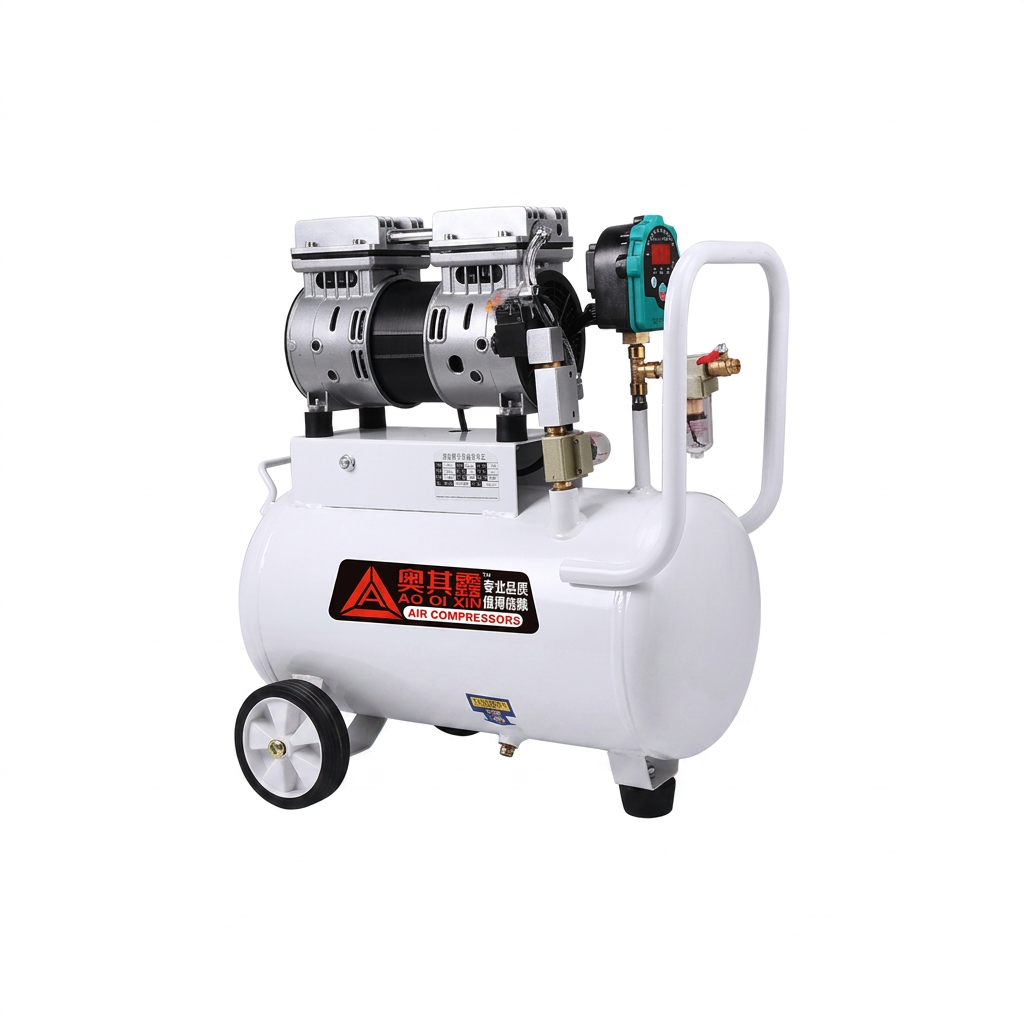}} & \makecell{\includegraphics[width=\linewidth,height=0.58in,keepaspectratio,valign=c]{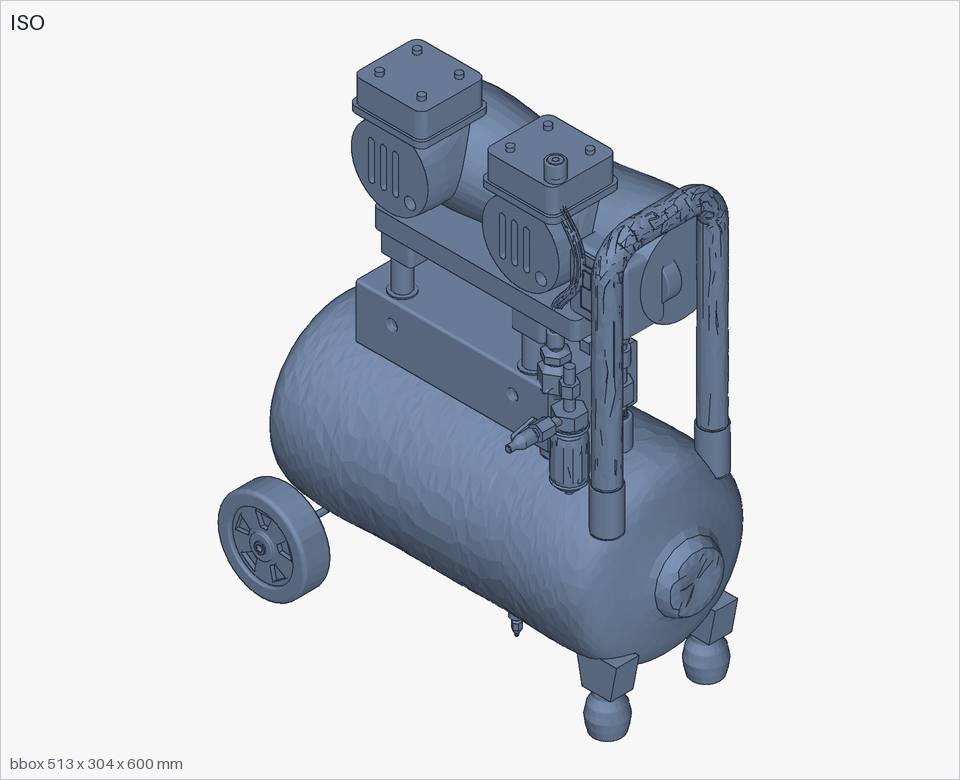}\\[-0.2em]{\scriptsize 83.2}} & \makecell{\includegraphics[width=\linewidth,height=0.58in,keepaspectratio,valign=c]{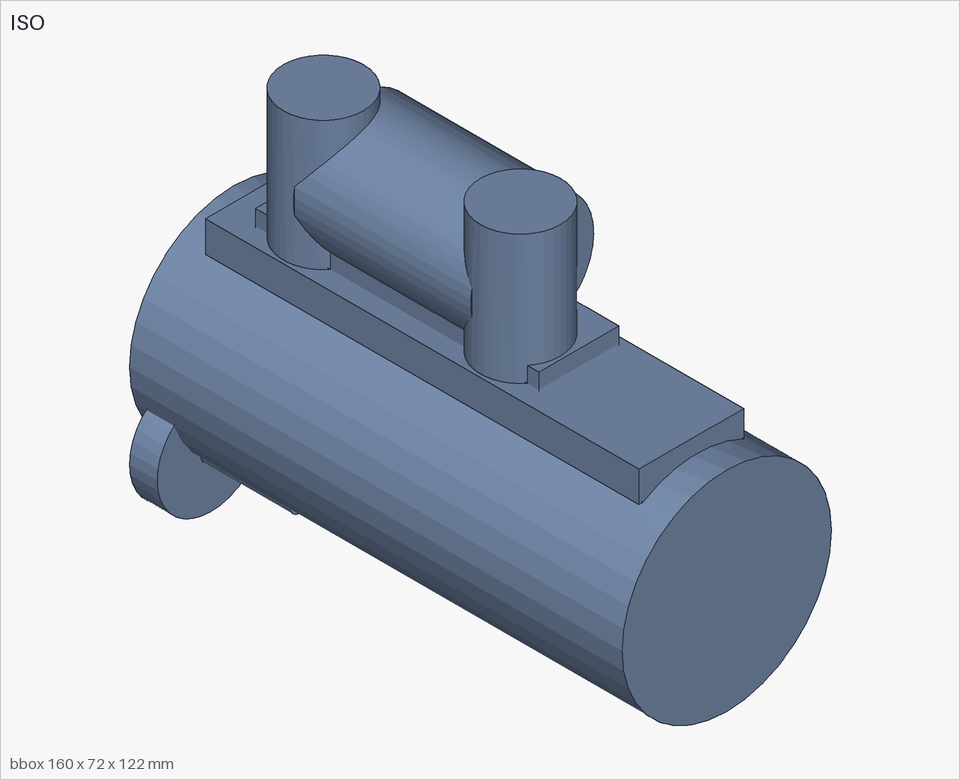}\\[-0.2em]{\scriptsize 47.1}} & \makecell{\includegraphics[width=\linewidth,height=0.58in,keepaspectratio,valign=c]{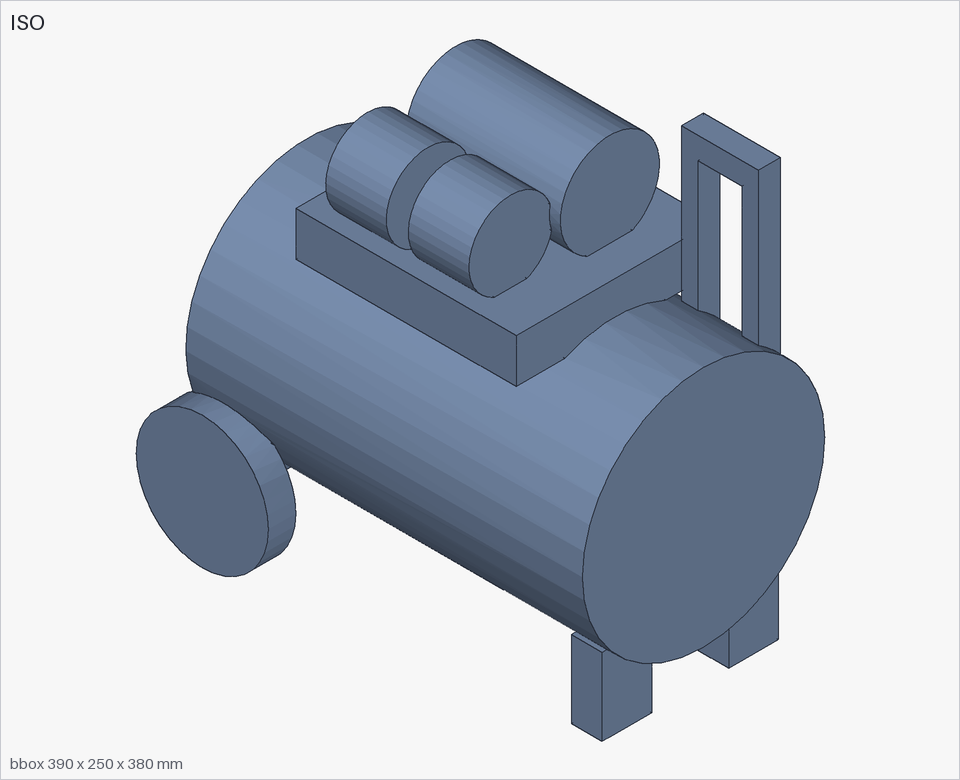}\\[-0.2em]{\scriptsize 59.6}} \\
\texttt{\scriptsize rcb\_000348219} & \makecell{\includegraphics[width=\linewidth,height=0.58in,keepaspectratio,valign=c]{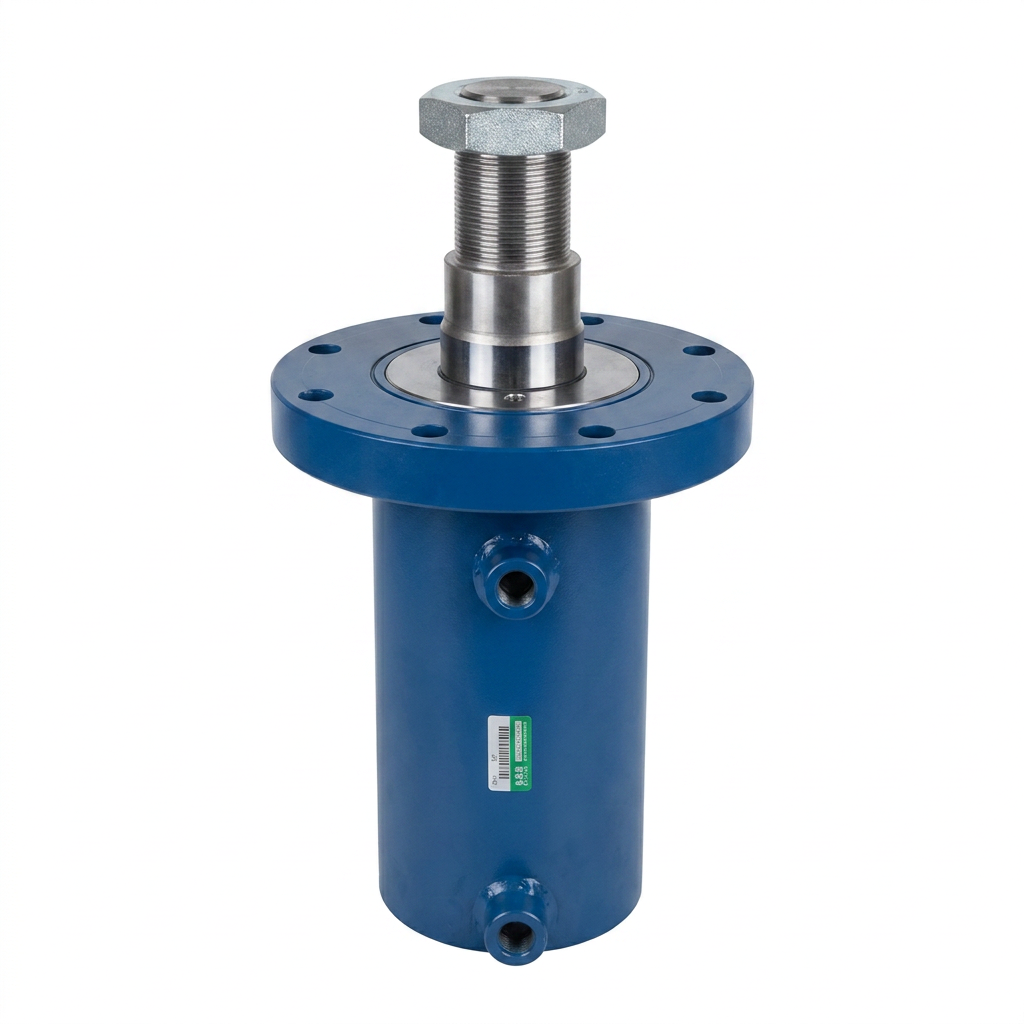}} & \makecell{\includegraphics[width=\linewidth,height=0.58in,keepaspectratio,valign=c]{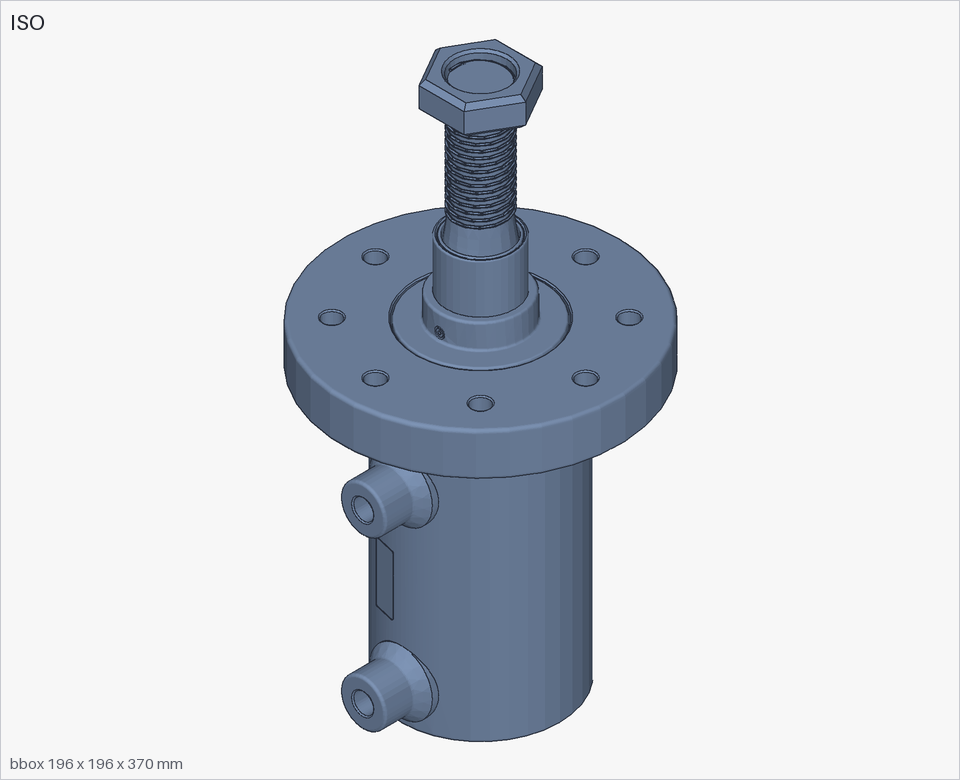}\\[-0.2em]{\scriptsize 89.4}} & \makecell{\includegraphics[width=\linewidth,height=0.58in,keepaspectratio,valign=c]{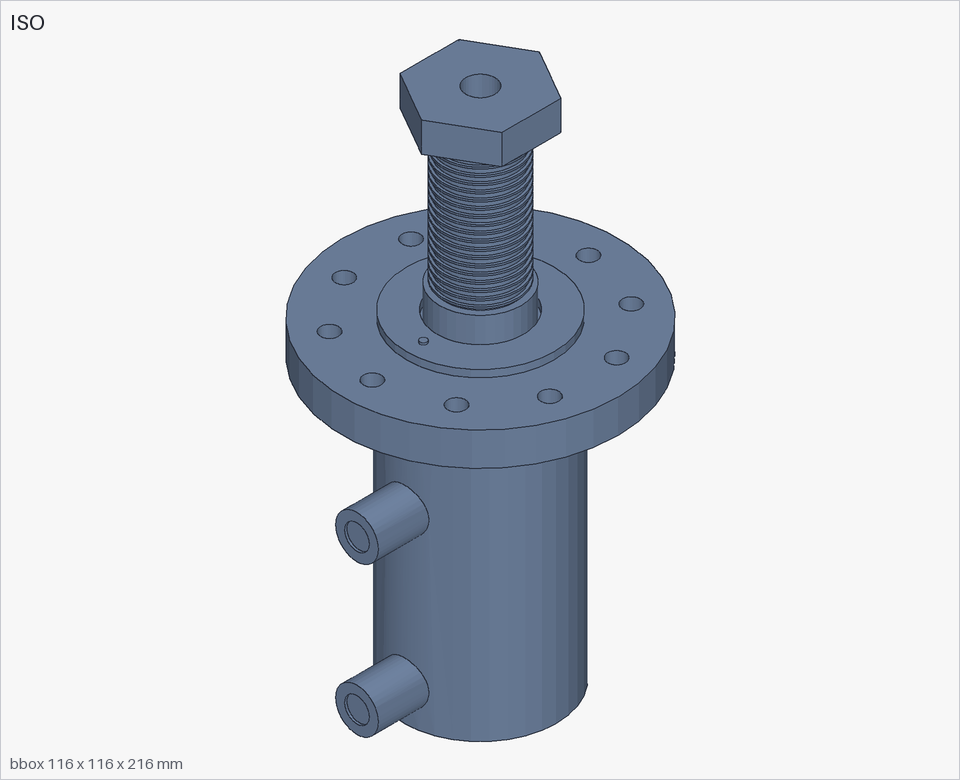}\\[-0.2em]{\scriptsize 88.2}} & \makecell{\includegraphics[width=\linewidth,height=0.58in,keepaspectratio,valign=c]{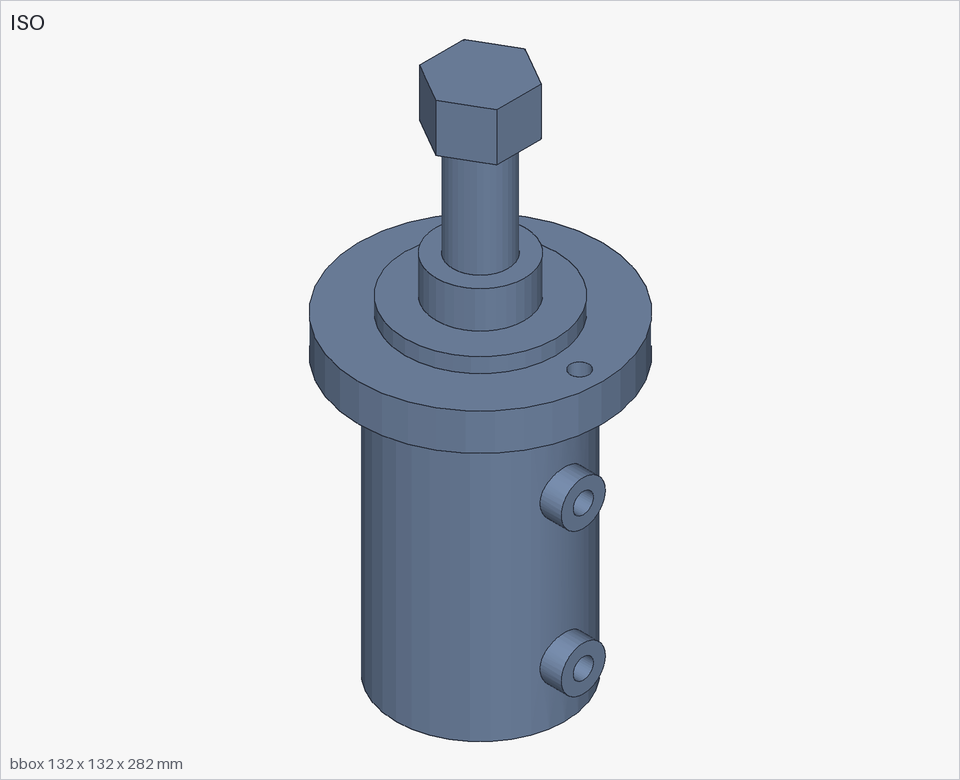}\\[-0.2em]{\scriptsize 74.8}} \\
\texttt{\scriptsize rcb\_000352539} & \makecell{\includegraphics[width=\linewidth,height=0.58in,keepaspectratio,valign=c]{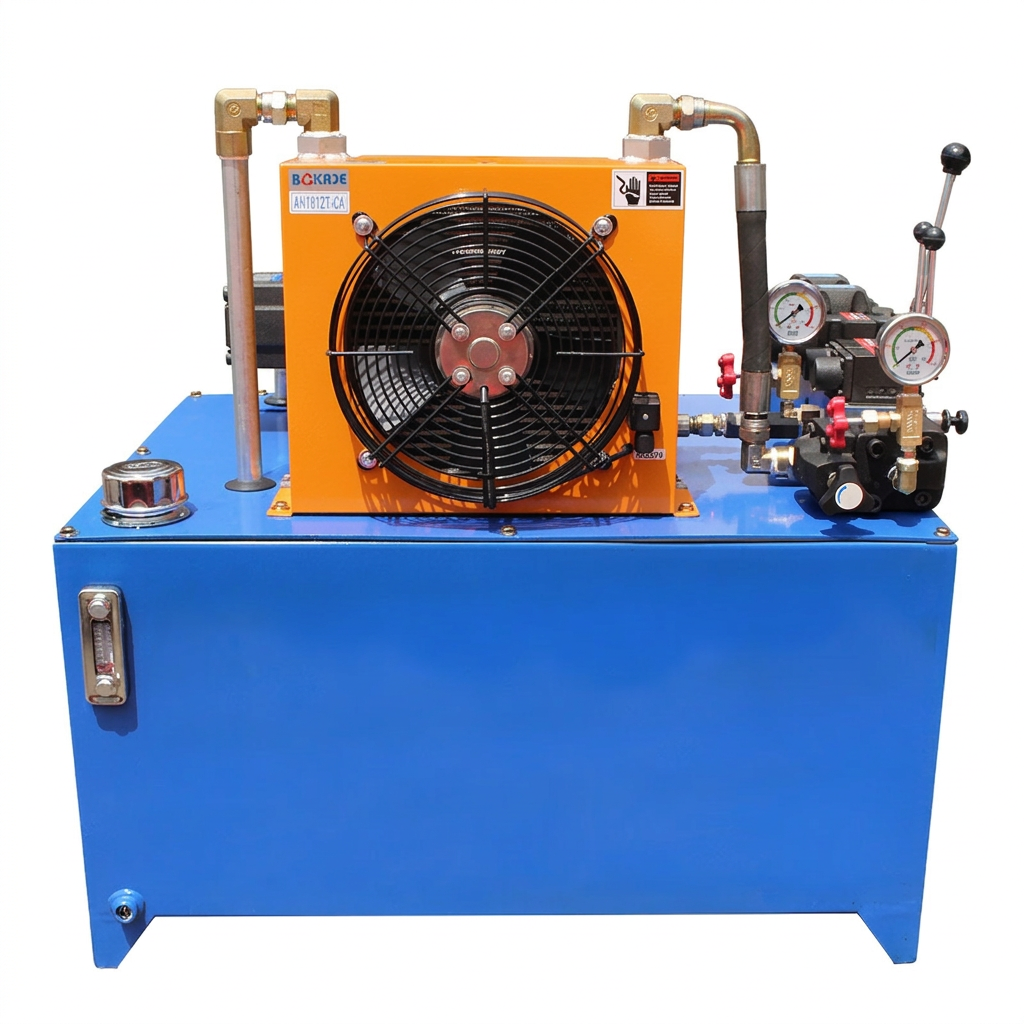}} & \makecell{\includegraphics[width=\linewidth,height=0.58in,keepaspectratio,valign=c]{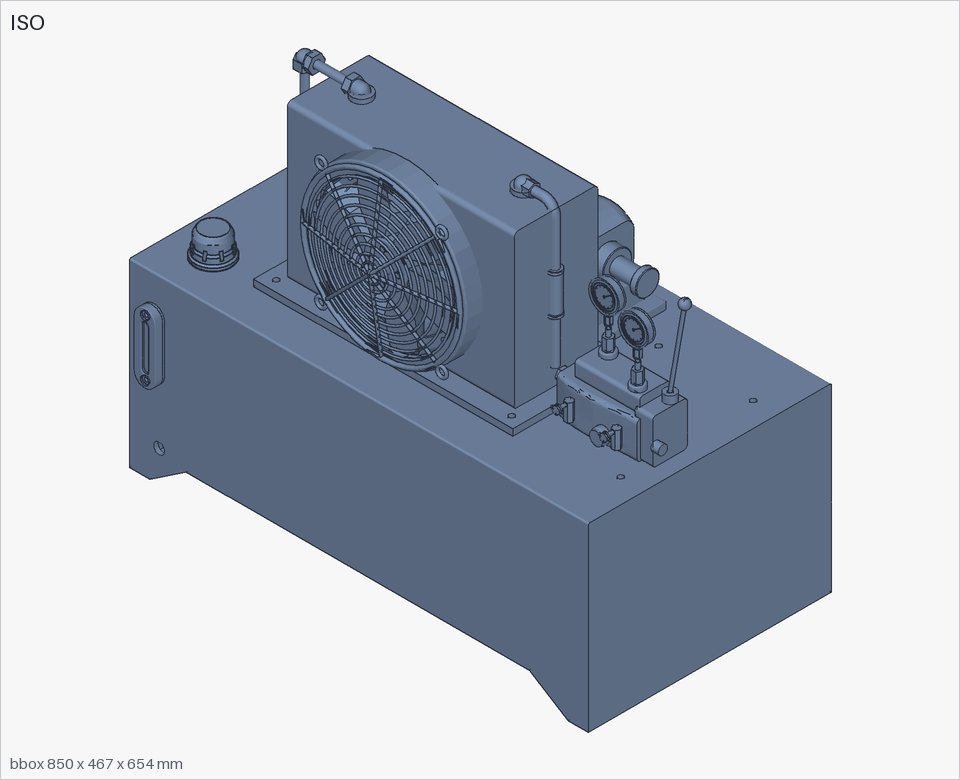}\\[-0.2em]{\scriptsize 86.8}} & \makecell{\includegraphics[width=\linewidth,height=0.58in,keepaspectratio,valign=c]{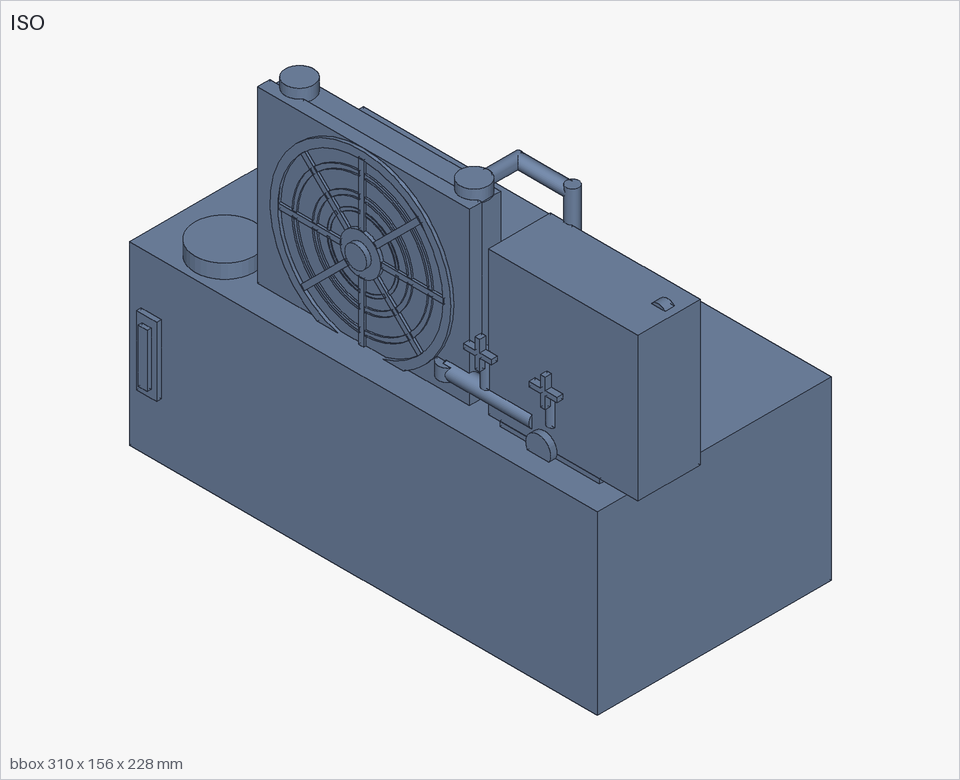}\\[-0.2em]{\scriptsize 64.5}} & \makecell{\includegraphics[width=\linewidth,height=0.58in,keepaspectratio,valign=c]{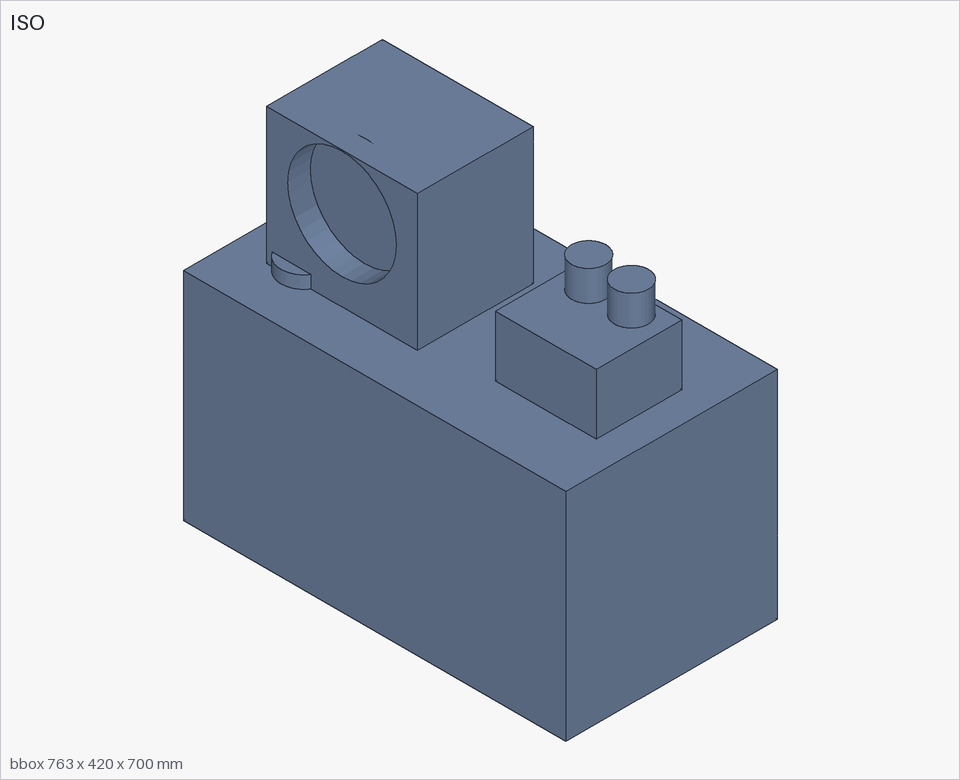}\\[-0.2em]{\scriptsize 45.2}} \\
\texttt{\scriptsize rcb\_000362639} & \makecell{\includegraphics[width=\linewidth,height=0.58in,keepaspectratio,valign=c]{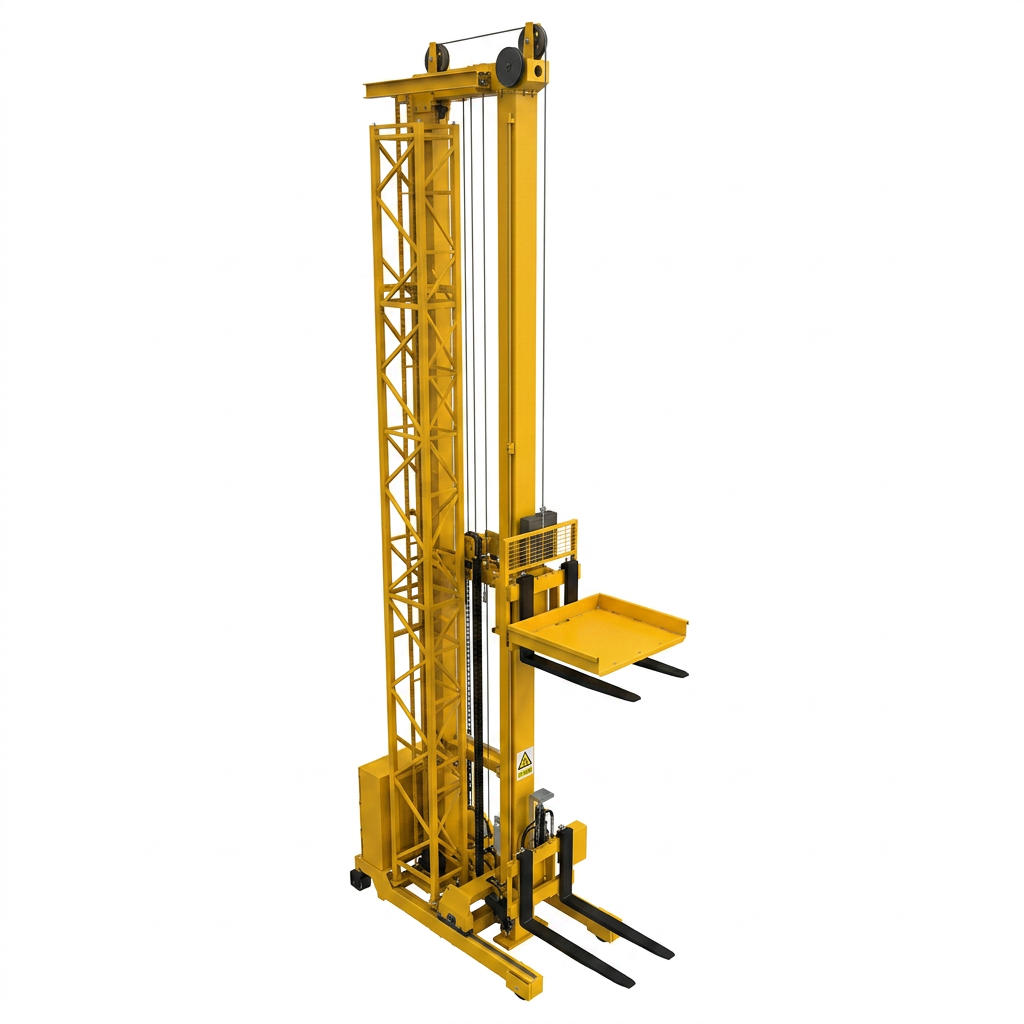}} & \makecell{\includegraphics[width=\linewidth,height=0.58in,keepaspectratio,valign=c]{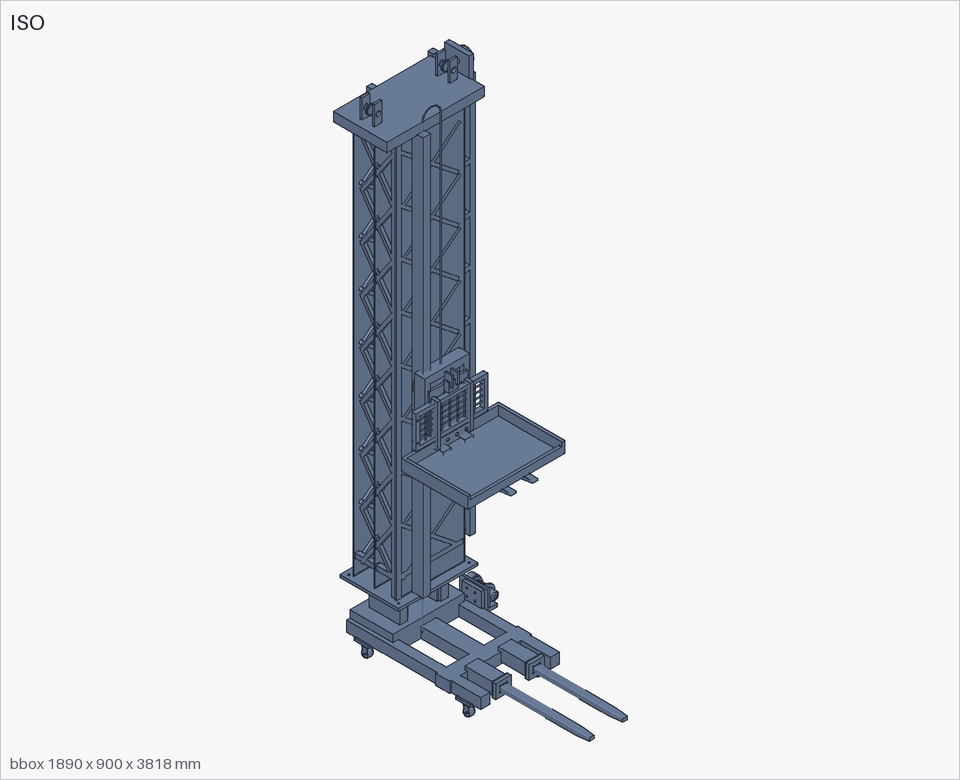}\\[-0.2em]{\scriptsize 82.6}} & \makecell{\includegraphics[width=\linewidth,height=0.58in,keepaspectratio,valign=c]{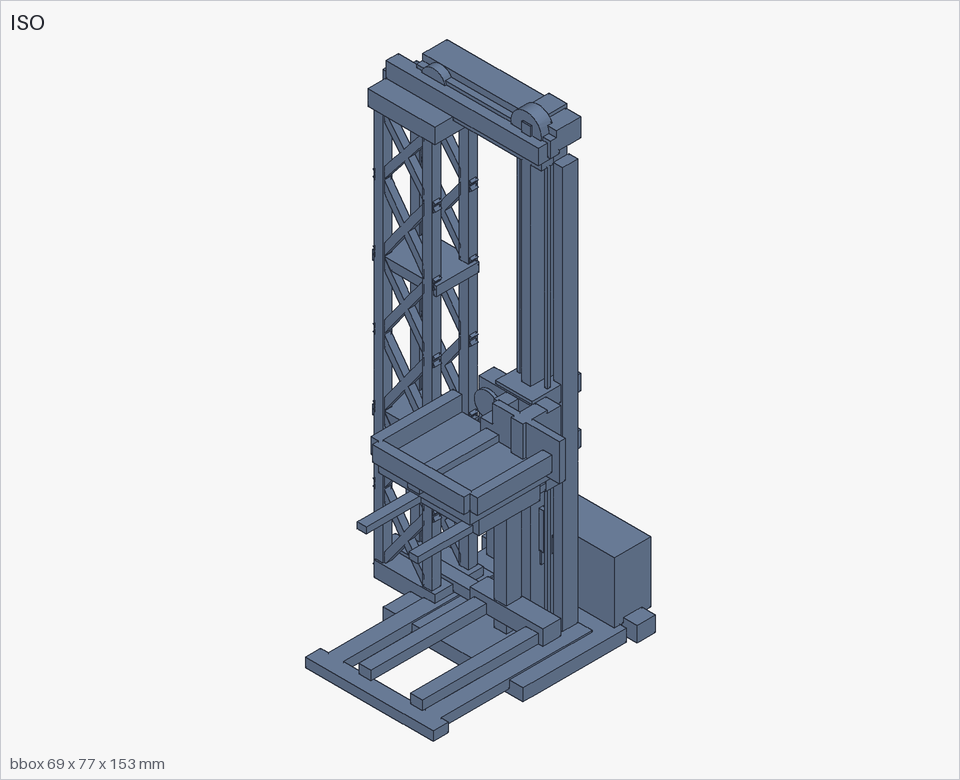}\\[-0.2em]{\scriptsize 79.1}} & \makecell{\includegraphics[width=\linewidth,height=0.58in,keepaspectratio,valign=c]{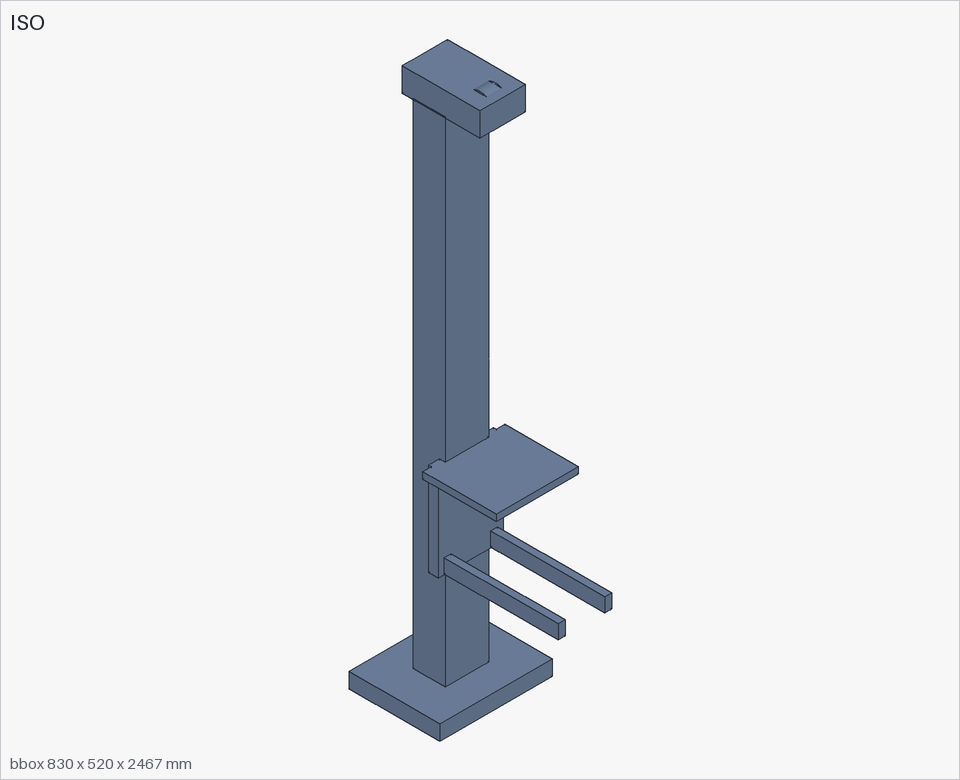}\\[-0.2em]{\scriptsize 41.2}} \\
\texttt{\scriptsize rcb\_000371197} & \makecell{\includegraphics[width=\linewidth,height=0.58in,keepaspectratio,valign=c]{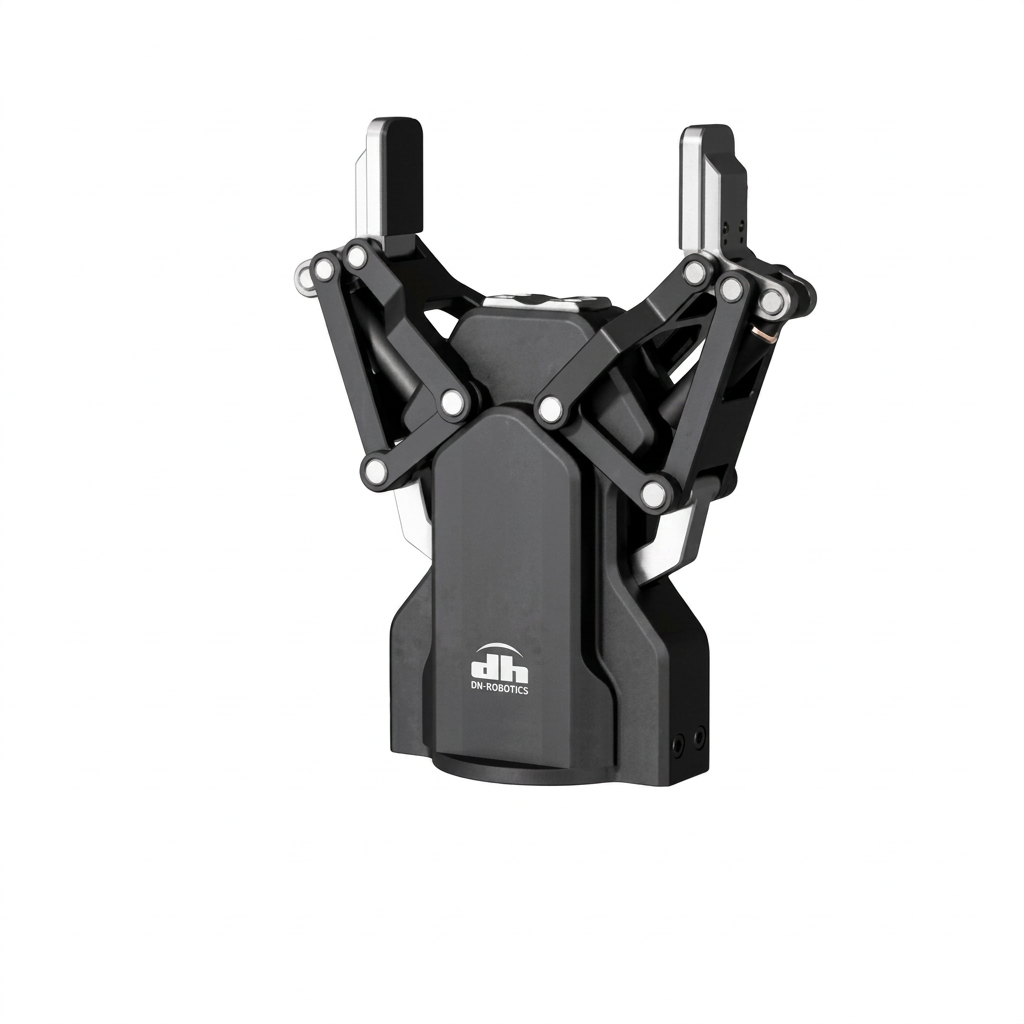}} & \makecell{\includegraphics[width=\linewidth,height=0.58in,keepaspectratio,valign=c]{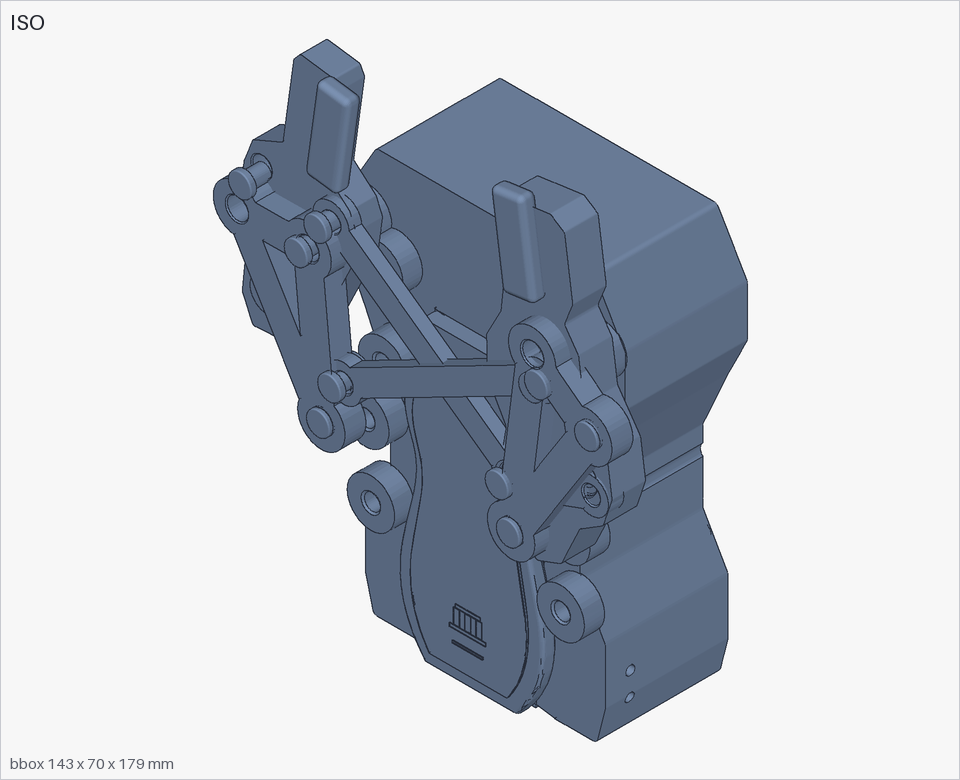}\\[-0.2em]{\scriptsize 83.3}} & \makecell{\includegraphics[width=\linewidth,height=0.58in,keepaspectratio,valign=c]{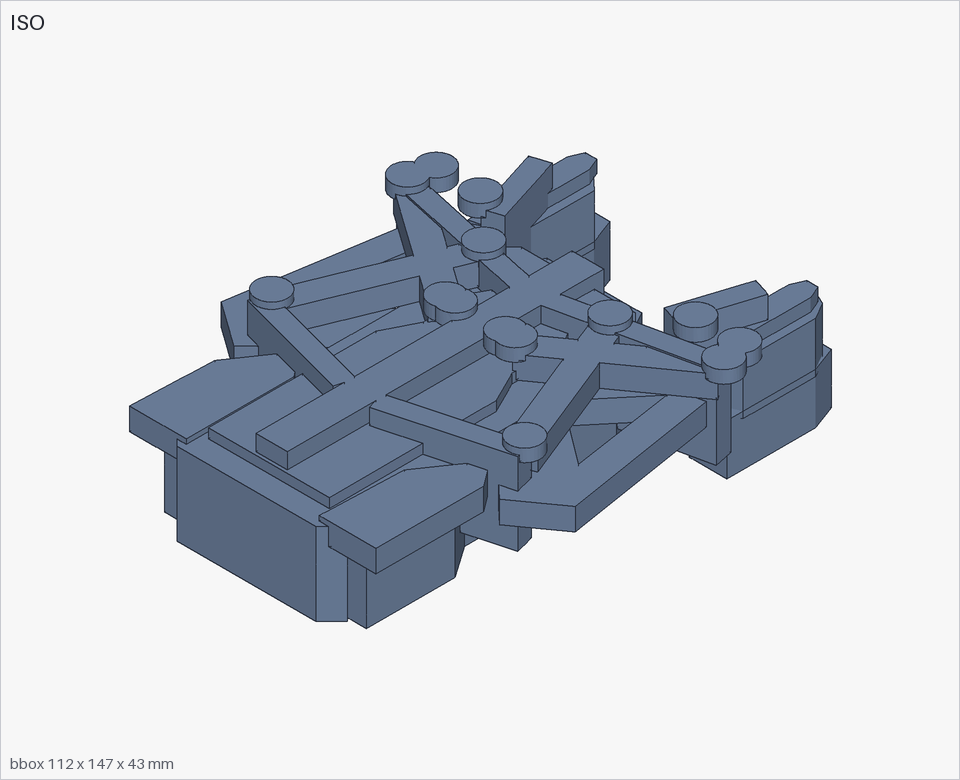}\\[-0.2em]{\scriptsize 72.3}} & \makecell{\includegraphics[width=\linewidth,height=0.58in,keepaspectratio,valign=c]{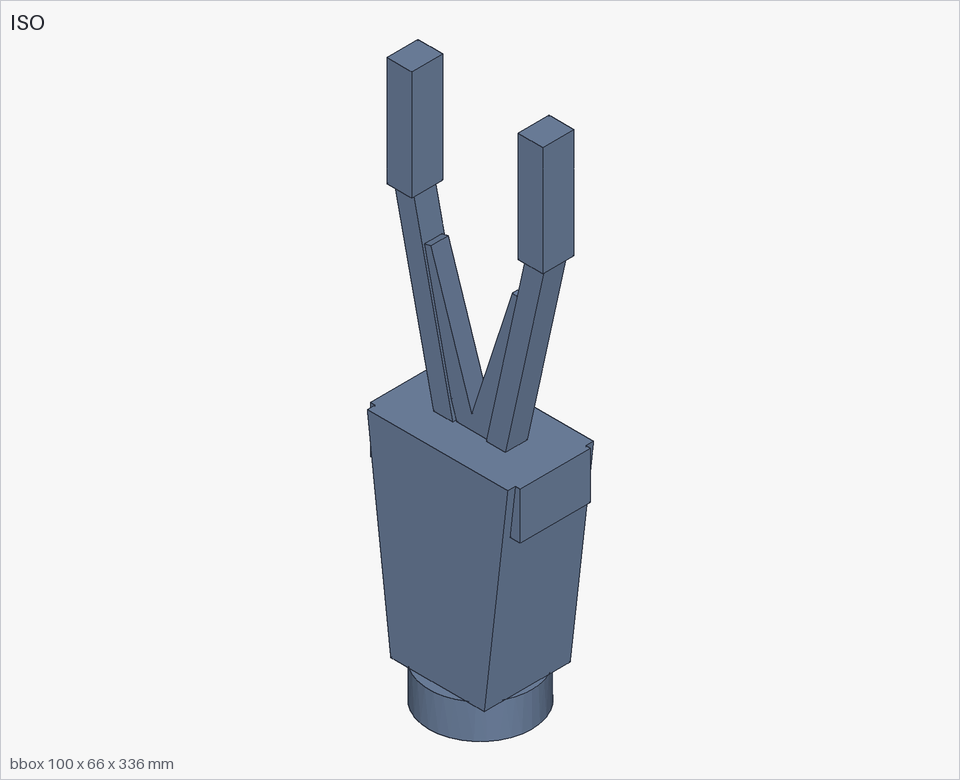}\\[-0.2em]{\scriptsize 45.6}} \\
\texttt{\scriptsize rcb\_000398895} & \makecell{\includegraphics[width=\linewidth,height=0.58in,keepaspectratio,valign=c]{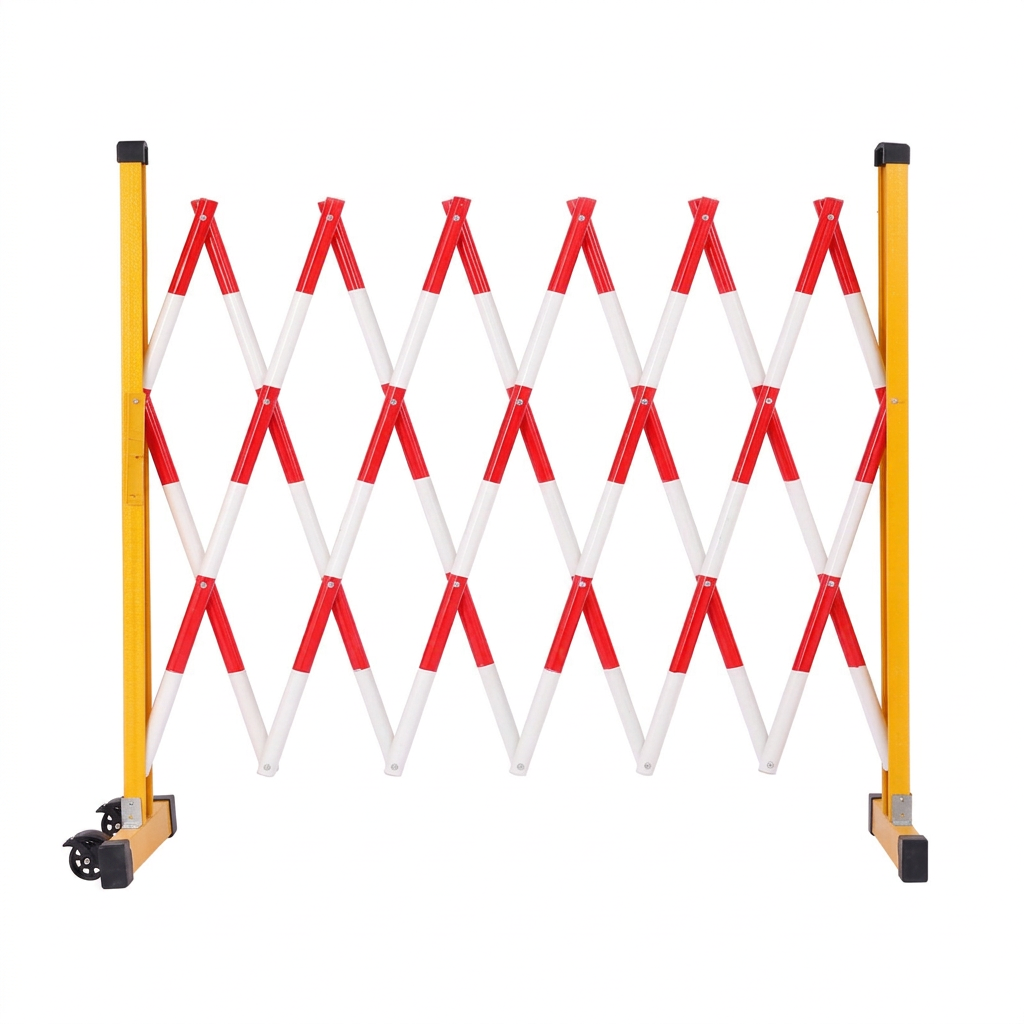}} & \makecell{\includegraphics[width=\linewidth,height=0.58in,keepaspectratio,valign=c]{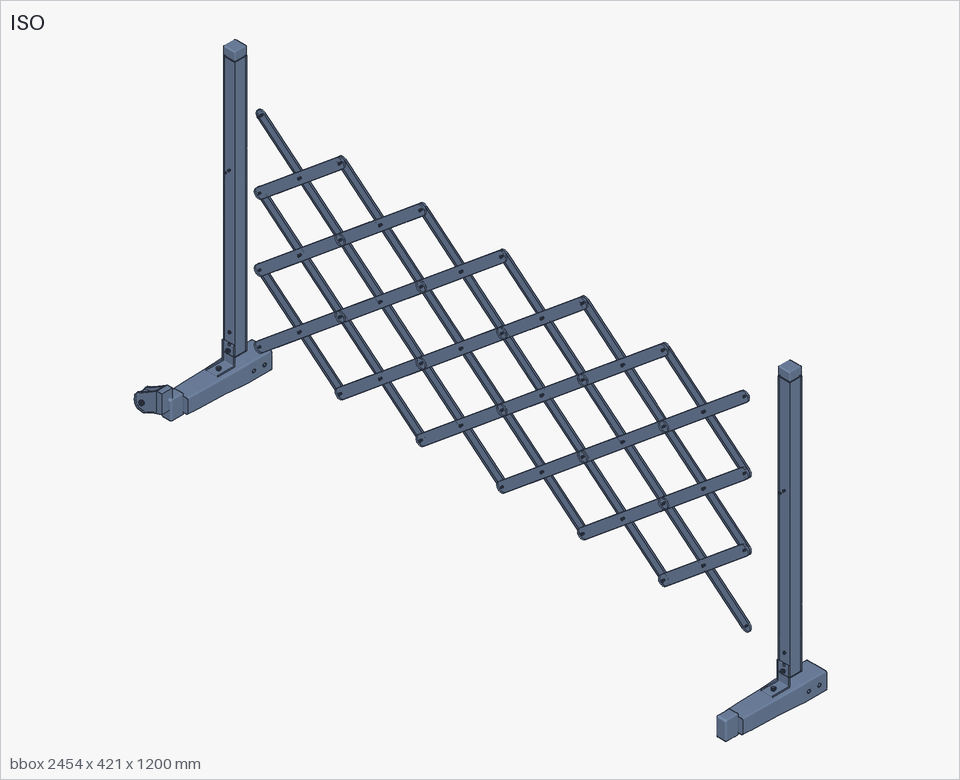}\\[-0.2em]{\scriptsize 77.9}} & \makecell{\includegraphics[width=\linewidth,height=0.58in,keepaspectratio,valign=c]{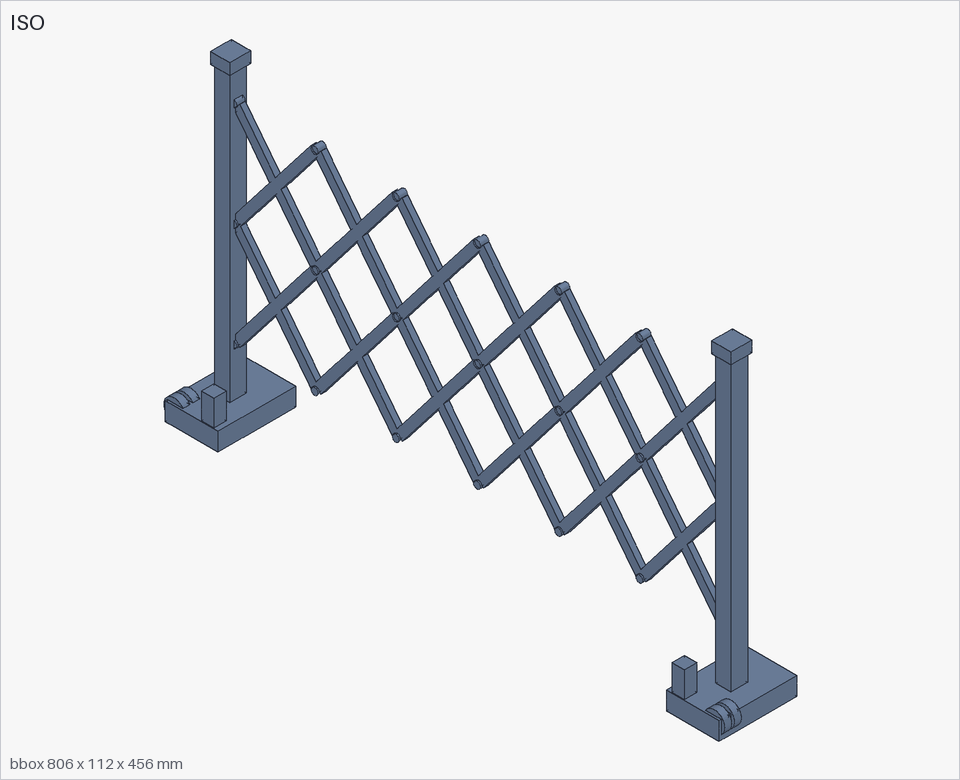}\\[-0.2em]{\scriptsize 83.7}} & \makecell{\includegraphics[width=\linewidth,height=0.58in,keepaspectratio,valign=c]{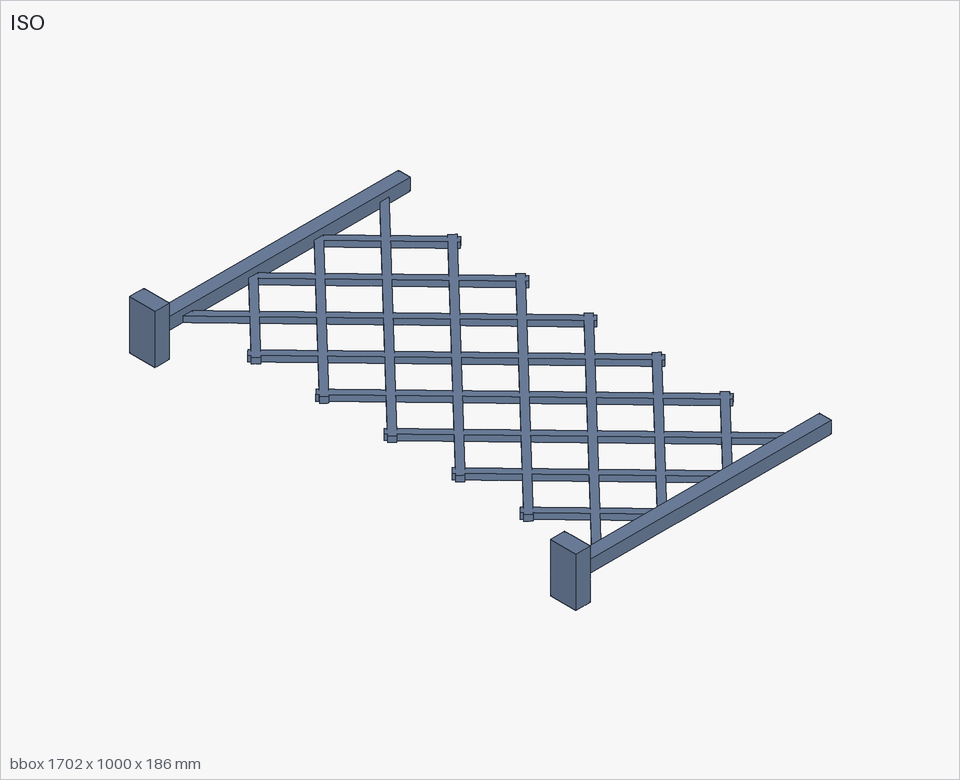}\\[-0.2em]{\scriptsize 56.0}} \\
\texttt{\scriptsize rcb\_000398996} & \makecell{\includegraphics[width=\linewidth,height=0.58in,keepaspectratio,valign=c]{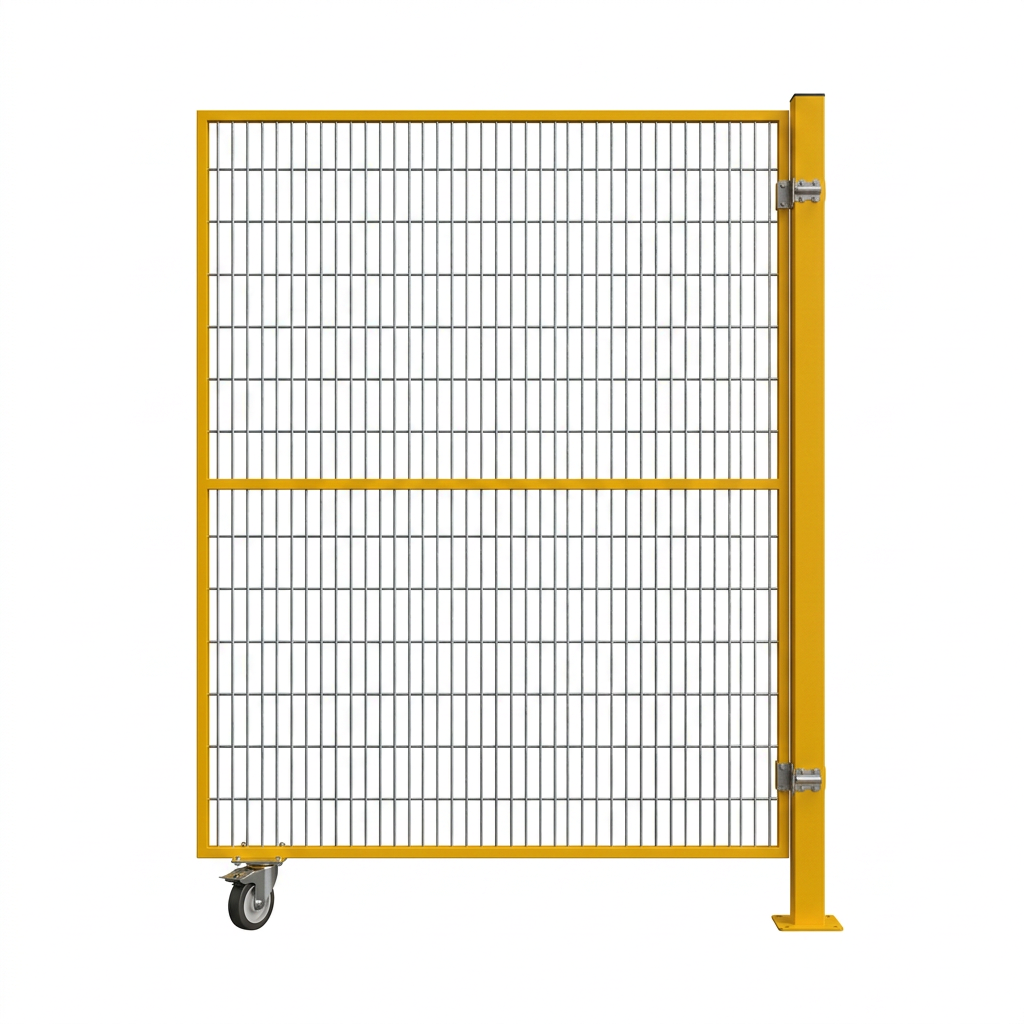}} & \makecell{\includegraphics[width=\linewidth,height=0.58in,keepaspectratio,valign=c]{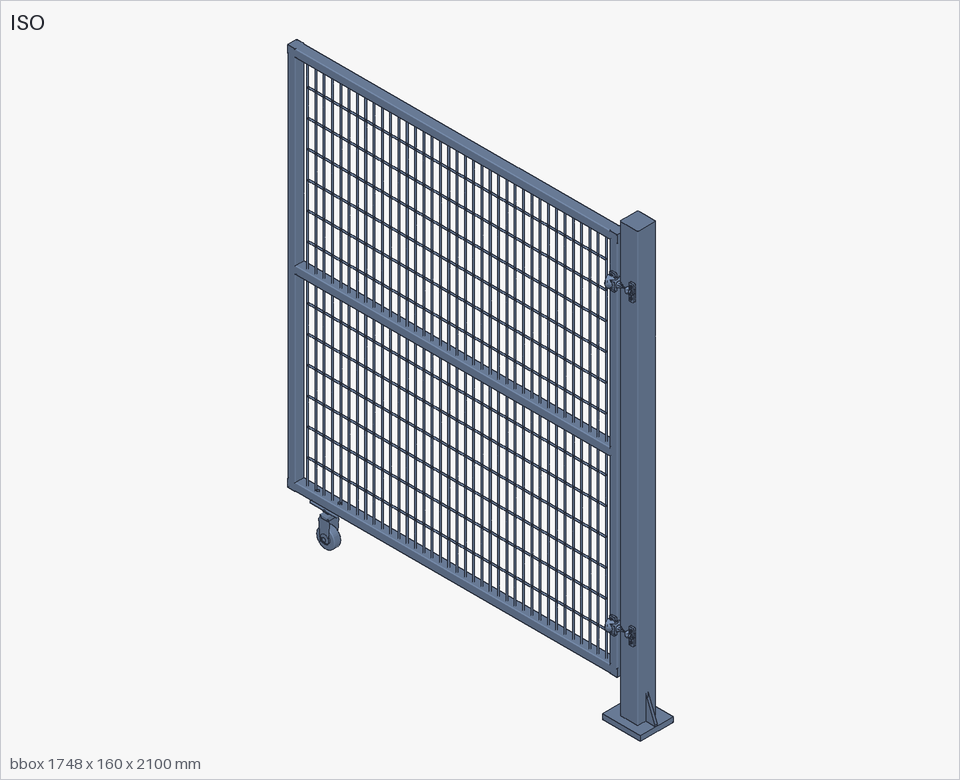}\\[-0.2em]{\scriptsize 90.3}} & \makecell{\includegraphics[width=\linewidth,height=0.58in,keepaspectratio,valign=c]{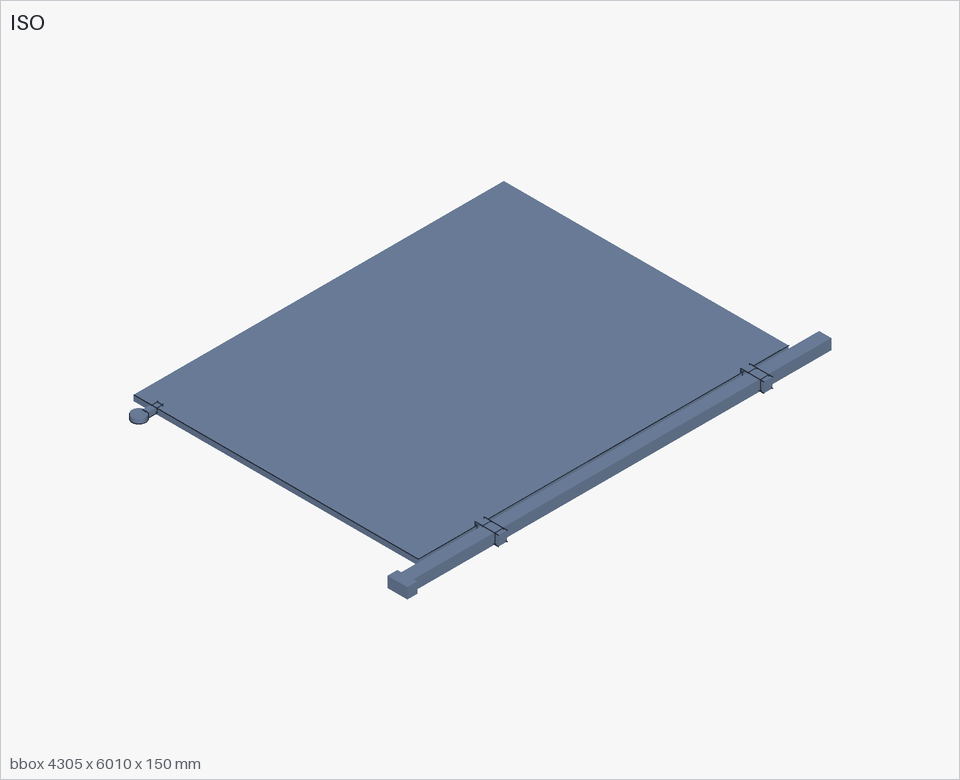}\\[-0.2em]{\scriptsize 45.2}} & \makecell{\includegraphics[width=\linewidth,height=0.58in,keepaspectratio,valign=c]{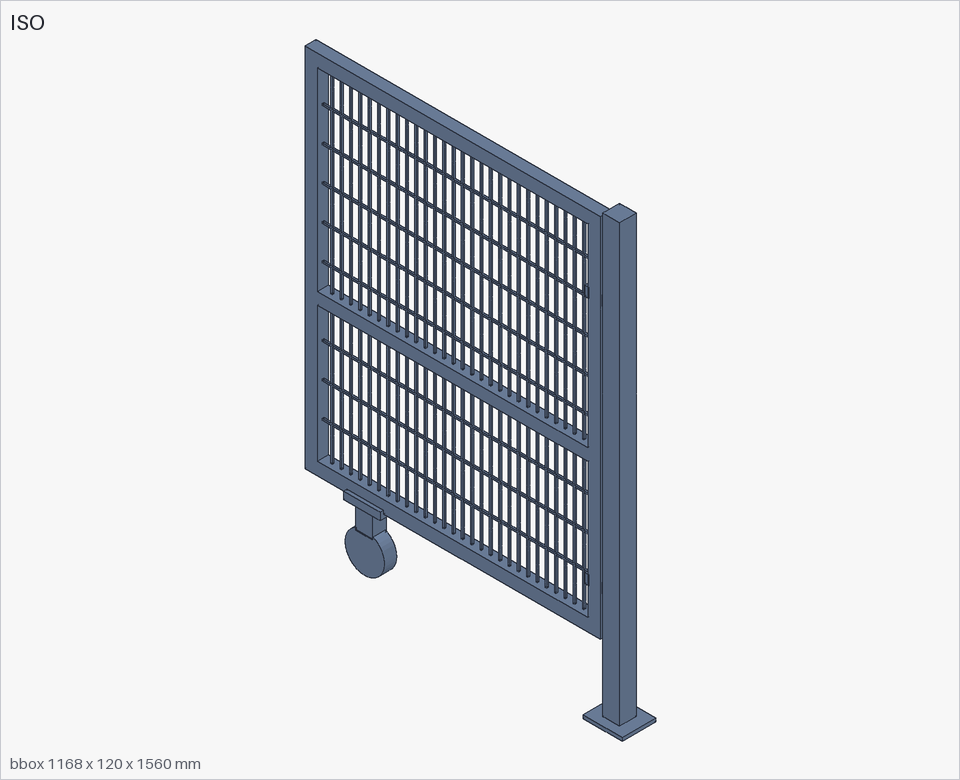}\\[-0.2em]{\scriptsize 85.9}} \\
\texttt{\scriptsize rcb\_000514100} & \makecell{\includegraphics[width=\linewidth,height=0.58in,keepaspectratio,valign=c]{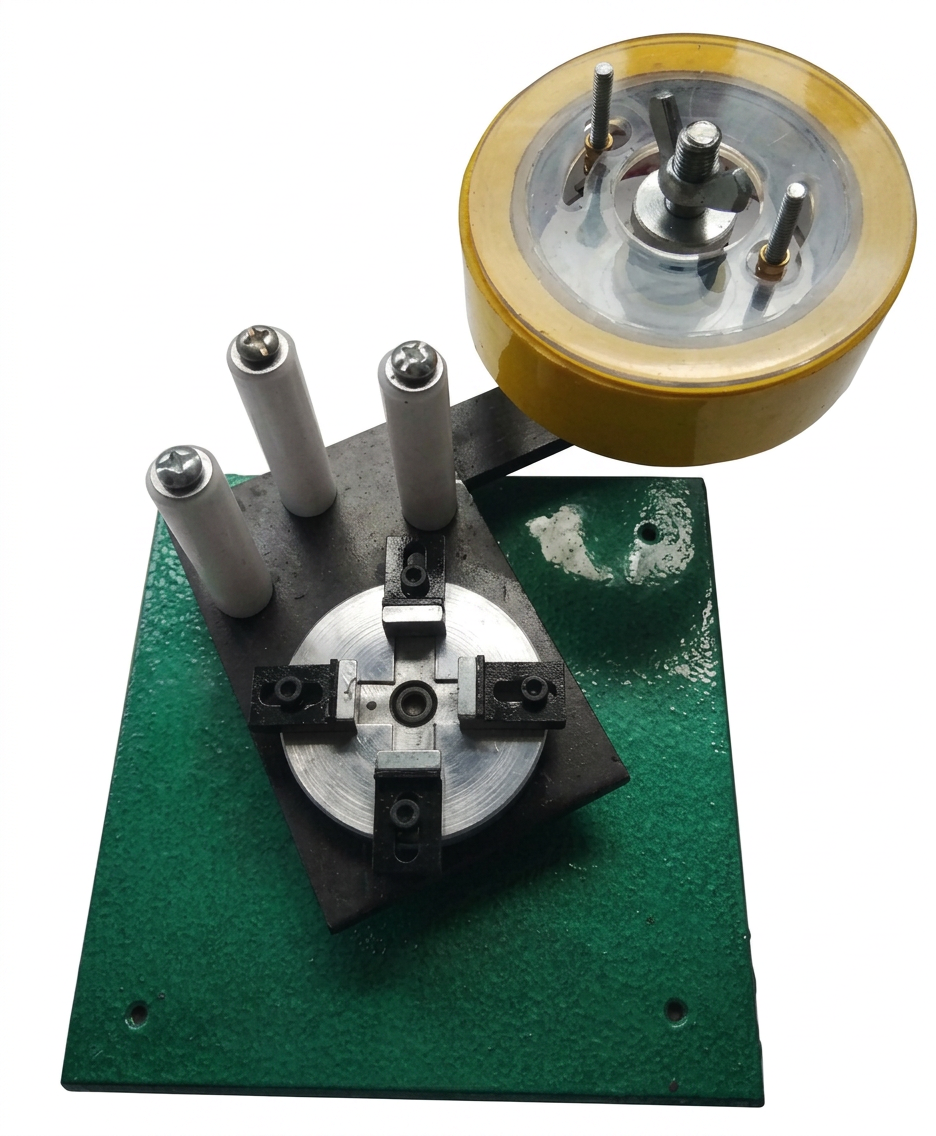}} & \makecell{\includegraphics[width=\linewidth,height=0.58in,keepaspectratio,valign=c]{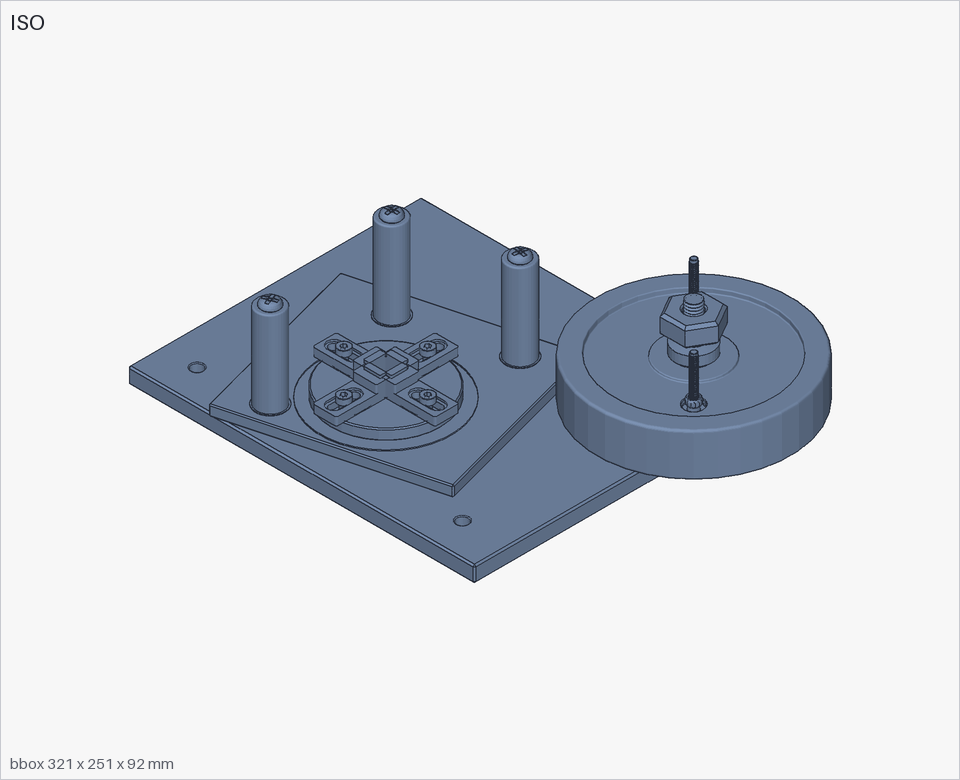}\\[-0.2em]{\scriptsize 85.5}} & \makecell{\includegraphics[width=\linewidth,height=0.58in,keepaspectratio,valign=c]{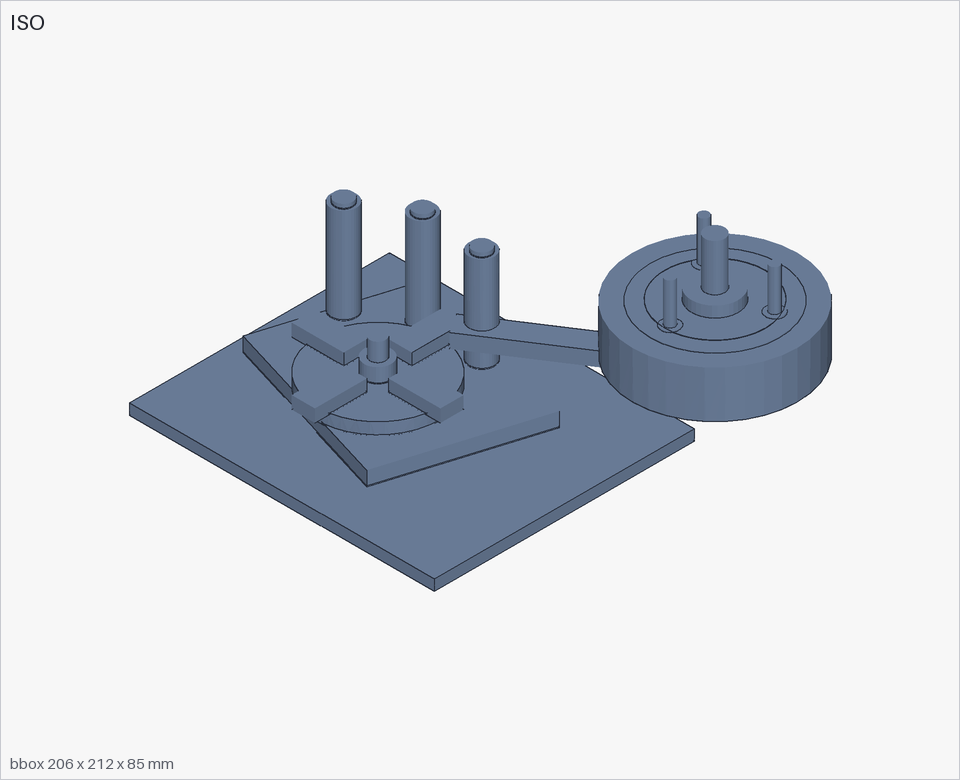}\\[-0.2em]{\scriptsize 77.0}} & \makecell{\includegraphics[width=\linewidth,height=0.58in,keepaspectratio,valign=c]{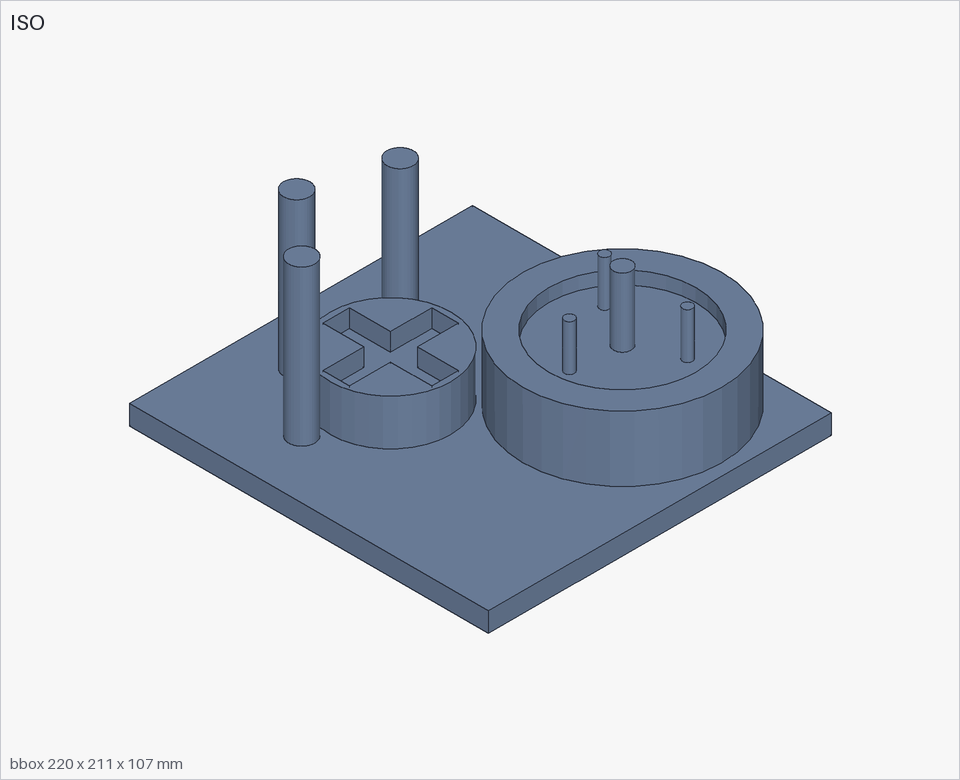}\\[-0.2em]{\scriptsize 52.8}} \\
\texttt{\scriptsize rcb\_000529447} & \makecell{\includegraphics[width=\linewidth,height=0.58in,keepaspectratio,valign=c]{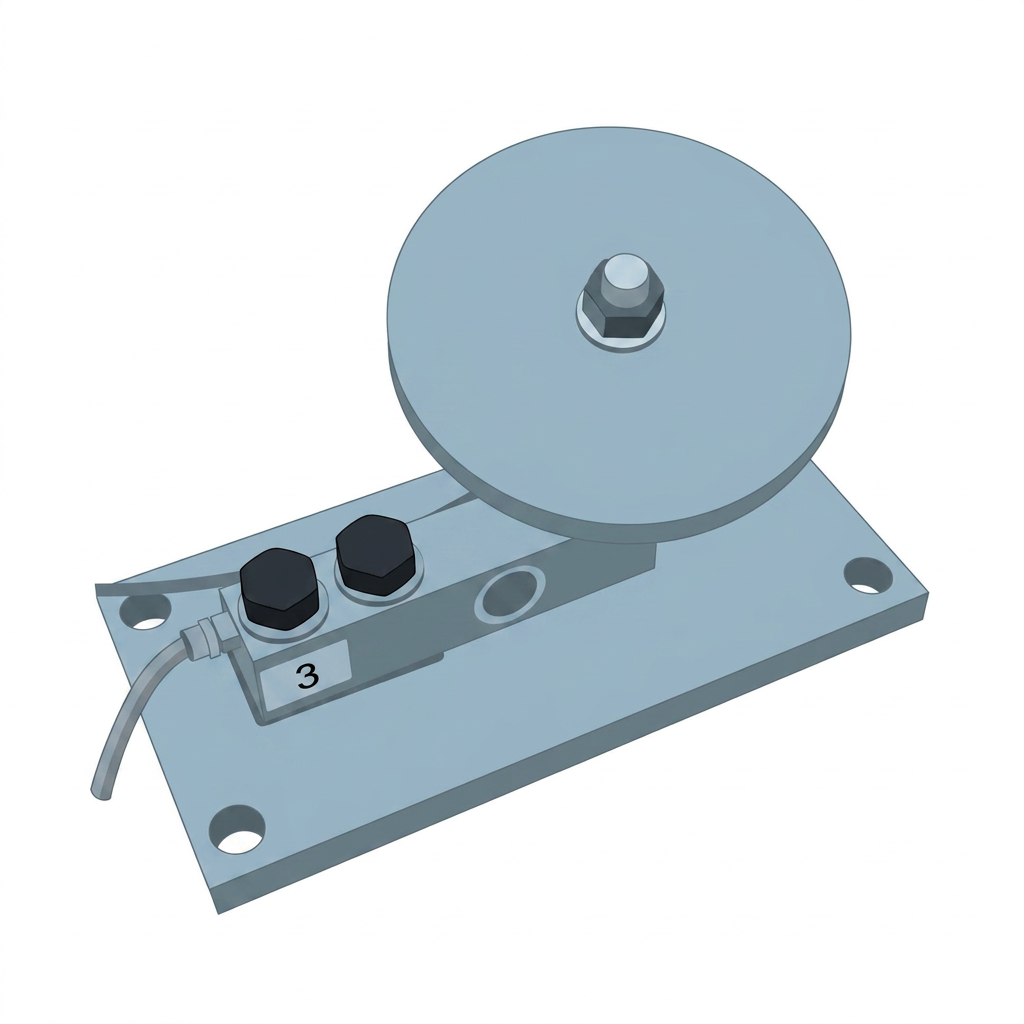}} & \makecell{\includegraphics[width=\linewidth,height=0.58in,keepaspectratio,valign=c]{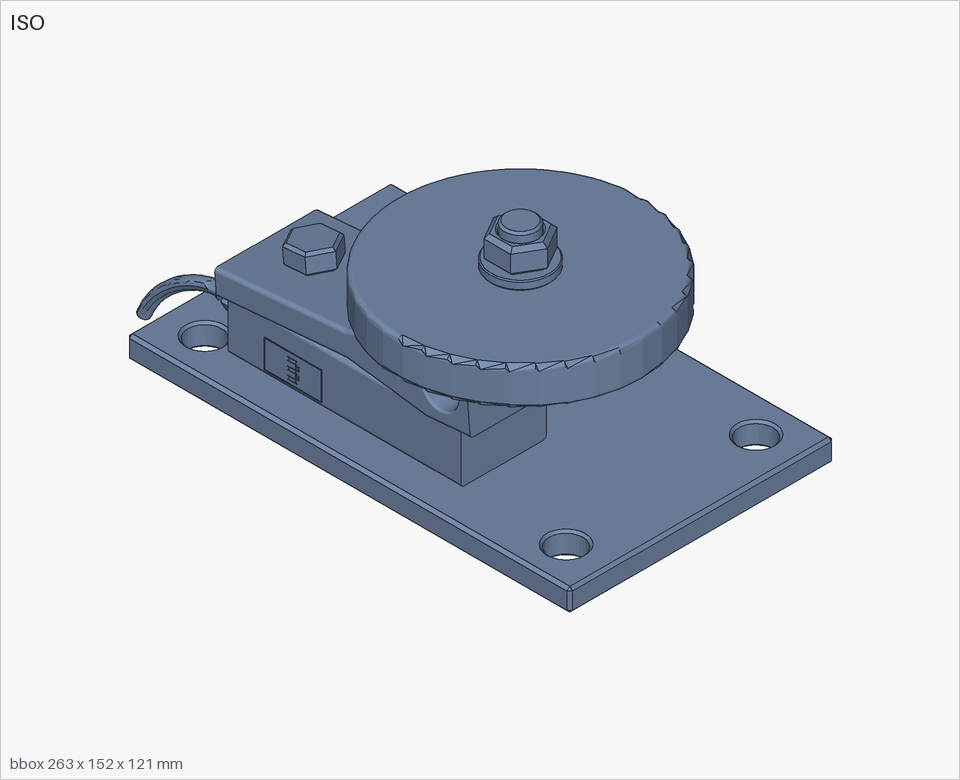}\\[-0.2em]{\scriptsize 87.6}} & \makecell{\includegraphics[width=\linewidth,height=0.58in,keepaspectratio,valign=c]{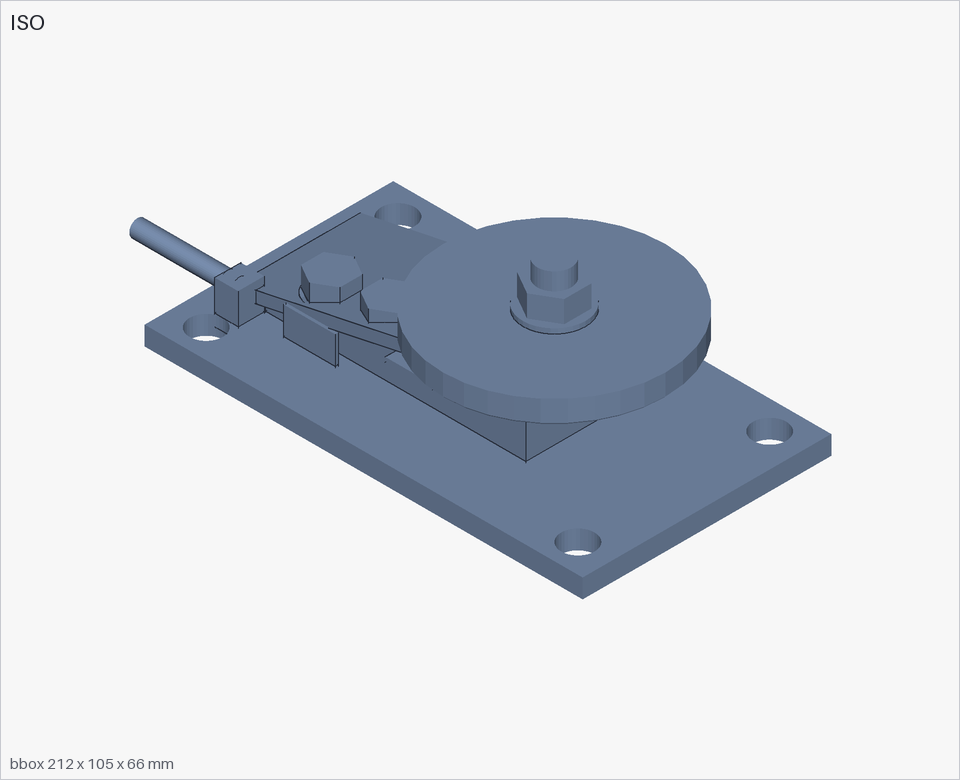}\\[-0.2em]{\scriptsize 80.4}} & \makecell{\includegraphics[width=\linewidth,height=0.58in,keepaspectratio,valign=c]{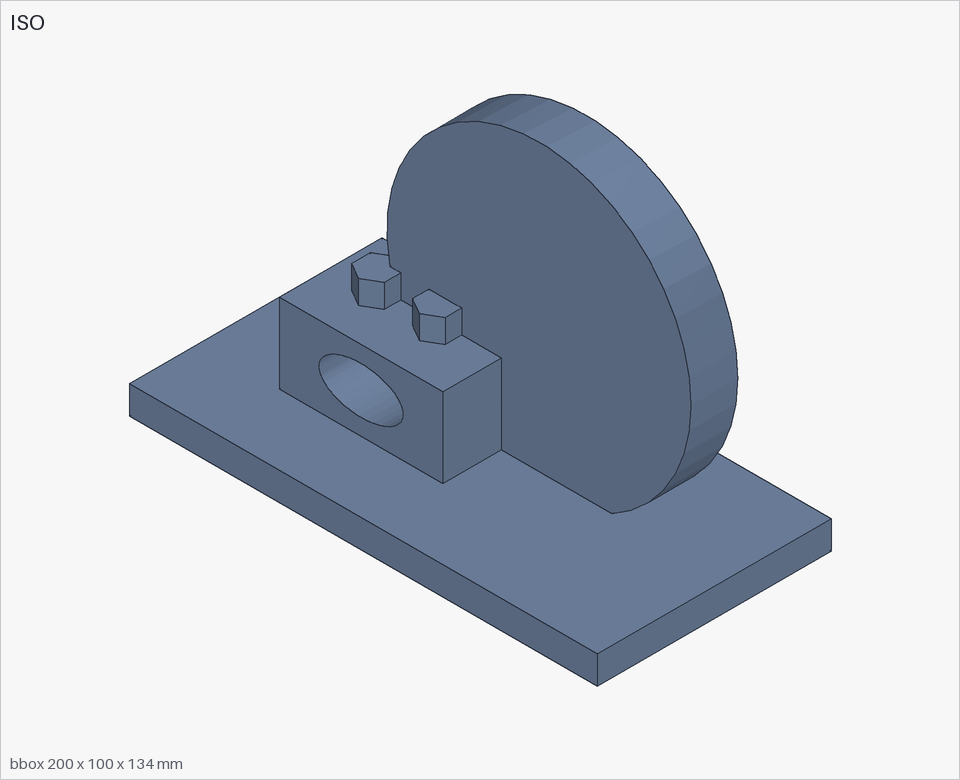}\\[-0.2em]{\scriptsize 52.8}} \\
\texttt{\scriptsize rcb\_000539368} & \makecell{\includegraphics[width=\linewidth,height=0.58in,keepaspectratio,valign=c]{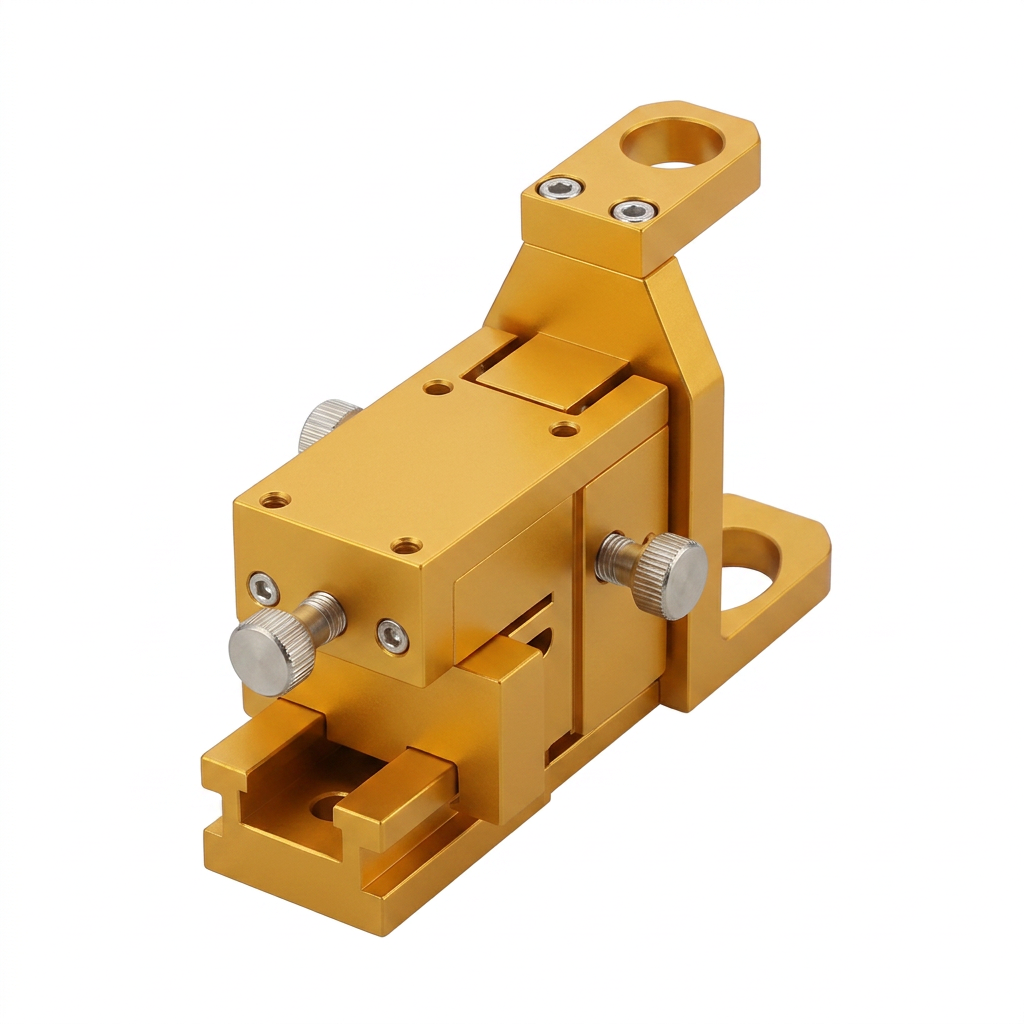}} & \makecell{\includegraphics[width=\linewidth,height=0.58in,keepaspectratio,valign=c]{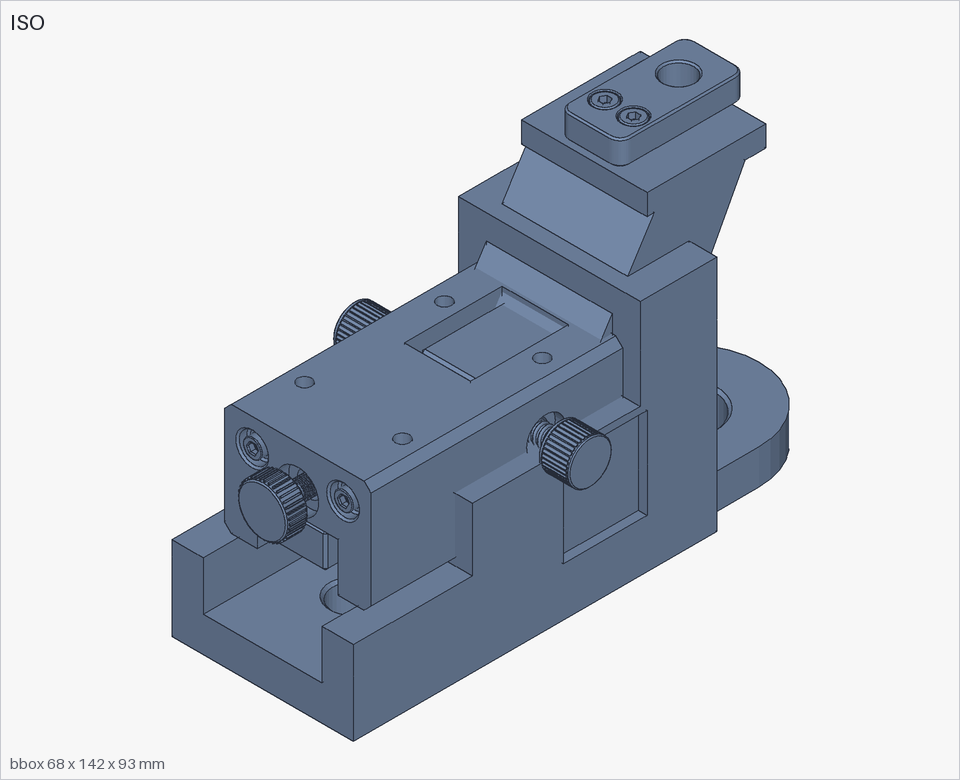}\\[-0.2em]{\scriptsize 83.1}} & \makecell{\includegraphics[width=\linewidth,height=0.58in,keepaspectratio,valign=c]{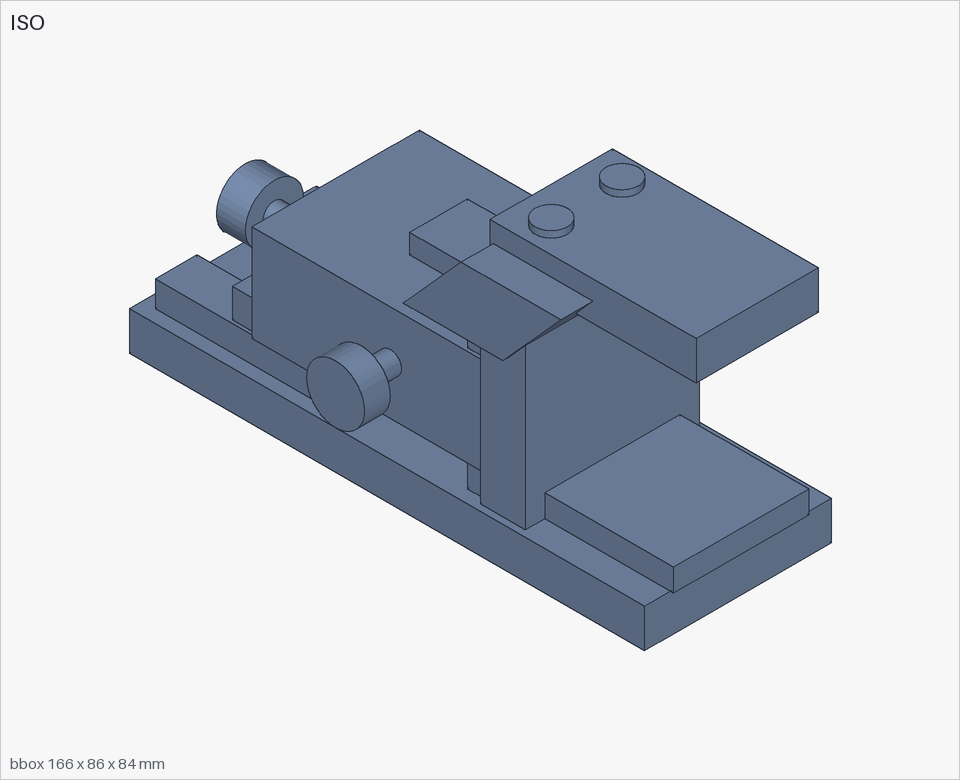}\\[-0.2em]{\scriptsize 60.2}} & \makecell{\includegraphics[width=\linewidth,height=0.58in,keepaspectratio,valign=c]{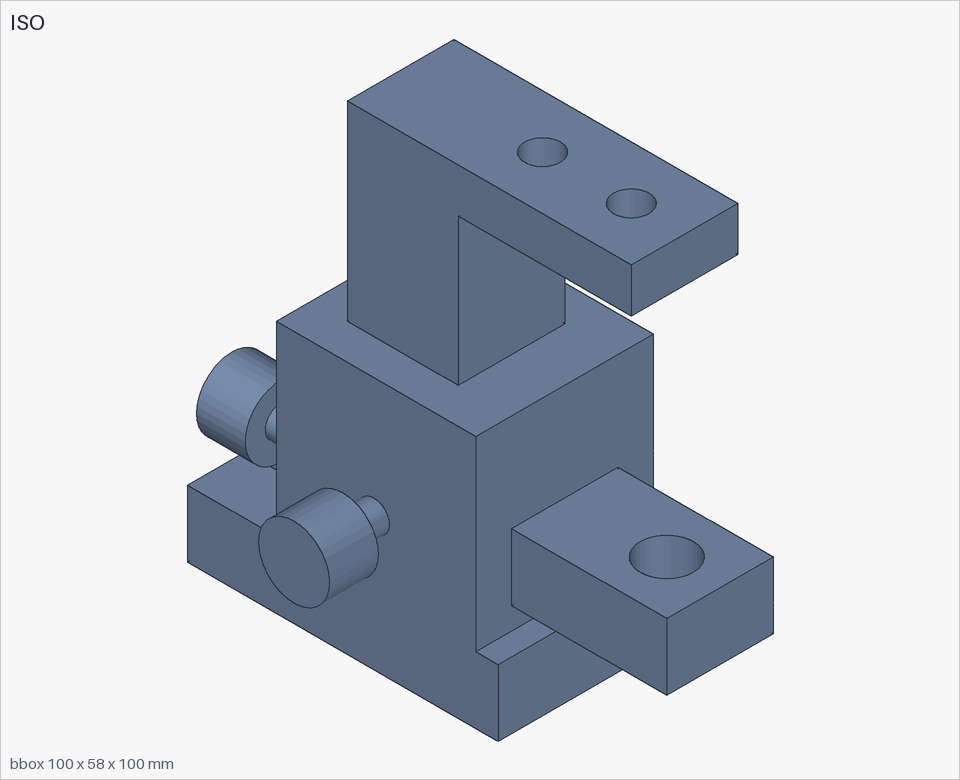}\\[-0.2em]{\scriptsize 60.0}} \\
\texttt{\scriptsize rcb\_000540054} & \makecell{\includegraphics[width=\linewidth,height=0.58in,keepaspectratio,valign=c]{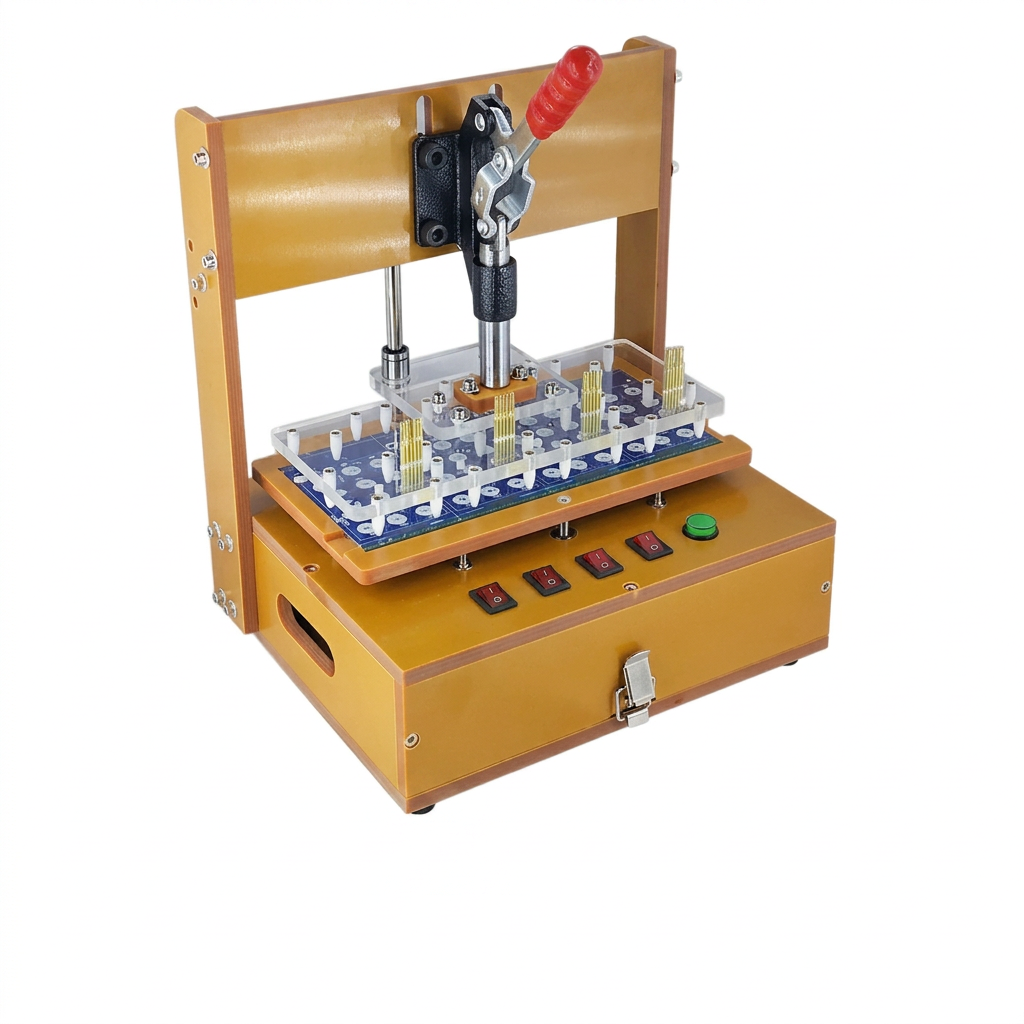}} & \makecell{\includegraphics[width=\linewidth,height=0.58in,keepaspectratio,valign=c]{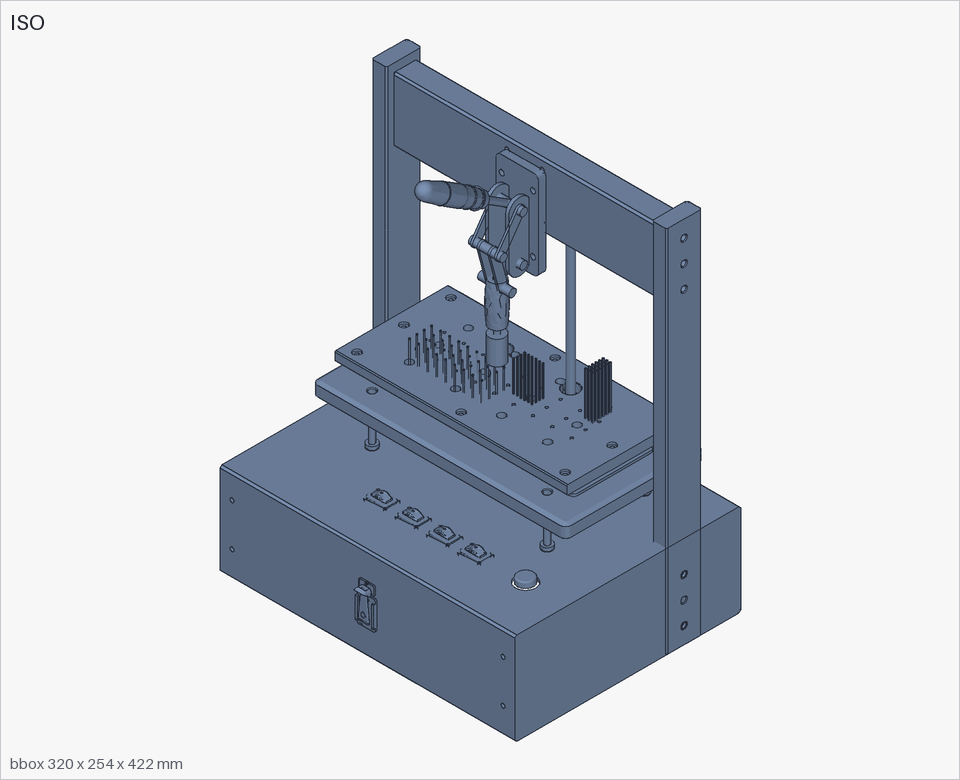}\\[-0.2em]{\scriptsize 88.0}} & \makecell{\includegraphics[width=\linewidth,height=0.58in,keepaspectratio,valign=c]{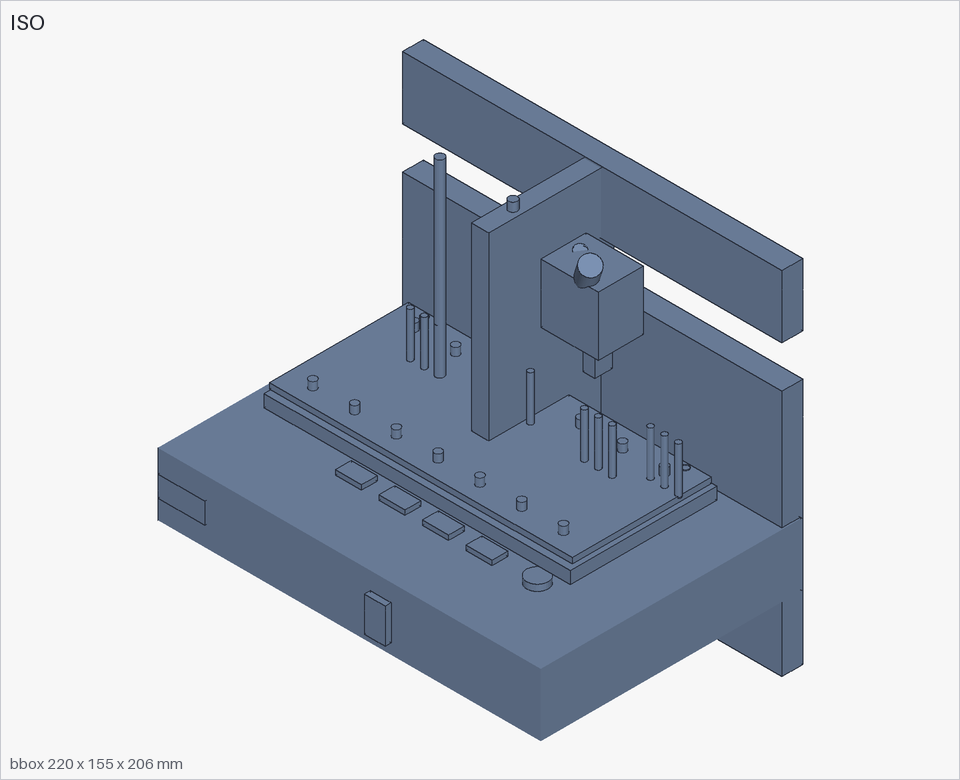}\\[-0.2em]{\scriptsize 79.4}} & \makecell{\includegraphics[width=\linewidth,height=0.58in,keepaspectratio,valign=c]{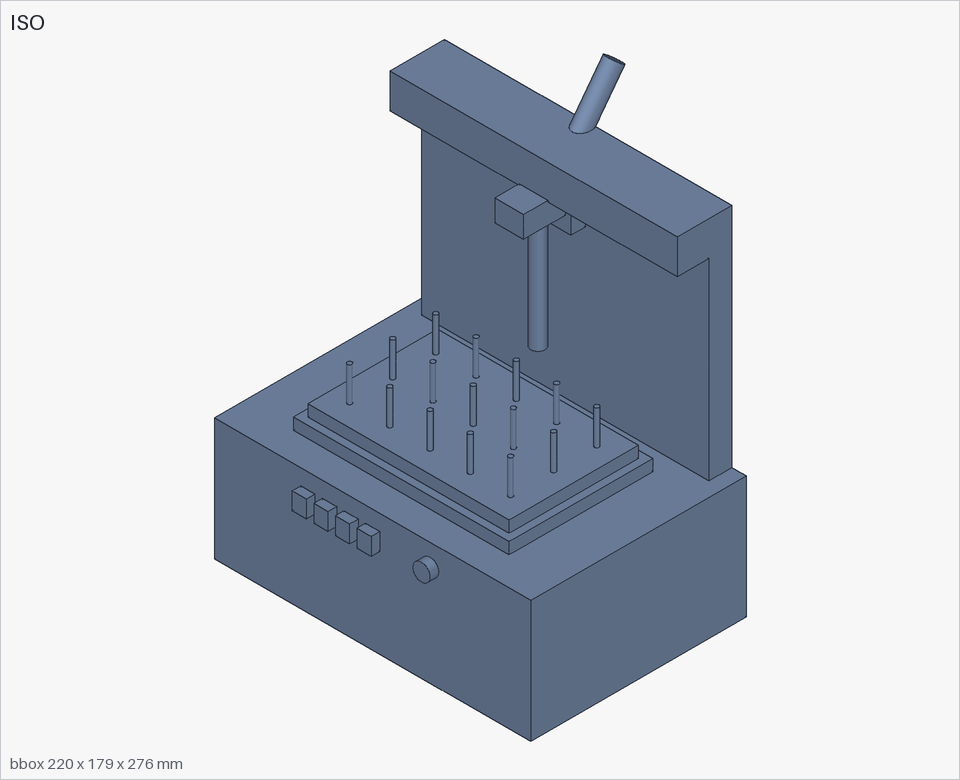}\\[-0.2em]{\scriptsize 72.3}} \\
\texttt{\scriptsize rcb\_000575344} & \makecell{\includegraphics[width=\linewidth,height=0.58in,keepaspectratio,valign=c]{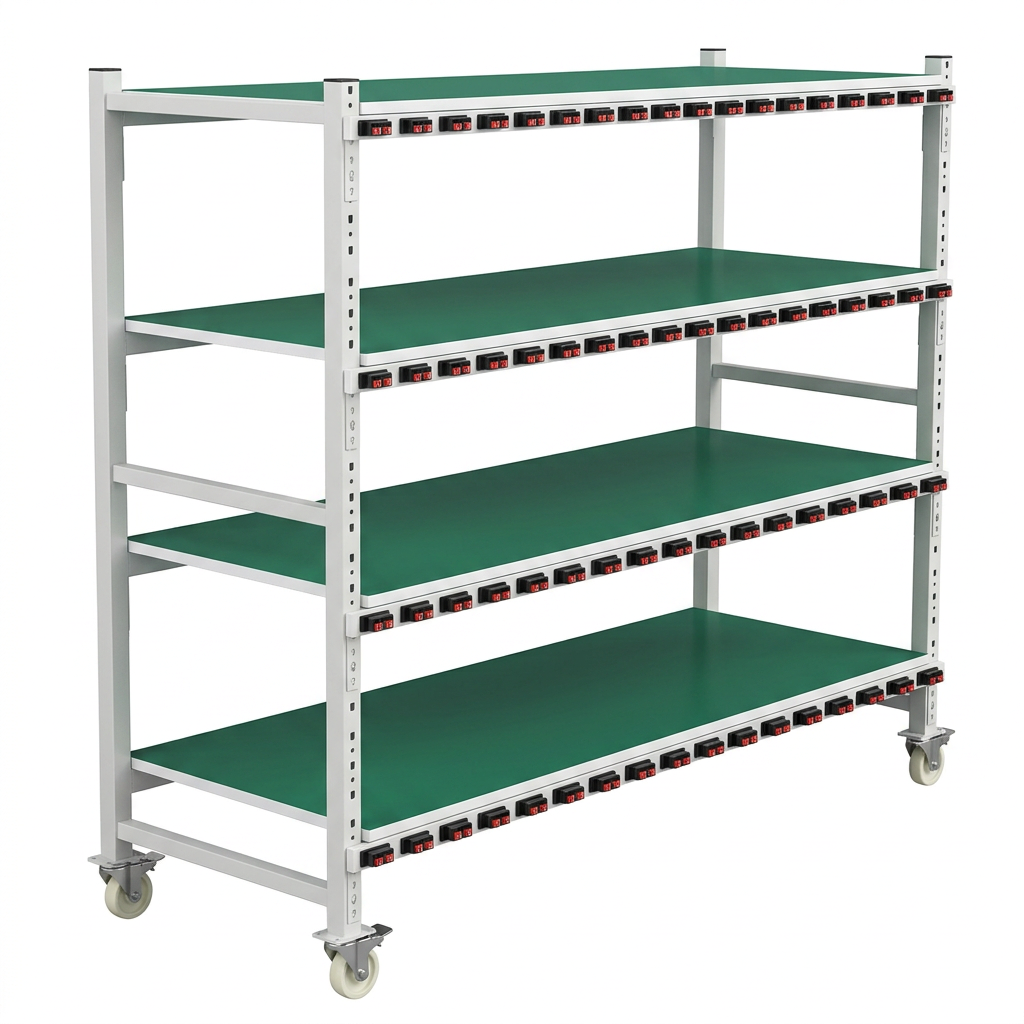}} & \makecell{\includegraphics[width=\linewidth,height=0.58in,keepaspectratio,valign=c]{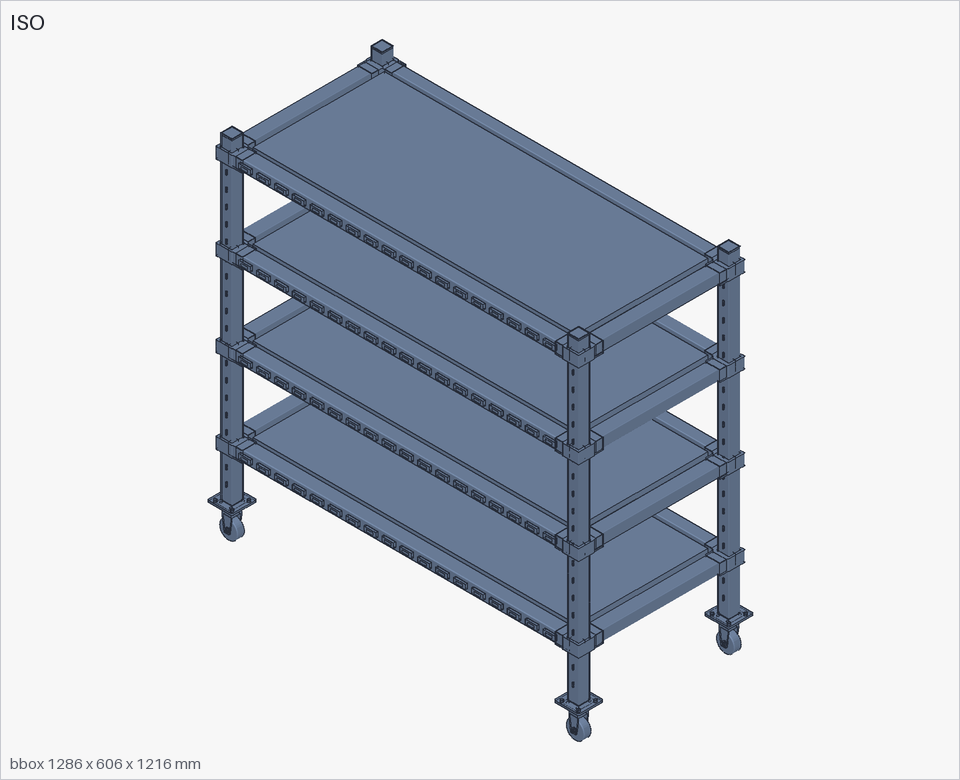}\\[-0.2em]{\scriptsize 88.5}} & \makecell{\includegraphics[width=\linewidth,height=0.58in,keepaspectratio,valign=c]{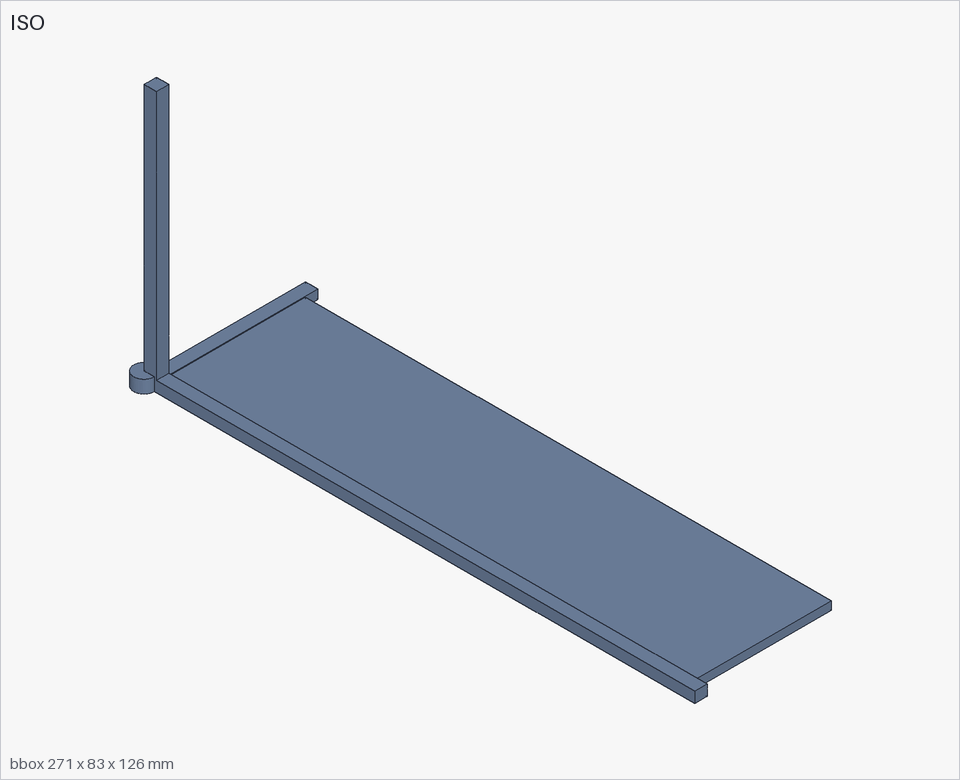}\\[-0.2em]{\scriptsize 20.1}} & \makecell{\includegraphics[width=\linewidth,height=0.58in,keepaspectratio,valign=c]{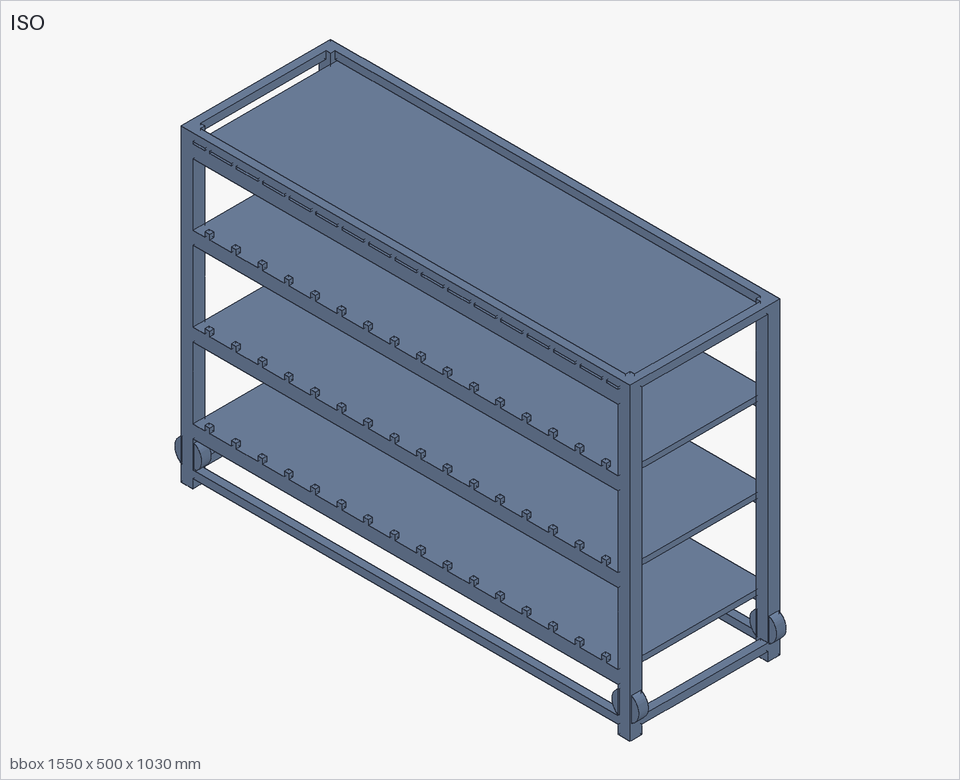}\\[-0.2em]{\scriptsize 75.1}} \\
\texttt{\scriptsize rcb\_000580852} & \makecell{\includegraphics[width=\linewidth,height=0.58in,keepaspectratio,valign=c]{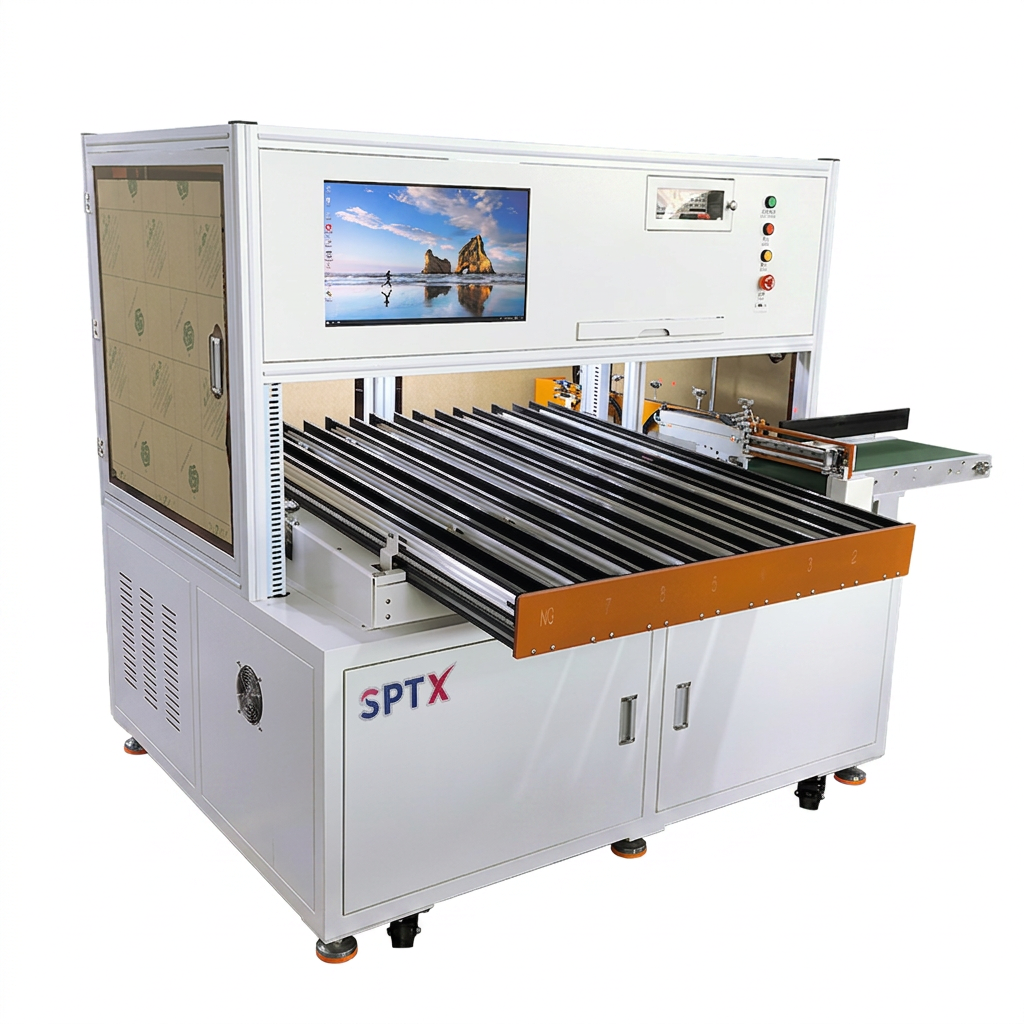}} & \makecell{\raisebox{0.18in}{\color{gray}\scriptsize no export}} & \makecell{\includegraphics[width=\linewidth,height=0.58in,keepaspectratio,valign=c]{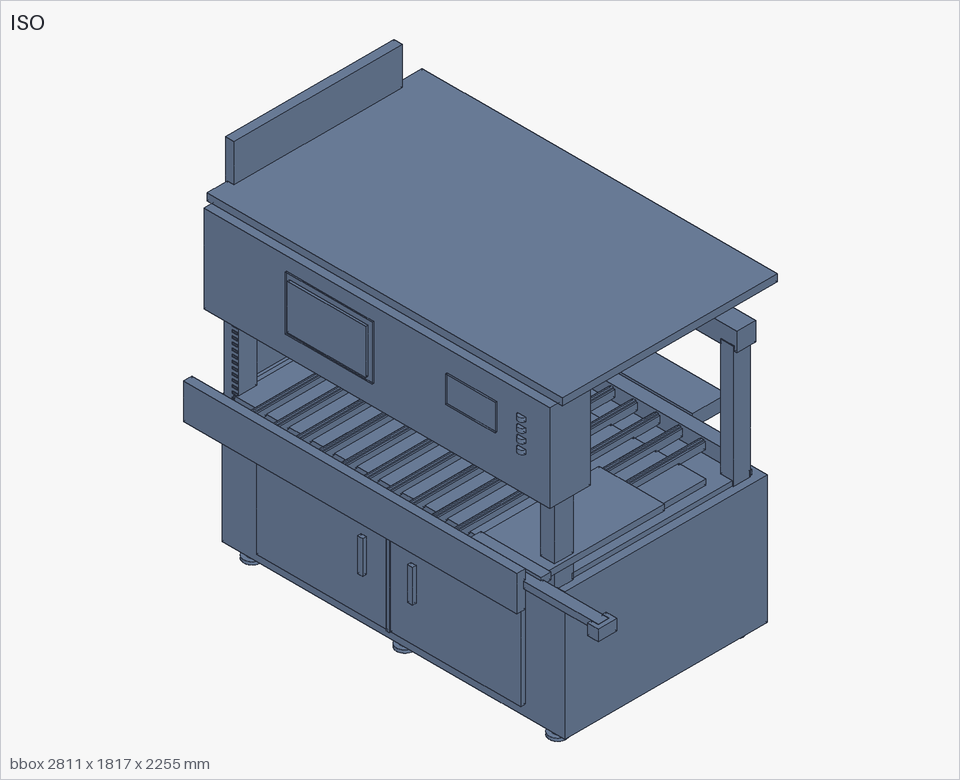}\\[-0.2em]{\scriptsize 74.2}} & \makecell{\includegraphics[width=\linewidth,height=0.58in,keepaspectratio,valign=c]{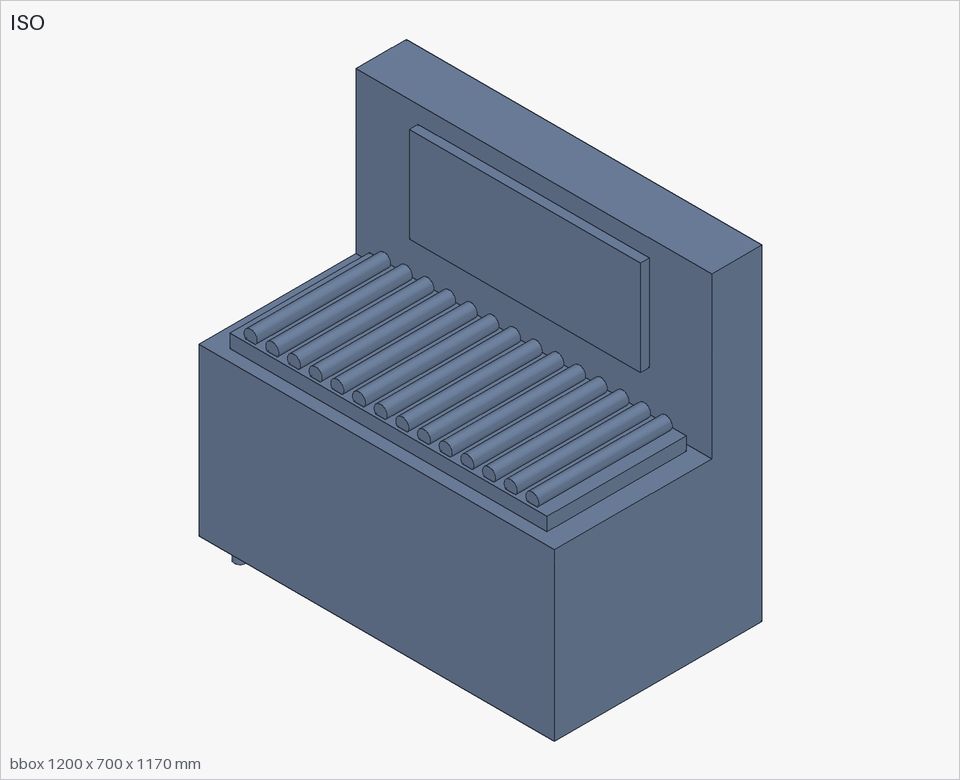}\\[-0.2em]{\scriptsize 41.7}} \\
\texttt{\scriptsize rcb\_000613854} & \makecell{\includegraphics[width=\linewidth,height=0.58in,keepaspectratio,valign=c]{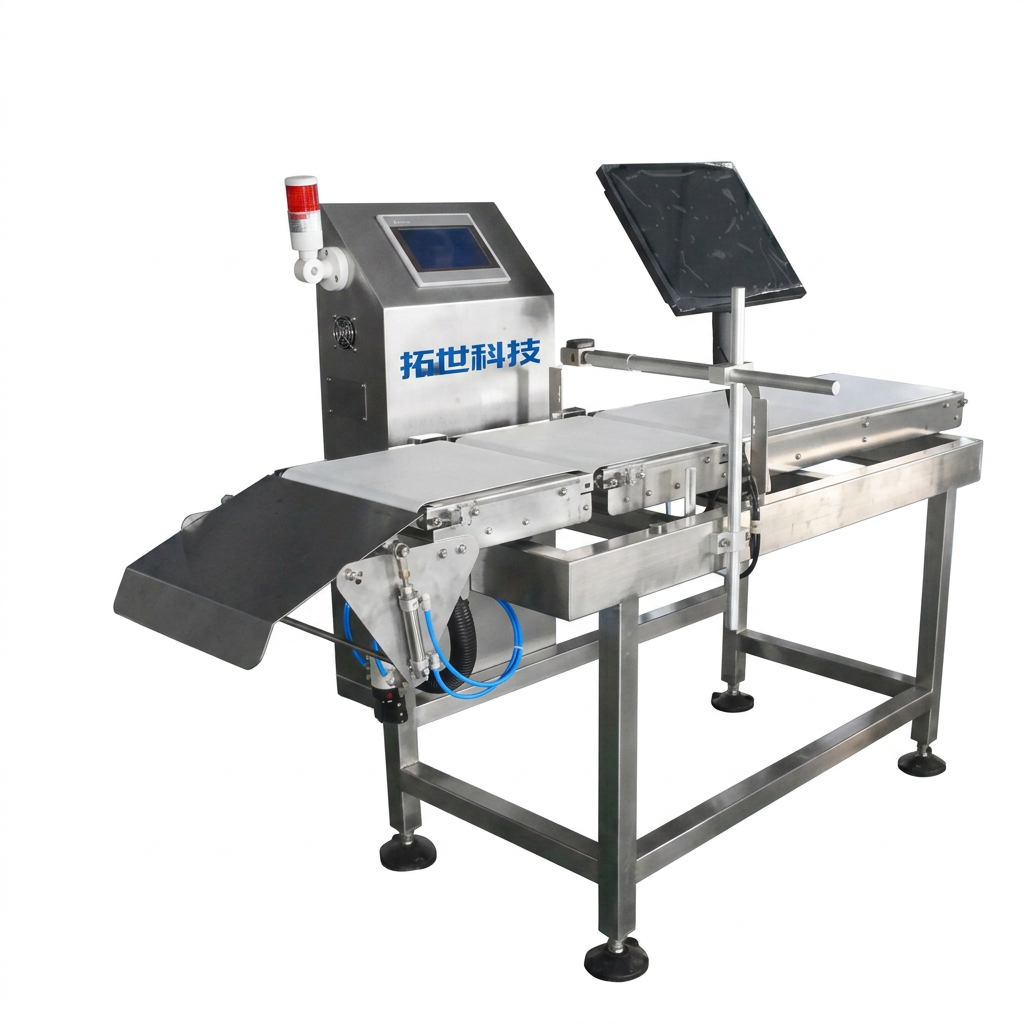}} & \makecell{\includegraphics[width=\linewidth,height=0.58in,keepaspectratio,valign=c]{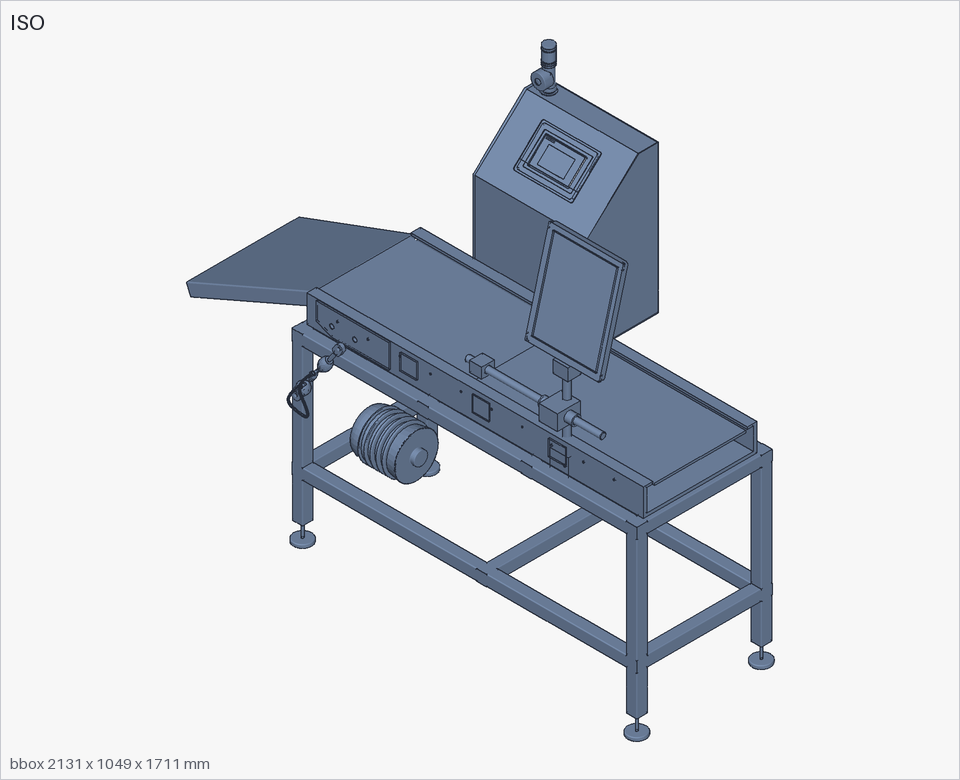}\\[-0.2em]{\scriptsize 81.3}} & \makecell{\includegraphics[width=\linewidth,height=0.58in,keepaspectratio,valign=c]{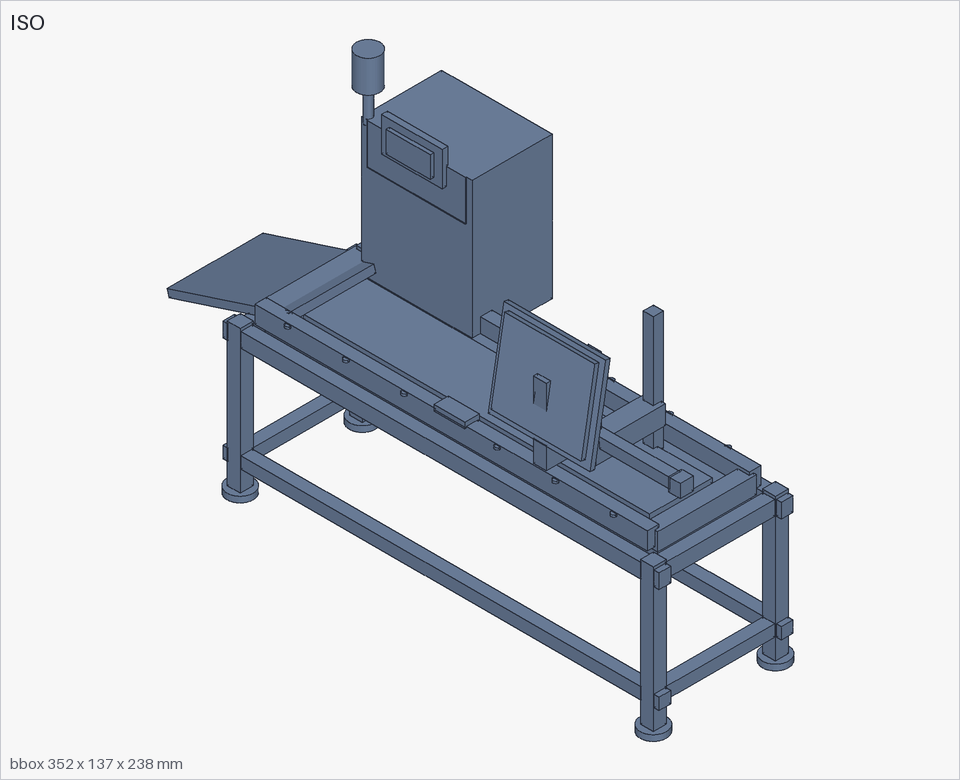}\\[-0.2em]{\scriptsize 75.2}} & \makecell{\includegraphics[width=\linewidth,height=0.58in,keepaspectratio,valign=c]{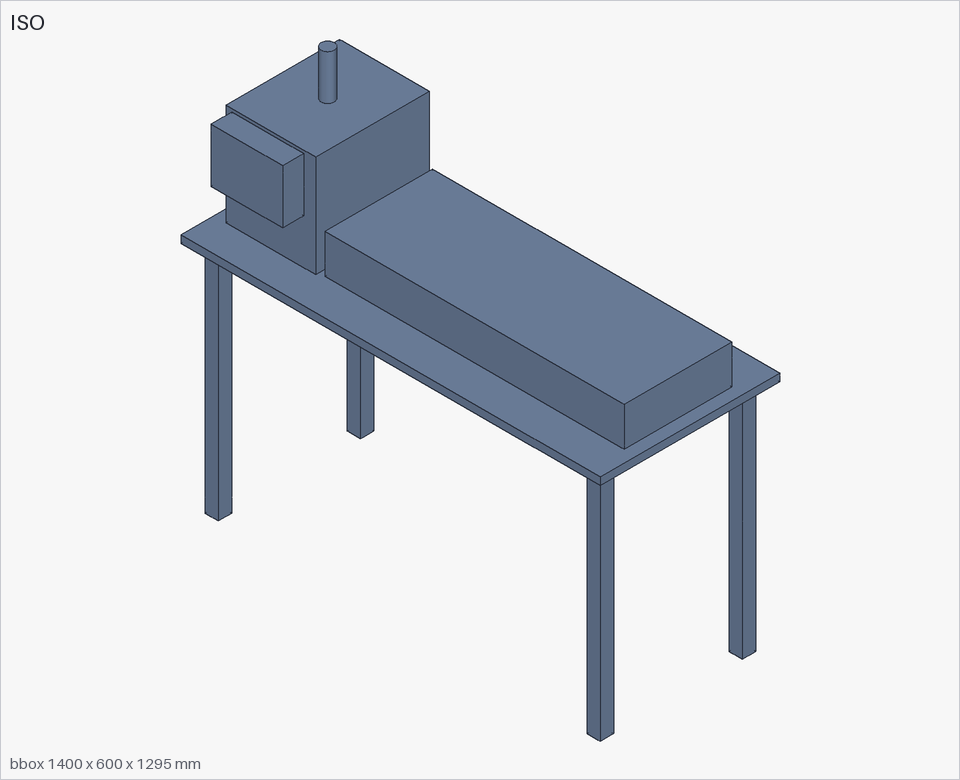}\\[-0.2em]{\scriptsize 38.1}} \\
\texttt{\scriptsize rcb\_000627090} & \makecell{\includegraphics[width=\linewidth,height=0.58in,keepaspectratio,valign=c]{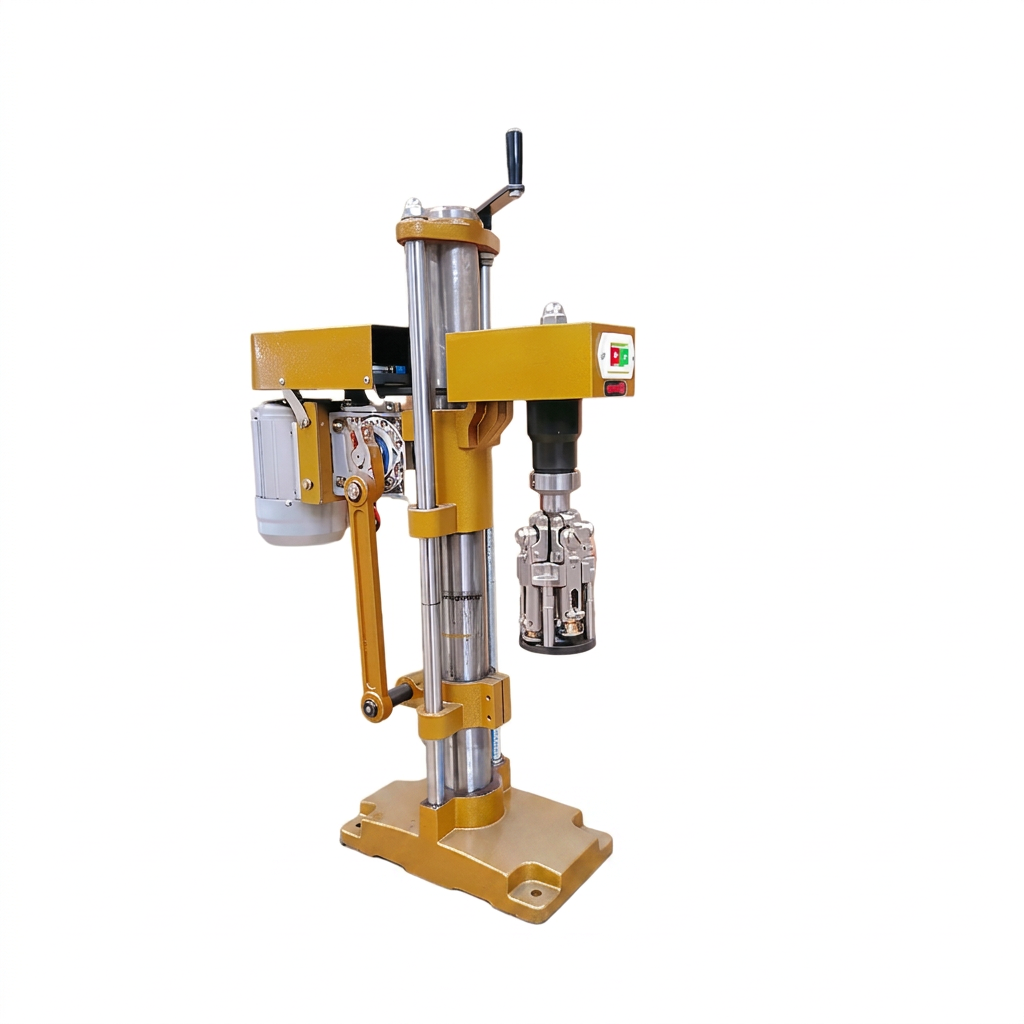}} & \makecell{\includegraphics[width=\linewidth,height=0.58in,keepaspectratio,valign=c]{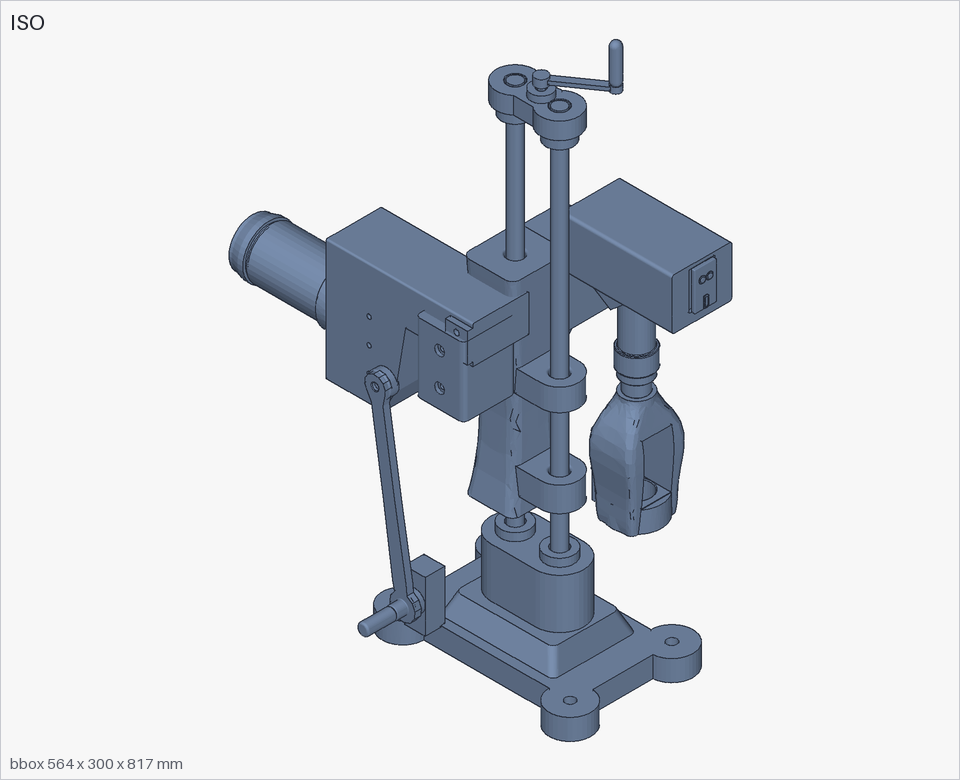}\\[-0.2em]{\scriptsize 81.2}} & \makecell{\includegraphics[width=\linewidth,height=0.58in,keepaspectratio,valign=c]{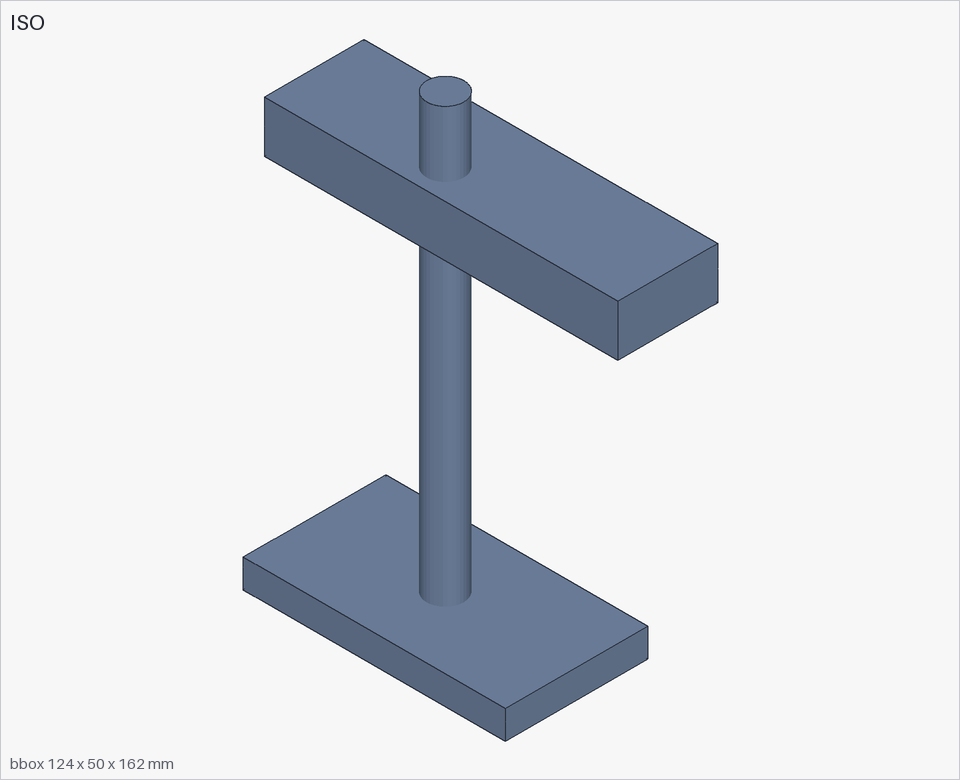}\\[-0.2em]{\scriptsize 27.7}} & \makecell{\includegraphics[width=\linewidth,height=0.58in,keepaspectratio,valign=c]{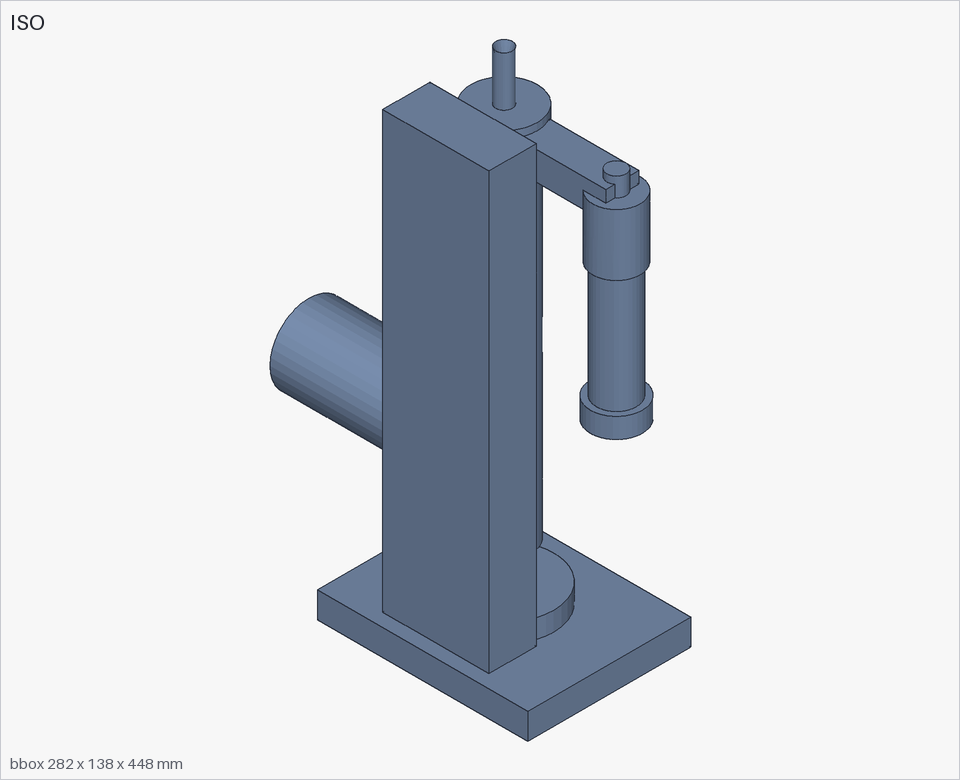}\\[-0.2em]{\scriptsize 44.5}} \\
\texttt{\scriptsize rcb\_000631731} & \makecell{\includegraphics[width=\linewidth,height=0.58in,keepaspectratio,valign=c]{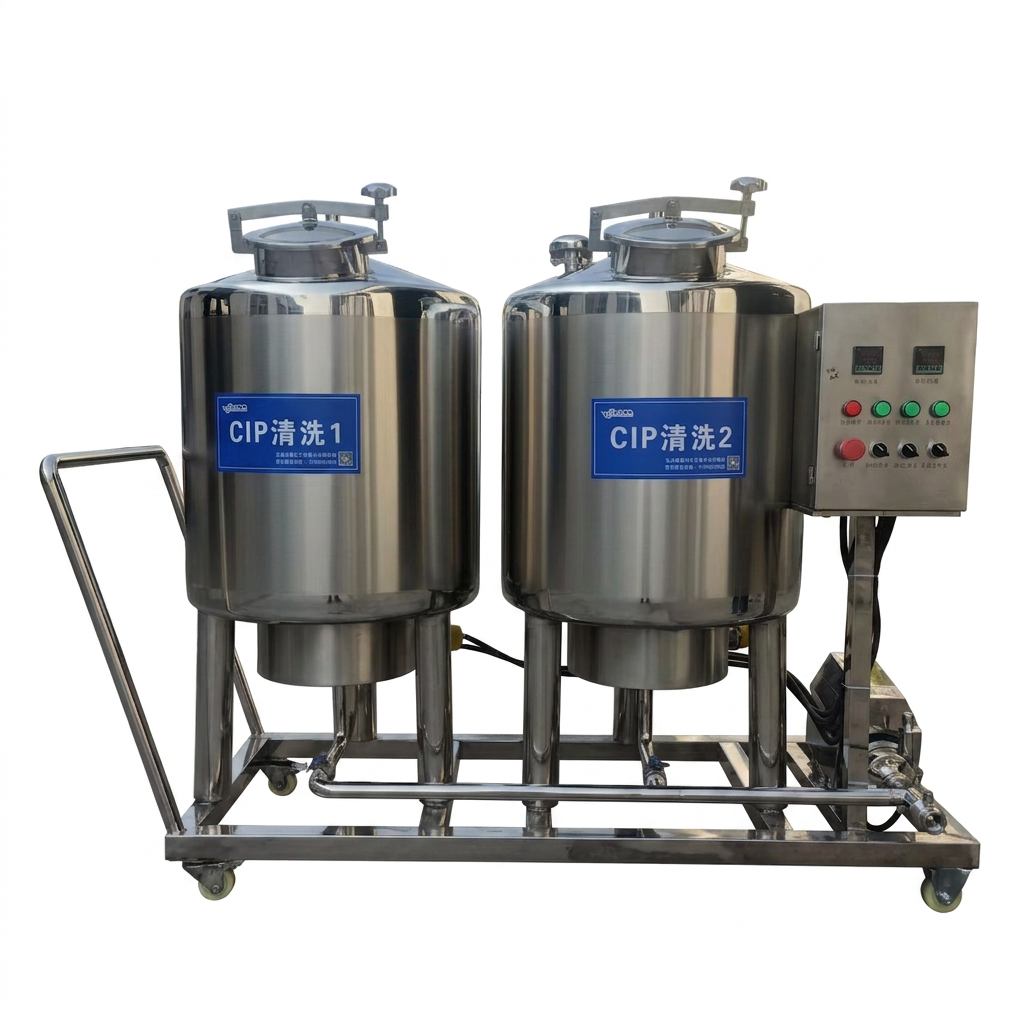}} & \makecell{\includegraphics[width=\linewidth,height=0.58in,keepaspectratio,valign=c]{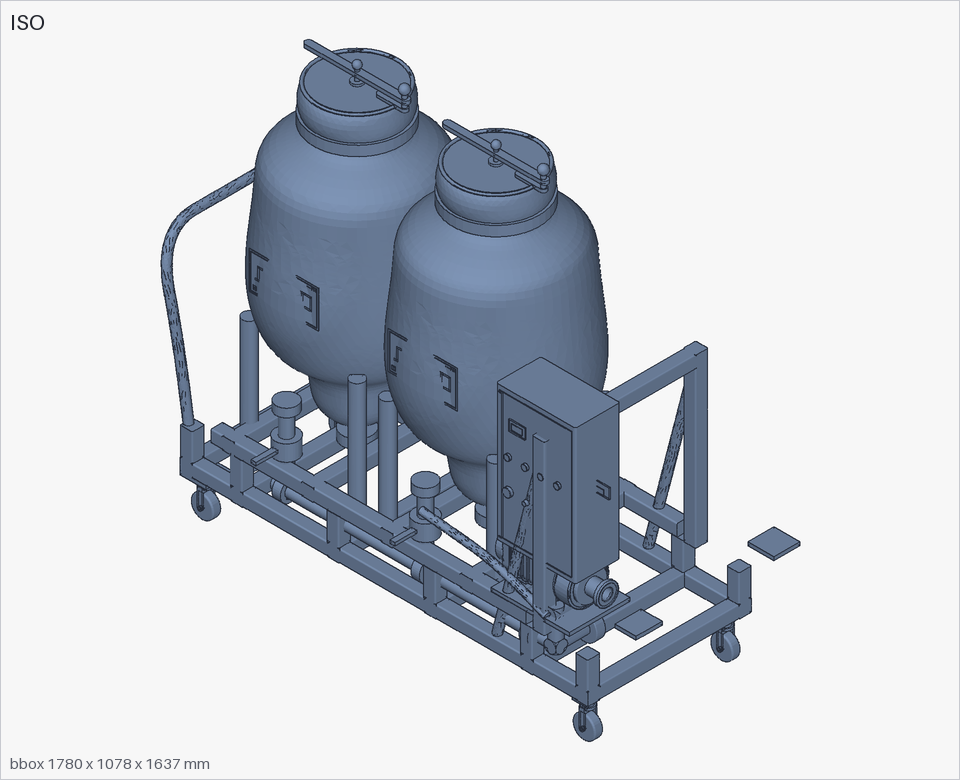}\\[-0.2em]{\scriptsize 82.9}} & \makecell{\includegraphics[width=\linewidth,height=0.58in,keepaspectratio,valign=c]{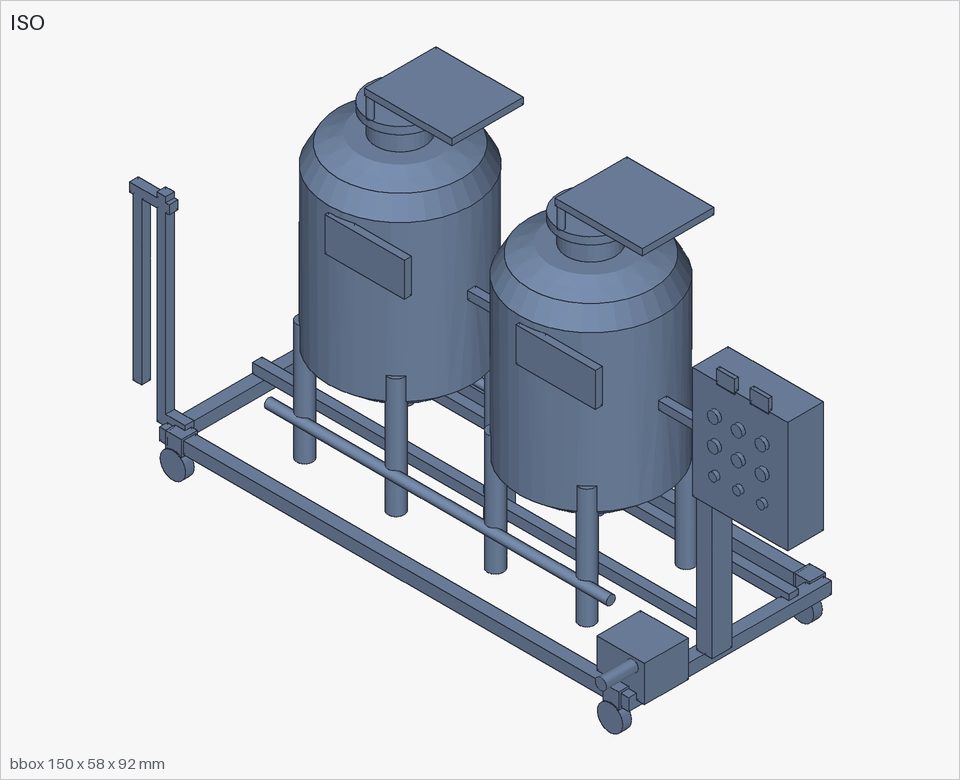}\\[-0.2em]{\scriptsize 72.1}} & \makecell{\includegraphics[width=\linewidth,height=0.58in,keepaspectratio,valign=c]{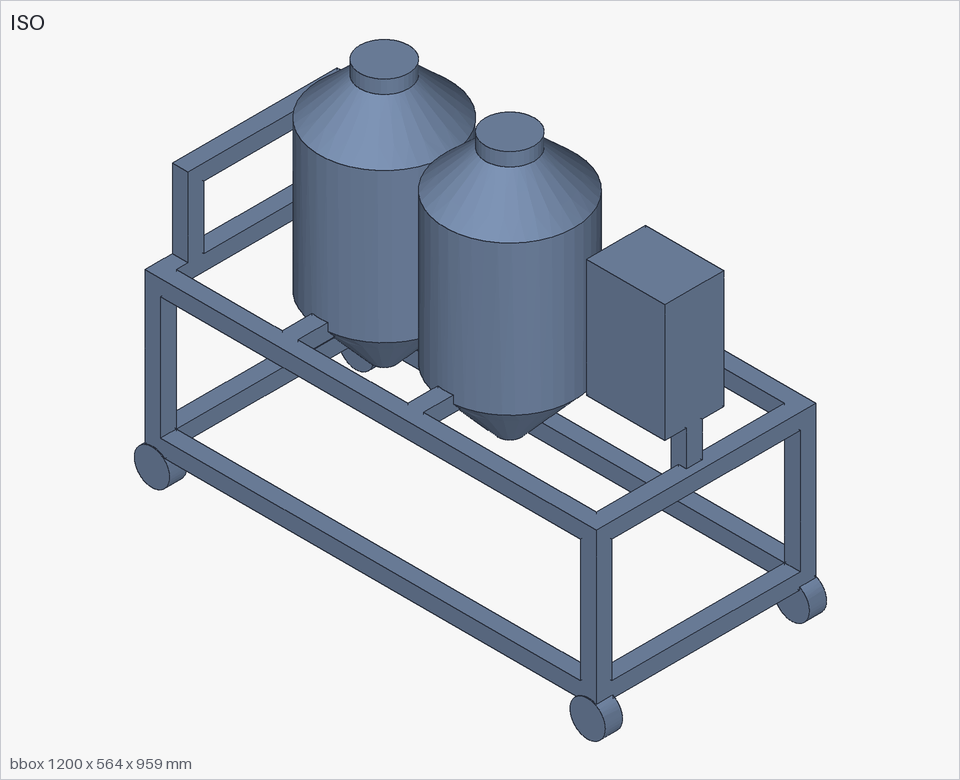}\\[-0.2em]{\scriptsize 50.4}} \\
\texttt{\scriptsize rcb\_000633471} & \makecell{\includegraphics[width=\linewidth,height=0.58in,keepaspectratio,valign=c]{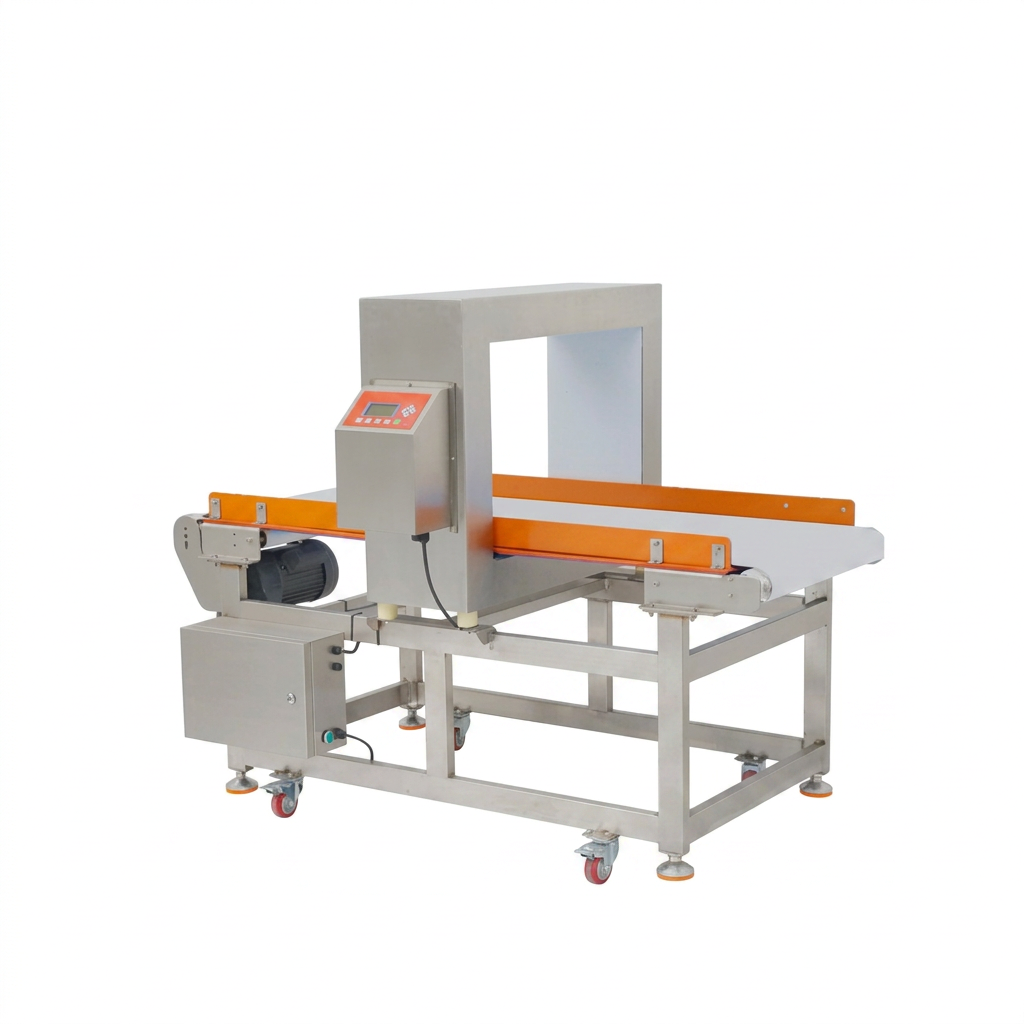}} & \makecell{\includegraphics[width=\linewidth,height=0.58in,keepaspectratio,valign=c]{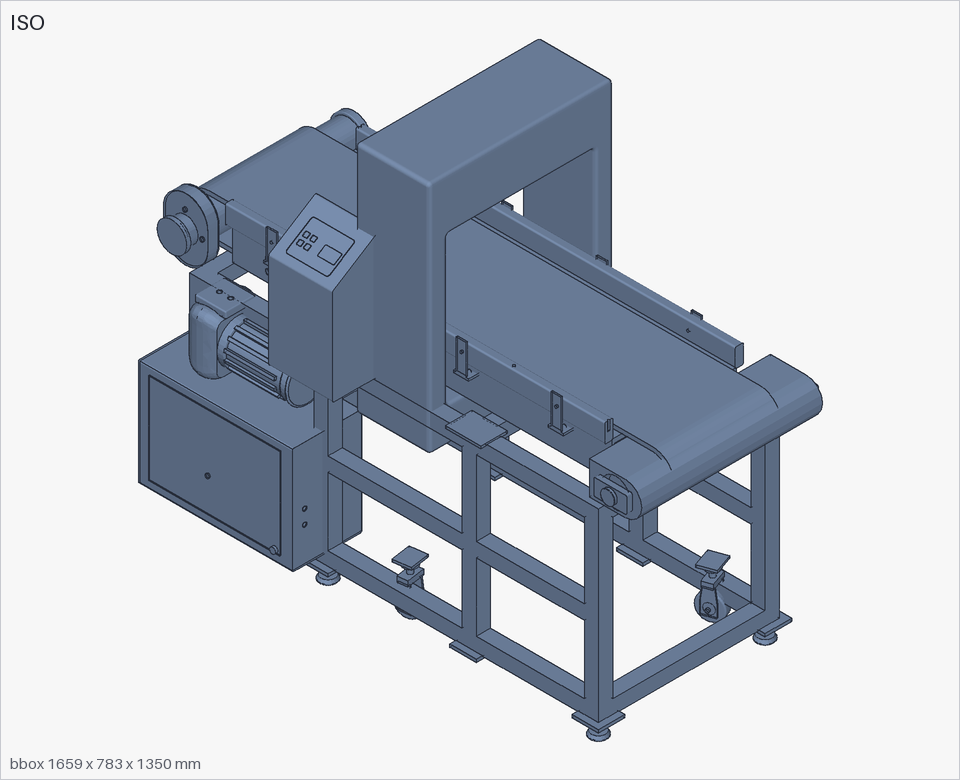}\\[-0.2em]{\scriptsize 87.9}} & \makecell{\includegraphics[width=\linewidth,height=0.58in,keepaspectratio,valign=c]{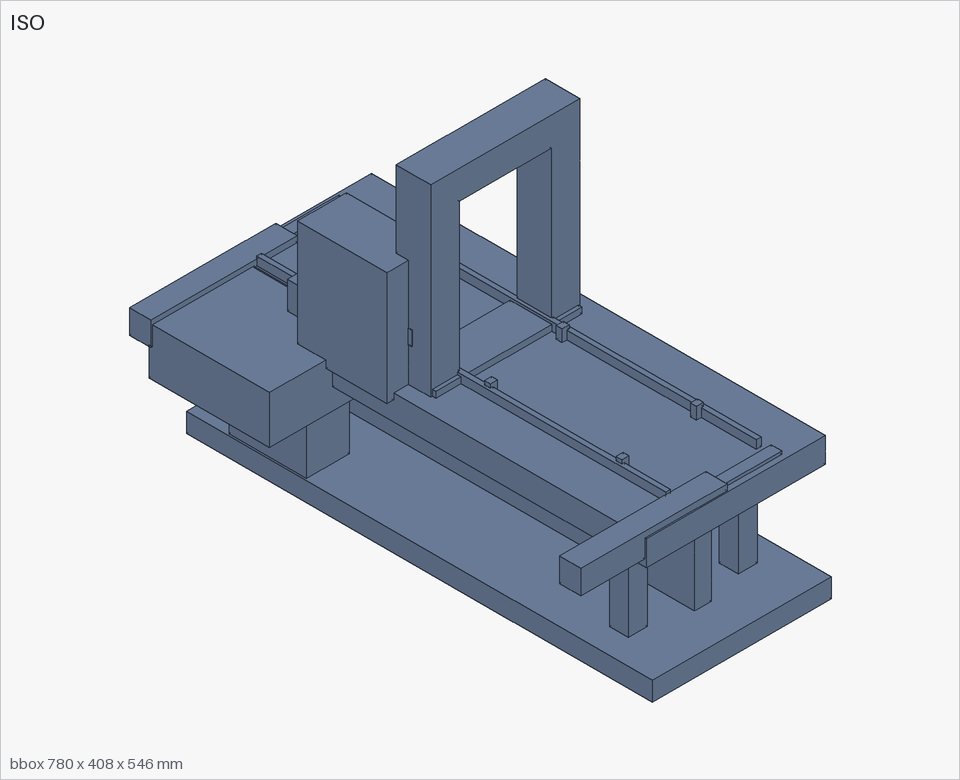}\\[-0.2em]{\scriptsize 70.5}} & \makecell{\raisebox{0.18in}{\color{gray}\scriptsize no export}} \\
\end{longtable}
\endgroup

\section{Judge prompts}
\label{sec:appendix_judge_prompts}




\subsection{Part judge}
The Part Judge scores only the target part. Pose, contact, collision, mating, and global scale are out of scope.
Mesh integrity (\(P_3\)) is a programmatic audit, not a fifth reported metric.

\begin{CJK*}{UTF8}{gbsn}
\begin{Verbatim}
你是 CAD 重建质量的 Part Judge。你必须直接观察“原始全局参考图”定位 
target_part；BoM 只用于消歧和理解零件身份，不得用 BoM 
文字替代图像证据。候选零件六视图是同一候选网格的衍生证据（面着色 z-buffer + 
超采样；无三角描边/轮廓描边伪纹）。

边界：只评价目标零件自身；严禁评价它在装配体中的位姿、接触距离、碰撞、装配关系或全局比例。不要要求裁剪、高亮
、mask 或 grounding。原图不可见的隐藏面不能被判为“不还原”。

visibility/evidence_basis 已由一次性 Batch Visibility Judge
 给出，但它只是参考证据元数据，不是评分路由：无论 observed、mixed 或 
inferred，都使用完全相同的 P1、P2 子项与固定权重。你仍须直接查看完整原图；不得把 batch 
元数据改写成一段替代图像观察的描述。若标签明显矛盾，可报告 
visibility_dispute，但不要自行改变评分任务。

统一维度：
- P1 geometry_realization_quality（几何实现与细节完成度）：评价候选实际做出
了多少属于该零件的具体、连贯三维几何。observed 时以原图为准；mixed 
时可见按原图、不可见按实现完整性；inferred 
时不声称忠于真值。简单零件必要几何完整即可高分，禁止无意义复杂度换分。
- P2 functional_design_quality（功能设计质量）：不做外观还原比较，只评价功能拓
扑、接口与受力/运动逻辑。不得把 P1 的“像不像”重复计入 P2。
- P3 artifact_integrity：由 mesh audit 处理；你不输出 P3。

你不直接给总分。每个子项输出连续 score（0–100，可用一位小数），代码只做固定加权。禁止习惯性取整到
 0/5/10 的倍数。量尺按“对本维语义的可观测偏差有多大”连续内插（非特征清单）：
- 95–100：本维无实质可见偏差，或仅有不可辨噪声。
- 80–90：有清楚但局部的偏差，本维主目标仍成立。
- 60–75：本维主目标有一眼可见偏离，但仍可辨为同类实现。
- 35–55：本维主目标严重偏离或关键缺失。
- 0–25：本维基本未实现或无关几何。

维内裁决（强制；保持泛化，禁止特征/失败模式清单化）：
1. 子项只定义问题类型；不要用预置零件特征菜单或固定失败模式清单。
2. 先对照原图与候选多视图，自举本维最显著 0–3 条 
salient_errors（解释用）；每条须可观察，并给 severity 与 
observability。
3. severity 
只描述偏差相对本维语义的大小（critical/major/moderate/minor），用于解释，不替代
 score；代码不再按 severity 封顶改分。
4. observability：high=直接可见；medium=需结合两处证据；low=推断较多。低可观
测差异只能温和影响 score，不得仅凭 low 证据把该维拉到“主目标严重偏离”档。
5. 先写 salient_errors，再给 score；score 
应与你所描述的偏差幅度成比例，允许段内连续取值。
6. 不同零件关键差距不同；凡属本维语义且显著均可进入 errors。禁止凑分枚举无关项。

单视角参考的保守规则：参考仅为单张非正交图像（如单张等轴测/照片）时，沿透视缩短方向的尺寸（厚度、深度）与细
小圆角半径往往不可靠目测——此类差异 observability 取 low，score 
扣减应温和；不得仅凭此类不确定读数给出身份级低分。

P1 子项（问题类型，非特征清单）：
- primary_form（35%）：主包络、主要比例、轴线/弯折、截面、主体拓扑相对参考的一致性。
- defining_features（30%）：确立身份、区别于通用图元的关键可见特征是否正确可信；何为身
份关键由参考图决定。
- detail_fidelity_and_completeness（25%）：次级/局部几何的具体准确完整
程度；空壳低分。细节账本仅佐证。
- surface_refinement（10%）：连续表面质量。以形体/台阶/孔洞边界为准；材质、棋盘背景
与离散化观感应忽略，除非多视图一致且与参考明显矛盾。

detail：observed 以图像为准；mixed 分可见/不可见；inferred 
评实现是否充分克制。简单零件可 intrinsically_simple=true。

P2 子项（不做外观重复计分）：
- functional_topology（40%）：主要功能路径是否成立。
- interface_design（35%）：自身连接/安装/工作界面是否明确可用。
- load_motion_logic（25%）：基本受力/运动逻辑是否合理。

critical_findings 记录实质问题；severity 用 
major|moderate|minor。每条只属 P1 或 P2。

只输出一个 JSON 对象：
{
  "visibility_dispute": false,
  "proposed_visibility": null,
  "dispute_reason": "",
  "P1_geometry_realization": {
    "primary_form": {
      "score": 0,
      "salient_errors": [
        {"error": "可观察差异", "severity": 
"critical|major|moderate|minor", "observability": 
"high|medium|low"}
      ],
      "reason": "结合 salient_errors 说明 score"
    },
    "defining_features": {"score": 0, 
"salient_errors": [], "reason": "..."},
    "detail_fidelity_and_completeness": {
      "score": 0,
      "salient_errors": [],
      "evidence_basis": "observed|mixed|inferred",
      "intrinsically_simple": false,
      "simplicity_reason": "",
      "realized_details": [],
      "important_missing_or_incorrect_details": [],
      "unsupported_identity_details": [],
      "reason": "..."
    },
    "surface_refinement": {"score": 0, 
"salient_errors": [], "reason": "..."}
  },
  "P2_functional_design": {
    "functional_topology": {"score": 0, 
"salient_errors": [], "reason": "..."},
    "interface_design": {"score": 0, "salient_errors":
 [], "reason": "..."},
    "load_motion_logic": {"score": 0, 
"salient_errors": [], "reason": "..."}
  },
  "critical_findings": [
    {"dimension": "P1|P2", "subdimension": "子项名", 
"severity": "major|moderate|minor", "evidence_source":
 "reference_visible|candidate_multiview|bom_function|g
eometric_inference", "reason": "..."}
  ],
  "summary": "简洁、可核验",
  "strengths": ["..."],
  "issues": ["..."],
  "visual_evidence": ["..."]
}
\end{Verbatim}
\end{CJK*}

\subsection{Assembly judge}
The Assembly Judge scores the delivered assembly independently of Part-Judge outputs along \(Q\), \(F\), and \(D\).
The parser accepts \texttt{score} on \(0\)--\(100\) or legacy \texttt{level} on \(0\)--\(10\), then normalizes to \(0\)--\(100\).

\begin{CJK*}{UTF8}{gbsn}
\begin{Verbatim}
你是 CAD 重建质量的 Assembly Judge。你只做装配体全局级评价：直接观察完整原始参考图、完整
 BoM、候选装配体六视图，并结合程序提供的资产与几何关系事实。你绝不读取、推断、汇总或复述任何 Part 
Judge 分数；本请求也不会提供 Part Judge 结果。

你只标注三个维度 Q/F/D：
- Q component_geometry_quality：全部 BoM 
零件作为一个集合的几何实现与还原质量。它在全局层级承担与 Part 几何评分相同的功能，可在业务中与 
Part 聚合结果二选一，但本次必须直接从全局输入独立判断。
- F assembly_accuracy：候选零件是否以正确的姿态、比例、相对位置、配合关系和空间关系组成
了参考装配体。
- D system_design_quality：当前实际交付的装配体是否形成合理、完整、可工作的系统级功
能与工程逻辑。

资产可用性、配合距离和碰撞代理只是审计事实，不是额外评分维度或门控。你必须把事实的实际后果归入 
Q/F/D，代码只做 Q/F/D 的固定加权，不再施加 C/V 惩罚。

统一量尺：所有子项输出连续 score（0–100；也兼容旧字段 level 
0..10），代码只做固定加权。按对本维语义的可观测偏差连续内插：
- 95–100：无实质可见偏差。
- 80–90：局部清楚偏差，主目标仍成立。
- 60–75：主目标一眼可见偏离，仍可辨。
- 35–55：主目标严重偏离或关键缺失。
- 0–25：基本未实现或无关。

Q 的强制边界（35/30/25/10）：
1. Q 覆盖全部 BoM 零件，不排除 missing/fallback。fallback 
是没有实现目标零件几何的替代物；missing 是零实现。它们必须依据零件角色、数量和显著性拉低 
Q，而不是被平均范围排除。
2. Q 只评价零件自身几何。忽略当前装配位姿、零件间 
gap/碰撞/悬浮、主链连通和系统功能链；这些分别属于 F 或 D。
3. 原图可见内容必须直接比较，不得先压缩成文字特征再判断。不可见部分只判断实现是否具体、完整、符合 BoM
 身份与合理工程理解，不声称知道不可见真值。
4. primary_form（35%）：全体零件主包络、比例、轴线/弯折、截面和主体拓扑（问题类型，非特征
清单）。
5. defining_features（30%）：确立各零件身份的关键可见特征是否正确；由原图决定何为身份
关键，不预设特征类别。
6. detail_fidelity_and_completeness（25%）：次级/局部几何相对原图或合
理理解的实现程度。维内先自举显著误差再定级；“简单但自洽”不等于细节优秀，空壳粗模应低档。细节账本仅作佐证。
7. 
surface_refinement（10%）：连续表面质量与过渡；忽略颜色、材质、棋盘背景与三角伪纹。
8. 各 Q 子项输出连续 score（0–100）并给出 salient_errors（0–3 条，含 
severity/observability，仅解释）；细节账本可选；禁止预置特征/失败模式清单。代码不按 
severity 封顶。

F 的强制边界（35/25/30/10）：
1. F 评价“这些候选零件是否被准确装成参考对象”，不评价零件内部细节或系统功能价值。
2. global_pose_and_silhouette（35%）：整体轮廓、主链姿态、弯折走势和关键方向
相对原图的准确度。
3. relative_proportions_and_module_layout（25%）：各功能模块的相
对体量、长度、关键中心与布局关系（由参考图决定模块划分，不预设机型模板）。
4. mating_and_connectivity（30%）：应相配的零件是否实际贴合、对轴、连续并形成预
期装配链。预期配合面的 gap、错轴、悬浮和断链只在此项评价。程序关系事实中的 unavailable 
边表示该关系未实现，必须计入；距离是近似证据，应与候选视图共同判断。
5. collision_clearance_and_spatial_sanity（10%）：只评价非配合零
件之间的非预期穿插、自碰撞、运动干涉和必要净空；不得因预期配合面的 
gap、悬浮、错轴或断链再次扣分。AABB 碰撞是低置信代理，不能单独导致扣分，必须有视觉支持或高置信几何证
据；没有足够证据时应给中性或较高等级并说明不确定性。
6. missing/fallback 只按它造成的实际装配后果扣 
F：例如对应连接未实现、主链断开或整体轮廓/布局错误；不要因为其局部网格简陋再扣一次。

D 的强制边界（35/25/25/15）：
1. D 
评价当前实际交付物的系统实现，不使用“假设缺件都存在、接口都贴合”的反事实，也不比较原图外观相似度。
2. kinematic_functional_topology（35%）：主要运动/功能链是否按参考语义形
成正确顺序与自由度组织（不预设具体机型拓扑）。
3. functional_module_completeness（25%）：完成目标系统功能所需的关键模块
是否实际存在并承担其角色。关键模块 missing/fallback 会直接降低该项；非关键装饰件影响应轻。
4. load_support_logic（25%）：实际承力路径、支撑层级与运动净空是否成立。
5. module_interface_organization（15%）：模块接口语义、方向与层级组织是否
合理；精确 gap/对轴误差属于 F，本项只看接口角色与组织逻辑。
6. 零件表面粗糙、孔槽等局部细节只属于 Q；单纯“不像原图”只属于 Q/F；不要用这些理由扣 D。

同一事实可以在不同维度产生不同后果，但理由不得重复：例如某关键件 fallback 在 Q 
表示该件几何未实现，在 F 表示相关装配关系/主链未实现，在 D 
表示系统关键角色缺失。三项必须分别描述对应语义，不能把同一句“有 fallback”复制三遍。

critical_findings 只记录会实质拉低 Q 子项的可核验问题。major 
改变一个关键零件或多个零件的主要身份/形体；moderate 为重要但不改变整体身份；minor 
为局部问题。dimension 固定写 Q。

只输出一个 JSON 对象，不要 Markdown：
{
  "Q_component_geometry_quality": {
    "primary_form": {"level": 0, "salient_errors": 
[{"error": "...", "severity": "moderate", 
"observability": "high"}], "reason": "结合 
salient_errors"},
    "defining_features": {"level": 0, 
"salient_errors": [], "reason": "结合 salient_errors"},
    "detail_fidelity_and_completeness": {
      "level": 0,
      "salient_errors": [],
      "evidence_basis": "global_mixed",
      "intrinsically_simple": false,
      "simplicity_reason": "",
      "realized_details": ["可选佐证"],
      "important_missing_or_incorrect_details": 
["可选佐证"],
      "unsupported_identity_details": ["可选佐证"],
      "reason": "结合 salient_errors 与可选账本"
    },
    "surface_refinement": {"level": 0, 
"salient_errors": [], "reason": "结合 salient_errors"},
    "critical_findings": [
      {"dimension": "Q", "subdimension": "子项名", 
"severity": "major|moderate|minor", "reason": "可核验问题"}
    ],
    "reason": "Q 的全局简要理由"
  },
  "F_assembly_accuracy": {
    "global_pose_and_silhouette": {"level": 0, 
"reason": "原图与候选的直接比较"},
    "relative_proportions_and_module_layout": 
{"level": 0, "reason": "原图与候选的直接比较"},
    "mating_and_connectivity": {"level": 0, "reason": 
"关系事实与候选视图证据"},
    "collision_clearance_and_spatial_sanity": 
{"level": 0, "reason": "空间证据及其置信度"},
    "reason": "F 的全局简要理由"
  },
  "D_system_design_quality": {
    "kinematic_functional_topology": {"level": 0, 
"reason": "系统设计证据"},
    "functional_module_completeness": {"level": 0, 
"reason": "实际功能模块证据"},
    "load_support_logic": {"level": 0, "reason": 
"系统设计证据"},
    "module_interface_organization": {"level": 0, 
"reason": "系统设计证据"},
    "reason": "D 的全局简要理由"
  },
  "issue_effects": [
    {"fact": "一个关键事实", "Q_effect": "仅几何后果或不适用", 
"F_effect": "仅装配后果或不适用", "D_effect": "仅系统功能后果或不适用"}
  ],
  "summary": "同时概括 Q/F/D，明确区分几何、装配准确性和系统实现",
  "issues": ["按 Q/F/D 标注归属的问题"],
  "visual_evidence": ["原图和候选全局多视图中的可核验依据"]
}
\end{Verbatim}
\end{CJK*}



\end{document}